\documentclass{article}

\PassOptionsToPackage{numbers, compress}{natbib}

\usepackage[final]{neurips_data_2023}

\usepackage[utf8]{inputenc}
\usepackage[T1]{fontenc}
\usepackage{hyperref}
\usepackage{url}
\usepackage{booktabs}
\usepackage{amsfonts}
\usepackage{nicefrac}
\usepackage{microtype}
\usepackage{xcolor}

\usepackage{graphicx}
\usepackage{subcaption}
\usepackage{wrapfig}
\usepackage{amsmath}
\usepackage{multirow}
\usepackage[version=4]{mhchem}
\usepackage{tikz}

\newcommand{\secref}[1]{Section~\ref{#1}}
\newcommand{\figref}[1]{Figure~\ref{#1}}
\newcommand{\tabref}[1]{Table~\ref{#1}}

\newcommand{\secsref}[2]{Sections~\ref{#1} and~\ref{#2}}
\newcommand{\figsref}[2]{Figures~\ref{#1} and~\ref{#2}}

\newcommand{\secssref}[3]{Sections~\ref{#1}, \ref{#2}, and~\ref{#3}}
\newcommand{\figssref}[3]{Figures~\ref{#1}, \ref{#2}, and~\ref{#3}}

\newcommand{\bsB}{\boldsymbol B}
\newcommand{\bsE}{\boldsymbol E}
\newcommand{\bsJ}{\boldsymbol J}

\title{Datasets and Benchmarks for Nanophotonic Structure and Parametric Design Simulations}
\author{%
  Jungtaek Kim \\
  University of Pittsburgh \\
  Pittsburgh, PA 15261, USA \\
  \texttt{jungtaek.kim@pitt.edu} \\
  \And
  Mingxuan Li \\
  University of Pittsburgh \\
  Pittsburgh, PA 15261, USA \\
  \texttt{mil152@pitt.edu} \\
  \AND
  Oliver Hinder \\
  University of Pittsburgh \\
  Pittsburgh, PA 15261, USA \\
  \texttt{ohinder@pitt.edu} \\
  \And
  Paul W. Leu \\
  University of Pittsburgh \\
  Pittsburgh, PA 15261, USA \\
  \texttt{pleu@pitt.edu} \\
}

\begin{document}

\maketitle

\begin{abstract}
Nanophotonic structures have versatile applications including solar cells, anti-reflective coatings, electromagnetic interference shielding, optical filters, and light emitting diodes. To design and understand these nanophotonic structures, electrodynamic simulations are essential. These simulations enable us to model electromagnetic fields over time and calculate optical properties. In this work, we introduce frameworks and benchmarks to evaluate nanophotonic structures in the context of parametric structure design problems. The benchmarks are instrumental in assessing the performance of optimization algorithms and identifying an optimal structure based on target optical properties. Moreover, we explore the impact of varying grid sizes in electrodynamic simulations, shedding light on how evaluation fidelity can be strategically leveraged in enhancing structure designs.
\end{abstract}

\section{Introduction}
\label{sec:introduction}

Nanophotonic structures play a crucial role for a wide range of real-world applications such as solar cells,
anti-reflective coatings,
electromagnetic interference shielding,
optical filters,
and light emitting diodes.\footnote{Our implementation can be found at \href{https://github.com/jungtaekkim/nanophotonic-structures}{https://github.com/jungtaekkim/nanophotonic-structures}.}
In particular,
several studies have demonstrated that 
nanophotonic structures can enhance target performance
for many applications~\citep{WangB2012nt,WangB2012ol,WangB2014oe,HaghanifarS2020optica,HaghanifarS2022oe,LiM2022oe}.
Electrodynamic simulations, which are based on Maxwell's equations, provide accurate predictions of the optical and electromagnetic properties of these structures~\citep{GriffithsDJ2005book}.
Leveraging the utility of these simulations, we can combine them with optimization procedures to design and discover new and improved structures.

In this paper,
we introduce datasets and benchmarks for nanophotonic structures,
focusing on parametric design simulations in relation to optical properties.
These benchmarks allow us to analyze the performance of optimization algorithms such as derivative-free algorithms~\citep{RiosLM2013jgo} and Bayesian optimization~\citep{GarnettR2023book}, 
when it comes to identifying optimal photonic structures based on a target optical property.
Our frameworks support two modes for fast prototyping of optimization methods:
discretized search space mode and surrogate model mode.
The discretized search space mode uses stored simulated results over grid query points;
the surrogate model mode 
estimates any query points within a continuous search space based on training with our datasets.  
These modes allow users to quickly test their optimization algorithms without running expensive electrodynamic simulations.
Our system facilitates the use of evaluation fidelity, which is controlled by a simulation resolution,
and the handling of multiple objectives.
Moreover, our frameworks can also provide electromagnetic fields as a function of position and time.

Our contributions are summarized as follows:
\begin{itemize}
    \item Development of a generic simulation scheme and pipeline for nanophotonic structures in Python, based on the open-source software, Meep, and licensed under the MIT license;
    \item Creation of datasets of a myriad of nanophotonic structures for electromagnetic interference shielding, anti-reflection, and solar cells;
    \item Investigation into the effects of altering grid sizes in electrodynamic simulations, providing insights into tradeoffs between computational time and simulation accuracy;
    \item Introduction of benchmarks specifically designed for the optimization of parametric structures, facilitating the evaluation and comparison of different optimization algorithms.
\end{itemize}

\section{Background}
\label{sec:related}

In this section, we delve into the  optical and electromagnetic properties of materials, explore nanophotonic structure designs, and review related literature.

\subsection{Optical and Electromagnetic Properties of Materials}

The optical and electromagnetic properties of materials
can be determined by solving Maxwell's equations~\citep{GriffithsDJ2005book}.
When studying materials featuring sizes smaller than wavelengths of interest,
geometrical optics becomes unsuitable, and ray-tracing methods are inaccurate~\citep{BornM2013book}.
In such cases, it is crucial to use a simulation method that
captures the wave-like nature of light to accurately address phenomena like interference and diffraction.
Classical electromagnetic theory can predict reflection, absorption, and transmission spectra by solving Maxwell's equations:
\begin{equation}
    \nabla \cdot \bsE = \frac{\rho}{\epsilon_0},
    \qquad
    \nabla \cdot \bsB = 0,
    \qquad
    \nabla \times \bsE = - \frac{\partial \bsB}{\partial t},
    \qquad
    \nabla \times \bsB = \mu_0 \bsJ + \mu_0 \epsilon_0 \frac{\partial \bsB}{\partial t},
\end{equation}
where $\bsE$ and $\bsB$ are electric and magnetic fields,
$\rho$ and $\bsJ$ are charge and current densities,
and $\epsilon_0$ and $\mu_0$ are the permittivity and permeability of free space.
The speed of light is defined as $c = 1 / \sqrt{\mu_0 \epsilon_0}$.

As shown in~\figref{fig:em_wave},
light consists of two synchronized waves:
\emph{electric fields} and \emph{magnetic fields}.
These fields are characterized by an angular frequency $\omega$, which can also be expressed as a frequency $f$, energy $E$, or wavelength $\lambda$, 
with the relationships $\omega = 2 \pi f$, $E = h f$, and $E = h c/\lambda$ where $h$ is the Planck constant (i.e., $6.63 \times 10^{-34}$~J-sec) and $c$ is the speed of light (i.e., $3 \times 10^{8}$~m/sec).
The interaction of light with materials is characterized by either a complex refractive index, $n(\omega) = n_r (\omega) + n_i (\omega)$, or a complex permittivity, $\epsilon (\omega) = \epsilon_r (\omega) + \epsilon_i (\omega)$,
both of which describe how light bends 
and is absorbed within that material.
The relative permittivities of the materials used in this paper are detailed in~\secref{sec:materials}.
An important aspect of light-material interactions is understanding the 
transmission, reflection, and absorption spectra from a specific light source as it interacts with a material at a given frequency, as depicted in~\figref{fig:ref_abs_tra}.
As governed by the law of conservation of energy~\citep{GriffithsDJ2005book},
the following equation is satisfied:
\begin{equation}
    R(\omega) + A(\omega) + T(\omega) = 1,
    \label{eqn:sum_optical_properties}
\end{equation}
where $R(\omega)$, $A(\omega)$, and $T(\omega)$ are the reflectance, absorbance, and transmittance of a material at a particular frequency $\omega$.

\subsection{Designs of Nanophotonic Structures for Optoelectronic Applications}

Numerous electrodynamic simulation techniques such as the transfer matrix method~\citep{BillardJ1967phdthesis}, finite element method~\citep{JinJM2015book}, rigorous coupled-wave analysis~\citep{MoharamMG1981josa}, and finite-difference time-domain (FDTD) method~\citep{TafloveA1980ieeetec} are important for nanophotonics research.
FDTD, in particular, is widely used across various domains to explore how electromagnetic waves interact with different materials.
Its applications span across photonic crystals, waveguides, plasmonics, and metamaterials.
In our research,
we harness the FDTD simulations for the design and optimization of optical devices,
focusing on applications in solar cells, anti-reflection, and transparent electromagnetic interference shielding.

\paragraph{Solar Cells.}

Nanomaterials are revolutionizing solar cell technology, promising significantly enhanced efficiency and reduced costs.
Certain nanostructures have the potential to surpass the Shockley-Queisser efficiency limit and capture light
beyond the Yablonovitch or Lambertian limits~\citep{PolmanA2012nm,ShockleyW1961jap,YablonovitchE1982ieeeted,CallahanDM2012nl}.
In this paper we investigate vertical nanowire arrays and nanosphere coatings, as they have proven abilities to enhance light trapping~\citep{KelzenbergMD2008nl,GrandidierJ2011am,WangB2015ne,WangB2016ne,HaghanifarS2022oe}.
Our evaluation of solar cells is based on their ultimate efficiency, excluding transport and radiative losses, effectively focusing on optimizing solar absorption.

\paragraph{Anti-Reflection.}

Light traveling from air to glass partially reflects due to the disparity in index of refraction. 
Anti-reflection thin film coatings can be used to achieve perfect anti-reflection at a single wavelength and normal incidence angle.
However, these structures are less effective for reflection across a range of wavelengths or incidence angles.
This challenge has been tackled with moth eye~\citep{WilsonSJ1982oaijo,StavengaDG2006prsb} and glasswing butterfly wing-inspired structures~\citep{SiddiqueRH2015nc,HaghanifarS2019mh},
which contain sub-wavelength structures with an effective refractive index that gradually transits between those of the air and the glass.  
These structures achieve broad-spectrum and wide-angle anti-reflection, enhancing light emission from displays and improving light intake for photodetectors and solar cells.

\paragraph{Electromagnetic Interference Shielding.}

As the usage of electronic devices has grown, there is a growing demand for strategies to shield these devices from 
external electromagnetic waves
and interference, which can lead to disruptions.
Furthermore,
transparent electromagnetic interference shielding has emerged as a significance field of interest in this context.
However, a considerable challenge is balancing effective shielding performance and transparency~\citep{WangH2019acsami,LiM2023acsami}.
Notably, multi-layer films have been identified as a viable solution, successfully attaining high levels of both transparency and shielding effectiveness~\citep{ManiyaraRA2016nc,WangH2019acsami}.
Furthermore, the integration of nanocone structures into these films enhances their transparency, offering an innovative approach for this complex challenge~\citep{LiM2022oe}.

\subsection{Related Work}

The field of molecular discovery has garnered extensive attention across numerous studies, leading to significant advancements~\citep{SanchezB2018science}.
One such contribution is the creation of a comprehensive dataset featuring a large array of organic molecules,
complete with details on their geometric, energetic, electronic, and thermodynamic properties~\citep{RamakrishnanR2014sd}.
This dataset serves as a valuable resource for framing and addressing the challenges associated with molecular discovery.
MoleculeNet, a tool developed using the DeepChem software,
further extends this by consolidating various public molecular property datasets~\citep{WuZ2018cs}.
Moreover, benchmarks specifically designed for de novo molecular design have been introduced to aid in solving inverse molecular discovery problems~\citep{BrownN2019jcim}.

The emergence of deep learning for nanophotonic device design, particularly in inverse design methods, has been discussed~\citep{SoS2020np}.
This era has also witnessed exploration into data-driven surrogate models for artificial electromagnetic materials,
utilizing neural networks to capture the intricate behaviors of metamaterials, nanophotonics, and color filters~\citep{DengY2021neuripsdb}.
A noteworthy study has conducted a comprehensive comparison of various graph neural networks, evaluating their effectiveness in learning the dynamics of simple physical systems~\citep{ThangamuthuA2022neuripsdb}.
Furthermore, a combination of Bayesian optimization and Bayesian neural networks has been effectively applied to tackle complex scientific challenges such as photonic crystal topology and quantum chemistry~\citep{KimS2022tmlr}.

In the fields of machine learning and optimization, the community often leverages diverse benchmark functions.
These include simple, cost-effective synthetic functions such as Branin, Rastrigin, and Ackley functions.
The benchmarks extend beyond these elementary examples, encompassing a variety of scenarios for applications like black-box optimization~\citep{HansenN2010gecco,TurnerR2020neuripscd},
hyperparameter optimization~\citep{EggenspergerK2021neuripsdb,PfistererF2022automlconf,SehicK2022automlconf},
and neural architecture search~\citep{YingC2019icml,DongX2019iclr,KleinA2019arxiv,DongX2021ieeetpami}.
Despite the diversity, a noticeable gap exists, as these benchmarks predominantly feature similar types of examples and problems,
offering limited insight into the complex challenges prevalent in real-world scientific and engineering contexts.

\section{Nanophotonic Structures and Their Design Problems}
\label{sec:main}

In this section we introduce the scheme of electrodynamic simulations,
our nanophotonic structures,
and their design problems.
These include specifying possible material choices and ranges of sizes for structure components.
Moreover, we detail the consideration of simulation fidelity and multiple objectives.
We also visualize how our objectives vary as we change parameters that define a structure.

\subsection{Electrodynamic Simulations}

\begin{wrapfigure}{r}{0.48\textwidth}
    \centering
    \vspace{-10pt}
    \includegraphics[width=0.46\textwidth]{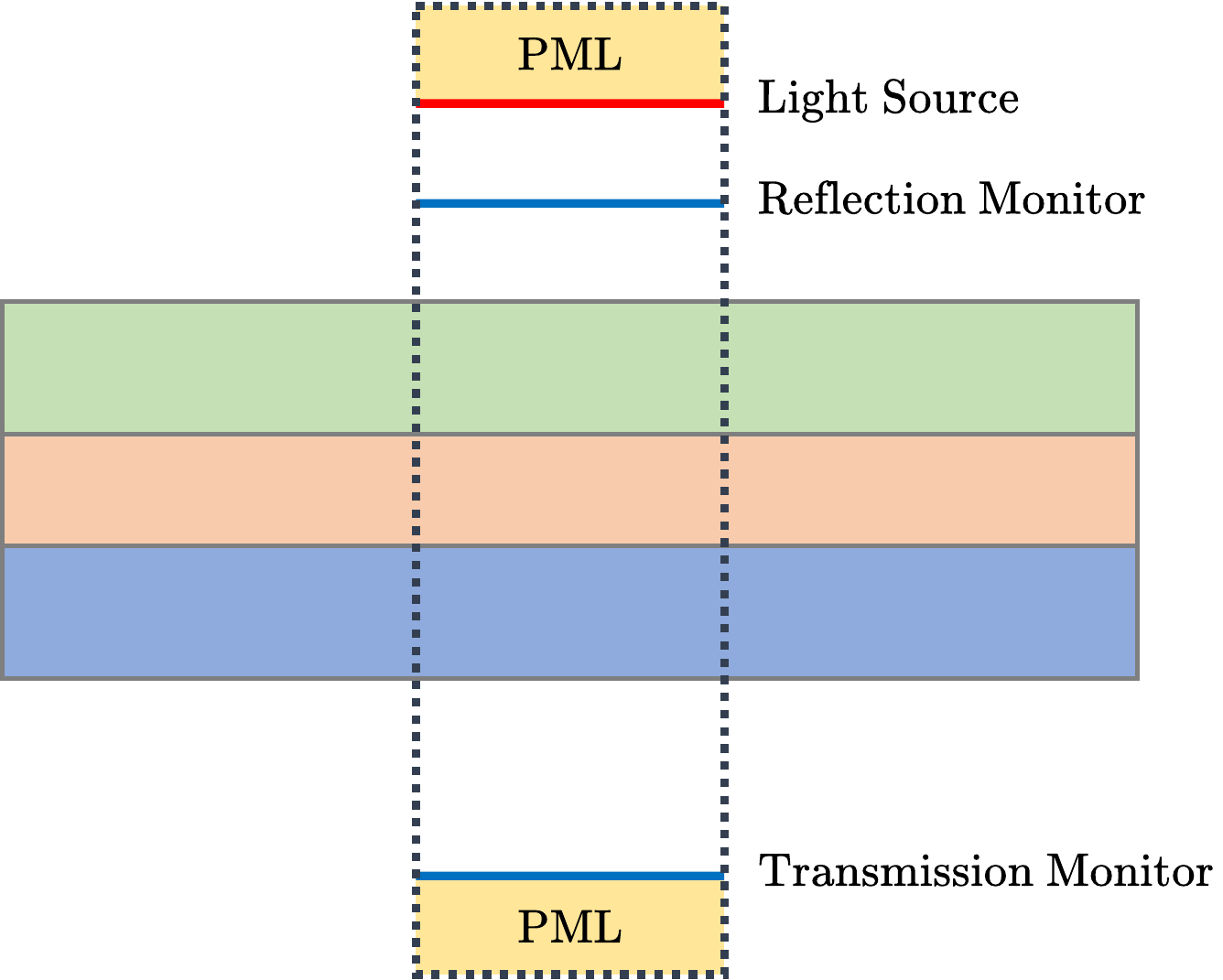}
    \caption{Schematic of nanophotonic structure simulations. PMLs, light source, and reflection and transmission monitors are located in a simulation cell indicated by the dotted rectangle.}
    \label{fig:schematic}
    \vspace{-15pt}
\end{wrapfigure}

We cover the scheme of electrodynamic simulations using the FDTD method -- a widely adopted technique in time-domain simulations for nanophotonic applications. This method discretizes both spatial and temporal dimensions, enabling the approximation of derivatives at discrete points. Our simulations focus on exploring the optical behaviors of structures in two and three dimensions through the FDTD method~\citep{TafloveA1980ieeetec}.

For a light source, we employ a Gaussian-pulse wave, monitoring the resulting electric fields $\bsE(t)$ and magnetic fields $\bsB(t)$.
These fields are subsequently Fourier-transformed to yield their frequency-dependent counterparts $\bsE(\omega)$ and $\bsB(\omega)$. 
A Yee grid, a computational mesh where the electric and magnetic field components are calculated, is used~\citep{YeeKS1966ieeeap}.
Our simulations are conducted using Meep, which is the open-source FDTD simulation software licensed under the GNU General Public License~\citep{OskooiAF2010cpc}.

\figref{fig:schematic} shows a schematic that illustrates the setup of our FDTD simulation,
where the simulation domain is confined within the simulation cell indicated by the dotted rectangle.
The size and type of this simulation cell are tailored to the specific nanophotonic structure investigated.
We utilize periodic boundary conditions for the sides of the simulation cell and perfectly matched layers (PMLs) for the top and bottom of the cell.
Periodic boundary conditions model semi-infinite arrays,
while the PML boundary conditions ensure that fields radiate to infinity instead of reflecting~\citep{BerengerJP1994jcp}.

Within the simulation cell, a light source is positioned near the top, while transmission and reflection monitors are situated near the bottom and below the source, respectively. 
The electromagnetic wave is chosen to be normally incident to the structure of interest.  
The monitors enable the calculation of flux from $\textrm{Re} \int \bsE^{*} (\omega) \times \bsB (\omega) / 2$ as a function of frequency, facilitating the derivation of reflectance $R(\omega)$, absorbance $A(\omega)$, and transmittance $T(\omega)$.
To accurately measure reflectance,
the total incident flux from an empty simulation cell is subtracted from the flux measured in the presence of the nanophotonic structure.
The positioning of the light source and monitors is chosen to ensure that they are appropriately spaced from the top and bottom boundaries of the simulation cell.

For our simulations, we consider three types of spectral ranges: a single wavelength 550~nm at the approximate center of the visible spectrum,
the visible spectrum from 380~nm to 750~nm with standard illuminant D65~\citep{CIE2022SID65},
and the AM1.5 global solar spectrum from 280~nm to 2500~nm~\citep{ASTMI2020SolarAM15}.

\subsection{Structures of Interest}

\begin{figure}[t]
    \centering
    \begin{subfigure}[b]{0.495\textwidth}
        \centering
        \includegraphics[width=0.49\textwidth]{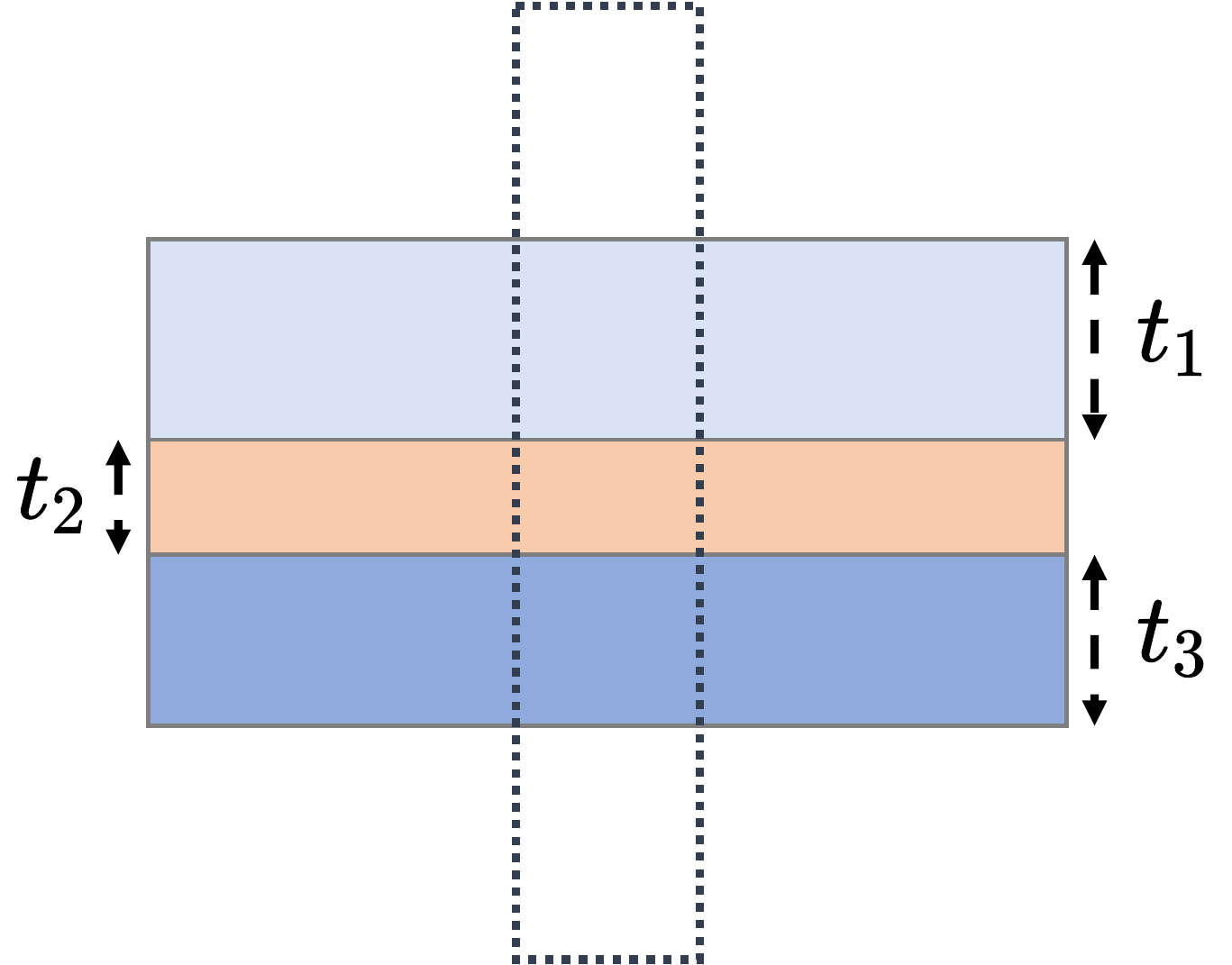}
        \includegraphics[width=0.49\textwidth]{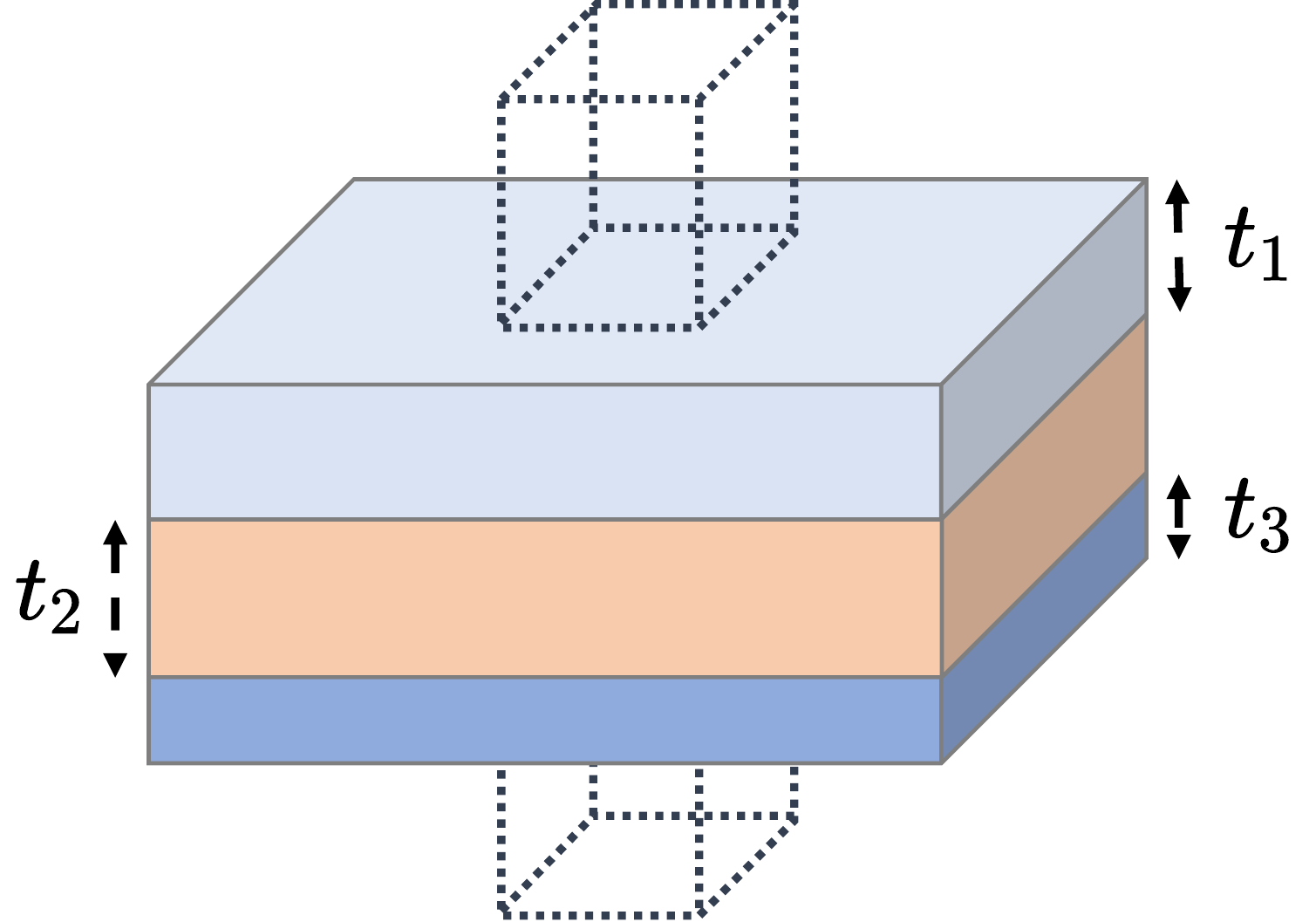}
        \caption{Three-layer film}
        \label{fig:threelayers}
    \end{subfigure}
    \begin{subfigure}[b]{0.495\textwidth}
        \centering
        \includegraphics[width=0.49\textwidth]{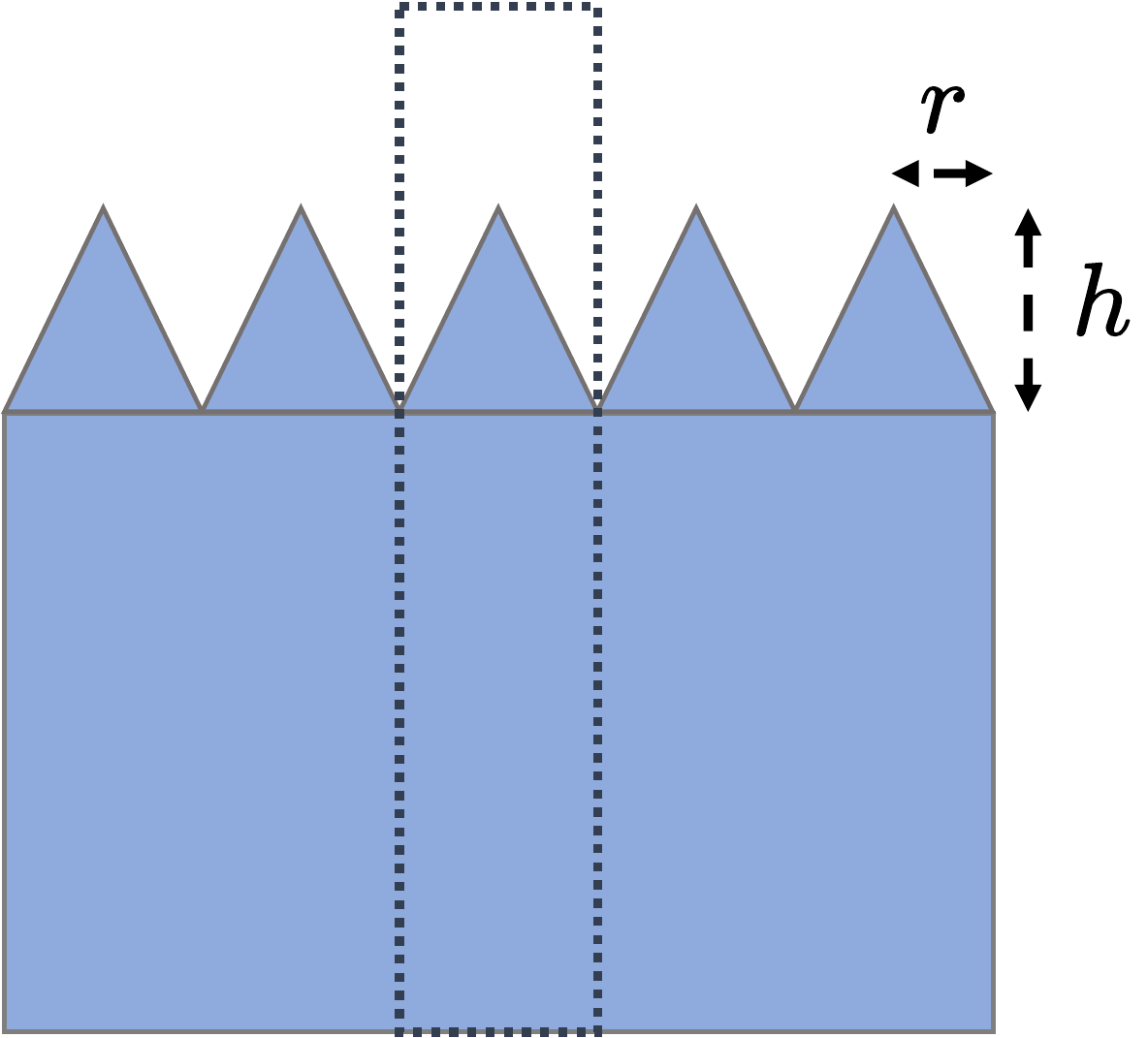}
        \includegraphics[width=0.49\textwidth]{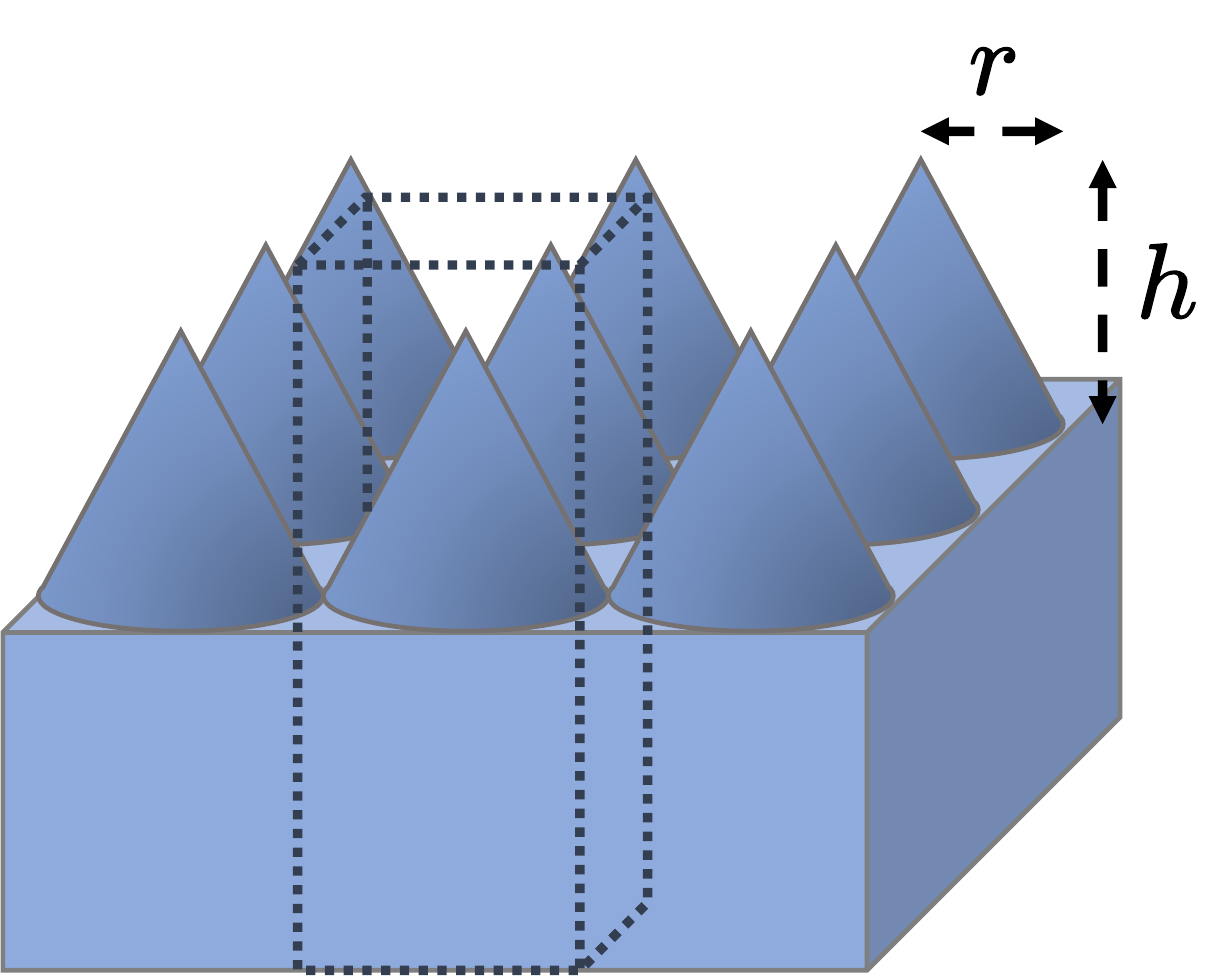}
        \caption{Anti-reflective nanocones}
        \label{fig:nanocones}
    \end{subfigure}
    \begin{subfigure}[b]{0.495\textwidth}
        \centering
        \includegraphics[width=0.49\textwidth]{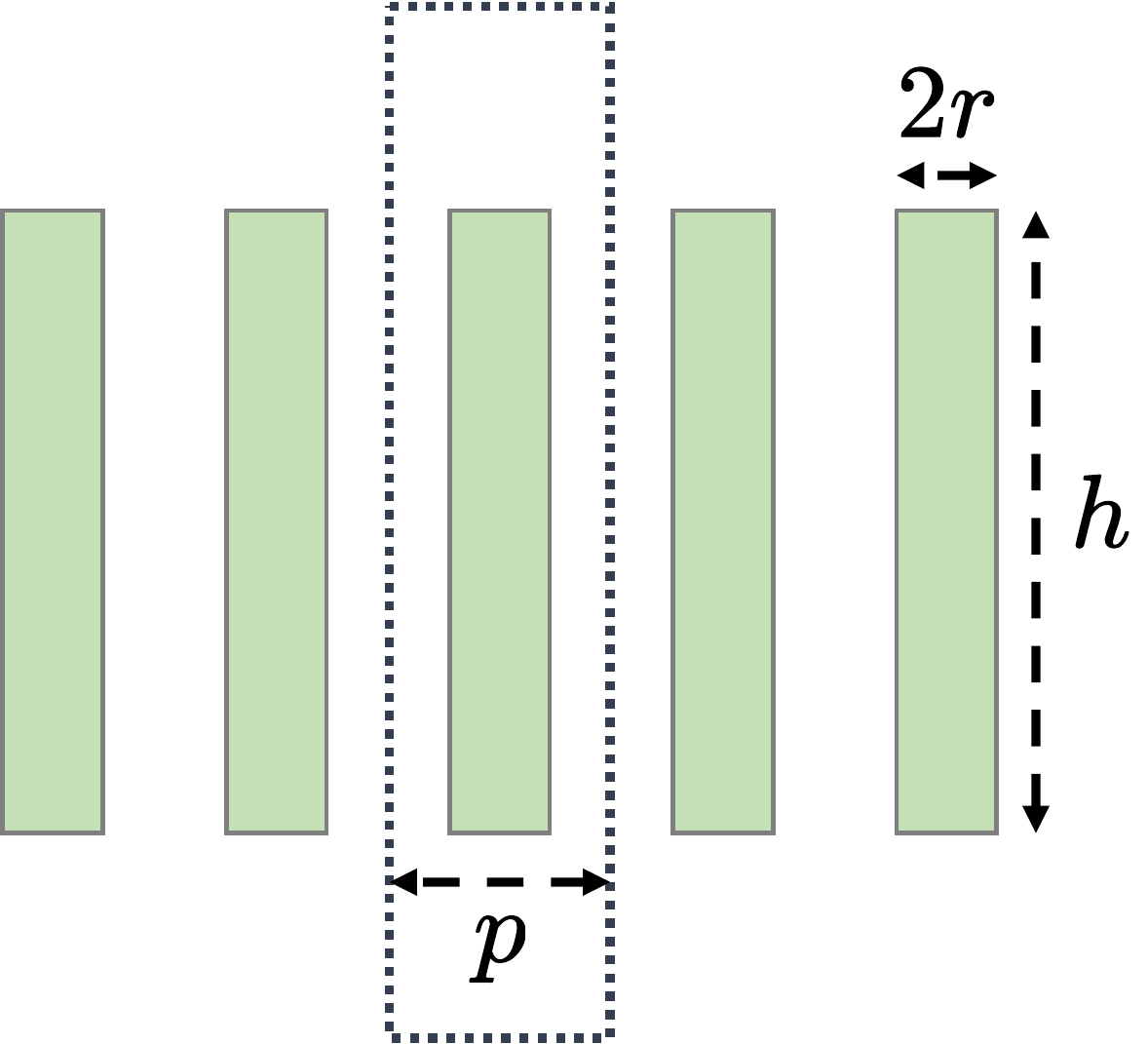}
        \includegraphics[width=0.49\textwidth]{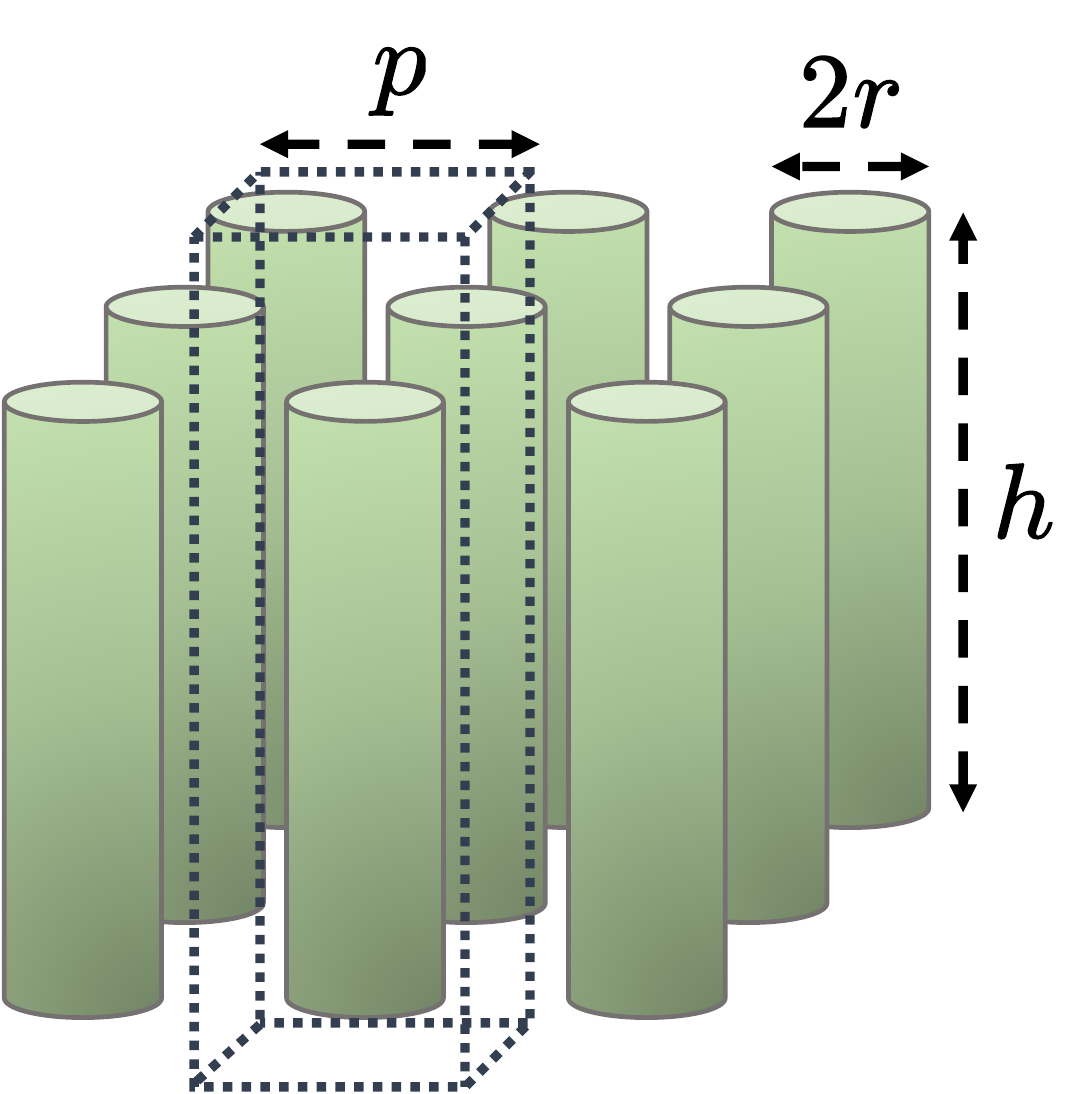}
        \caption{Vertical nanowires}
        \label{fig:nanowires}
    \end{subfigure}
    \begin{subfigure}[b]{0.495\textwidth}
        \centering
        \includegraphics[width=0.49\textwidth]{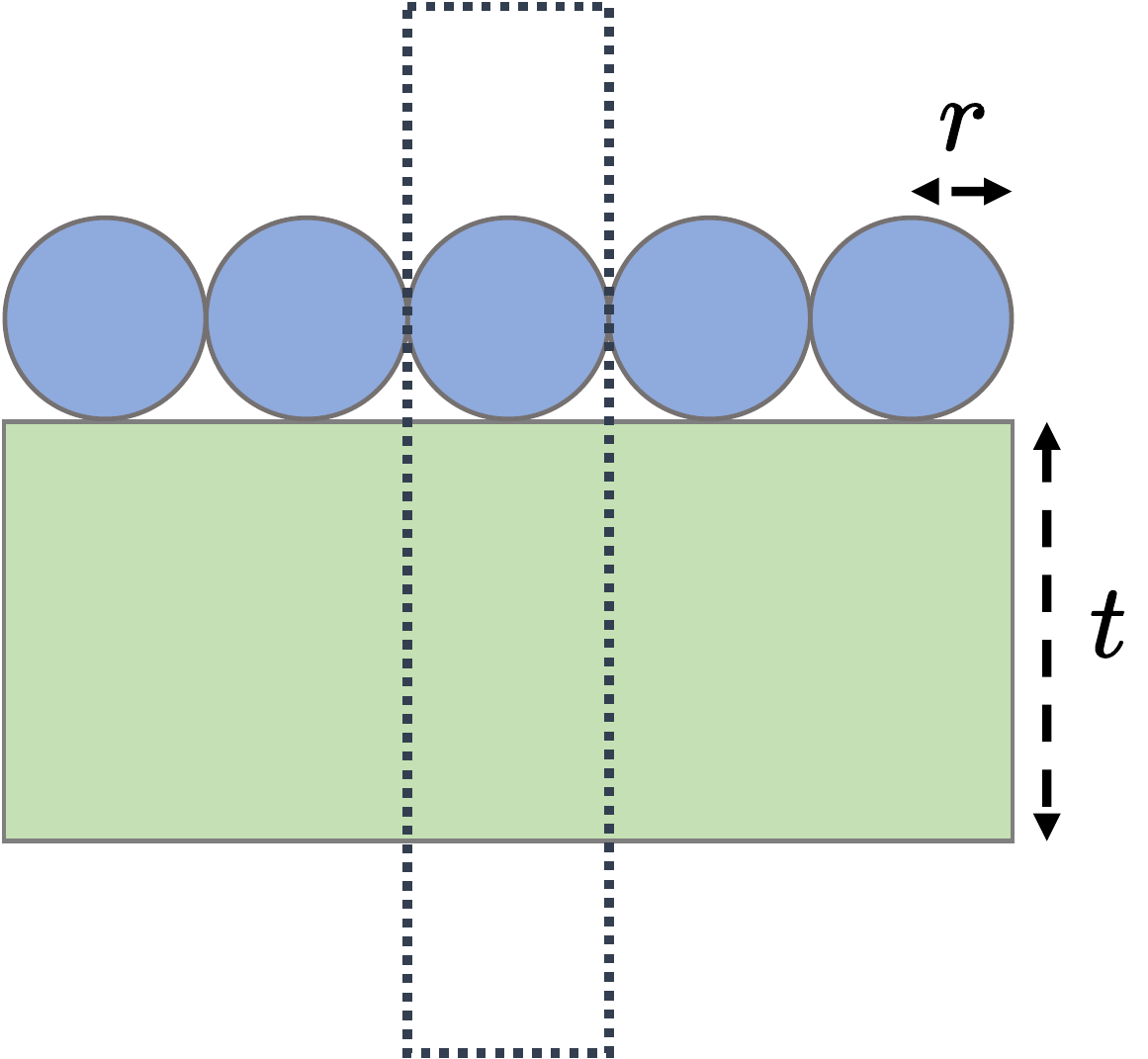}
        \includegraphics[width=0.49\textwidth]{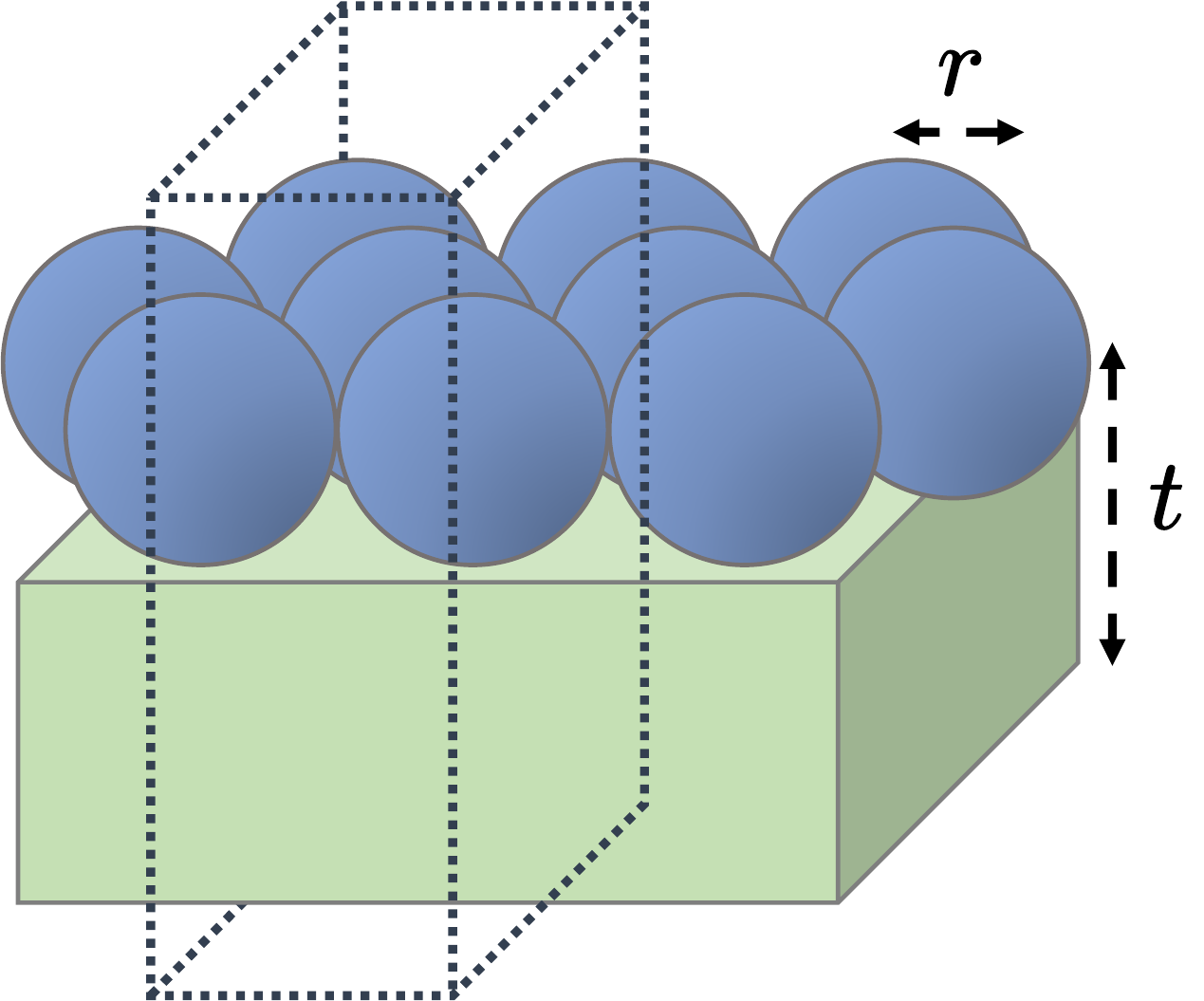}
        \caption{Close-packed nanospheres}
        \label{fig:nanospheres}
    \end{subfigure}
    \begin{subfigure}[b]{0.495\textwidth}
        \centering
        \includegraphics[width=0.49\textwidth]{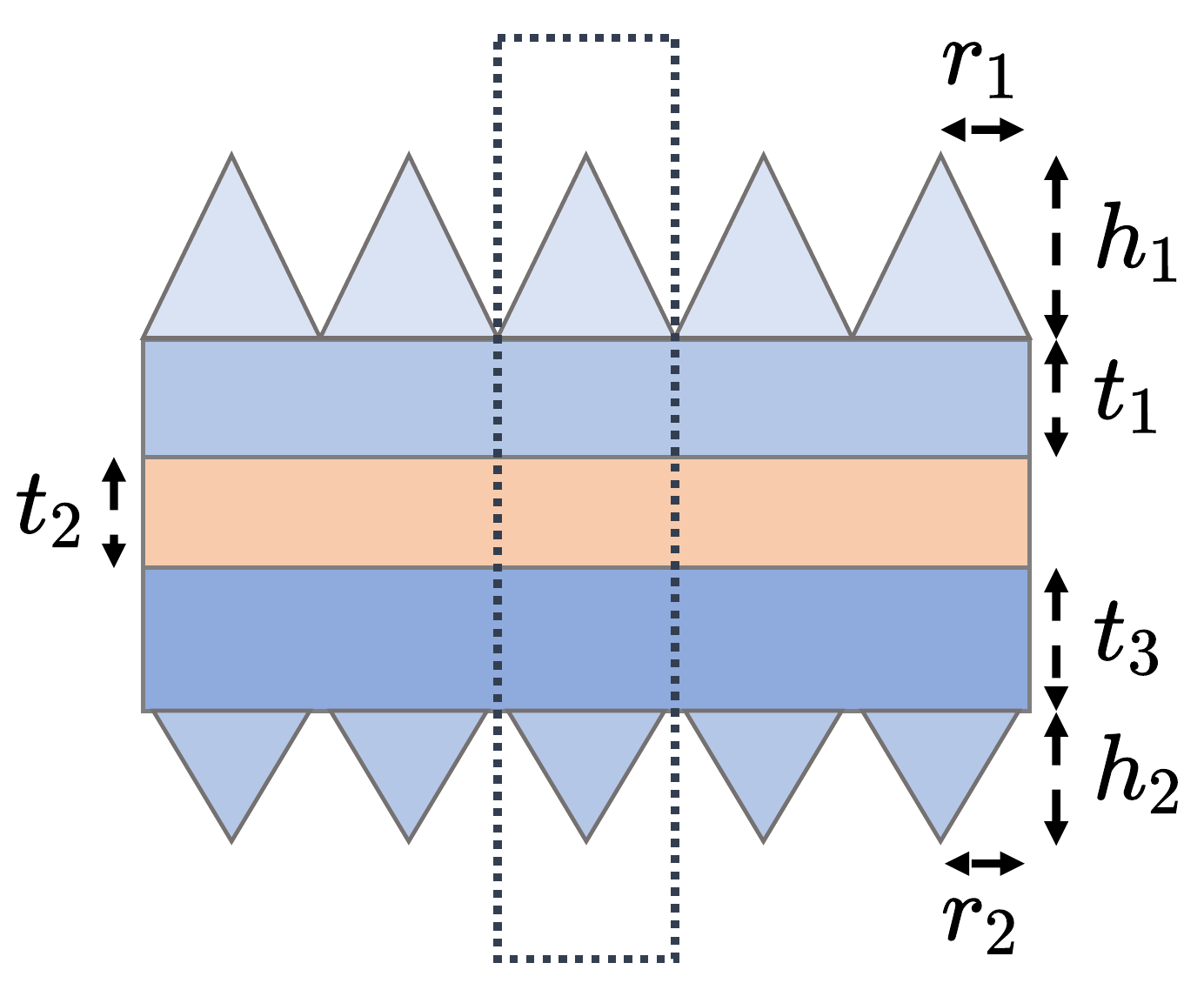}
        \includegraphics[width=0.49\textwidth]{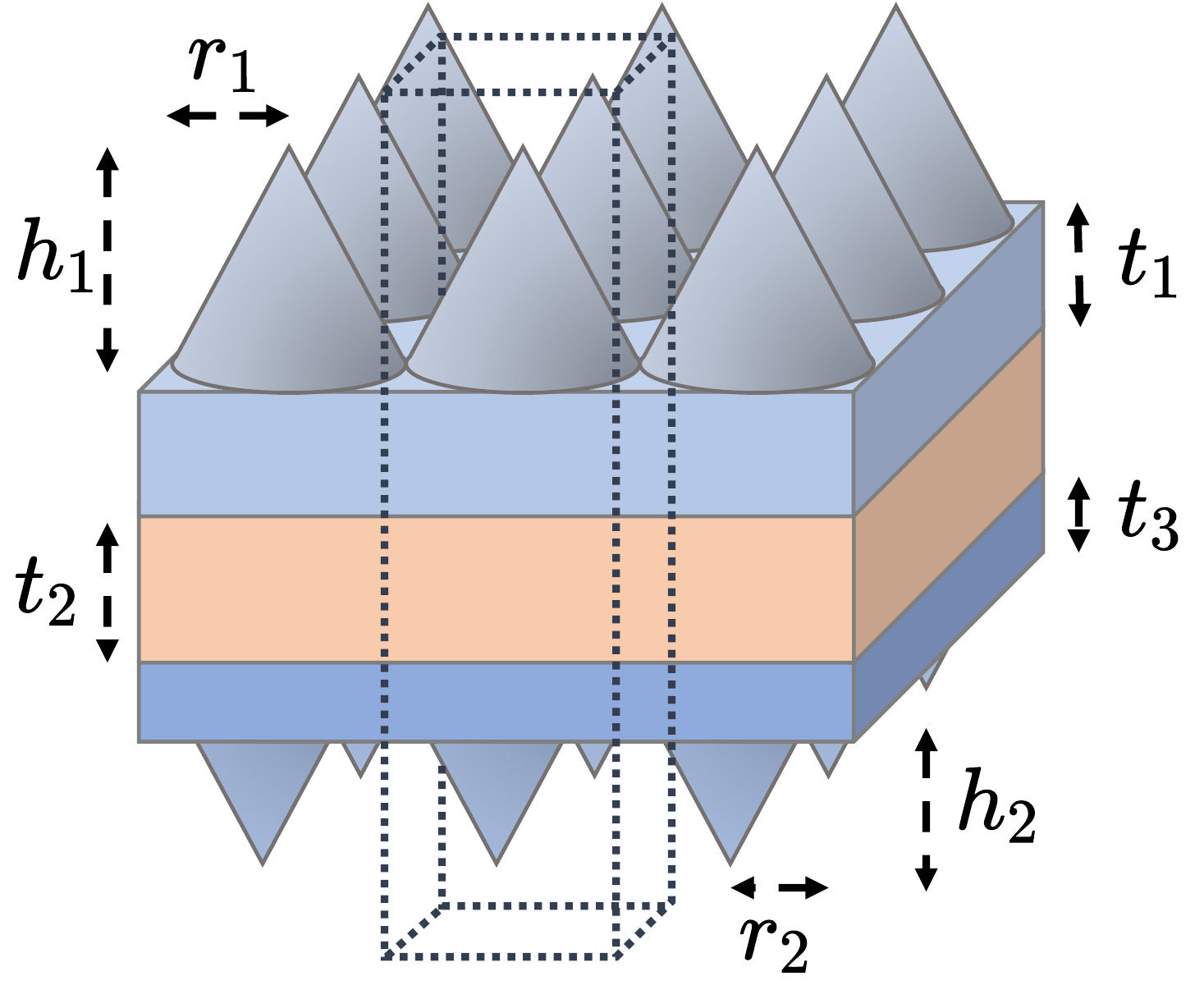}
        \caption{Three-layer film with double-sided nanocones}
        \label{fig:doublenanocones}
    \end{subfigure}
    \begin{subfigure}[b]{0.495\textwidth}
        \centering
        \includegraphics[width=0.49\textwidth]{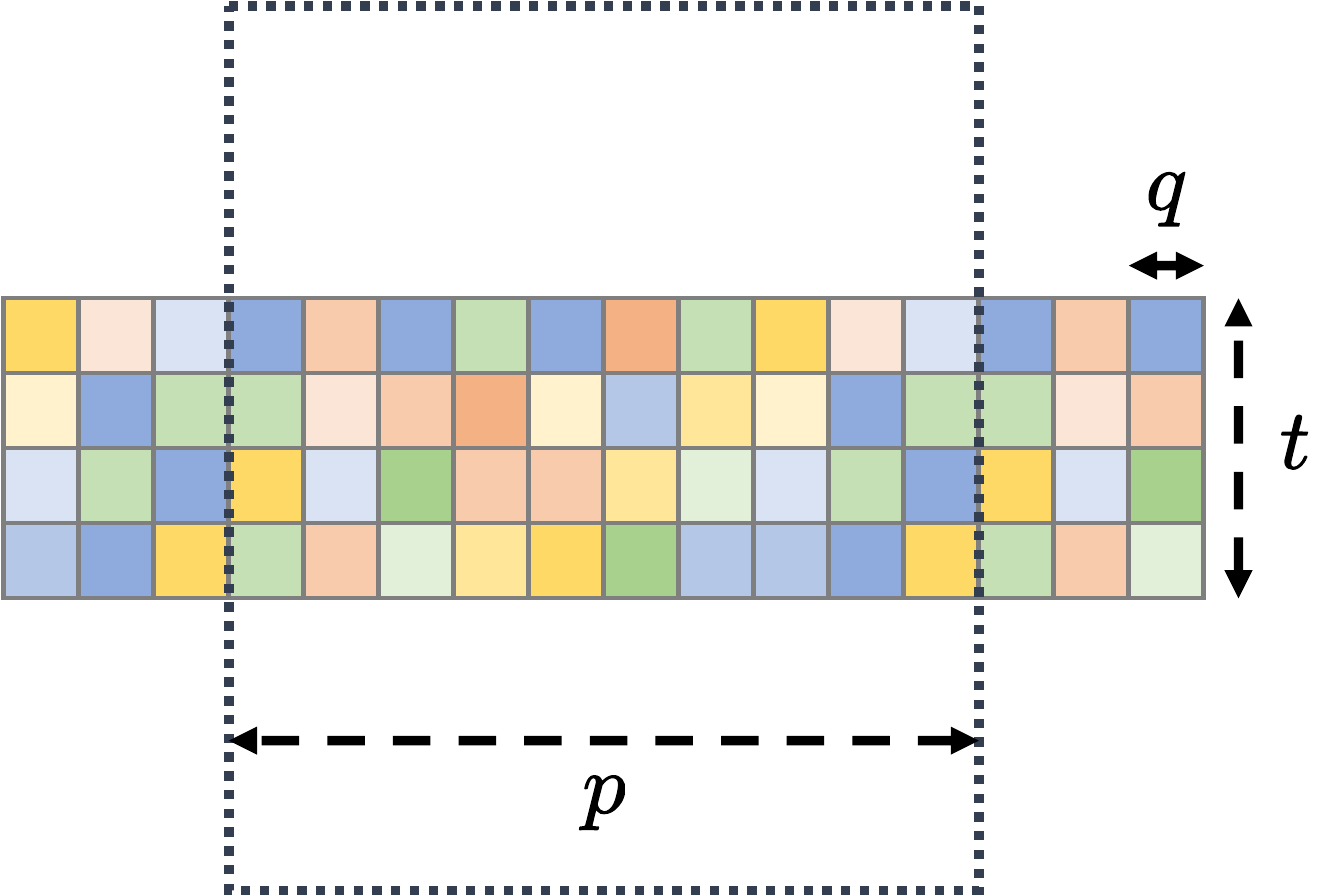}
        \includegraphics[width=0.49\textwidth]{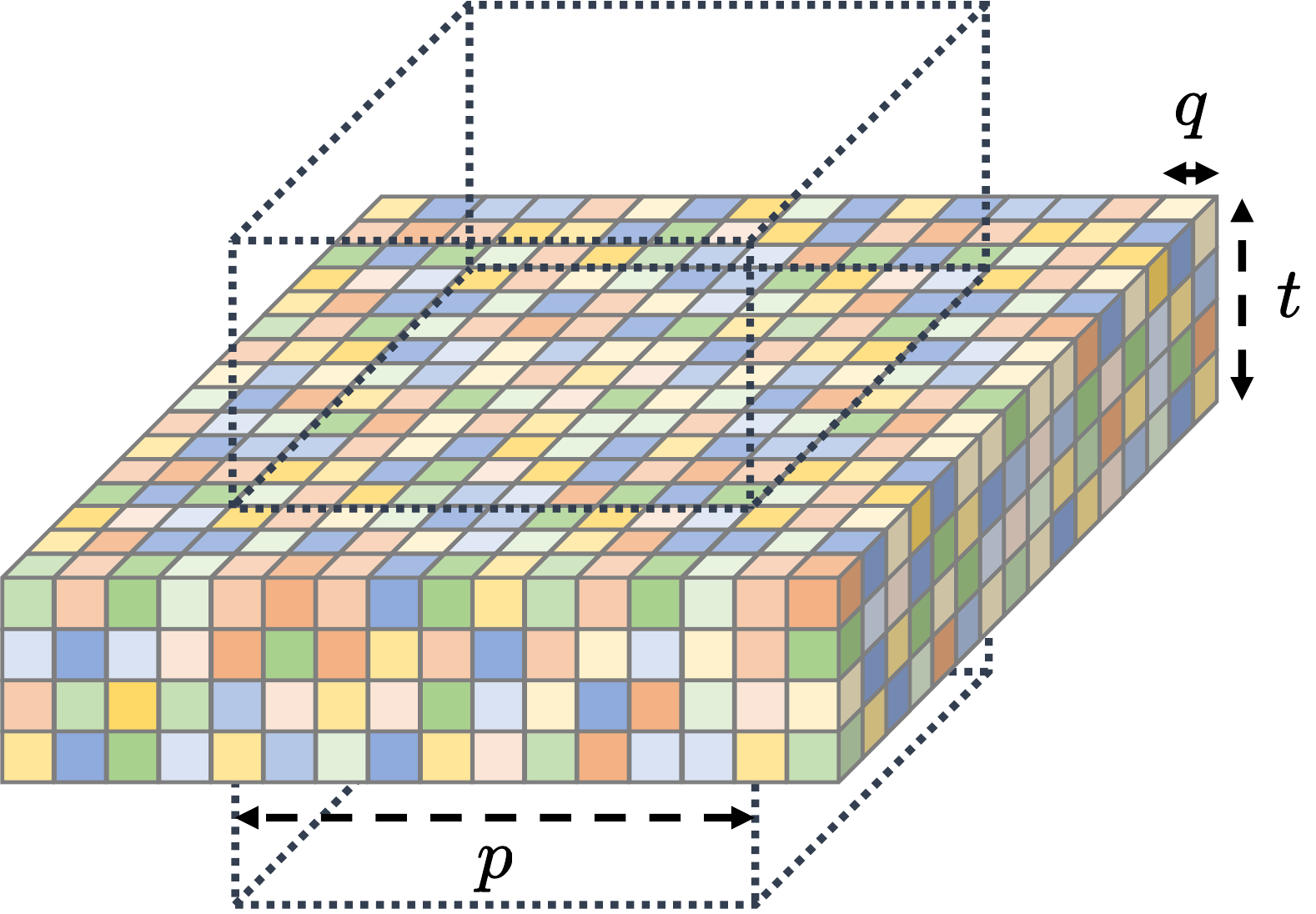}
        \caption{Combinatorial system with material blocks}
        \label{fig:combinatorial}
    \end{subfigure}
    \caption{Two- and three-dimensional nanophotonic structures. Different colors indicate different materials in each figure and a dotted region represents a simulation cell.}
    \label{fig:structures_2d_3d}
\end{figure}
\begin{table}[t]
    \caption{Parameters, materials, and the search spaces of the parameters for nanophotonic structures. We assume that only one of \ce{AZO}, \ce{cSi}, \ce{ITO}, \ce{TiO2}, and \ce{ZnO} should be used for one structural configuration in the three-layer film with or without double-sided nanocones.}
    \label{tab:search_spaces}
    \begin{center}
    \small
    \setlength{\tabcolsep}{3pt}
    \begin{tabular}{lcccc}
        \toprule
        \multirow{2}{*}{\textbf{Structure}} & \multirow{2}{*}{\textbf{Parameter}} & \multirow{2}{*}{\textbf{Materials}} & \textbf{Lower} & \textbf{Upper} \\
        &&& \textbf{Bound (nm)} & \textbf{Bound (nm)} \\
        \midrule
        \multirow{3}{*}{Three-layer film} & $t_1$ & \ce{AZO}, \ce{cSi}, \ce{ITO}, \ce{TiO2}, \ce{ZnO} & 10 & 100 \\
        & $t_2$ & \ce{Ag}, \ce{Au}, \ce{Cu}, \ce{Ni} & 3 & 20 \\
        & $t_3$ & \ce{AZO}, \ce{cSi}, \ce{ITO}, \ce{TiO2}, \ce{ZnO} & 10 & 100 \\
        \midrule
        \multirow{2}{*}{Anti-reflective nanocones} & $r$ & \multirow{2}{*}{Fused silica} & 5 & 150 \\
        & $h$ && 1 & 300 \\
        \midrule
        \multirow{3}{*}{Vertical nanowires} & $g$ & \multirow{3}{*}{\ce{cSi}, \ce{CH3NH3PbI3}, \ce{GaAs}} & 1 & 200 \\
        & $r$ && 5 & 200 \\
        & $h$ && 200 & 200 \\
        \midrule
        \multirow{2}{*}{Close-packed nanospheres} & $t$ & \ce{cSi}, \ce{CH3NH3PbI3}, \ce{GaAs} & 100 & 400 \\
        & $r$ & \ce{TiO2} & 10 & 200 \\
        \midrule
        & $t_1$ & \ce{AZO}, \ce{cSi}, \ce{ITO}, \ce{TiO2}, \ce{ZnO} & 10 & 50 \\
        & $t_2$ & \ce{Ag}, \ce{Au}, \ce{Cu}, \ce{Ni} & 3 & 20 \\
        Three-layer film & $t_3$ & \ce{AZO}, \ce{cSi}, \ce{ITO}, \ce{TiO2}, \ce{ZnO} & 10 & 50 \\
        with double-sided & $r_1$ & \multirow{2}{*}{\ce{AZO}, \ce{cSi}, \ce{ITO}, \ce{TiO2}, \ce{ZnO}} & 20 & 50 \\
        nanocones & $h_1$ && 50 & 100 \\
        & $r_2$ & \multirow{2}{*}{\ce{AZO}, \ce{cSi}, \ce{ITO}, \ce{TiO2}, \ce{ZnO}} & 20 & 50 \\
        & $h_2$ && 50 & 100 \\
        \midrule
        Combinatorial system & \multirow{2}{*}{--} & \ce{Ag}, Air, \ce{Au}, \ce{AZO}, \ce{cSi}, \ce{CH3NH3PbI3}, & \multirow{2}{*}{--} & \multirow{2}{*}{--} \\
        with material blocks && \ce{Cu}, \ce{GaAs}, \ce{ITO}, \ce{Ni}, \ce{TiO2}, \ce{ZnO} && \\
        \bottomrule
    \end{tabular}
    \end{center}
\end{table}

In this section, we describe the specifics of the nanophotonic structures studied in this paper.
The details of simulation cell sizes are presented in~\secref{sec:details_cell_sizes}.
\tabref{tab:search_spaces} provides the summary of the design space including parameters, materials, and parameters' ranges for each of these structures.

\paragraph{Three-Layer Film for Electromagnetic Interference Shielding.}

This film structure is aimed at mitigating electromagnetic interference, incorporating three distinct layers; see \figref{fig:threelayers}.
The central layer is metallic,
potentially consisting of silver (\ce{Ag}), gold (\ce{Au}), copper (\ce{Cu}), or nickel (\ce{Ni}).
The adjacent layers can be selected from materials such as titanium dioxide (\ce{TiO2}), crystalline silicon (\ce{cSi}), zinc oxide (\ce{ZnO}), indium tin oxide (\ce{ITO}), or aluminum-doped zinc oxide (\ce{AZO}).
We simplify our material selection by assuming the materials for the outer layers are identical.
The three layers are characterized by their thicknesses: $t_1$, $t_2$, and $t_3$.
We use the electromagnetic wave of wavelength $\lambda =$ 550 nm,
which approximately corresponds to green light.

The objectives of this system include maximizing transmittance over $t_1$, $t_2$, and $t_3$ and tackling a multi-objective optimization problem for maximizing both electromagnetic interference shielding effectiveness and transmittance.
The formula for shielding effectiveness is $S = 20 \log_{10} (1 + \eta_0 t_2/2\rho)$,
where $\eta_0 =$ 376.73 $\Omega$ is the free space impedance and $\rho$ is the bulk metal resistivity.

\paragraph{Graded Index of Refraction Structures for Anti-Reflection.}

Here, we include a semi-infinite array of glass nanocones attached to a glass substrate,
as illustrated in~\figref{fig:nanocones}.
The nanocones serve to gradually transition the index of refraction from air to glass,
with fused silica as the glass of choice.
This structure is defined by two parameters: the nanocones' height $h$ and radius $r$. Assuming a close-packed square array, we determine the pitch of the nanocones is $p = 2r$.
Our goal is to minimize solar reflection;
see~\secref{sec:solar_reflection} for the details of solar reflection.

\paragraph{Vertical Nanowires for Solar Cells.}

We explore arrays of vertical nanowires composed of crystalline silicon (\ce{cSi}),
gallium arsenide (\ce{GaAs}),
or a perovskite structure of methylammonium lead iodide (\ce{CH3NH3PbI3}),
as shown in~\figref{fig:nanowires}.
The structures are defined by the nanowire radius $r$, height $h$, and pitch $p$, with a fixed $h =$ 200 nm.
The relationship $2r \leq p$ has to be satisfied, which makes the diameter less than or equal to the pitch.
We redefine this constraint using 
$g = p - 2r$, where $g \geq 0$.
The nanowires are arranged in a square array, with optimization focused on maximizing solar absorption or ultimate efficiency;
refer to~\secref{sec:sa_ue} for further details.

\paragraph{Close-Packed Nanospheres for Solar Cells.}

This system features titanium dioxide (\ce{TiO2}) nanospheres in a hexagonal array on top of a semiconductor thin film with possible materials including crystalline silicon (\ce{cSi}), gallium arsenide (\ce{GaAs}),
or a perovskite structure of methylammonium lead iodide (\ce{CH3NH3PbI3}).
The system, which is visualized in~\figref{fig:nanospheres},
is defined by the semiconductor layer thickness $t$ and nanosphere radius $r$.
Similar to the vertical nanowires,
the optimization aims at maximizing solar absorption or ultimate efficiency;
refer to~\secref{sec:sa_ue} for details.

\paragraph{Three-Layer Film with Nanocones for Electromagnetic Interference Shielding.}

Building upon the three-layer film concept,
this structure incorporates double-sided nanocones for enhanced electromagnetic interference shielding;
see~\figref{fig:doublenanocones}.
It comprises three layers and nanocones on both sides,
where material options for the three-layer film are identical to the previously discussed three-layer film and nanocones are made of one of titanium dioxide (\ce{TiO2}),
crystalline silicon (\ce{cSi}), zinc oxide (\ce{ZnO}),
indium tin oxide (\ce{ITO}),
and aluminum-doped zinc oxide (\ce{AZO}).
Seven parameters define this structure:
three layer thicknesses $t_1$, $t_2$, $t_3$, and two sets of cone heights $h_1$, $h_2$ and radii $r_1$, $r_2$.
Objective for the optimization of this structure
can be either maximization of visible transmission or
simultaneous maximization of visible transmission and shielding effectiveness,
where light of the visible spectrum is injected;
refer to the description of the aforementioned three-layer film.

\paragraph{Combinatorial System with Material Blocks.}

We introduce a novel and challenging problem:
a combinatorial system composed of material blocks;
see \figref{fig:combinatorial}.
The size of each block within the structure is $(q, q)$ for two-dimensional systems or $(q, q, q)$ for three-dimensional ones.
Construction of this structure is accomplished by strategically selecting specific materials for all the blocks placed in a repeating unit.
This unit consists of $p / q$ blocks along the $x$ and $y$ directions,
and $t / q$ blocks along the $z$ direction,
where $p$ denotes the repeating unit's pitch size
and $t$ represents the maximum thickness of the structure.
Consequently, the repeating unit is comprised of either $p / q \times t / q$ blocks for two-dimensional structures or $p / q \times p / q \times t / q$ blocks for three-dimensional ones.

In this paper, we choose specific values:
$p =$ 200 nm, $t =$ 40 nm, and $q =$ 10 nm.
The range of materials considered for the material blocks includes silver (\ce{Ag}), air, gold (\ce{Au}),
aluminum-doped zinc oxide (\ce{AZO}),
crystalline silicon (\ce{cSi}),
methylammonium lead iodide perovskite (\ce{CH3NH3PbI3}),
copper (\ce{Cu}), gallium arsenide (\ce{GaAs}),
indium tin oxide (\ce{ITO}), nickel (\ce{Ni}),
titanium dioxide (\ce{TiO2}), and zinc oxide (\ce{ZnO}).
The AM1.5 solar spectrum is used for simulating this structure.

\subsection{Simulation Fidelity and Multiple Objectives}

As described above,
the accuracy of simulation results are heavily influenced by the fidelity of the simulations.
Simulations with a low fidelity level are faster but less accurate, while simulations with a high fidelity level are more accurate but slower.
There is an inherent tradeoff between speed and accuracy.
Additionally, our frameworks can generate objective functions such as shielding effectiveness and transparency for three-layer films, both with and without nanocones.
The introduction of multiple objectives adds a layer of complexity to the optimization problem.
In summary, utilizing fidelity levels and multiple objectives enables us to expand our research,
potentially exploring multi-fidelity optimization~\citep{ForresterAIJ2007prsa,KandasamyK2017icml,BelakariaS2020aaai} and multi-objective optimization~\citep{KnowlesJ2006ieeetec,HernandezLobatoD2016neurips,BelakariaS2019neurips}.

\begin{figure}[t]
    \centering
    \begin{subfigure}[b]{0.32\textwidth}
        \centering
        \includegraphics[width=\textwidth]{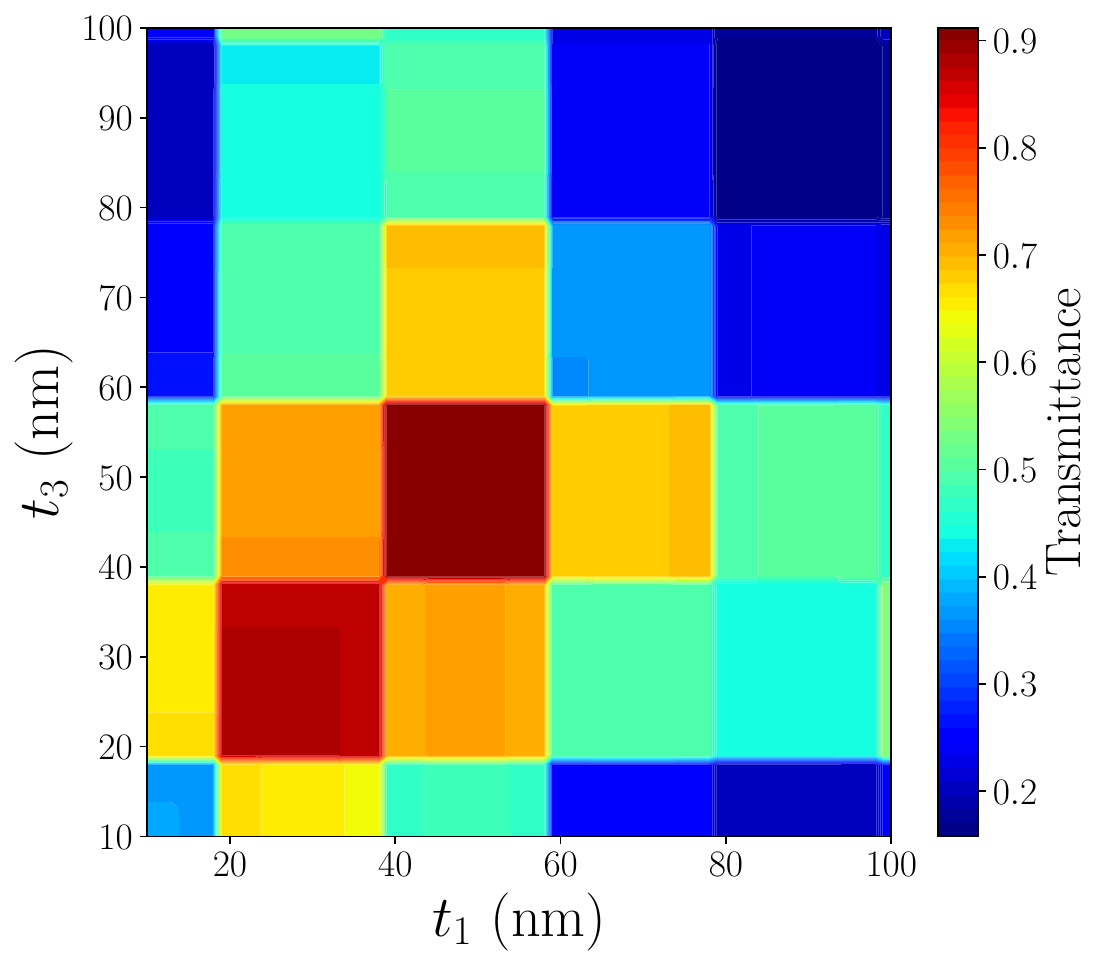}
        \caption{Low fidelity}
    \end{subfigure}
    \begin{subfigure}[b]{0.32\textwidth}
        \centering
        \includegraphics[width=\textwidth]{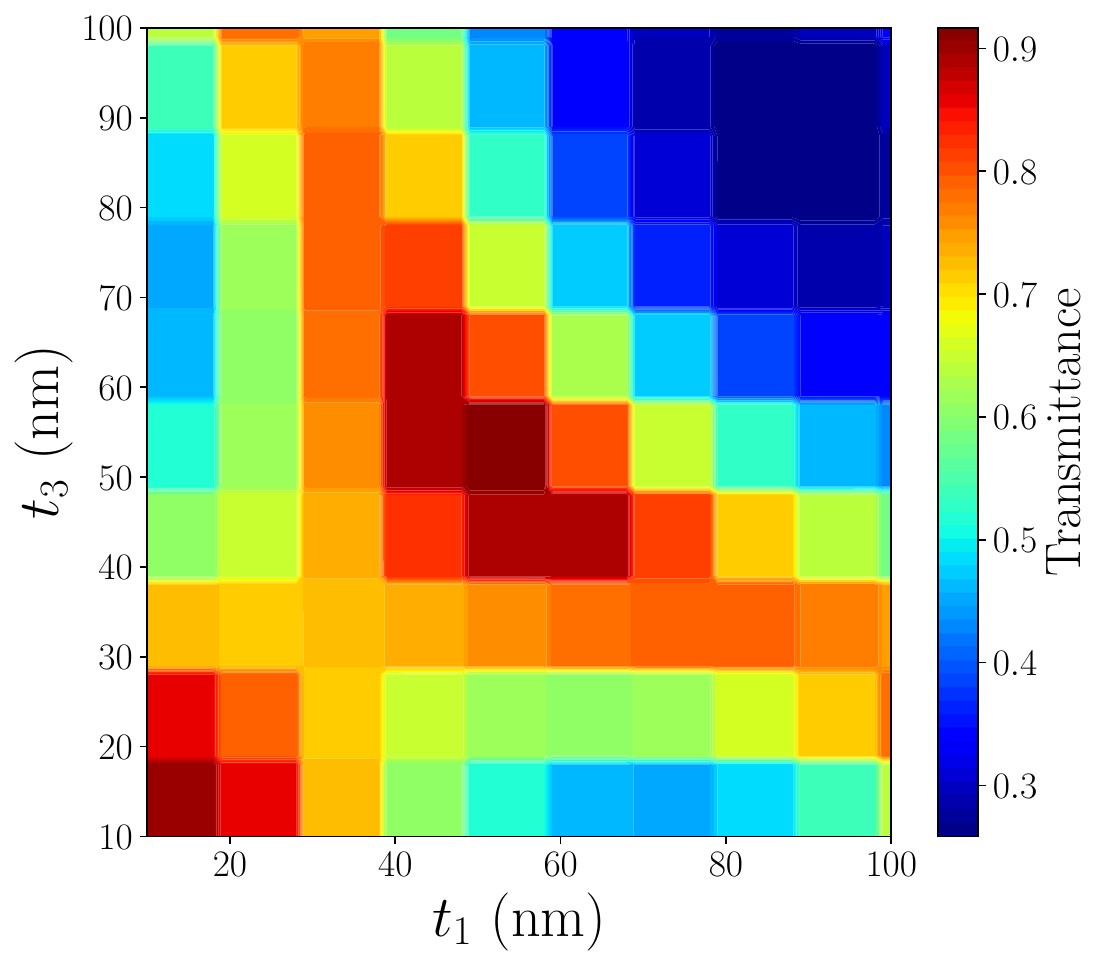}
        \caption{Medium fidelity}
    \end{subfigure}
    \begin{subfigure}[b]{0.32\textwidth}
        \centering
        \includegraphics[width=\textwidth]{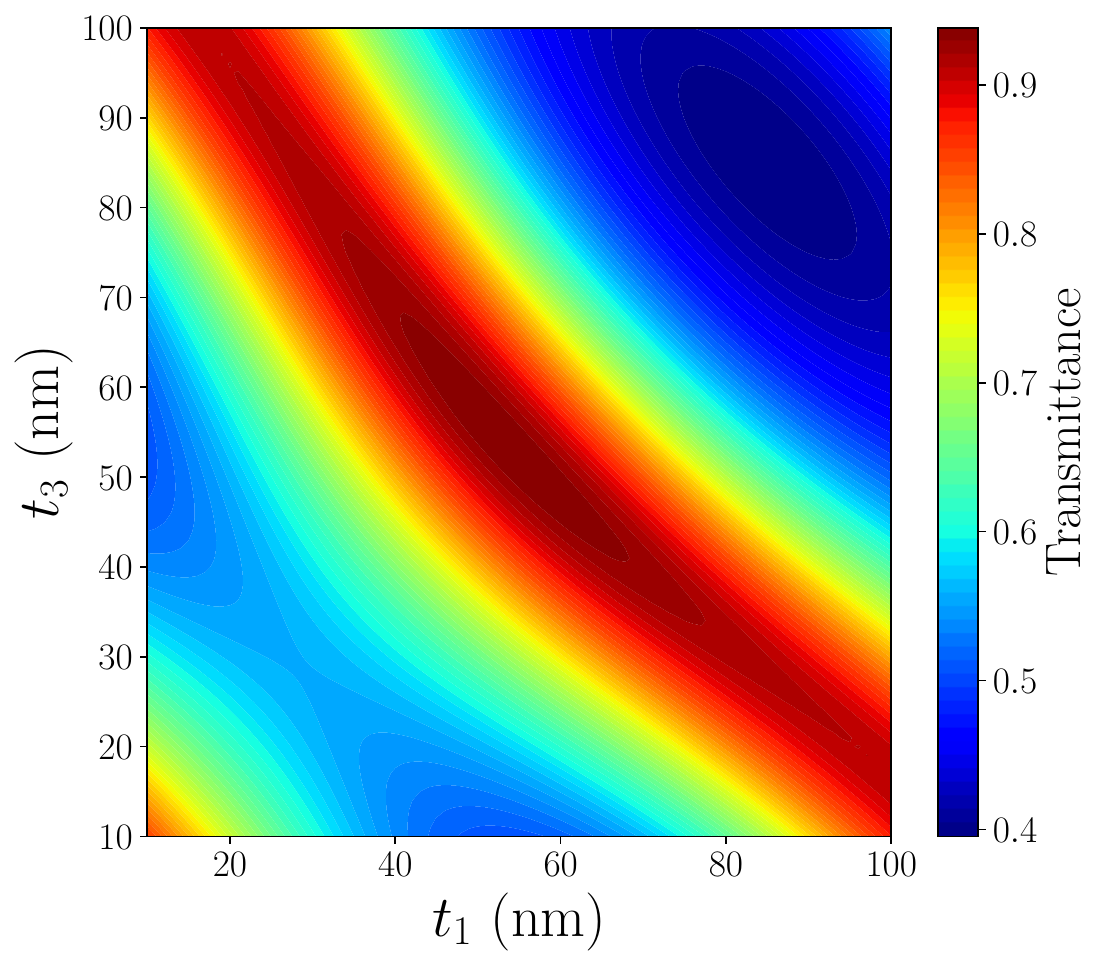}
        \caption{High fidelity}
    \end{subfigure}
    \caption{Visualization of the transmittance of the three-layer film made of \ce{TiO2}/\ce{Ag}/\ce{TiO2} for three different fidelity levels, where the second layer's thickness $t_2$ is 3 nm.}
    \label{fig:selected_threelayers}
\end{figure}
\begin{figure}[t]
    \centering
    \begin{subfigure}[b]{0.32\textwidth}
        \centering
        \includegraphics[width=\textwidth]{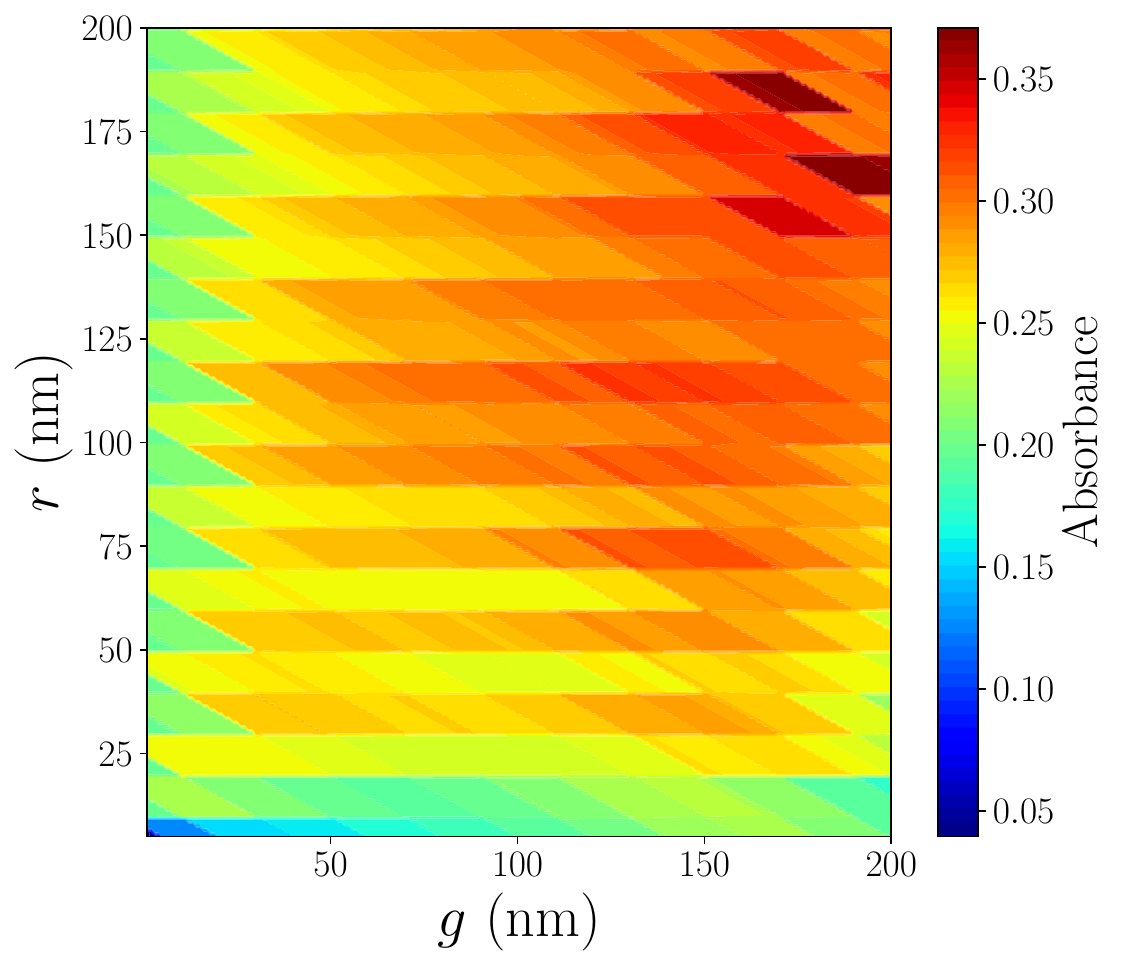}
        \caption{Low fidelity}
    \end{subfigure}
    \begin{subfigure}[b]{0.32\textwidth}
        \centering
        \includegraphics[width=\textwidth]{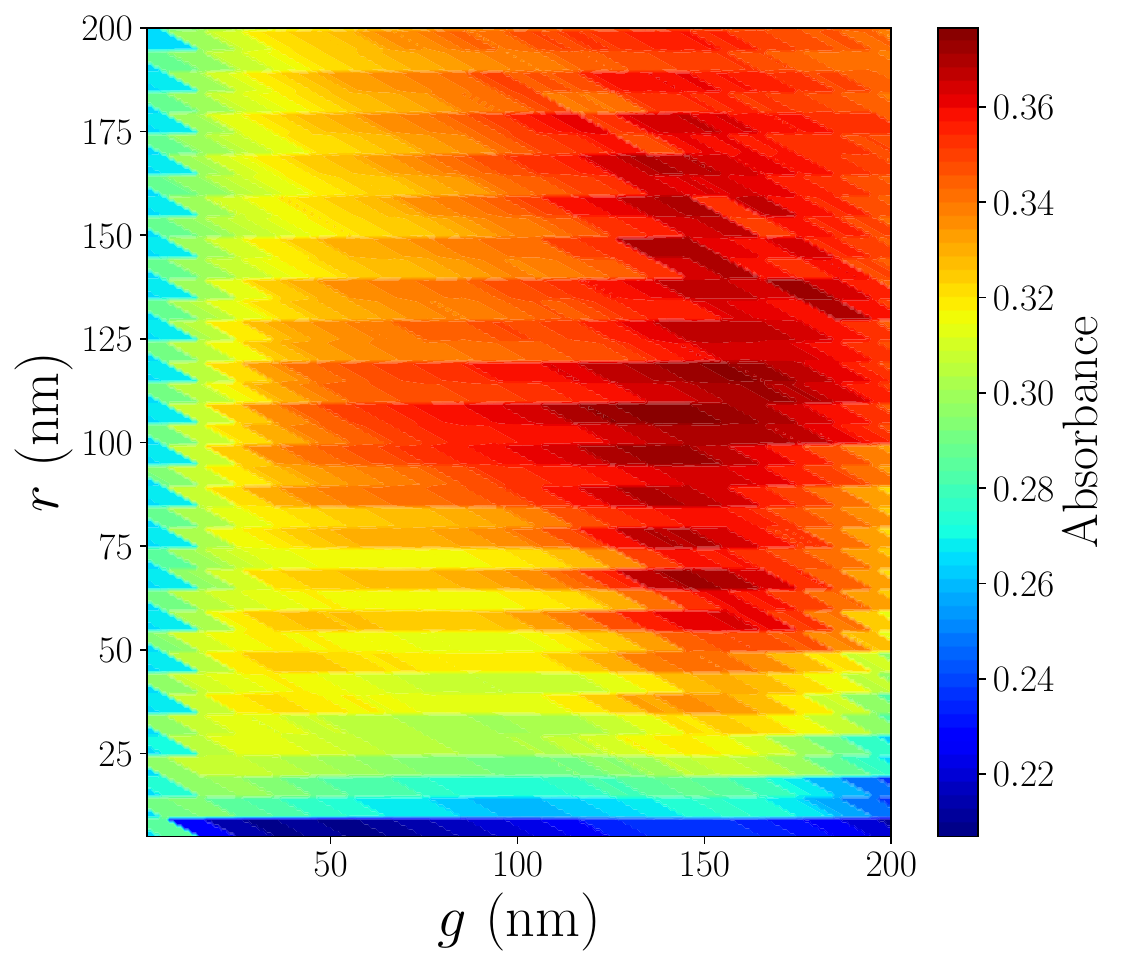}
        \caption{Medium fidelity}
    \end{subfigure}
    \begin{subfigure}[b]{0.32\textwidth}
        \centering
        \includegraphics[width=\textwidth]{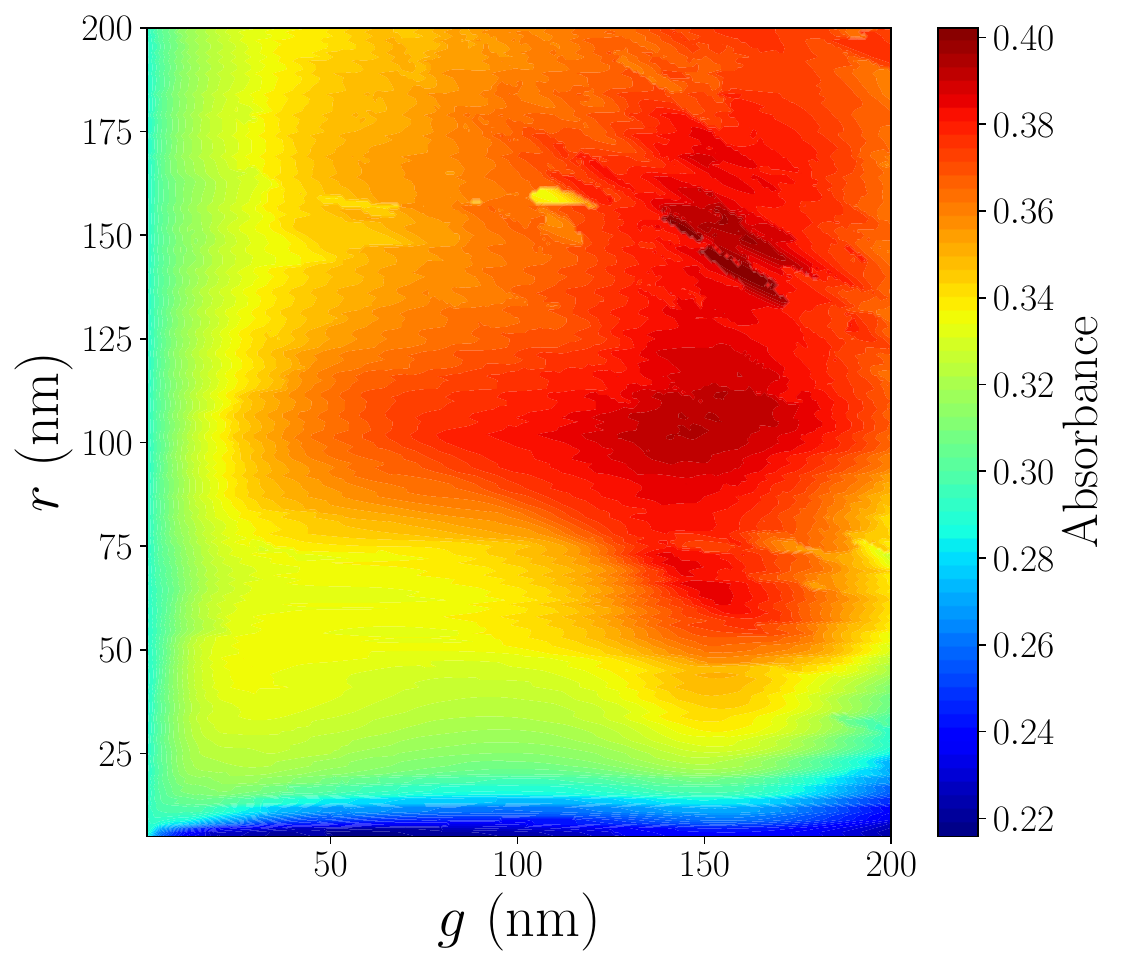}
        \caption{High fidelity}
    \end{subfigure}
    \caption{Visualization of the absorbance of the vertical nanowires made of \ce{cSi} for three different fidelity levels, where the nanowire height $h$ is 200 nm.}
    \label{fig:selected_nanowires}
\end{figure}

\subsection{Visualization}
\label{sec:visualization}

\begin{wrapfigure}{r}{0.48\textwidth}
    \centering
    \vspace{-12pt}
    \includegraphics[width=0.23\textwidth,trim={1.52in 0.48in 1.40in 0.48in},clip]{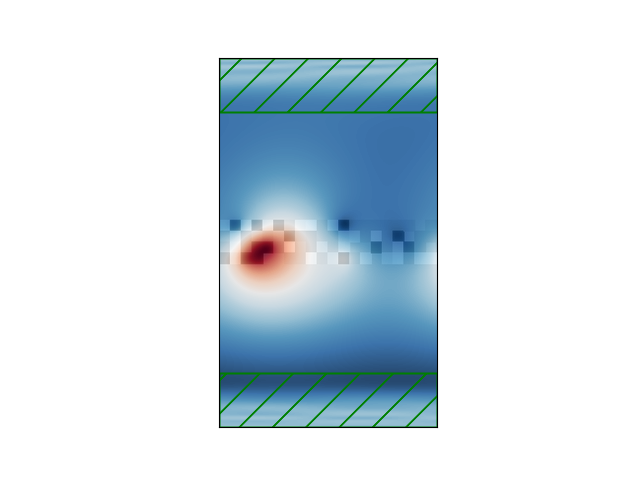}
    \includegraphics[width=0.23\textwidth,trim={1.52in 0.48in 1.40in 0.48in},clip]{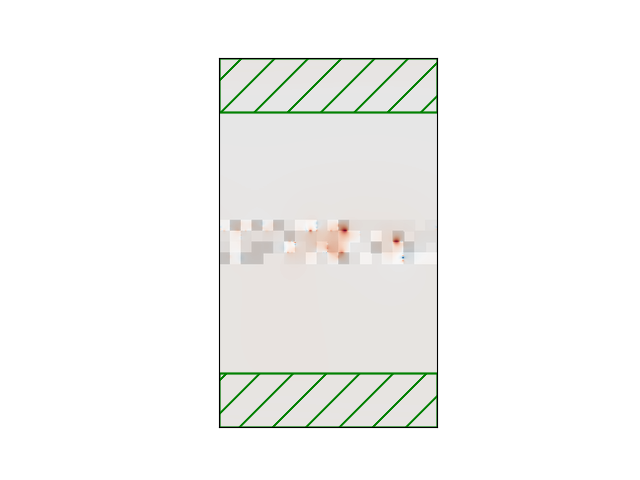}
    \caption{Examples of E-fields (left) and H-fields (right) out of plane for the combinatorial system with material blocks.}
    \label{fig:e_h_fields_combinatorial}
    \vspace{-15pt}
\end{wrapfigure}

For each type of nanophotonic structure featured in our frameworks,
we visualize how specific optical properties change across different structural parameters,
as shown in~\figssref{fig:selected_threelayers}{fig:selected_nanowires}{fig:selected_nanocones_nanospheres_doublenanocones}.
It is important to note that the accuracy and smoothness of the optical property values are dependent on the fidelity level.  
Results from lower fidelity simulations tend to be less accurate and show more erratic fluctuations compared to results from their higher fidelity counterparts.
This difference is evident in the comparisons provided in~\figsref{fig:selected_threelayers}{fig:selected_nanowires}.
Additional visualization for more diverse examples is available in~\secref{sec:additional_visualization}, where the data and examples provided in this paper pertain to two-dimensional nanophotonic structures.

\begin{figure}[t]
    \centering
    \begin{subfigure}[b]{0.32\textwidth}
        \centering
        \includegraphics[width=\textwidth]{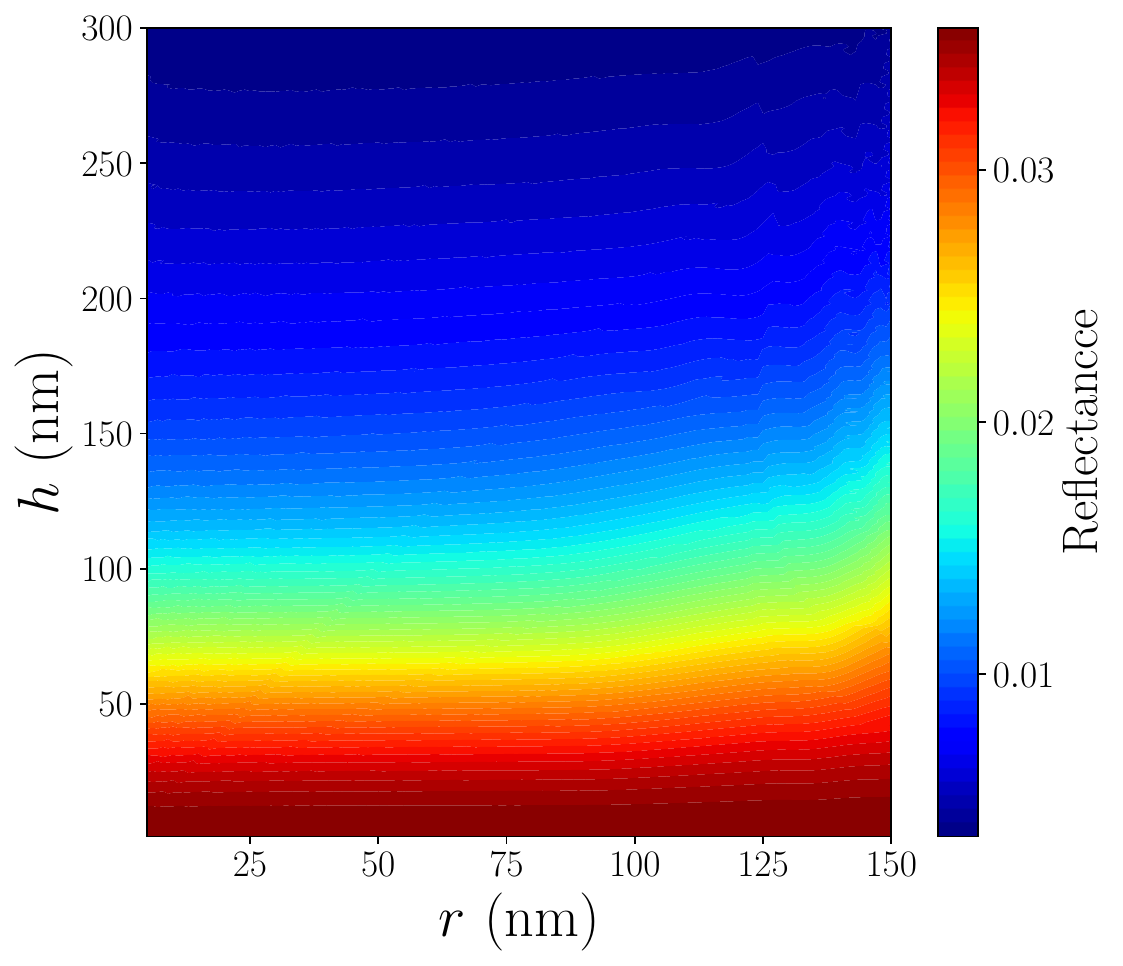}
        \caption{Anti-reflective nanocones}
    \end{subfigure}
    \begin{subfigure}[b]{0.32\textwidth}
        \centering
        \includegraphics[width=\textwidth]{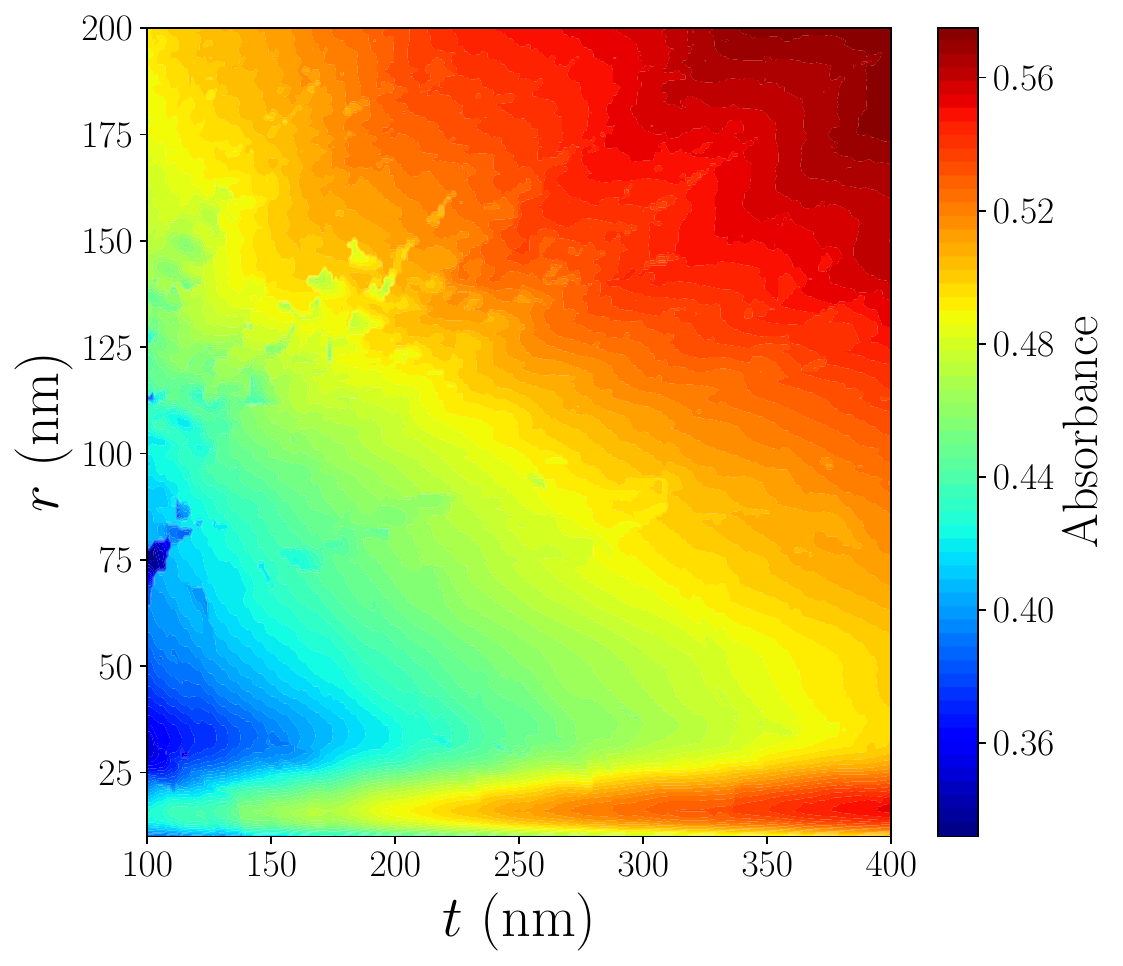}
        \caption{Close-packed nanospheres}
    \end{subfigure}
    \begin{subfigure}[b]{0.32\textwidth}
        \centering
        \includegraphics[width=\textwidth]{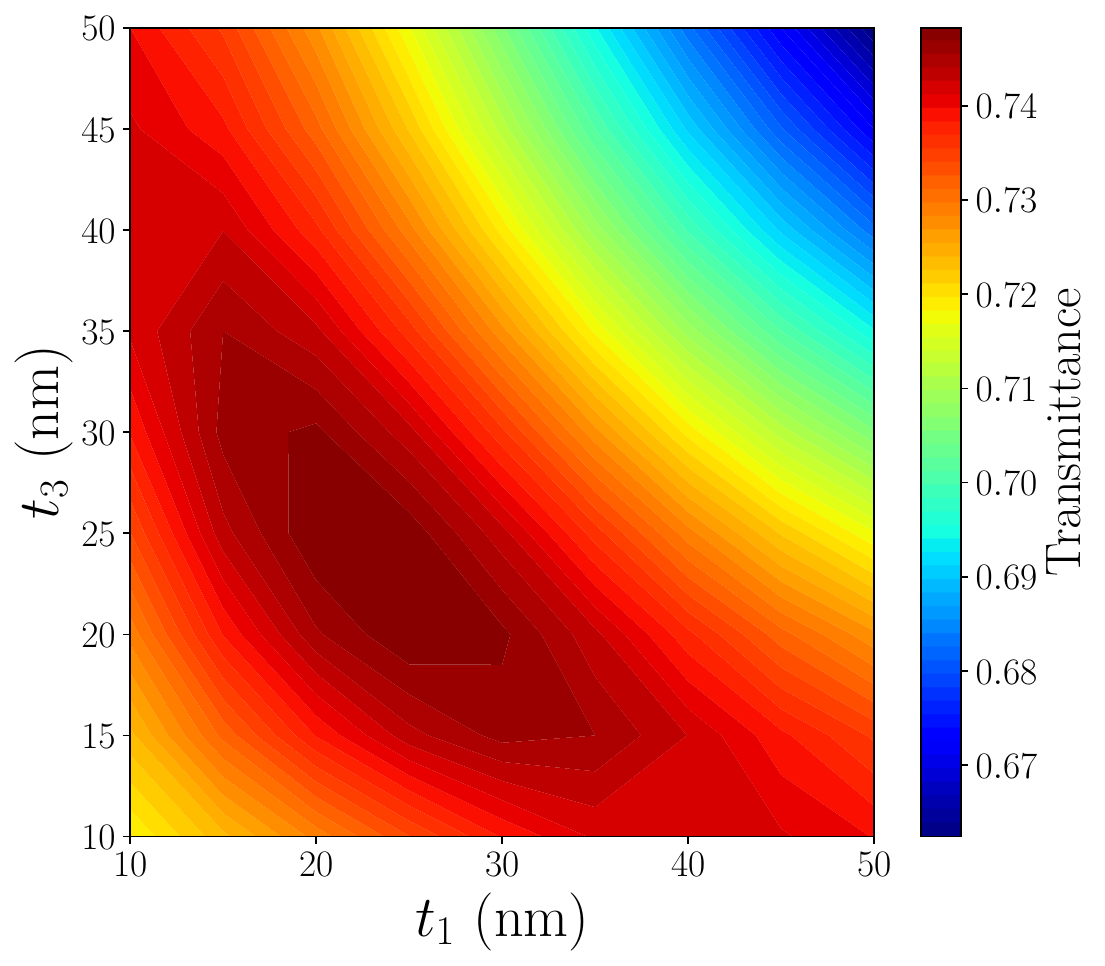}
        \caption{Double-sided nanocones}
    \end{subfigure}
    \caption{Visualization of the target properties of anti-reflective nanocones, close-packed nanospheres, and three-layer film with nanocones, made of fused silica, \ce{cSi}/\ce{TiO2}, and \ce{TiO2}/Ag/\ce{TiO2}/\ce{TiO2}/\ce{TiO2}, respectively. For the film with nanocones, $t_2 =$ 3 nm, $r_1 = r_2 =$ 20 nm, $h_1 = h_2 =$ 50 nm.}
    \label{fig:selected_nanocones_nanospheres_doublenanocones}
\end{figure}
\begin{figure}[t]
    \centering
    \begin{subfigure}[b]{0.19\textwidth}
        \centering
        \includegraphics[width=0.48\textwidth,trim={2.66in 0.48in 2.58in 0.48in},clip]{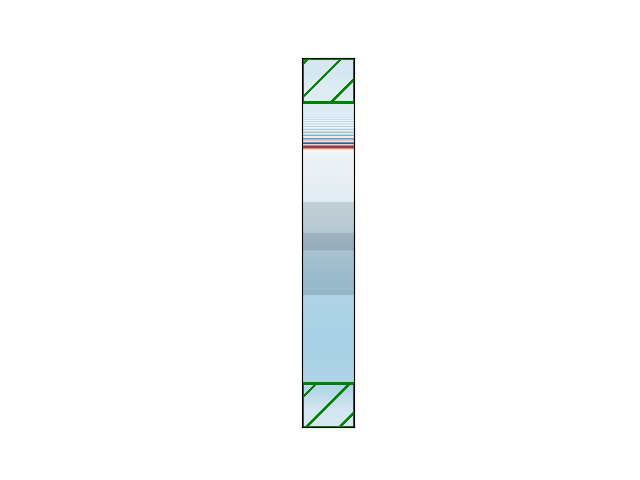}
        \includegraphics[width=0.48\textwidth,trim={2.66in 0.48in 2.58in 0.48in},clip]{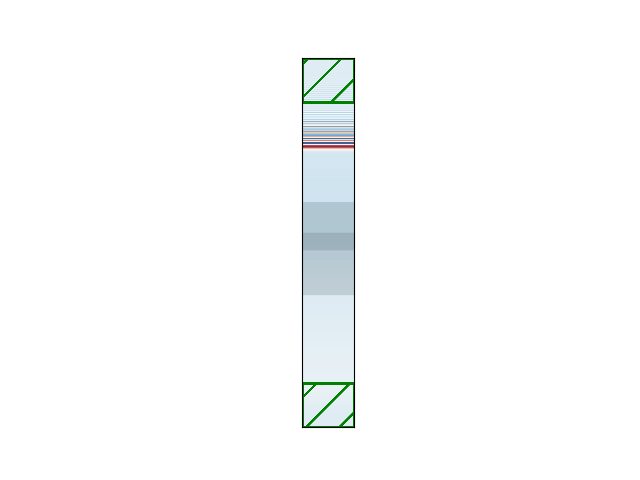}
        \caption{Three-layer}
    \end{subfigure}
    \begin{subfigure}[b]{0.19\textwidth}
        \centering
        \includegraphics[width=0.48\textwidth,trim={2.66in 0.48in 2.58in 0.48in},clip]{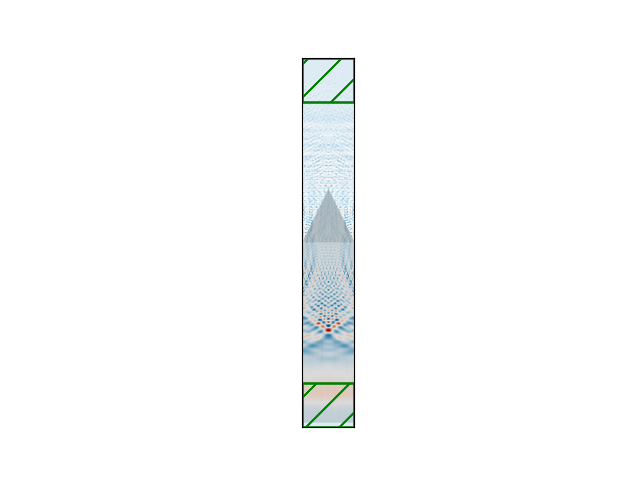}
        \includegraphics[width=0.48\textwidth,trim={2.66in 0.48in 2.58in 0.48in},clip]{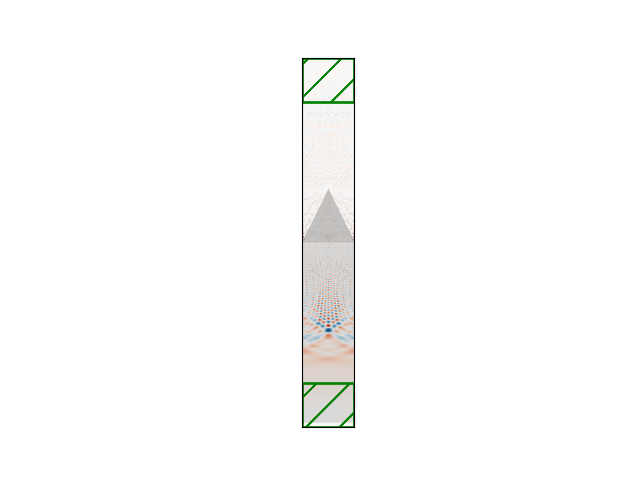}
        \caption{Nanocones}
    \end{subfigure}
    \begin{subfigure}[b]{0.19\textwidth}
        \centering
        \includegraphics[width=0.48\textwidth,trim={2.66in 0.48in 2.58in 0.48in},clip]{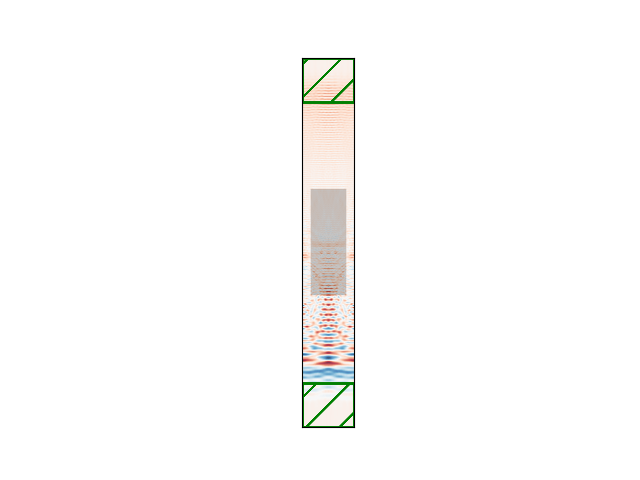}
        \includegraphics[width=0.48\textwidth,trim={2.66in 0.48in 2.58in 0.48in},clip]{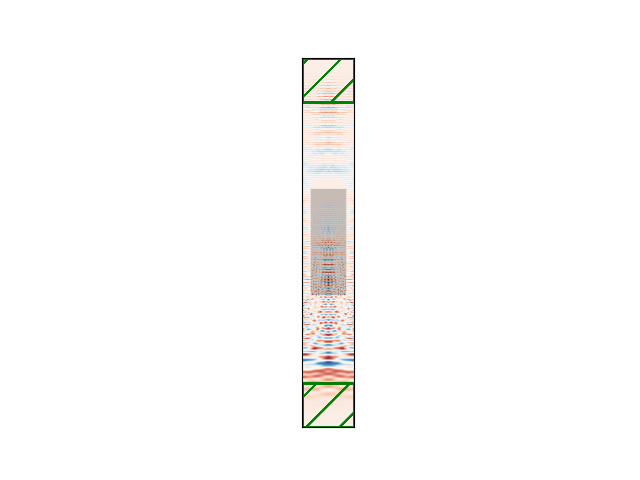}
        \caption{Nanowires}
    \end{subfigure}
    \begin{subfigure}[b]{0.19\textwidth}
        \centering
        \includegraphics[width=0.48\textwidth,trim={2.66in 0.48in 2.58in 0.48in},clip]{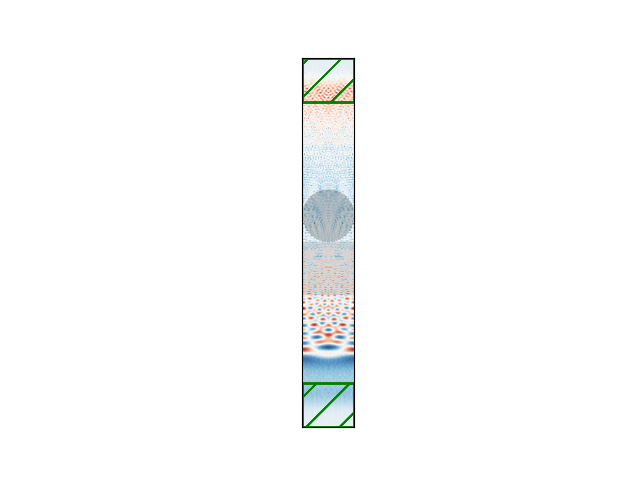}
        \includegraphics[width=0.48\textwidth,trim={2.66in 0.48in 2.58in 0.48in},clip]{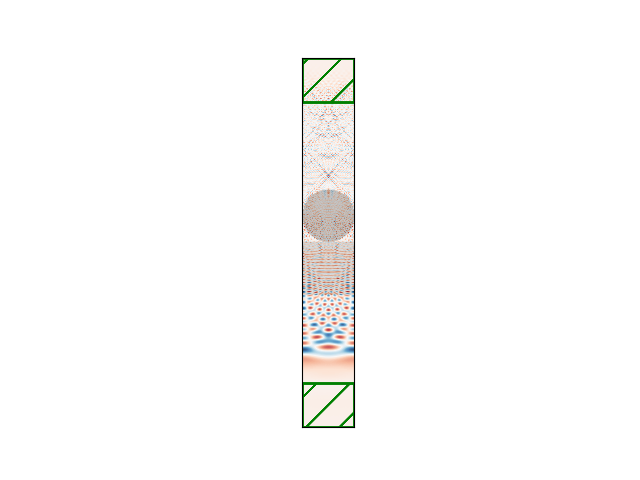}
        \caption{Nanospheres}
    \end{subfigure}
    \begin{subfigure}[b]{0.19\textwidth}
        \centering
        \includegraphics[width=0.48\textwidth,trim={2.66in 0.48in 2.58in 0.48in},clip]{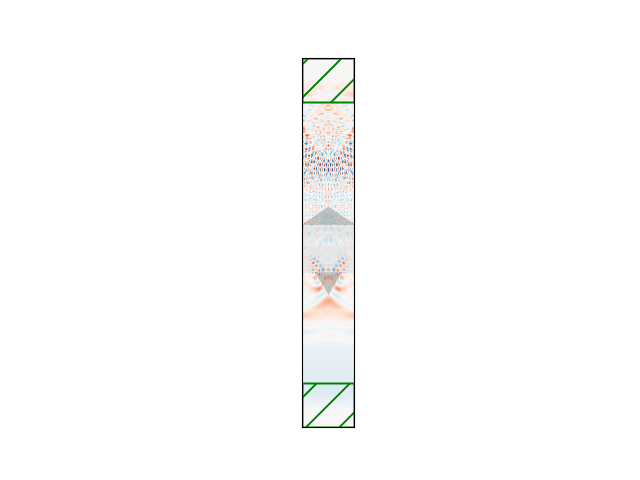}
        \includegraphics[width=0.48\textwidth,trim={2.66in 0.48in 2.58in 0.48in},clip]{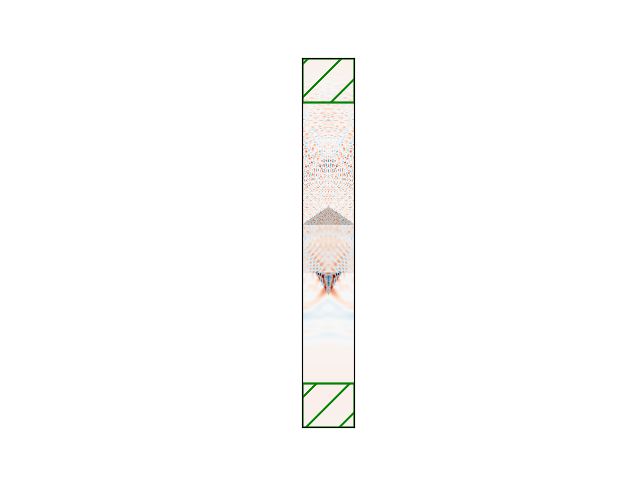}
        \caption{Double-sided}
    \end{subfigure}
    \caption{Examples of E-fields (left of each panel) and H-fields (right of each panel) out of plane for the structures studied in this work. The results for the combinatorial system are shown in~\figref{fig:e_h_fields_combinatorial}.}
    \label{fig:e_h_fields}
\end{figure}

In addition to optical properties, our frameworks are also equipped to simulate and provide data on electric and magnetic fields
(i.e., E- and H-fields),
across various positions and time steps within a simulation cell.
\figsref{fig:e_h_fields_combinatorial}{fig:e_h_fields} provide the visual representation of these fields, clearly showing how they are influenced and altered by the presence of nanophotonic structures.

\section{Datasets and Benchmarks for Nanophotonic Structures and Their Designs}

Following~\secref{sec:main}, we present our datasets and benchmarks defined with several aforementioned nanophotonic structures and their design problems in this section.

\subsection{Datasets}
\label{sec:datasets}

Our frameworks provide datasets and processes for generating datasets,
tailored for nanophotonic structures and their associated parametric design simulations.
To generate these datasets,
it is necessary to define feasible search spaces for parametric structure designs.
Building on~\tabref{tab:search_spaces},
\tabref{tab:statistics} presents
the number of parameters associated with each structure,
the increment value for each structure,
and the number of possible configurations conducted for each structure.

Our frameworks provide datasets of two primary modalities:
\begin{itemize}
    \item Reflection, absorption, and transmission spectra, accompanied by fidelity information: This includes measurements of the three key properties across various structural configurations, considering fidelity information. The fidelity level, which can be low, medium, or high, is determined by the simulation resolution as defined by the spacing of the Yee grid.
    \item Electric and magnetic fields: These E- and H-fields are measured across different positions and simulation time steps as a light source emits a Gaussian-pulse light until it decays sufficiently. The resolution of the simulation and the size of the simulation cell determine the number of grid points used for these positional and temporal measurements.
\end{itemize}

By leveraging the diverse components within our datasets,
users can explore and test a variety of tasks, ranging from multi-fidelity and multi-objective optimization to direct property predictions.
In particular,
our datasets can serve as a valuable testbed for evaluating a variety of machine learning models, such as training them to predict optical spectra or electromagnetic fields,
based on material properties and geometric configurations.

To provide readily accessible datasets for practitioners,
we select some material combinations from~\tabref{tab:search_spaces}
and then generate datasets.
Specifically,
for the three-layer films with and without double-sided nanocones,
we assume that the top and bottom layers, and potentially the top and bottom nanocones are made of the identical material.
Due to the vast number of possible configurations for the combinatorial system with material blocks,
we do not create datasets for this structure,
but instead provide a specific feature to work with this structure; see~\secref{sec:optimization_modes} for further details.

\begin{table}[t]
    \caption{Details of our discretization of the search spaces for the datasets and discretized search space optimization mode. Elaborate description on the number of configurations can be found in~\secref{sec:details_datasets}.}
    \label{tab:statistics}
    \begin{center}
    \small
    \begin{tabular}{lccc}
        \toprule
        \textbf{Structure} & \textbf{\#Parameters} & \textbf{Increment (nm)} & \textbf{\#Configurations} \\
        \midrule
        Three-layer film & 3 & 1 & 149,058 \\
        Anti-reflective nanocones & 2 & 1 & 43,800 \\
        Vertical nanowires & 3 & 1 & 39,200 \\
        Close-packed nanospheres & 2 & 1 & 57,491 \\
        Three-layer film with double-sided nanocones & 7 & 5 & 1,920,996 \\
        Combinatorial system with material blocks & -- & -- & 12$^{\textrm{80}}$ or 12$^{\textrm{1600}}$ \\
        \bottomrule
    \end{tabular}
    \end{center}
\end{table}

\subsection{Optimization Modes}
\label{sec:optimization_modes}

Our benchmarks support two modes for fast prototyping of optimization methods as follows.

\paragraph{Discretized Search Space Mode.}
In this mode, a search space is discretized with the increment specified in~\tabref{tab:statistics}. At each configuration a simulation is run and the outcome of the simulation is recorded. Access to these simulation results is available online.

\paragraph{Surrogate Model Mode.}
Based on the dataset collected from the discretized search space, we fit a surrogate model. This allows us to evaluate any structural configuration in the search space using the trained surrogate model.
All the specifics of this process are detailed in~\secref{sec:details_regression_models}.
It is crucial to note that the predictions made by the surrogate model are not actual evaluations and might contain errors. A comprehensive discussion of these limitations and their implications can be found in~\secref{sec:limitations}.

Note that we currently support these two modes only for the selection of some material combinations.
Also, we do not support these modes for the combinatorial system with material blocks because the search space of this system is extremely huge and therefore it is difficult to obtain its datasets and build an accurate surrogate model.

\paragraph{Simulation Mode.}

Users can also directly run these simulations using our frameworks based on Meep and perform an optimization algorithm by conducting a simulation at every iteration.
This can be more time consuming to execute but is preferred if the user has sufficient compute available.
This allows users to carry out an optimization procedure on the complete space specified in~\tabref{tab:search_spaces} including exploring different material choices.

\subsection{Experiments with the Surrogate Model Mode}
\label{sec:experiments}

\begin{figure}[t]
    \centering
    \begin{subfigure}[b]{0.32\textwidth}
        \centering
        \includegraphics[width=\textwidth]{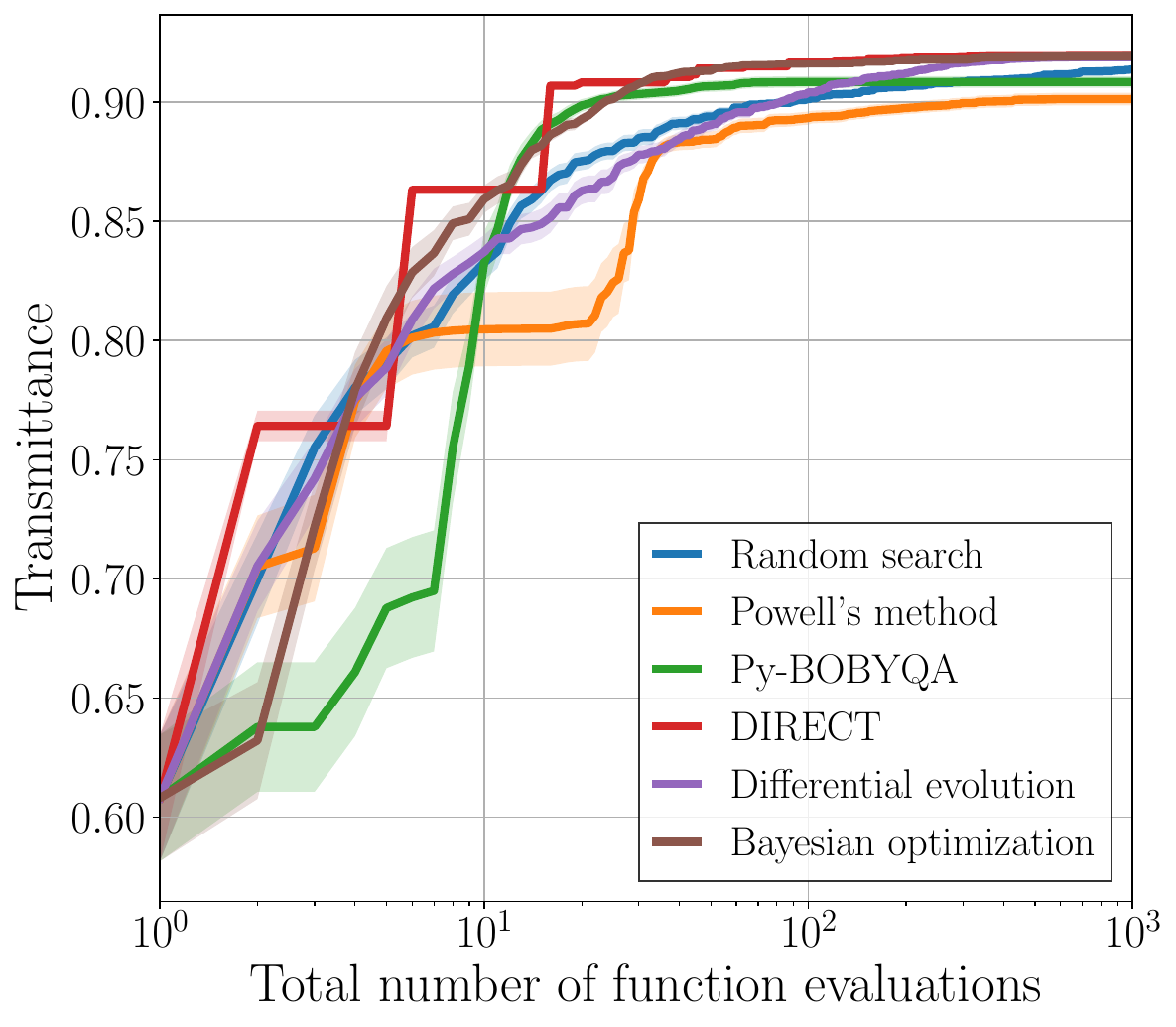}
        \caption{Three-layer film, \ce{TiO2}/\ce{Ag}/\ce{TiO2}}
        \label{fig:results_threelayers}
    \end{subfigure}
    \begin{subfigure}[b]{0.32\textwidth}
        \centering
        \includegraphics[width=\textwidth]{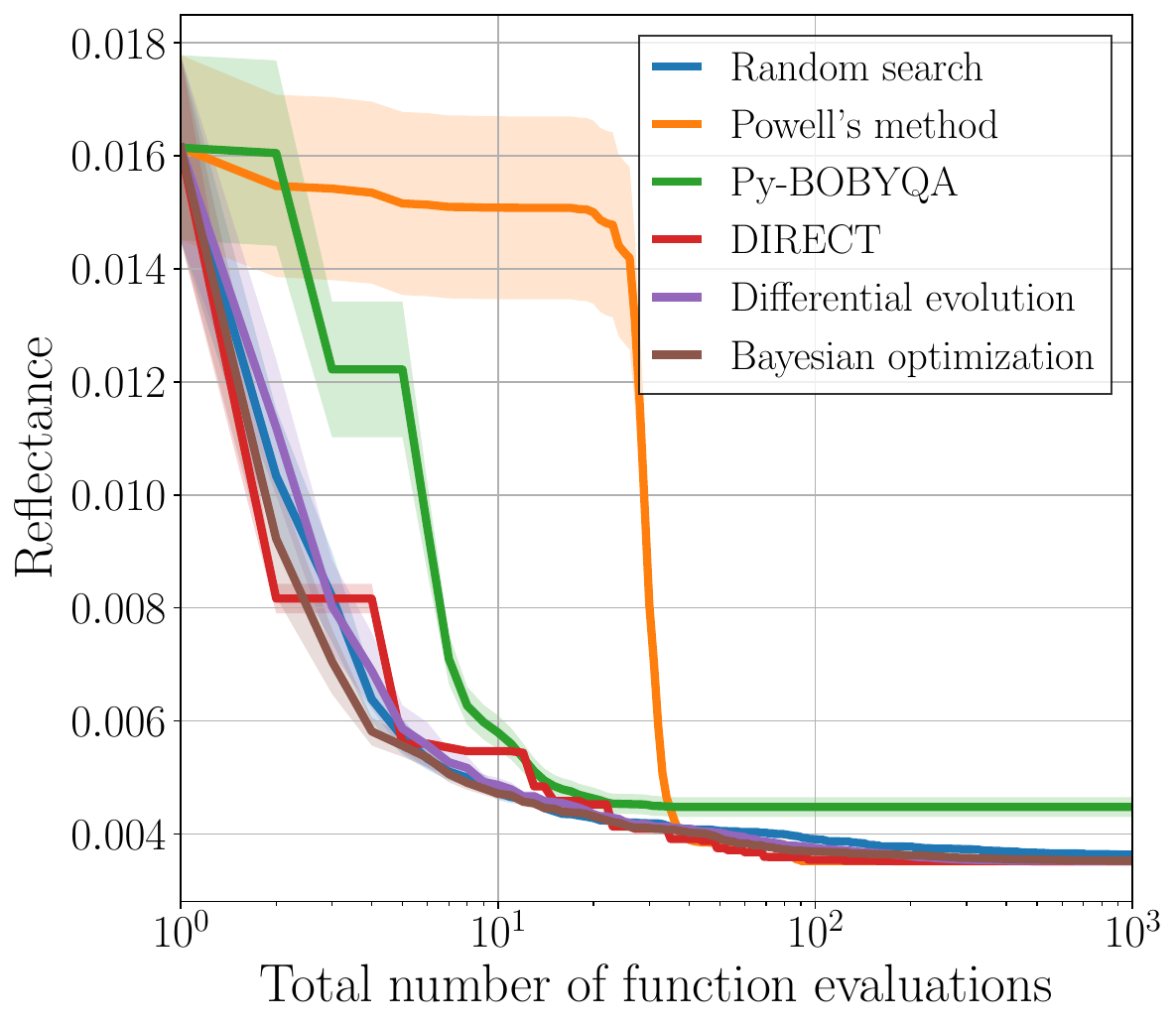}
        \caption{AR nanocones, fused silica}
        \label{fig:results_nanocones}
    \end{subfigure}
    \begin{subfigure}[b]{0.32\textwidth}
        \centering
        \includegraphics[width=\textwidth]{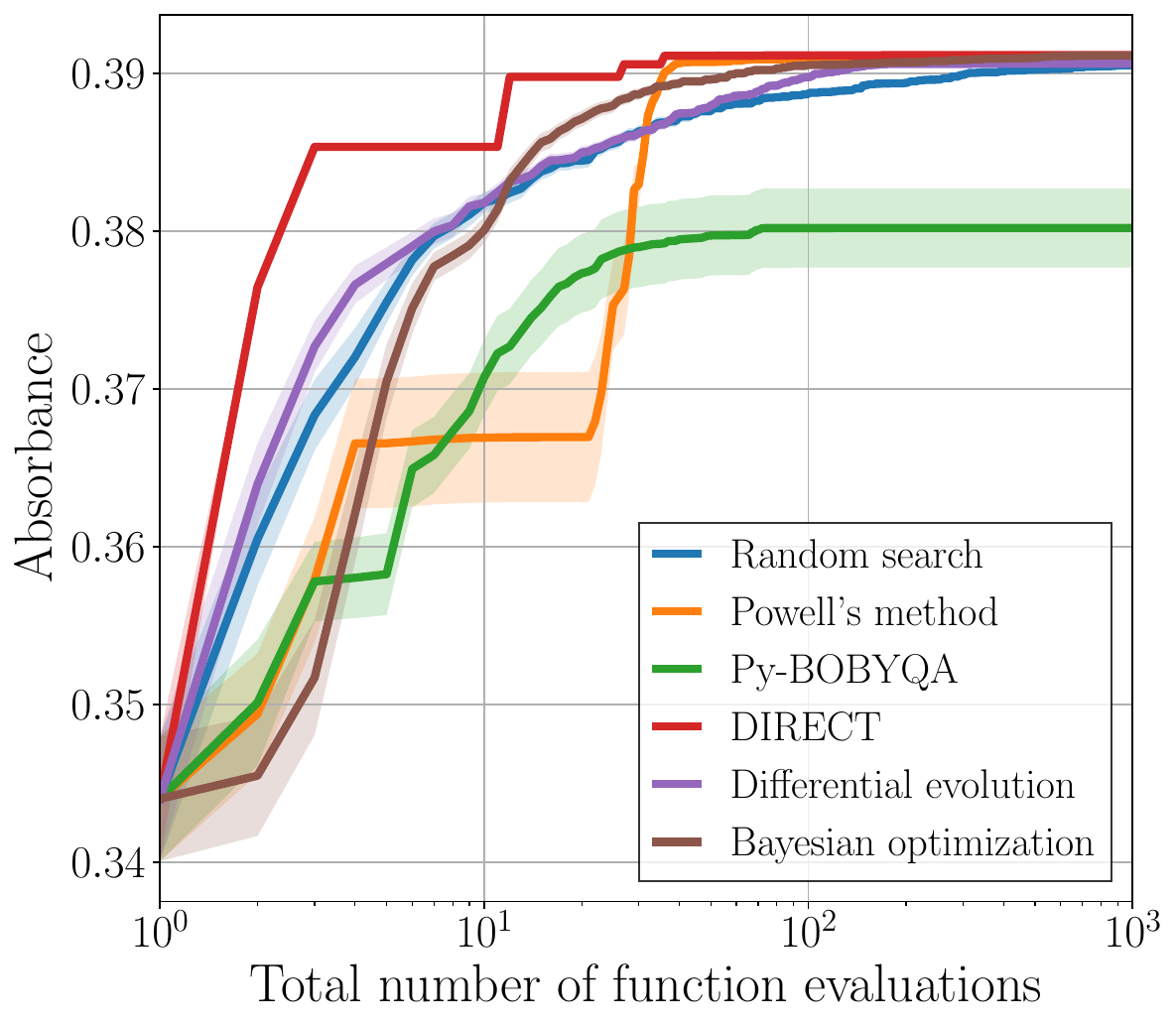}
        \caption{Vertical nanowires, \ce{cSi}}
        \label{fig:results_nanowires}
    \end{subfigure}
    \begin{subfigure}[b]{0.32\textwidth}
        \centering
        \includegraphics[width=\textwidth]{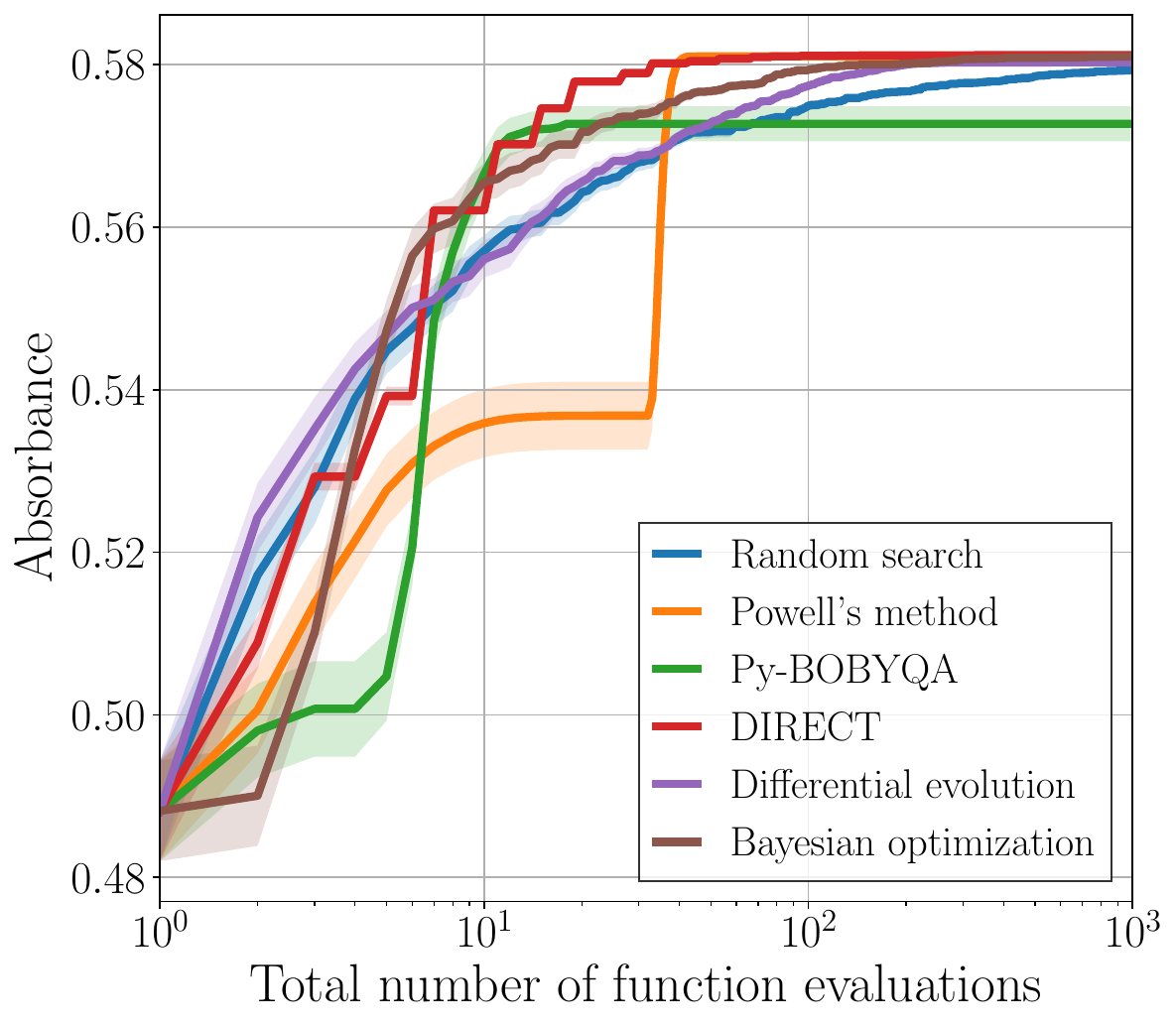}
        \caption{CP nanospheres, \ce{cSi}/\ce{TiO2}}
        \label{fig:results_nanospheres}
    \end{subfigure}
    \begin{subfigure}[b]{0.32\textwidth}
        \centering
        \includegraphics[width=\textwidth]{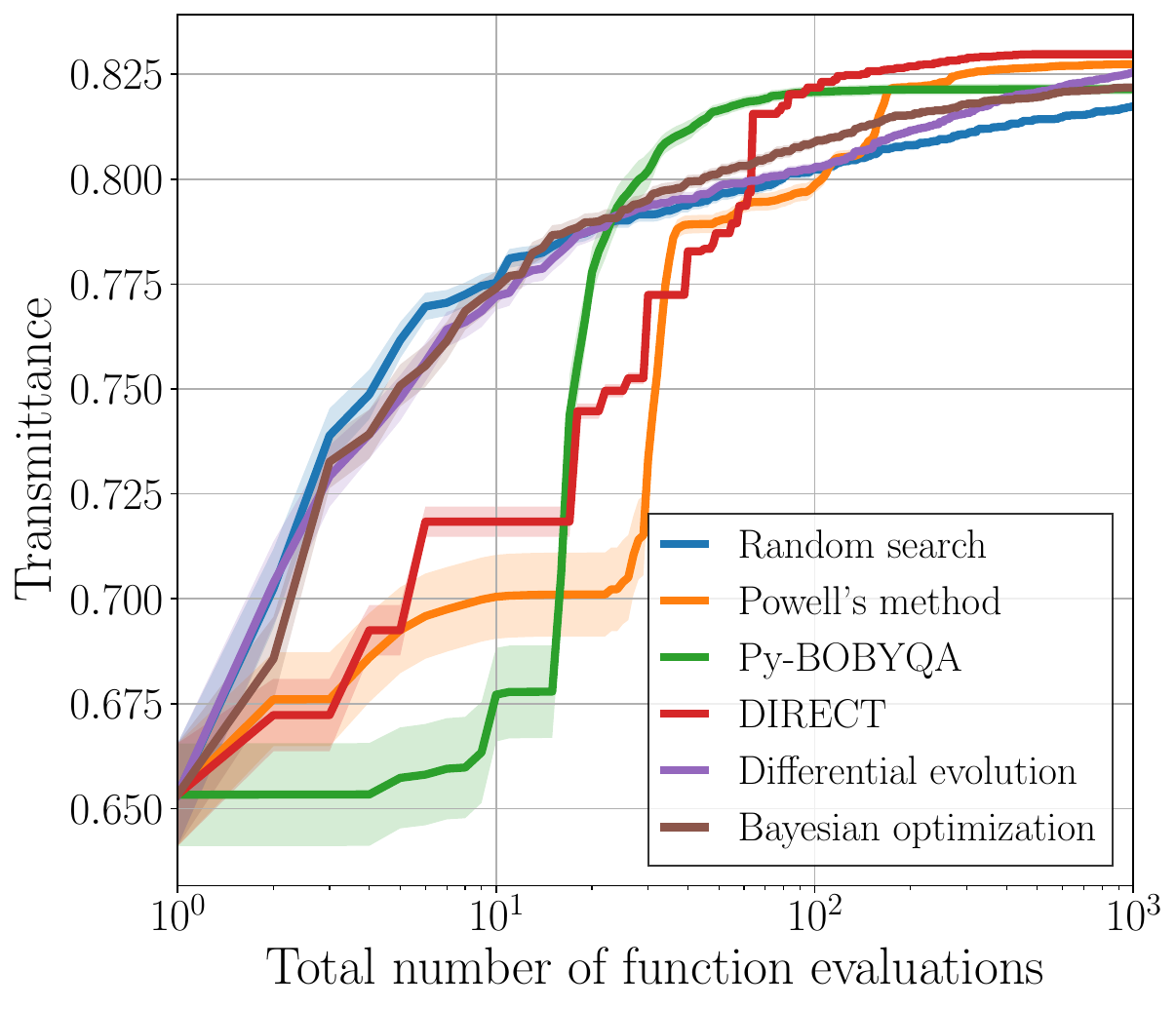}
        \caption{DS cones, \ce{TiO2}/\ce{Ag}/\ce{TiO2}/--/--}
        \label{fig:results_doublenanocones}
    \end{subfigure}
    \caption{Results of experiments on nanophotonic structure optimization. Each experiment is conducted 50 times and the mean and standard error are depicted. AR nanocones, CP nanospheres, DS cones indicate the anti-reflective nanocones, close-packed nanospheres, and three-layer film with double-sided nanocones, respectively. Also, \ce{TiO2}/\ce{Ag}/\ce{TiO2}/--/-- stands for \ce{TiO2}/\ce{Ag}/\ce{TiO2}/\ce{TiO2}/\ce{TiO2}.}
    \label{fig:optimization_experiments}
\end{figure}

In this section, we evaluate the performance of various optimization methods applied to parametric structure optimization.
We consider a range of approaches including
random search,
Powell's method~\citep{PowellMJD1964tcj},
Py-BOBYQA~\citep{CartisC2019acmtoms},
DIRECT~\citep{JonesDR1993jota},
differential evolution~\citep{StornR1997jgo},
and Bayesian optimization~\citep{GarnettR2023book}.
For the experiments shown in~\figref{fig:optimization_experiments},
we specifically focus on assessing the effectiveness of our benchmarks in the surrogate model mode that is defined on a continuous search space.
Most of the algorithms are implemented using SciPy~\citep{VirtanenP2020nm}
and NumPy~\citep{HarrisCR2020nature} and some algorithms are executed using the open-source versions of them;
see~\secref{sec:details_experiments} for their missing details.
Each experiment is independently repeated 50 times, with consistent random seeds maintained across all methods.
The detailed description of experimental setup and implementation can be found in~\secsref{sec:details_regression_models}{sec:details_experiments}.

The experimental results presented in~\figref{fig:optimization_experiments} show that global optimization methods such as DIRECT, differential evolution, and Bayesian optimization generally perform well across different structures.
DIRECT, in particular, consistently outperforms the other techniques.
Bayesian optimization and differential evolution also exhibit strong performance, although they can be slower than or comparable to DIRECT.
However, in the case of the three-layer film with nanocones,
we find that DIRECT tends to be slower than both Bayesian optimization and differential evolution; see~\figref{fig:results_doublenanocones}.
Conversely, local optimization methods such as Powell's method and Py-BOBYQA generally take longer to converge and are prone to getting stuck in local optima.
While Powell's method converges to the same solutions as DIRECT in some cases such as~\figssref{fig:results_nanocones}{fig:results_nanowires}{fig:results_nanospheres},
both local search methods struggle to escape local optima in the other instances.
This is the expected behavior of local optimization methods.
Interestingly, random search, despite its simplicity, demonstrates effectiveness and even outperforms the local optimization methods in particular cases.
However, its performance drops in the higher-dimensional problem; see~\figref{fig:results_doublenanocones}.

On the other hand,
for the combinatorial system with material blocks,
we conduct experiments with random search using the simulation mode to evaluate configurations directly.
The results and the analysis on these results are described in~\secref{sec:combinatorial_material_blocks}.

\section{Conclusion}
\label{sec:conclusion}

In this paper we introduced several nanophotonic structures such as the three-layer film,
anti-reflective nanocones,
vertical nanowires,
close-packed nanospheres,
three-layer film with double-sided nanocones,
and combinatorial system with material blocks for real-world applications.
To construct our datasets and benchmarks, we devised a generic simulation scheme and pipeline for nanophotonic structure and parametric design simulations.
Finally, the datasets and benchmarks are proposed by modeling, simulating, and optimizing nanophotonic structures.
Furthermore,
the future directions, limitations, and societal impacts of our work are discussed in~\secssref{sec:future_directions}{sec:limitations}{sec:societal_impacts}.

\begin{ack}
This research was partly funded by the National Science Foundation (NSF) under grant ECCS 1552712. We also received support from the MDS-Rely Center, which is funded by the NSF's Industry--University Cooperative Research Center (IUCRC) program through awards EEC-2052662 and EEC-2052776. OH acknowledges support from the National Science Foundation and United States-Israel Binational Science Foundation (NSF-BSF) program under NSF grant 2239527 and from Air Force Office of Scientific Research (AFOSR) grant FA9550-23-1-0242. Additionally, the University of Pittsburgh Center for Research Computing provided essential resources for this work, particularly the H2P cluster. This cluster operates with support from the NSF award OAC-2117681.
\end{ack}

\bibliographystyle{abbrvnat}
\bibliography{kjt}

\begin{thebibliography}{80}
\providecommand{\natexlab}[1]{#1}
\providecommand{\url}[1]{\texttt{#1}}
\expandafter\ifx\csname urlstyle\endcsname\relax
  \providecommand{\doi}[1]{doi: #1}\else
  \providecommand{\doi}{doi: \begingroup \urlstyle{rm}\Url}\fi

\bibitem[Aguilar et~al.(2019)Aguilar, {de Castro}, Godoy, and
  Dias]{AguilarO2019ome}
O.~Aguilar, S.~{de Castro}, M.~P.~F. Godoy, and M.~R.~S. Dias.
\newblock Optoelectronic characterization of
  {Zn}$_{\textrm{1-x}}${Cd}$_{\textrm{x}}${O} thin films as an alternative to
  photonic crystals in organic solar cells.
\newblock \emph{Optical Materials Express}, 9\penalty0 (9):\penalty0
  3638--3648, 2019.

\bibitem[{{ASTM} International}(2020)]{ASTMI2020SolarAM15}
{{ASTM} International}.
\newblock Tables for reference solar spectral irradiances: Direct normal and
  hemispherical on 37 tilted surface.
\newblock \url{https://doi.org/10.1520/g0173-03r20}, 2020.

\bibitem[Belakaria et~al.(2019)Belakaria, Deshwal, and
  Doppa]{BelakariaS2019neurips}
S.~Belakaria, A.~Deshwal, and J.~R. Doppa.
\newblock Max-value entropy search for multi-objective {Bayesian} optimization.
\newblock In \emph{Advances in Neural Information Processing Systems
  (NeurIPS)}, volume~32, pages 7825--7835, Vancouver, British Columbia, Canada,
  2019.

\bibitem[Belakaria et~al.(2020)Belakaria, Deshwal, and
  Doppa]{BelakariaS2020aaai}
S.~Belakaria, A.~Deshwal, and J.~R. Doppa.
\newblock Multi-fidelity multi-objective {Bayesian} optimization: An output
  space entropy search approach.
\newblock In \emph{Proceedings of the AAAI Conference on Artificial
  Intelligence (AAAI)}, pages 10035--10043, New York, New York, USA, 2020.

\bibitem[Berenger(1994)]{BerengerJP1994jcp}
J.-P. Berenger.
\newblock A perfectly matched layer for the absorption of electromagnetic
  waves.
\newblock \emph{Journal of Computational Physics}, 114\penalty0 (2):\penalty0
  185--200, 1994.

\bibitem[Billard(1967)]{BillardJ1967phdthesis}
J.~Billard.
\newblock \emph{Contribution {\`a} l'{\'e}tude de la propagation des ondes
  {\'e}lectromagn{\'e}tiques planes dans certains milieux mat{\'e}riels}.
\newblock PhD thesis, Centre national de la recherche scientifique, 1967.

\bibitem[Born and Wolf(2013)]{BornM2013book}
M.~Born and E.~Wolf.
\newblock \emph{Principles of optics: electromagnetic theory of propagation,
  interference and diffraction of light}.
\newblock Elsevier, 2013.

\bibitem[Brown et~al.(2019)Brown, Fiscato, Segler, and Vaucher]{BrownN2019jcim}
N.~Brown, M.~Fiscato, M.~H.~S. Segler, and A.~C. Vaucher.
\newblock {GuacaMol}: benchmarking models for de novo molecular design.
\newblock \emph{Journal of Chemical Information and Modeling}, 59\penalty0
  (3):\penalty0 1096--1108, 2019.

\bibitem[Callahan et~al.(2012)Callahan, Munday, and Atwater]{CallahanDM2012nl}
D.~M. Callahan, J.~N. Munday, and H.~A. Atwater.
\newblock Solar cell light trapping beyond the ray optic limit.
\newblock \emph{Nano Letters}, 12\penalty0 (1):\penalty0 214--218, 2012.

\bibitem[Cartis et~al.(2019)Cartis, Fiala, Marteau, and
  Roberts]{CartisC2019acmtoms}
C.~Cartis, J.~Fiala, B.~Marteau, and L.~Roberts.
\newblock Improving the flexibility and robustness of model-based
  derivative-free optimization solvers.
\newblock \emph{ACM Transactions on Mathematical Software}, 45\penalty0
  (3):\penalty0 1--41, 2019.

\bibitem[Deng et~al.(2021)Deng, Dong, Ren, Khatib, Soltani, Tarokh, Padilla,
  and Malof]{DengY2021neuripsdb}
Y.~Deng, J.~Dong, S.~Ren, O.~Khatib, M.~Soltani, V.~Tarokh, W.~Padilla, and
  J.~Malof.
\newblock Benchmarking data-driven surrogate simulators for artificial
  electromagnetic materials.
\newblock In \emph{Advances in Neural Information Processing Systems
  (NeurIPS)}, Virtual, 2021.
\newblock Datasets and Benchmarks Track.

\bibitem[Dong and Yang(2019)]{DongX2019iclr}
X.~Dong and Y.~Yang.
\newblock {NAS-Bench-201}: Extending the scope of reproducible neural
  architecture search.
\newblock In \emph{Proceedings of the International Conference on Learning
  Representations (ICLR)}, New Orleans, Louisiana, USA, 2019.

\bibitem[Dong et~al.(2021)Dong, Liu, Musial, and Gabrys]{DongX2021ieeetpami}
X.~Dong, L.~Liu, K.~Musial, and B.~Gabrys.
\newblock {NATS-Bench}: Benchmarking {NAS} algorithms for architecture topology
  and size.
\newblock \emph{IEEE Transactions on Pattern Analysis and Machine
  Intelligence}, 44\penalty0 (7):\penalty0 3634--3646, 2021.

\bibitem[Eggensperger et~al.(2021)Eggensperger, M{\"u}ller, Mallik, Feurer,
  Sass, Klein, Awad, Lindauer, and Hutter]{EggenspergerK2021neuripsdb}
K.~Eggensperger, P.~M{\"u}ller, N.~Mallik, M.~Feurer, R.~Sass, A.~Klein,
  N.~Awad, M.~Lindauer, and F.~Hutter.
\newblock {HPOBench}: A collection of reproducible multi-fidelity benchmark
  problems for {HPO}.
\newblock In \emph{Advances in Neural Information Processing Systems
  (NeurIPS)}, Virtual, 2021.
\newblock Datasets and Benchmarks Track.

\bibitem[Forrester et~al.(2007)Forrester, S{\'o}bester, and
  Keane]{ForresterAIJ2007prsa}
A.~I.~J. Forrester, A.~S{\'o}bester, and A.~J. Keane.
\newblock Multi-fidelity optimization via surrogate modelling.
\newblock \emph{Proceedings of the Royal Society A: Mathematical, Physical and
  Engineering Sciences}, 463\penalty0 (2088):\penalty0 3251--3269, 2007.

\bibitem[Garnett(2023)]{GarnettR2023book}
R.~Garnett.
\newblock \emph{{Bayesian Optimization}}.
\newblock Cambridge University Press, 2023.

\bibitem[Grandidier et~al.(2011)Grandidier, Callahan, Munday, and
  Atwater]{GrandidierJ2011am}
J.~Grandidier, D.~M. Callahan, J.~N. Munday, and H.~A. Atwater.
\newblock Light absorption enhancement in thin-film solar cells using
  whispering gallery modes in dielectric nanospheres.
\newblock \emph{Advanced Materials}, 23\penalty0 (10):\penalty0 1272--1276,
  2011.

\bibitem[Griffiths(2005)]{GriffithsDJ2005book}
D.~J. Griffiths.
\newblock \emph{Introduction to electrodynamics}.
\newblock American Association of Physics Teachers, 2005.

\bibitem[Haghanifar and Leu(2022)]{HaghanifarS2022oe}
S.~Haghanifar and P.~W. Leu.
\newblock Detailed balance analysis of vertical {GaAs} nanowire array solar
  cells: exceeding the {Shockley Queisser} limit.
\newblock \emph{Optics Express}, 30\penalty0 (10):\penalty0 16145--16158, 2022.

\bibitem[Haghanifar et~al.(2019)Haghanifar, McCourt, Cheng, Wuenschell,
  Ohodnicki, and Leu]{HaghanifarS2019mh}
S.~Haghanifar, M.~McCourt, B.~Cheng, J.~Wuenschell, P.~Ohodnicki, and P.~W.
  Leu.
\newblock Creating glasswing butterfly-inspired durable antifogging
  superomniphobic supertransmissive, superclear nanostructured glass through
  {Bayesian} learning and optimization.
\newblock \emph{Materials Horizons}, 6\penalty0 (8):\penalty0 1632--1642, 2019.

\bibitem[Haghanifar et~al.(2020)Haghanifar, McCourt, Cheng, Wuenschell,
  Ohodnicki, and Leu]{HaghanifarS2020optica}
S.~Haghanifar, M.~McCourt, B.~Cheng, J.~Wuenschell, P.~Ohodnicki, and P.~W.
  Leu.
\newblock Discovering high-performance broadband and broad angle antireflection
  surfaces by machine learning.
\newblock \emph{Optica}, 7\penalty0 (7):\penalty0 784--789, 2020.

\bibitem[Hammond et~al.(2021)Hammond, Oskooi, Johnson, and
  Ralph]{HammondAM2021oe}
A.~M. Hammond, A.~Oskooi, S.~G. Johnson, and S.~E. Ralph.
\newblock Photonic topology optimization with semiconductor-foundry design-rule
  constraints.
\newblock \emph{Optics Express}, 29\penalty0 (15):\penalty0 23916--23938, 2021.

\bibitem[Hammond et~al.(2022)Hammond, Oskooi, Chen, Lin, Johnson, and
  Ralph]{HammondAM2022oe}
A.~M. Hammond, A.~Oskooi, M.~Chen, Z.~Lin, S.~G. Johnson, and S.~E. Ralph.
\newblock High-performance hybrid time/frequency-domain topology optimization
  for large-scale photonics inverse design.
\newblock \emph{Optics Express}, 30\penalty0 (3):\penalty0 4467--4491, 2022.

\bibitem[Hansen et~al.(2010)Hansen, Auger, Ros, Finck, and
  Po{\v{s}}{\'\i}k]{HansenN2010gecco}
N.~Hansen, A.~Auger, R.~Ros, S.~Finck, and P.~Po{\v{s}}{\'\i}k.
\newblock Comparing results of 31 algorithms from the black-box optimization
  benchmarking {BBOB-2009}.
\newblock In \emph{Proceedings of the Annual Conference on Genetic and
  Evolutionary Computation (GECCO)}, pages 1689--1696, Portland, Oregon, USA,
  2010.

\bibitem[Harris et~al.(2020)Harris, Millman, {van der Walt}, Gommers, Virtanen,
  Cournapeau, Wieser, Taylor, Berg, Smith, Kern, Picus, Hoyer, {van Kerkwijk},
  Brett, Haldane, {del R\'{i}o}, Wiebe, Peterson, G\'{e}rard-Marchant,
  Sheppard, Reddy, Weckesser, Abbasi, Gohlke, and Oliphant]{HarrisCR2020nature}
C.~R. Harris, K.~J. Millman, S.~J. {van der Walt}, R.~Gommers, P.~Virtanen,
  D.~Cournapeau, E.~Wieser, J.~Taylor, S.~Berg, N.~J. Smith, R.~Kern, M.~Picus,
  S.~Hoyer, M.~H. {van Kerkwijk}, M.~Brett, A.~Haldane, J.~F. {del R\'{i}o},
  M.~Wiebe, P.~Peterson, P.~G\'{e}rard-Marchant, K.~Sheppard, T.~Reddy,
  W.~Weckesser, H.~Abbasi, C.~Gohlke, and T.~E. Oliphant.
\newblock Array programming with {NumPy}.
\newblock \emph{Nature}, 585:\penalty0 357--362, 2020.

\bibitem[Hern{\'a}ndez-Lobato et~al.(2016)Hern{\'a}ndez-Lobato,
  {Hern\'andez-Lobato}, Shah, and Adams]{HernandezLobatoD2016neurips}
D.~Hern{\'a}ndez-Lobato, J.~M. {Hern\'andez-Lobato}, A.~Shah, and R.~P. Adams.
\newblock Predictive entropy search for multi-objective {Bayesian}
  optimization.
\newblock In \emph{Proceedings of the International Conference on Machine
  Learning (ICML)}, pages 1492--1501, New York, New York, USA, 2016.

\bibitem[{International Commission on Illumination
  ({CIE})}(2022)]{CIE2022SID65}
{International Commission on Illumination ({CIE})}.
\newblock {CIE} standard illuminant {D65}.
\newblock \url{https://doi.org/10.25039/CIE.DS.hjfjmt59}, 2022.

\bibitem[Ioffe and Szegedy(2015)]{IoffeS2015icml}
S.~Ioffe and C.~Szegedy.
\newblock Batch normalization: Accelerating deep network training by reducing
  internal covariate shift.
\newblock In \emph{Proceedings of the International Conference on Machine
  Learning (ICML)}, pages 448--456, Lille, France, 2015.

\bibitem[Jin(2015)]{JinJM2015book}
J.-M. Jin.
\newblock \emph{The finite element method in electromagnetics}.
\newblock John Wiley \& Sons, 2015.

\bibitem[Jones et~al.(1993)Jones, Perttunen, and Stuckman]{JonesDR1993jota}
D.~R. Jones, C.~D. Perttunen, and B.~E. Stuckman.
\newblock {Lipschitzian} optimization without the {Lipschitz} constant.
\newblock \emph{Journal of Optimization Theory and Applications}, 79\penalty0
  (1):\penalty0 157--181, 1993.

\bibitem[Kandasamy et~al.(2017)Kandasamy, Dasarathy, Schneider, and
  P{\'o}czos]{KandasamyK2017icml}
K.~Kandasamy, G.~Dasarathy, J.~Schneider, and B.~P{\'o}czos.
\newblock Multi-fidelity {Bayesian} optimisation with continuous
  approximations.
\newblock In \emph{Proceedings of the International Conference on Machine
  Learning (ICML)}, pages 1799--1808, Sydney, Australia, 2017.

\bibitem[Kelzenberg et~al.(2008)Kelzenberg, Turner-Evans, Kayes, Filler,
  Putnam, Lewis, and Atwater]{KelzenbergMD2008nl}
M.~D. Kelzenberg, D.~B. Turner-Evans, B.~M. Kayes, M.~A. Filler, M.~C. Putnam,
  N.~S. Lewis, and H.~A. Atwater.
\newblock Photovoltaic measurements in single-nanowire silicon solar cells.
\newblock \emph{Nano Letters}, 8\penalty0 (2):\penalty0 710--714, 2008.

\bibitem[Kim and Choi(2023)]{KimJ2023joss}
J.~Kim and S.~Choi.
\newblock {BayesO}: A {Bayesian} optimization framework in {Python}.
\newblock \emph{Journal of Open Source Software}, 8\penalty0 (90):\penalty0
  5320, 2023.

\bibitem[Kim et~al.(2022)Kim, Lu, Loh, Smith, Snoek, and
  Soljacic]{KimS2022tmlr}
S.~Kim, P.~Y. Lu, C.~Loh, J.~Smith, J.~Snoek, and M.~Soljacic.
\newblock Deep learning for {Bayesian} optimization of scientific problems with
  high-dimensional structure.
\newblock \emph{Transactions on Machine Learning Research}, 2022.

\bibitem[Kingma and Ba(2015)]{KingmaDP2015iclr}
D.~P. Kingma and J.~L. Ba.
\newblock {Adam}: A method for stochastic optimization.
\newblock In \emph{Proceedings of the International Conference on Learning
  Representations (ICLR)}, San Diego, California, USA, 2015.

\bibitem[Klein and Hutter(2019)]{KleinA2019arxiv}
A.~Klein and F.~Hutter.
\newblock Tabular benchmarks for joint architecture and hyperparameter
  optimization.
\newblock \emph{arXiv preprint arXiv:1905.04970}, 2019.

\bibitem[Knowles(2006)]{KnowlesJ2006ieeetec}
J.~Knowles.
\newblock {ParEGO}: a hybrid algorithm with on-line landscape approximation for
  expensive multiobjective optimization problems.
\newblock \emph{IEEE Transactions on Evolutionary Computation}, 10\penalty0
  (1):\penalty0 50--66, 2006.

\bibitem[K{\"o}nig et~al.(2014)K{\"o}nig, Ledin, Kerszulis, Mahmoud, El-Sayed,
  Reynolds, and Tsukruk]{KonigTAF2014acsn}
T.~A.~F. K{\"o}nig, P.~A. Ledin, J.~Kerszulis, M.~A. Mahmoud, M.~A. El-Sayed,
  J.~R. Reynolds, and V.~V. Tsukruk.
\newblock Electrically tunable plasmonic behavior of nanocube--polymer
  nanomaterials induced by a redox-active electrochromic polymer.
\newblock \emph{ACS Nano}, 8\penalty0 (6):\penalty0 6182--6192, 2014.

\bibitem[Li et~al.(2022)Li, McCourt, Galante, and Leu]{LiM2022oe}
M.~Li, M.~J. McCourt, A.~J. Galante, and P.~W. Leu.
\newblock {Bayesian} optimization of nanophotonic electromagnetic shielding
  with very high visible transparency.
\newblock \emph{Optics Express}, 30\penalty0 (18):\penalty0 33182--33194, 2022.

\bibitem[Li et~al.(2023)Li, Zarei, Mohammadi, Walker, LeMieux, and
  Leu]{LiM2023acsami}
M.~Li, M.~Zarei, K.~Mohammadi, S.~B. Walker, M.~LeMieux, and P.~W. Leu.
\newblock Silver meshes for record-performance transparent electromagnetic
  interference shielding.
\newblock \emph{ACS Applied Materials \& Interfaces}, 15\penalty0
  (25):\penalty0 30591--30599, 2023.

\bibitem[Malitson(1965)]{MalitsonIH1965josa}
I.~H. Malitson.
\newblock Interspecimen comparison of the refractive index of fused silica.
\newblock \emph{Journal of the Optical Society of America}, 55\penalty0
  (10):\penalty0 1205--1209, 1965.

\bibitem[Maniyara et~al.(2016)Maniyara, Mkhitaryan, Chen, Ghosh, and
  Pruneri]{ManiyaraRA2016nc}
R.~A. Maniyara, V.~K. Mkhitaryan, T.~L. Chen, D.~S. Ghosh, and V.~Pruneri.
\newblock An antireflection transparent conductor with ultralow optical loss (<
  2\%) and electrical resistance (< 6 $\omega$ sq$^-1$).
\newblock \emph{Nature Communications}, 7\penalty0 (1):\penalty0 13771, 2016.

\bibitem[Moharam and Gaylord(1981)]{MoharamMG1981josa}
M.~G. Moharam and T.~K. Gaylord.
\newblock Rigorous coupled-wave analysis of planar-grating diffraction.
\newblock \emph{Journal of the Optical Society of America}, 71\penalty0
  (7):\penalty0 811--818, 1981.

\bibitem[Oskooi et~al.(2010)Oskooi, Roundy, Ibanescu, Bermel, Joannopoulos, and
  Johnson]{OskooiAF2010cpc}
A.~F. Oskooi, D.~Roundy, M.~Ibanescu, P.~Bermel, J.~D. Joannopoulos, and S.~G.
  Johnson.
\newblock {MEEP}: A flexible free-software package for electromagnetic
  simulations by the {FDTD} method.
\newblock \emph{Computer Physics Communications}, 181\penalty0 (3):\penalty0
  687--702, 2010.

\bibitem[Paszke et~al.(2019)Paszke, Gross, Massa, Lerer, Bradbury, Chanan,
  Killeen, Lin, Gimelshein, Antiga, Desmaison, {K\"{o}pf}, Yang, DeVito,
  Raison, Tejani, Chilamkurthy, Steiner, Fang, Bai, and
  Chintala]{PaszkeA2019neurips}
A.~Paszke, S.~Gross, F.~Massa, A.~Lerer, J.~Bradbury, G.~Chanan, T.~Killeen,
  Z.~Lin, N.~Gimelshein, L.~Antiga, A.~Desmaison, A.~{K\"{o}pf}, E.~Yang,
  Z.~DeVito, M.~Raison, A.~Tejani, S.~Chilamkurthy, B.~Steiner, L.~Fang,
  J.~Bai, and S.~Chintala.
\newblock {PyTorch}: An imperative style, high-performance deep learning
  library.
\newblock In \emph{Advances in Neural Information Processing Systems
  (NeurIPS)}, volume~32, pages 8026--8037, Vancouver, British Columbia, Canada,
  2019.

\bibitem[Pfisterer et~al.(2022)Pfisterer, Schneider, Moosbauer, Binder, and
  Bischl]{PfistererF2022automlconf}
F.~Pfisterer, L.~Schneider, J.~Moosbauer, M.~Binder, and B.~Bischl.
\newblock {YAPHO Gym} -- an efficient multi-objective multi-fidelity benchmark
  for hyperparameter optimization.
\newblock In \emph{Proceedings of the International Conference on Automated
  Machine Learning (AutoML-Conf)}, Baltimore, Maryland, USA, 2022.

\bibitem[Phillips et~al.(2015)Phillips, Rashed, Treharne, Kay, Yates, Mitrovic,
  Weerakkody, Hall, and Durose]{PhillipsLJ2015db}
L.~J. Phillips, A.~M. Rashed, R.~E. Treharne, J.~Kay, P.~Yates, I.~Z. Mitrovic,
  A.~Weerakkody, S.~Hall, and K.~Durose.
\newblock Dispersion relation data for methylammonium lead triiodide perovskite
  deposited on a (100) silicon wafer using a two-step vapour-phase reaction
  process.
\newblock \emph{Data in Brief}, 5:\penalty0 926--928, 2015.

\bibitem[Polman and Atwater(2012)]{PolmanA2012nm}
A.~Polman and H.~A. Atwater.
\newblock Photonic design principles for ultrahigh-efficiency photovoltaics.
\newblock \emph{Nature Materials}, 11\penalty0 (3):\penalty0 174--177, 2012.

\bibitem[Powell(1964)]{PowellMJD1964tcj}
M.~J.~D. Powell.
\newblock An efficient method for finding the minimum of a function of several
  variables without calculating derivatives.
\newblock \emph{The Computer Journal}, 7\penalty0 (2):\penalty0 155--162, 1964.

\bibitem[Raki{\'c} and Majewski(1996)]{RakicAD1996jap}
A.~D. Raki{\'c} and M.~L. Majewski.
\newblock Modeling the optical dielectric function of {GaAs} and {AlAs}:
  Extension of {Adachi’s} model.
\newblock \emph{Journal of Applied Physics}, 80\penalty0 (10):\penalty0
  5909--5914, 1996.

\bibitem[Raki{\'c} et~al.(1998)Raki{\'c}, Djuri{\v{s}}i{\'c}, Elazar, and
  Majewski]{RakicAD1998ao}
A.~D. Raki{\'c}, A.~B. Djuri{\v{s}}i{\'c}, J.~M. Elazar, and M.~L. Majewski.
\newblock Optical properties of metallic films for vertical-cavity
  optoelectronic devices.
\newblock \emph{Applied Optics}, 37\penalty0 (22):\penalty0 5271--5283, 1998.

\bibitem[Ramakrishnan et~al.(2014)Ramakrishnan, Dral, Rupp, and {von
  Lilienfeld}]{RamakrishnanR2014sd}
R.~Ramakrishnan, P.~O. Dral, M.~Rupp, and O.~A. {von Lilienfeld}.
\newblock Quantum chemistry structures and properties of 134 kilo molecules.
\newblock \emph{Scientific Data}, 1\penalty0 (1):\penalty0 1--7, 2014.

\bibitem[Rios and Sahinidis(2013)]{RiosLM2013jgo}
L.~M. Rios and N.~V. Sahinidis.
\newblock Derivative-free optimization: a review of algorithms and comparison
  of software implementations.
\newblock \emph{Journal of Global Optimization}, 56:\penalty0 1247--1293, 2013.

\bibitem[Sanchez-Lengeling and Aspuru-Guzik(2018)]{SanchezB2018science}
B.~Sanchez-Lengeling and A.~Aspuru-Guzik.
\newblock Inverse molecular design using machine learning: Generative models
  for matter engineering.
\newblock \emph{Science}, 361\penalty0 (6400):\penalty0 360--365, 2018.

\bibitem[Schinke et~al.(2015)Schinke, Peest, Schmidt, Brendel, Bothe, Vogt,
  Kr{\"o}ger, Winter, Schirmacher, Lim, Nguyen, and
  MacDonald]{SchinkeC2015aipa}
C.~Schinke, P.~C. Peest, J.~Schmidt, R.~Brendel, K.~Bothe, M.~R. Vogt,
  I.~Kr{\"o}ger, S.~Winter, A.~Schirmacher, S.~Lim, H.~T. Nguyen, and
  D.~MacDonald.
\newblock Uncertainty analysis for the coefficient of band-to-band absorption
  of crystalline silicon.
\newblock \emph{AIP Advances}, 5\penalty0 (6):\penalty0 067168, 2015.

\bibitem[Schubert et~al.(2022)Schubert, Cheung, Williamson, Spyra, and
  Alexander]{SchubertMF2022acsp}
M.~F. Schubert, A.~K.~C. Cheung, I.~A.~D. Williamson, A.~Spyra, and D.~H.
  Alexander.
\newblock Inverse design of photonic devices with strict foundry fabrication
  constraints.
\newblock \emph{ACS Photonics}, 9\penalty0 (7):\penalty0 2327--2336, 2022.

\bibitem[{\v{S}}ehi{\'c} et~al.(2022){\v{S}}ehi{\'c}, Gramfort, Salmon, and
  Nardi]{SehicK2022automlconf}
K.~{\v{S}}ehi{\'c}, A.~Gramfort, J.~Salmon, and L.~Nardi.
\newblock {LassoBench}: A high-dimensional hyperparameter optimization
  benchmark suite for {Lasso}.
\newblock In \emph{Proceedings of the International Conference on Automated
  Machine Learning (AutoML-Conf)}, Baltimore, Maryland, USA, 2022.

\bibitem[Shkondin et~al.(2017)Shkondin, Takayama, Panah, Liu, Larsen, Mar,
  Jensen, and Lavrinenko]{ShkondinE2017ome}
E.~Shkondin, O.~Takayama, M.~E.~A. Panah, P.~Liu, P.~V. Larsen, M.~D. Mar,
  F.~Jensen, and A.~V. Lavrinenko.
\newblock Large-scale high aspect ratio {Al}-doped {ZnO} nanopillars arrays as
  anisotropic metamaterials.
\newblock \emph{Optical Materials Express}, 7\penalty0 (5):\penalty0
  1606--1627, 2017.

\bibitem[Shockley and Queisser(1961)]{ShockleyW1961jap}
W.~Shockley and H.~J. Queisser.
\newblock Detailed balance limit of efficiency of p-n junction solar cells.
\newblock \emph{Journal of Applied Physics}, 32\penalty0 (3):\penalty0
  510--519, 1961.

\bibitem[Siddique et~al.(2015)Siddique, Gomard, and
  H{\"o}lscher]{SiddiqueRH2015nc}
R.~H. Siddique, G.~Gomard, and H.~H{\"o}lscher.
\newblock The role of random nanostructures for the omnidirectional
  anti-reflection properties of the glasswing butterfly.
\newblock \emph{Nature Communications}, 6\penalty0 (1):\penalty0 6909, 2015.

\bibitem[Siefke et~al.(2016)Siefke, Kroker, Pfeiffer, Puffky, Dietrich, Franta,
  Ohl{\'\i}dal, Szeghalmi, Kley, and T{\"u}nnermann]{SiefkeT2016aom}
T.~Siefke, S.~Kroker, K.~Pfeiffer, O.~Puffky, K.~Dietrich, D.~Franta,
  I.~Ohl{\'\i}dal, A.~Szeghalmi, E.-B. Kley, and A.~T{\"u}nnermann.
\newblock Materials pushing the application limits of wire grid polarizers
  further into the deep ultraviolet spectral range.
\newblock \emph{Advanced Optical Materials}, 4\penalty0 (11):\penalty0
  1780--1786, 2016.

\bibitem[So et~al.(2020)So, Badloe, Noh, Bravo-Abad, and Rho]{SoS2020np}
S.~So, T.~Badloe, J.~Noh, J.~Bravo-Abad, and J.~Rho.
\newblock Deep learning enabled inverse design in nanophotonics.
\newblock \emph{Nanophotonics}, 9\penalty0 (5):\penalty0 1041--1057, 2020.

\bibitem[Stavenga et~al.(2006)Stavenga, Foletti, Palasantzas, and
  Arikawa]{StavengaDG2006prsb}
D.~G. Stavenga, S.~Foletti, G.~Palasantzas, and K.~Arikawa.
\newblock Light on the moth-eye corneal nipple array of butterflies.
\newblock \emph{Proceedings of the Royal Society B: Biological Sciences},
  273\penalty0 (1587):\penalty0 661--667, 2006.

\bibitem[Storn and Price(1997)]{StornR1997jgo}
R.~Storn and K.~Price.
\newblock Differential evolution -- a simple and efficient heuristic for global
  optimization over continuous spaces.
\newblock \emph{Journal of Global Optimization}, 11\penalty0 (4):\penalty0
  341--359, 1997.

\bibitem[Taflove(1980)]{TafloveA1980ieeetec}
A.~Taflove.
\newblock Application of the finite-difference time-domain method to sinusoidal
  steady-state electromagnetic-penetration problems.
\newblock \emph{IEEE Transactions on Electromagnetic Compatibility},
  EMC-22\penalty0 (3):\penalty0 191--202, 1980.

\bibitem[Thangamuthu et~al.(2022)Thangamuthu, Kumar, Bishnoi, Bhattoo,
  Krishnan, and Ranu]{ThangamuthuA2022neuripsdb}
A.~Thangamuthu, G.~Kumar, S.~Bishnoi, R.~Bhattoo, N.~M.~A. Krishnan, and
  S.~Ranu.
\newblock Unravelling the performance of physics-informed graph neural networks
  for dynamical systems.
\newblock In \emph{Advances in Neural Information Processing Systems
  (NeurIPS)}, volume~35, pages 3691--3702, New Orleans, Louisiana, USA, 2022.
\newblock Datasets and Benchmarks Track.

\bibitem[Treharne et~al.(2011)Treharne, Seymour-Pierce, Durose, Hutchings,
  Roncallo, and Lane]{TreharneRE2011jopcs}
R.~E. Treharne, A.~Seymour-Pierce, K.~Durose, K.~Hutchings, S.~Roncallo, and
  D.~Lane.
\newblock Optical design and fabrication of fully sputtered {CdTe/CdS} solar
  cells.
\newblock In \emph{Journal of Physics: Conference Series}, volume 286, page
  012038, 2011.

\bibitem[Turner et~al.(2020)Turner, Eriksson, McCourt, Kiili, Laaksonen, Xu,
  and Guyon]{TurnerR2020neuripscd}
R.~Turner, D.~Eriksson, M.~McCourt, J.~Kiili, E.~Laaksonen, Z.~Xu, and
  I.~Guyon.
\newblock {Bayesian} optimization is superior to random search for machine
  learning hyperparameter tuning: Analysis of the black-box optimization
  challenge 2020.
\newblock In \emph{Proceedings of the NeurIPS Competition and Demonstration
  Track}, pages 3--26, Virtual, 2020.

\bibitem[Virtanen et~al.(2020)Virtanen, Gommers, Oliphant, Haberland, Reddy,
  Cournapeau, Burovski, Peterson, Weckesser, Bright, {van der Walt}, Brett,
  Wilson, Millman, Mayorov, Nelson, Jones, Kern, Larson, Carey, Polat, Feng,
  Moore, VanderPlas, Laxalde, Perktold, Cimrman, Henriksen, Quintero, Harris,
  Archibald, Ribeiro, Pedregosa, {van Mulbregt}, and {SciPy 1.0
  Contributors}]{VirtanenP2020nm}
P.~Virtanen, R.~Gommers, T.~E. Oliphant, M.~Haberland, T.~Reddy, D.~Cournapeau,
  E.~Burovski, P.~Peterson, W.~Weckesser, J.~Bright, S.~J. {van der Walt},
  M.~Brett, J.~Wilson, K.~J. Millman, N.~Mayorov, A.~R.~J. Nelson, E.~Jones,
  R.~Kern, E.~Larson, C.~J. Carey, I.~Polat, Y.~Feng, E.~W. Moore,
  J.~VanderPlas, D.~Laxalde, J.~Perktold, R.~Cimrman, I.~Henriksen, E.~A.
  Quintero, C.~R. Harris, A.~M. Archibald, A.~H. Ribeiro, F.~Pedregosa, P.~{van
  Mulbregt}, and {SciPy 1.0 Contributors}.
\newblock {SciPy} 1.0: fundamental algorithms for scientific computing in
  {Python}.
\newblock \emph{Nature Methods}, 17:\penalty0 261--272, 2020.

\bibitem[Wang and Leu(2012{\natexlab{a}})]{WangB2012nt}
B.~Wang and P.~W. Leu.
\newblock Enhanced absorption in silicon nanocone arrays for photovoltaics.
\newblock \emph{Nanotechnology}, 23\penalty0 (19):\penalty0 194003,
  2012{\natexlab{a}}.

\bibitem[Wang and Leu(2012{\natexlab{b}})]{WangB2012ol}
B.~Wang and P.~W. Leu.
\newblock Tunable and selective resonant absorption in vertical nanowires.
\newblock \emph{Optics Letters}, 37\penalty0 (18):\penalty0 3756--3758,
  2012{\natexlab{b}}.

\bibitem[Wang and Leu(2015)]{WangB2015ne}
B.~Wang and P.~W. Leu.
\newblock High index of refraction nanosphere coatings for light trapping in
  crystalline silicon thin film solar cells.
\newblock \emph{Nano Energy}, 13:\penalty0 226--232, 2015.

\bibitem[Wang et~al.(2014)Wang, Stevens, and Leu]{WangB2014oe}
B.~Wang, E.~Stevens, and P.~W. Leu.
\newblock Strong broadband absorption in {GaAs} nanocone and nanowire arrays
  for solar cells.
\newblock \emph{Optics Express}, 22\penalty0 (102):\penalty0 A386--A395, 2014.

\bibitem[Wang et~al.(2016)Wang, Gao, and Leu]{WangB2016ne}
B.~Wang, T.~Gao, and P.~W. Leu.
\newblock Broadband light absorption enhancement in ultrathin film crystalline
  silicon solar cells with high index of refraction nanosphere arrays.
\newblock \emph{Nano Energy}, 19:\penalty0 471--475, 2016.

\bibitem[Wang et~al.(2019)Wang, Ji, Zhang, Zhang, Zhang, Lu, Tan, and
  Guo]{WangH2019acsami}
H.~Wang, C.~Ji, C.~Zhang, Y.~Zhang, Z.~Zhang, Z.~Lu, J.~Tan, and L.~J. Guo.
\newblock Highly transparent and broadband electromagnetic interference
  shielding based on ultrathin doped {Ag} and conducting oxides hybrid film
  structures.
\newblock \emph{ACS Applied Materials \& Interfaces}, 11\penalty0
  (12):\penalty0 11782--11791, 2019.

\bibitem[Wilson and Hutley(1982)]{WilsonSJ1982oaijo}
S.~J. Wilson and M.~C. Hutley.
\newblock The optical properties of `moth eye' antireflection surfaces.
\newblock \emph{Optica Acta: International Journal of Optics}, 29\penalty0
  (7):\penalty0 993--1009, 1982.

\bibitem[Wu et~al.(2018)Wu, Ramsundar, Feinberg, Gomes, Geniesse, Pappu,
  Leswing, and Pande]{WuZ2018cs}
Z.~Wu, B.~Ramsundar, E.~N. Feinberg, J.~Gomes, C.~Geniesse, A.~S. Pappu,
  K.~Leswing, and V.~Pande.
\newblock {MoleculeNet}: a benchmark for molecular machine learning.
\newblock \emph{Chemical Science}, 9\penalty0 (2):\penalty0 513--530, 2018.

\bibitem[Yablonovitch and Cody(1982)]{YablonovitchE1982ieeeted}
E.~Yablonovitch and G.~D. Cody.
\newblock Intensity enhancement in textured optical sheets for solar cells.
\newblock \emph{IEEE Transactions on Electron Devices}, 29\penalty0
  (2):\penalty0 300--305, 1982.

\bibitem[Yee(1966)]{YeeKS1966ieeeap}
K.~S. Yee.
\newblock Numerical solution of initial boundary value problems involving
  {Maxwell's} equations in isotropic media.
\newblock \emph{IEEE Transactions on Antennas and Propagation}, 14\penalty0
  (3):\penalty0 302--307, 1966.

\bibitem[Ying et~al.(2019)Ying, Klein, Christiansen, Real, Murphy, and
  Hutter]{YingC2019icml}
C.~Ying, A.~Klein, E.~Christiansen, E.~Real, K.~Murphy, and F.~Hutter.
\newblock {NAS-Bench-101}: Towards reproducible neural architecture search.
\newblock In \emph{Proceedings of the International Conference on Machine
  Learning (ICML)}, pages 7105--7114, Long Beach, California, USA, 2019.

\end{thebibliography}

\newpage
\appendix

\renewcommand{\thefigure}{S\arabic{figure}}
\renewcommand{\thetable}{S\arabic{table}}
\renewcommand{\theequation}{S\arabic{equation}}
\renewcommand{\thefootnote}{S\arabic{footnote}}
\setcounter{table}{0}
\setcounter{figure}{0}
\setcounter{equation}{0}
\setcounter{footnote}{0}

\section{Electromagnetic Wave of Light}

\begin{figure}[ht]
    \centering
    \colorlet{Ecol}{magenta!90!black}
\colorlet{Bcol}{blue!80!black}

\begin{tikzpicture}[
    x=(-15:0.9), y=(90:0.9), z=(-150:1.1),
    line cap=round, line join=round,
    axis/.style={black, thick, ->},
    vector/.style={>=stealth, ->}
]
    \normalsize
    \def\A{1.5}
    \def\nNodes{3}
    \def\nVectorsPerNode{6}
    \def\N{\nNodes * 40}
    \def\xmax{\nNodes * pi / 2 * 1.01}
    \pgfmathsetmacro\nVectors{
        (\nVectorsPerNode + 1) * \nNodes
    }
    
    \def\drawENode{
    \draw[
        Ecol,
        very thick,
        variable=\t,
        domain=\iOffset * pi / 2:(\iOffset + 1) * pi / 2 * 1.01,
        samples=40
    ]
        plot(\t, {\A * sin(\t * 360 / pi)}, 0);
        \foreach \k [evaluate={\t=\k * pi / 2 / (\nVectorsPerNode + 1);
        \angle=\k * 90 / (\nVectorsPerNode + 1);}]
        in {1, ..., \nVectorsPerNode}{
            \draw[vector, thick, Ecol!50] (\iOffset * pi / 2 + \t, 0, 0) -- ++(0, {\A * sin(2 * \angle + \iOffset * 180)}, 0);
        }
    }
    \def\drawBNode{
    \draw[
        Bcol,
        very thick,
        variable=\t,
        domain=\iOffset * pi / 2:(\iOffset + 1) * pi / 2 * 1.01,
        samples=40
    ]
        plot(\t, 0, {\A * sin(\t * 360 / pi)});
        \foreach \k [evaluate={\t=\k * pi / 2 / (\nVectorsPerNode + 1);
        \angle=\k * 90 / (\nVectorsPerNode + 1);}]
        in {1,...,\nVectorsPerNode}{
            \draw[vector, thick, Bcol!50] (\iOffset * pi / 2 + \t, 0, 0) -- ++(0, 0, {\A * sin(2 * \angle + \iOffset * 180)});
    }
    }
    
    \draw[axis] (0, 0, 0) -- ++(\xmax * 1.1, 0, 0) node[right] {$z$};
    \draw[axis] (0, -\A * 1, 0) -- (0, \A * 1, 0) node[right] {$x$};
    \draw[axis] (0, 0, -\A * 1) -- (0, 0, \A * 1) node[left] {$y$};
    
    \foreach \iNode [evaluate={\iOffset=\iNode-1;}] in {1, ..., \nNodes}{
        \ifodd
            \iNode \drawBNode \drawENode
        \else
            \drawENode \drawBNode
        \fi
    }
\end{tikzpicture}
    \caption{Electromagnetic wave of light. Magenta and blue waves correspond to electric and magnetic fields, respectively.}
    \label{fig:em_wave}
\end{figure}
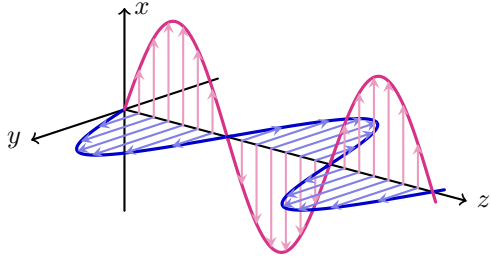

\figref{fig:em_wave} shows the electromagnetic wave of light,
which is composed of electric and magnetic fields.
The oscillations of these two fields are perpendicular
to each other and the direction of wave propagation.

\section{Reflection, Absorption, and Transmission of Light}

\begin{figure}[ht]
    \centering
    \includegraphics[width=0.35\textwidth]{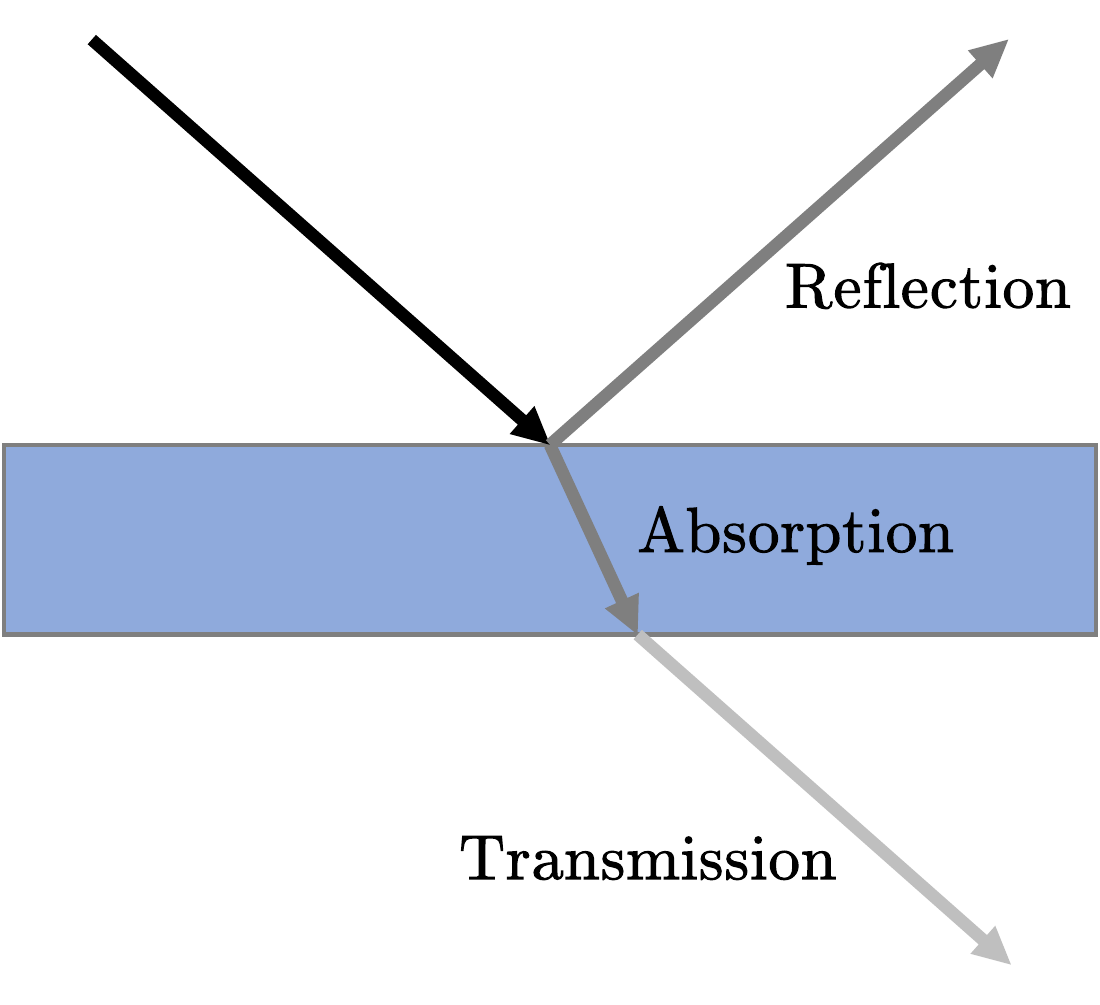}
    \caption{Light interacting with some material.}
    \label{fig:ref_abs_tra}
\end{figure}

\figref{fig:ref_abs_tra} depicts the properties of light,
where light interacts with some material.
As in~\eqref{eqn:sum_optical_properties},
the sum of reflectance, absorbance, and transmittance is one.

\section{Solar Reflection}
\label{sec:solar_reflection}

Solar reflection $R_{\textrm{solar}}$ over $d$ parameters $x_1, x_2, \ldots, x_d$ is defined as the following:
\begin{equation}
    R_{\textrm{solar}}(x_1, x_2, \ldots, x_d) = \frac{\int b_s(\lambda) R(x_1, x_2, \ldots, x_d; \lambda) \ \mathrm{d}\lambda}{\int b_s(\lambda) \ \mathrm{d}\lambda},
\end{equation}
where $R(x_1, x_2, \ldots, x_d; \lambda)$ is a reflection function of wavelength $\lambda$
and $b_s (\lambda)$ is the photon flux density of the AM1.5 spectrum~\citep{ASTMI2020SolarAM15} at wavelength $\lambda$.

\section{Solar Absorption and Ultimate Efficiency}
\label{sec:sa_ue}

Solar absorption $A_{\textrm{solar}}$ and ultimate efficiency $\eta_{\textrm{ue}}$ over $d$ parameters $x_1, x_2, \ldots, x_d$ are defined as the following:
\begin{align}
    A_{\textrm{solar}}(x_1, x_2, \ldots, x_d) &= \frac{\int_{E_g}^{\infty} b_s(E) A(x_1, x_2, \ldots, x_d; E) \ \mathrm{d}E}{\int_{0}^{\infty} b_s(E) \ \mathrm{d}E},\\
    \eta_{\textrm{ue}}(x_1, x_2, \ldots, x_d) &= \frac{\int_{E_g}^{\infty} I(E) A(x_1, x_2, \ldots, x_d; E) \frac{E_g}{E} \ \mathrm{d}E}{\int_{0}^{\infty} I(E) \ \mathrm{d}E},
\end{align}
where $b_s(E)$ is a photon flux density,
$I(E)$ is the irradiance,
and $E_g$ is the materials' energy band gap.
Note that ultimate efficiency assumes that the temperature of the solar cell is 0 K,
and $\int_0^{\infty} I(E) \ \mathrm{d}E =$ 1,000 W/m$^2$ for the AM1.5 global spectrum~\citep{ASTMI2020SolarAM15}.
Each material can only absorb light with energy above its band gap $E_g$
so the integration in the numerator is only for photons with energy greater than $E_g$.
$E_g =$ 1.12 eV ($\lambda =$ 1107 nm), for crystalline silicon (\ce{cSi}),
$E_g =$ 1.43 eV ($\lambda =$ 867 nm) for gallium arsenide (\ce{GaAs}),
and $E_g =$ 1.51 eV ($\lambda =$ 821 nm) for a perovskite structure of methylammonium lead iodide (\ce{CH3NH3PbI3}).

Consequently, maximizing $A_{\textrm{solar}}$ is equivalent to maximizing $\eta_{\textrm{ue}}$:
\begin{equation}
    \max A_{\textrm{solar}}(x_1, x_2, \ldots, x_d) = \max \eta_{\textrm{ue}}(x_1, x_2, \ldots, x_d).
\end{equation}

\section{Material Properties}
\label{sec:materials}

\begin{figure}[ht]
    \centering
    \begin{subfigure}[b]{0.245\textwidth}
        \centering
        \includegraphics[width=0.999\textwidth]{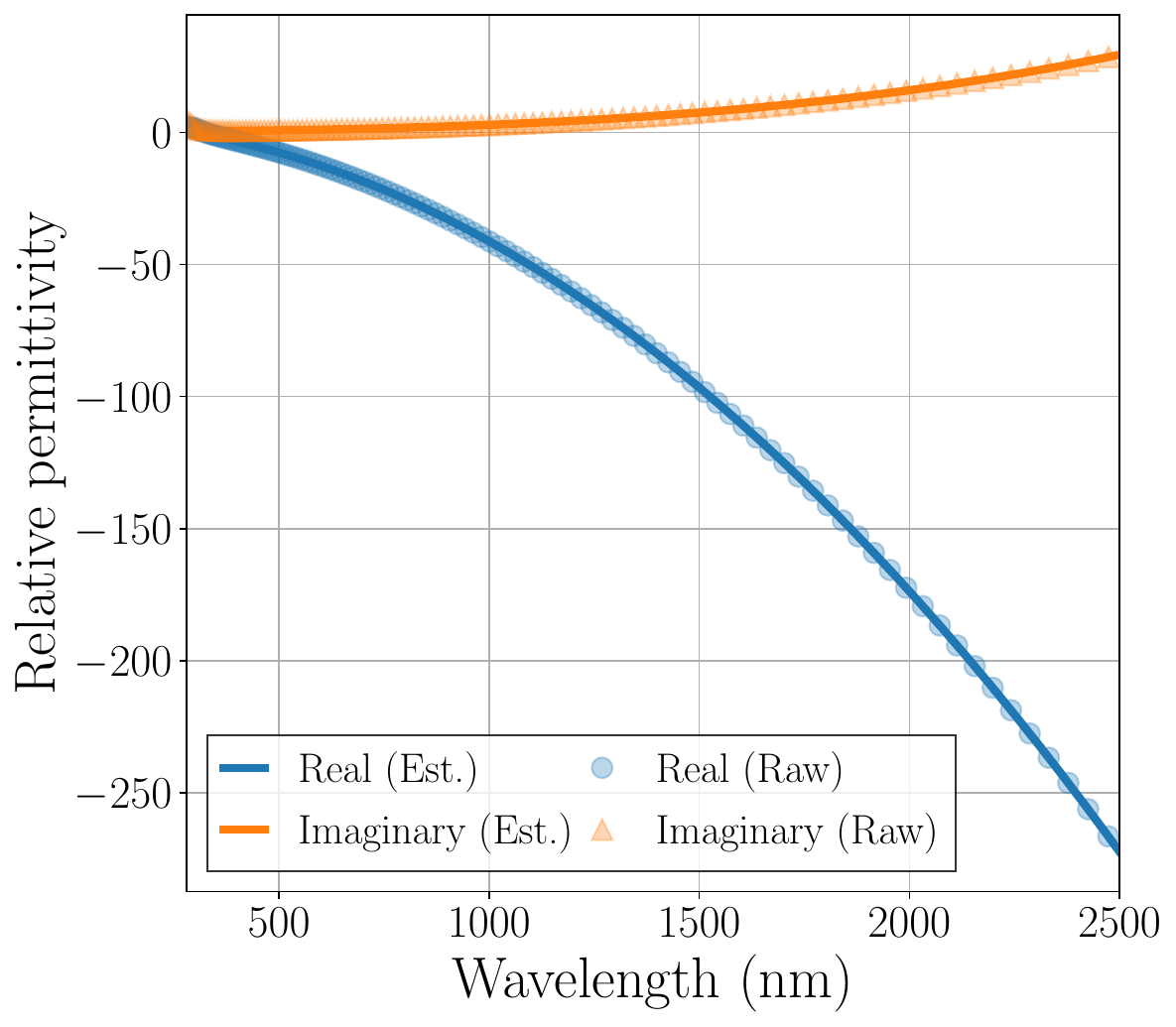}
        \caption{\ce{Ag}}
    \end{subfigure}
    \begin{subfigure}[b]{0.245\textwidth}
        \centering
        \includegraphics[width=0.999\textwidth]{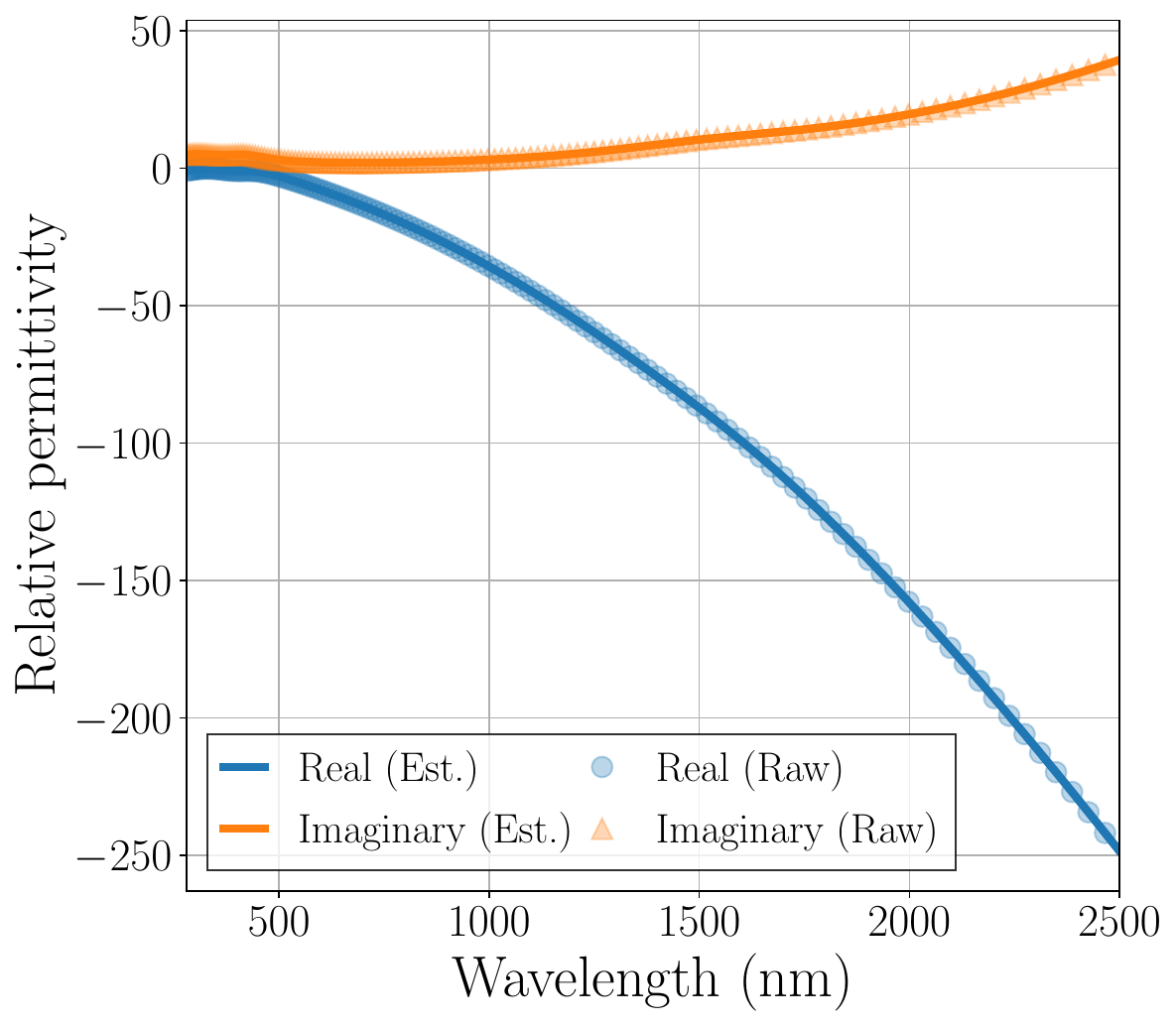}
        \caption{\ce{Au}}
    \end{subfigure}
    \begin{subfigure}[b]{0.245\textwidth}
        \centering
        \includegraphics[width=0.999\textwidth]{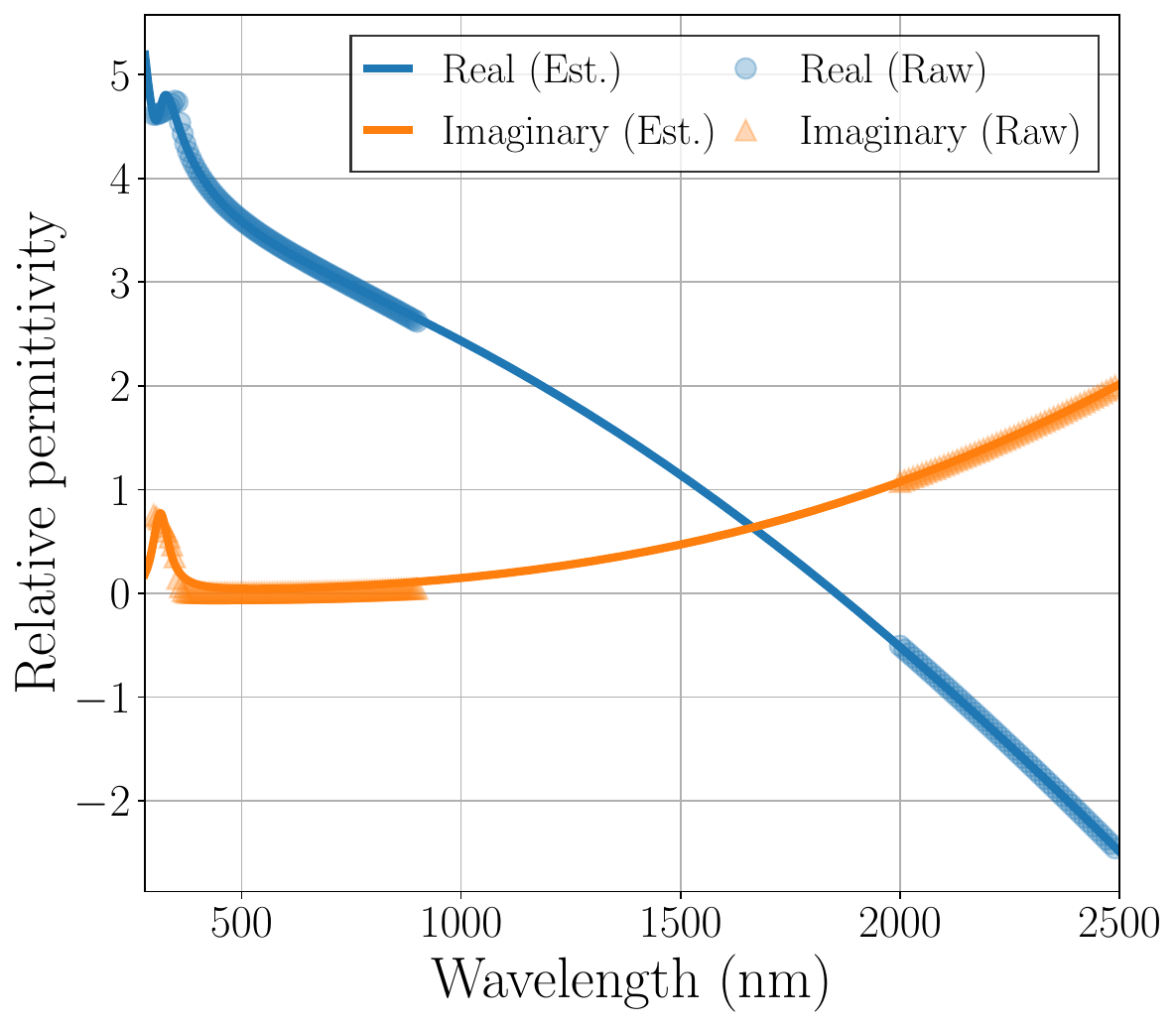}
        \caption{\ce{AZO}}
        \label{fig:azo}
    \end{subfigure}
    \begin{subfigure}[b]{0.245\textwidth}
        \centering
        \includegraphics[width=0.999\textwidth]{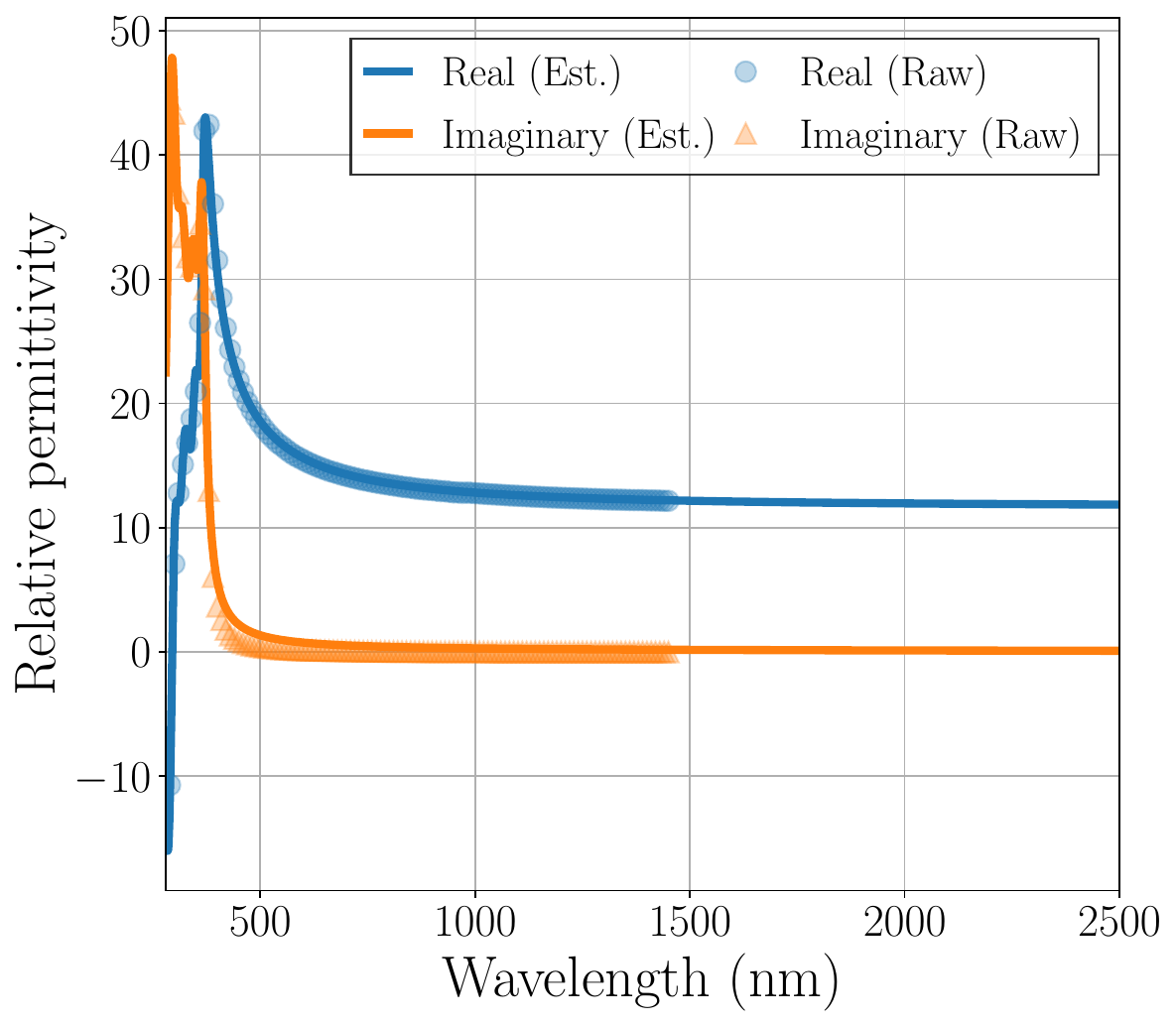}
        \caption{\ce{cSi}}
    \end{subfigure}
    \begin{subfigure}[b]{0.245\textwidth}
        \centering
        \includegraphics[width=0.999\textwidth]{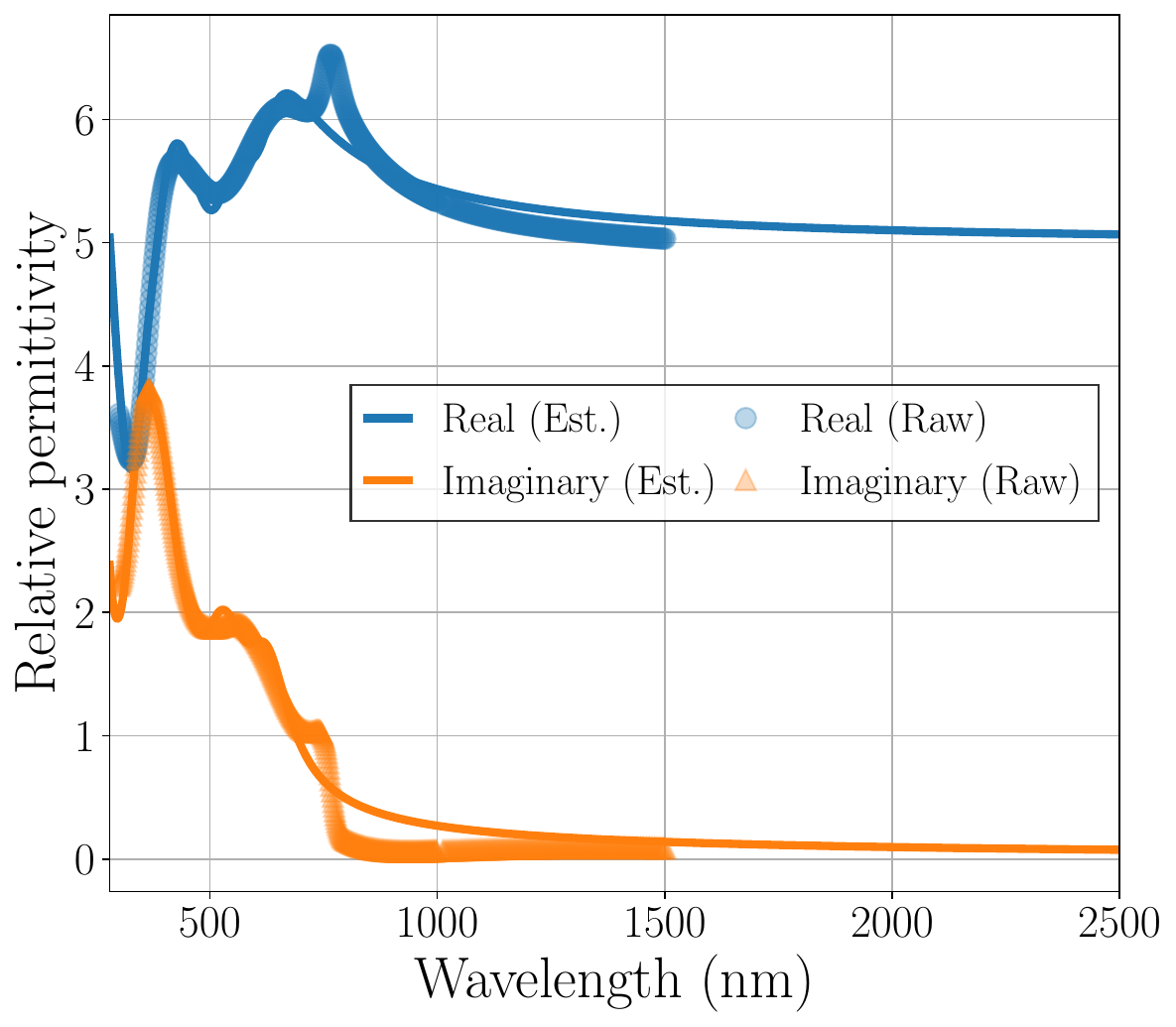}
        \caption{\ce{CH3NH3PbI3}}
    \end{subfigure}
    \begin{subfigure}[b]{0.245\textwidth}
        \centering
        \includegraphics[width=0.999\textwidth]{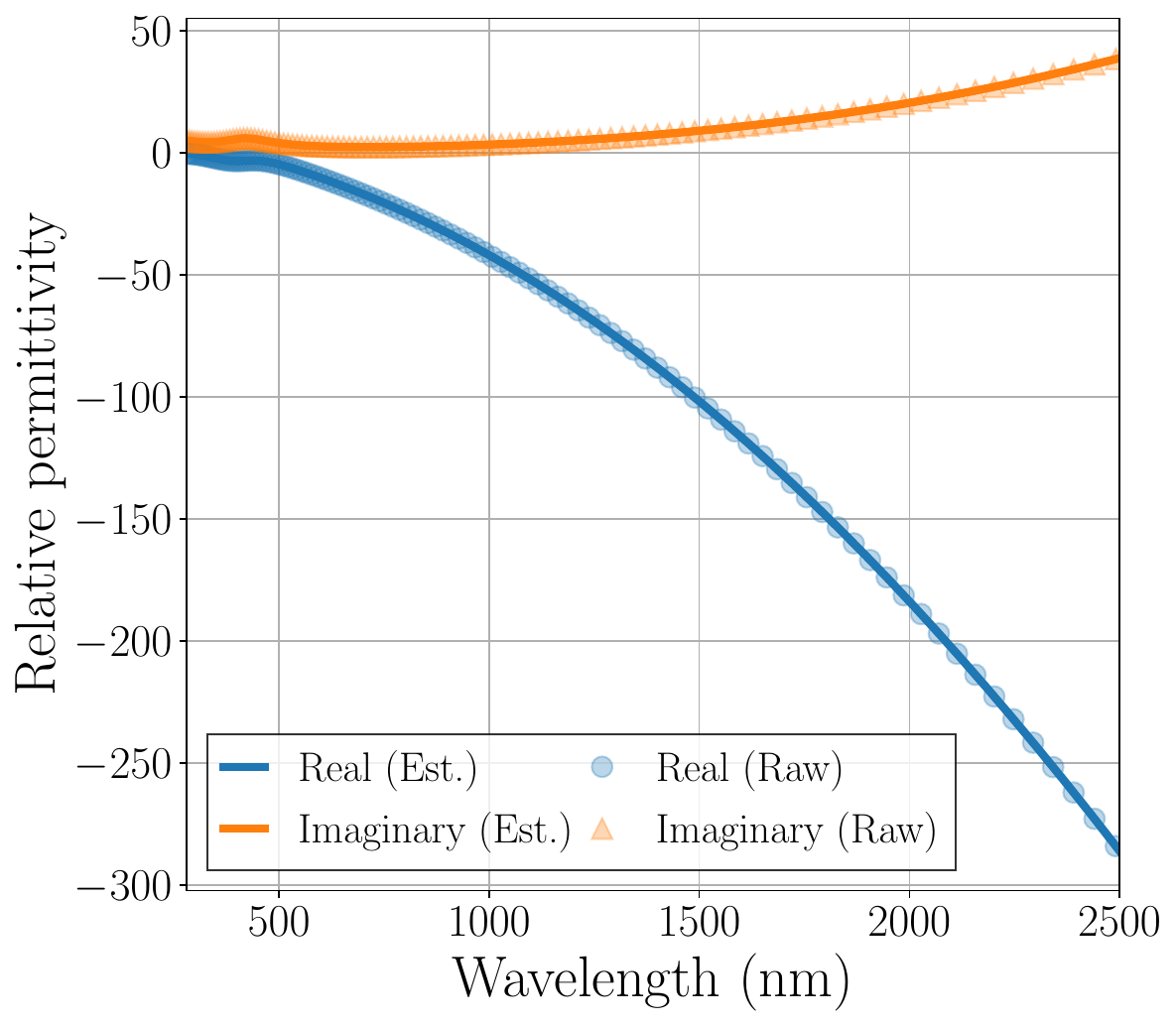}
        \caption{\ce{Cu}}
    \end{subfigure}
    \begin{subfigure}[b]{0.245\textwidth}
        \centering
        \includegraphics[width=0.999\textwidth]{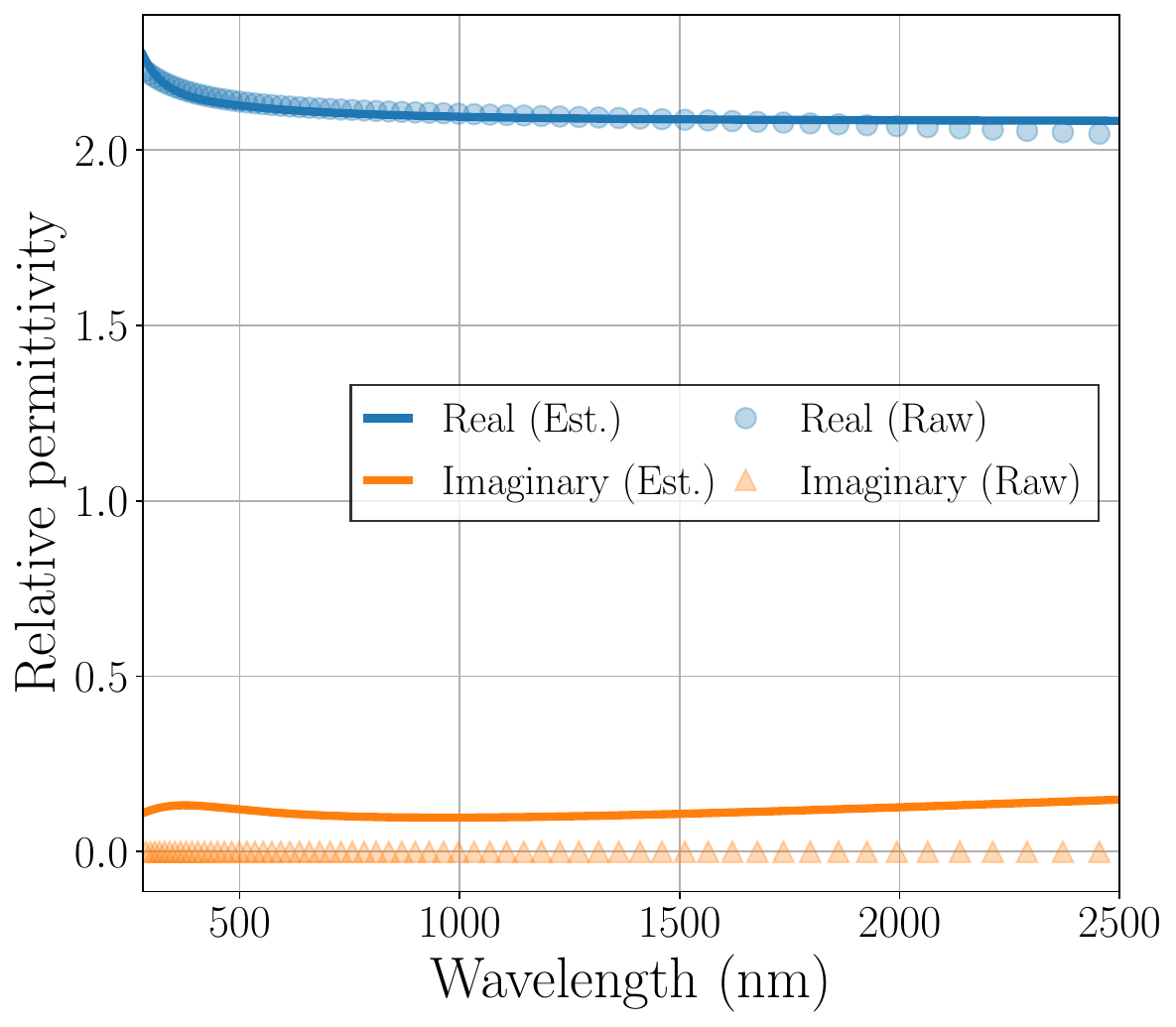}
        \caption{Fused silica}
        \label{fig:fused_silica}
    \end{subfigure}
    \begin{subfigure}[b]{0.245\textwidth}
        \includegraphics[width=0.999\textwidth]{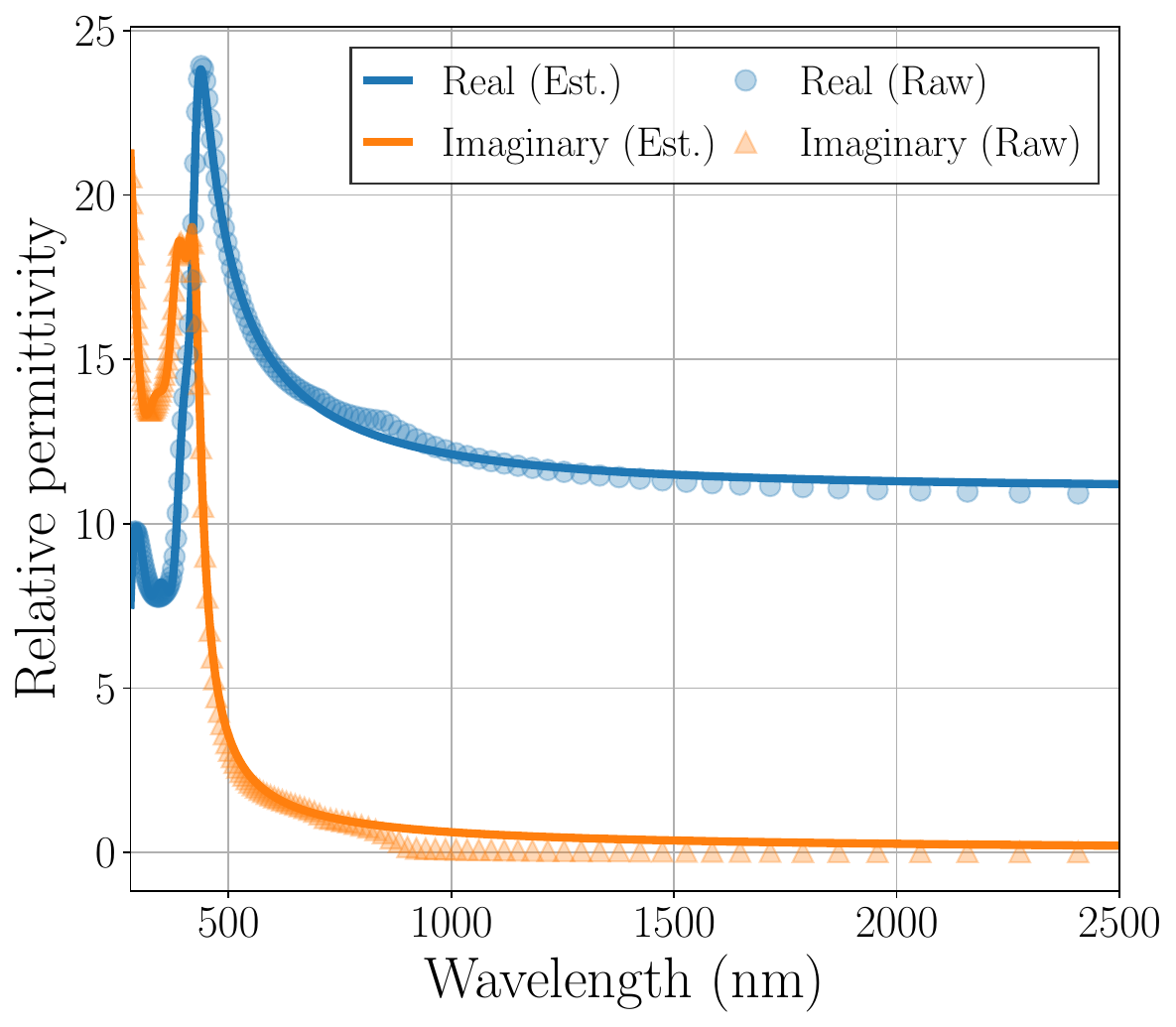}
        \caption{\ce{GaAs}}
    \end{subfigure}
    \begin{subfigure}[b]{0.245\textwidth}
        \centering
        \includegraphics[width=0.999\textwidth]{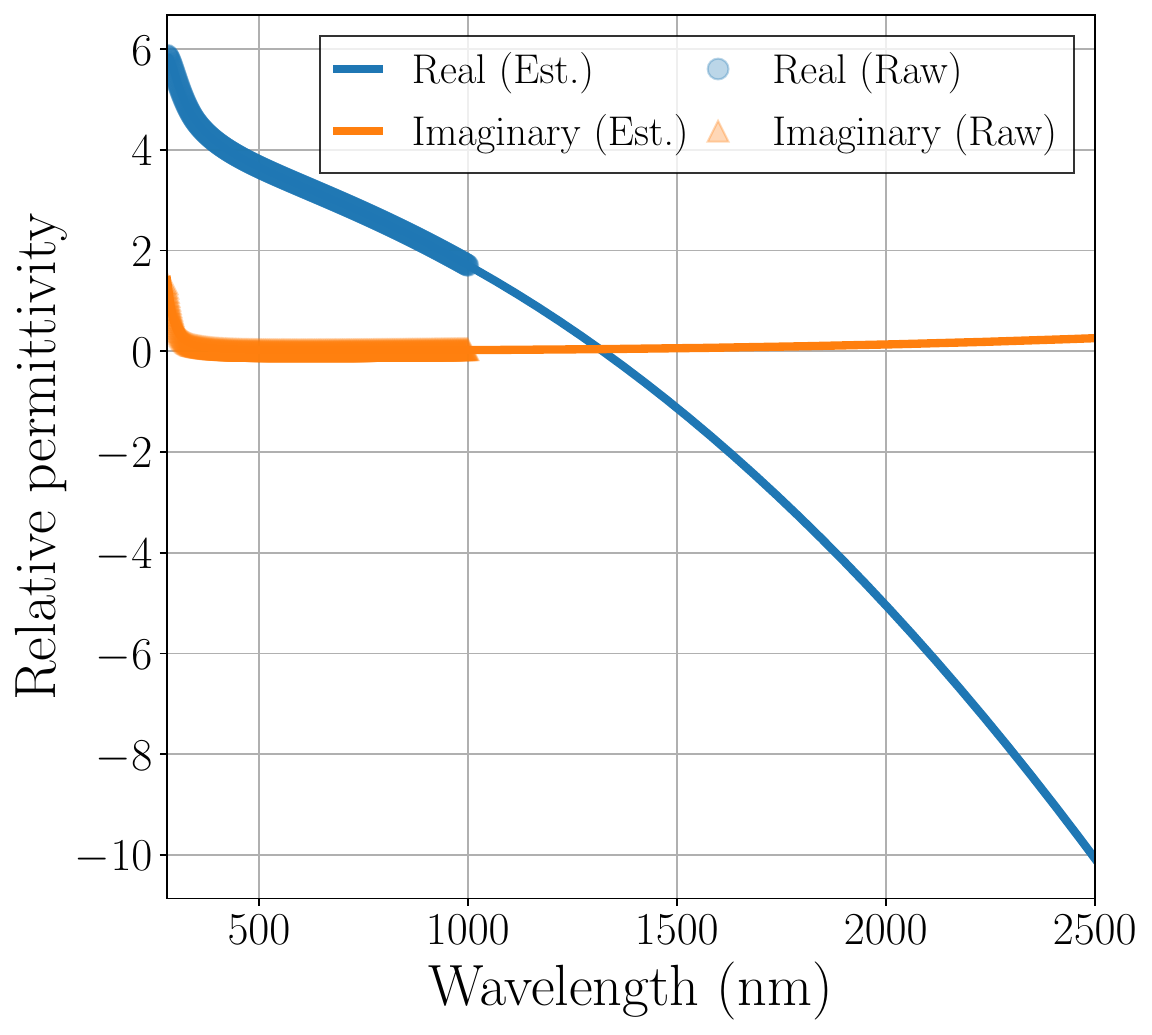}
        \caption{\ce{ITO}}
    \end{subfigure}
    \begin{subfigure}[b]{0.245\textwidth}
        \centering
        \includegraphics[width=0.999\textwidth]{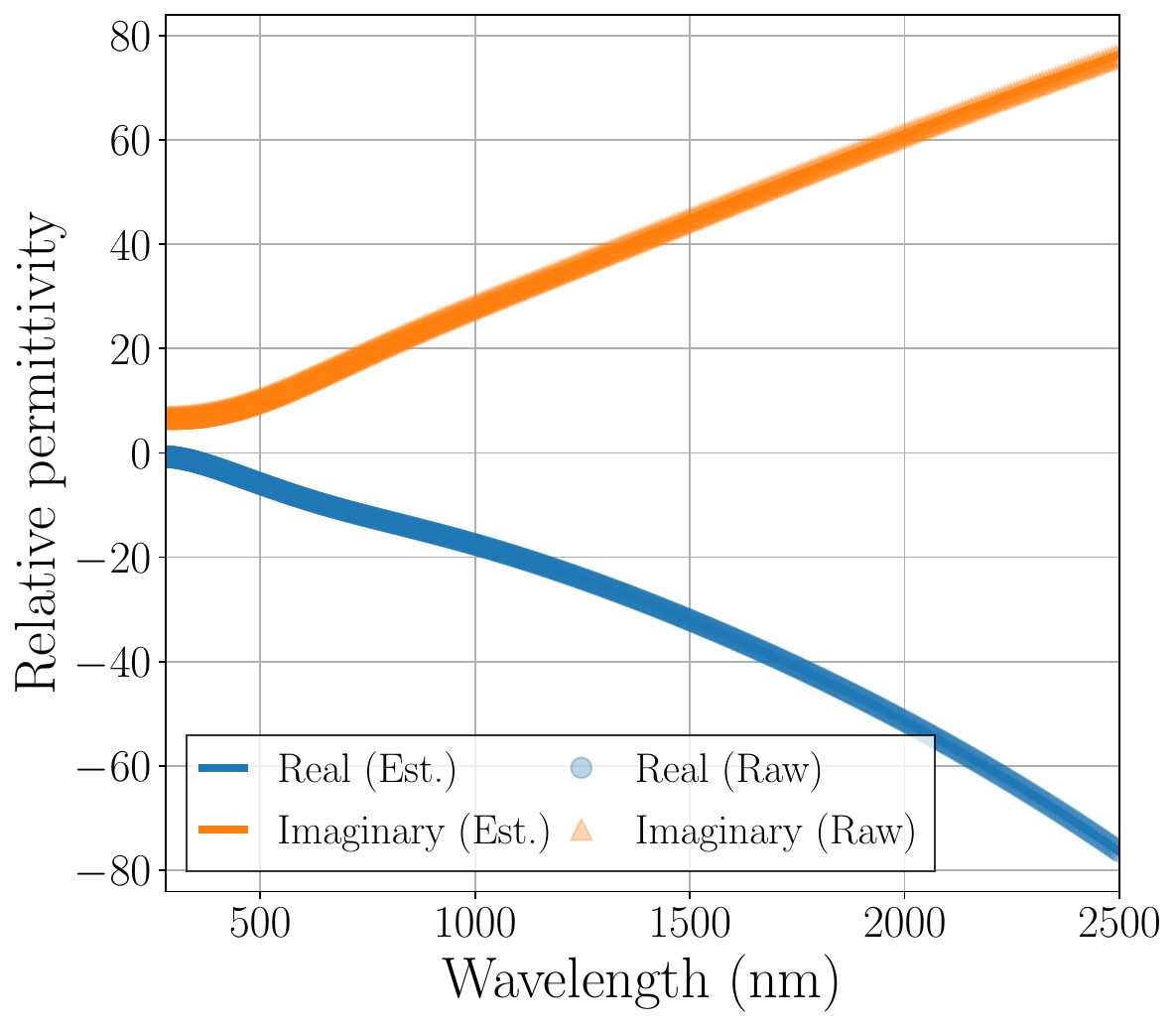}
        \caption{\ce{Ni}}
    \end{subfigure}
    \begin{subfigure}[b]{0.245\textwidth}
        \centering
        \includegraphics[width=0.999\textwidth]{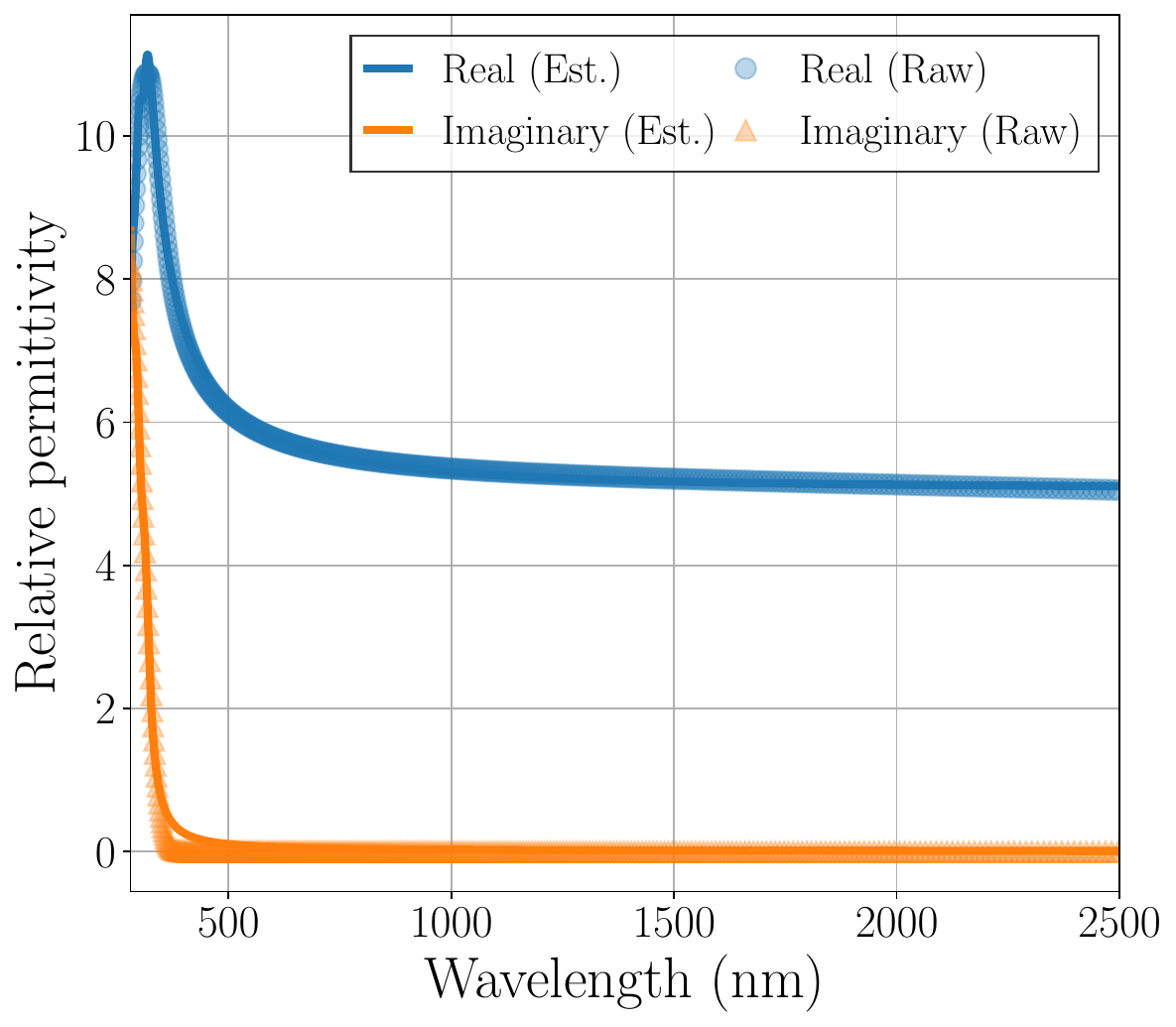}
        \caption{\ce{TiO2}}
    \end{subfigure}
    \begin{subfigure}[b]{0.245\textwidth}
        \centering
        \includegraphics[width=0.999\textwidth]{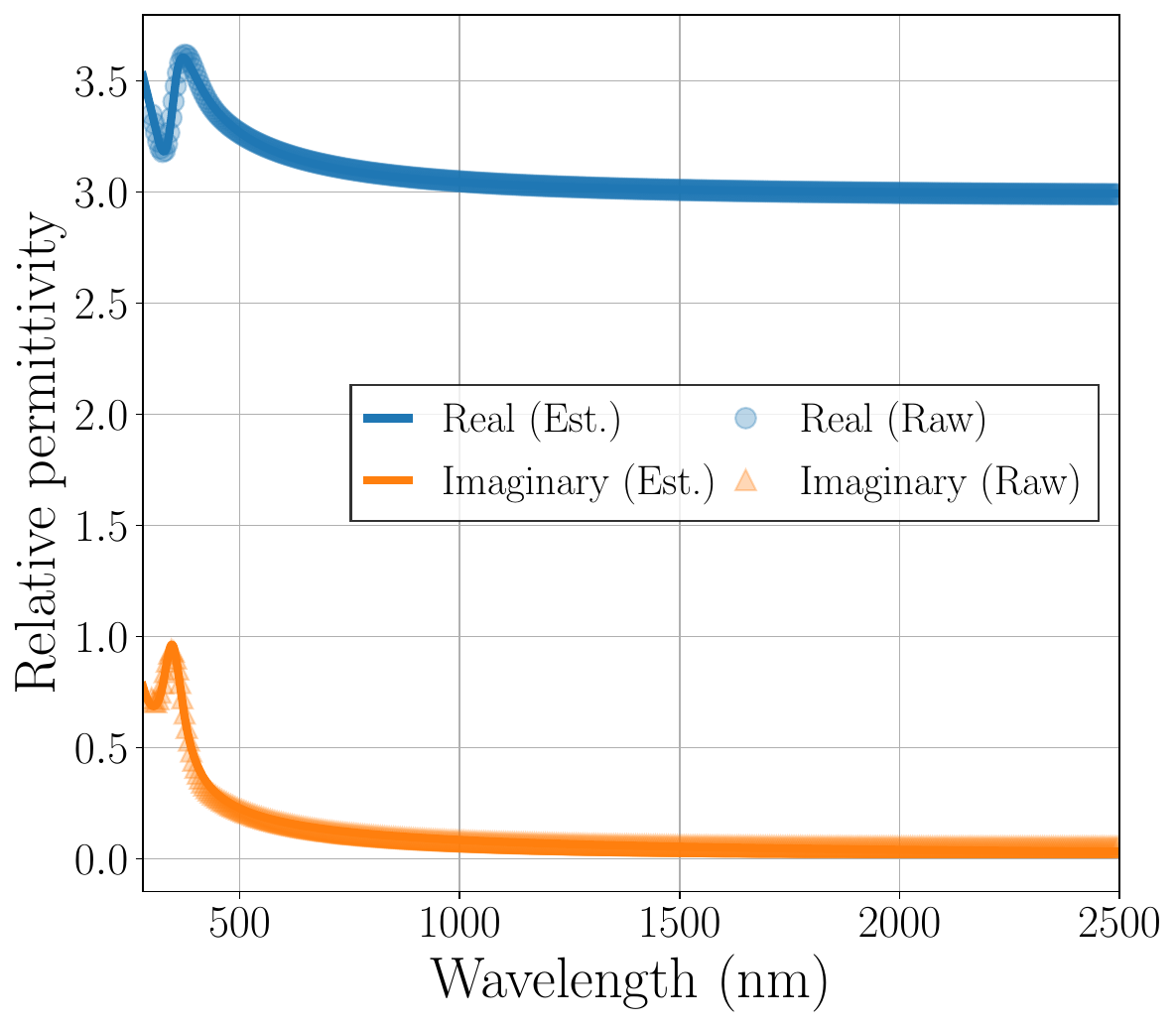}
        \caption{\ce{ZnO}}
    \end{subfigure}
    \caption{Relative permittivities of various materials versus different wavelengths. We utilize the raw values of complex permittivity and employ a gradient-based optimization technique to fit them to the Drude-Lorentz susceptibility model. The estimated complex permittivities presented here cover the solar spectrum range from 280 nm to 2500 nm.}
    \label{fig:materials}
\end{figure}

In~\figref{fig:materials},
we report relative permittivities across different wavelengths for the materials employed in this work.
The range of the solar spectrum we focus on spans from 280 nm to 2500 nm. To derive permittivity values throughout this range, we apply a gradient-based optimization method to fit the raw complex permittivity data to the Drude-Lorentz susceptibility model, following the methodology outlined in the Meep documentation~\citep{OskooiAF2010cpc}.

For \ce{AZO}, we combine two different sets of raw permittivity data to create a comprehensive dataset. This approach aims to enhance the accuracy of the fitted model across the entire solar spectrum. The effectiveness of this approach and the results of the fitted model are illustrated in~\figref{fig:azo}.

With regard to fused silica, a minor offset is intentionally introduced to the optical extinction coefficient of its complex refractive index.
This adjustment is necessary to stabilize the simulations, especially those at low and medium fidelity levels,
which are otherwise susceptible to divergence or failure. In this research, we choose an offset value of 0.04, and the associated results and adjustments are detailed in~\figref{fig:fused_silica}.

The materials examined in this paper include silver (\ce{Ag})~\citep{RakicAD1998ao},
gold (\ce{Au})~\citep{RakicAD1998ao},
aluminum-doped zinc oxide (\ce{AZO})~\citep{TreharneRE2011jopcs,ShkondinE2017ome},
crystalline silicon (\ce{cSi})~\citep{SchinkeC2015aipa},
a perovskite structure of methylammonium lead iodide (\ce{CH3NH3PbI3})~\citep{PhillipsLJ2015db},
copper (\ce{Cu})~\citep{RakicAD1998ao},
fused silica~\citep{MalitsonIH1965josa},
gallium arsenide (\ce{GaAs})~\citep{RakicAD1996jap},
indium tin oxide (\ce{ITO})~\citep{KonigTAF2014acsn},
nickel (\ce{Ni})~\citep{RakicAD1998ao},
titanium dioxide (\ce{TiO2})~\citep{SiefkeT2016aom},
and zinc oxide (\ce{ZnO})~\citep{AguilarO2019ome}.
Note that the respective references of the materials mentioned above indicate the sources of the raw data of the relative permittivities.

\section{Details of Simulation Cell Sizes}
\label{sec:details_cell_sizes}

\begin{table}[ht]
    \caption{Details of the simulation cell sizes declared for our simulations. A unit depth $m$ determines the spacing of geometries. All values are in nanometers. For simplicity, denote that $t^\dagger = t_2 + \max(2t_1, 2t_3) + \max(2h_1, 2h_2)$.}
    \label{tab:simulation_cell_sizes}
    \begin{center}
    \small
    \setlength{\tabcolsep}{1pt}
    \begin{tabular}{lcc}
        \toprule
        \textbf{Structure} & \textbf{Two-Dimensional Structure} & \textbf{Three-Dimensional Structure} \\
        \midrule
        Three-layer film & $(10, t_2 + \max(2t_1, 2t_3) + 6m)$ & $(10, 10, t_2 + \max(2t_1, 2t_3) + 6m)$ \\
        Anti-reflective nanocones & $(2r, 2h + 6m)$ & $(2r, 2r, 2h + 6m)$ \\
        Vertical nanowires & $(p, h + 6m)$ & $(p, p, h + 6m)$ \\
        Close-packed nanospheres & $(2r, \max(4r, 2t) + 6m)$ & $(2r, 2r\sqrt{3}, \max(4r, 2t) + 6m)$ \\
        Three-layer film with nanocones & $(\max(2r_1, 2r_2), t^\dagger + 6m)$ & $(\max(2r_1, 2r_2), \max(2r_1, 2r_2), t^\dagger + 6m)$ \\
        Combinatorial system & $(p, t + 6m)$ & $(p, p, t + 6m)$ \\
        \bottomrule
    \end{tabular}
    \end{center}
\end{table}

\tabref{tab:simulation_cell_sizes} represents the sizes of the simulation cells employed for nanophotonic structures.
A unit depth $m$ determines spacing between geometries such as the light source, transmission monitor, reflection monitor, and photonic structures.
More precisely,
supposing that we define the depth of PML as $m$,
distance between the light source and the reflection monitor is set as $m$
and the size of the simulation cell is also dependent on $m$;
see~\tabref{tab:simulation_cell_sizes} for that dependency.
Unless otherwise noted,
$m =$ 50 nm for our simulations.

\section{Details of Datasets}
\label{sec:details_datasets}

The number of possible configurations shown in~\tabref{tab:statistics} is determined by considering the search space presented in~\tabref{tab:search_spaces} and increment for the corresponding structure.
For example,
suppose that we are given two parameters where the first parameter can vary between 3 and 5, and the second between 10 and 14, increasing in steps of 1.
In this example,
the first parameter has three possible values \{3, 4, 5\},
and the second has five values \{10, 11, 12, 13, 14\}.
This results in a total of 15 unique configurations that can be derived from these parameters.

\section{Additional Dataset Visualization}
\label{sec:additional_visualization}

\begin{figure}[p]
    \centering
    \begin{subfigure}[b]{\textwidth}
        \centering
        \includegraphics[width=0.31\textwidth]{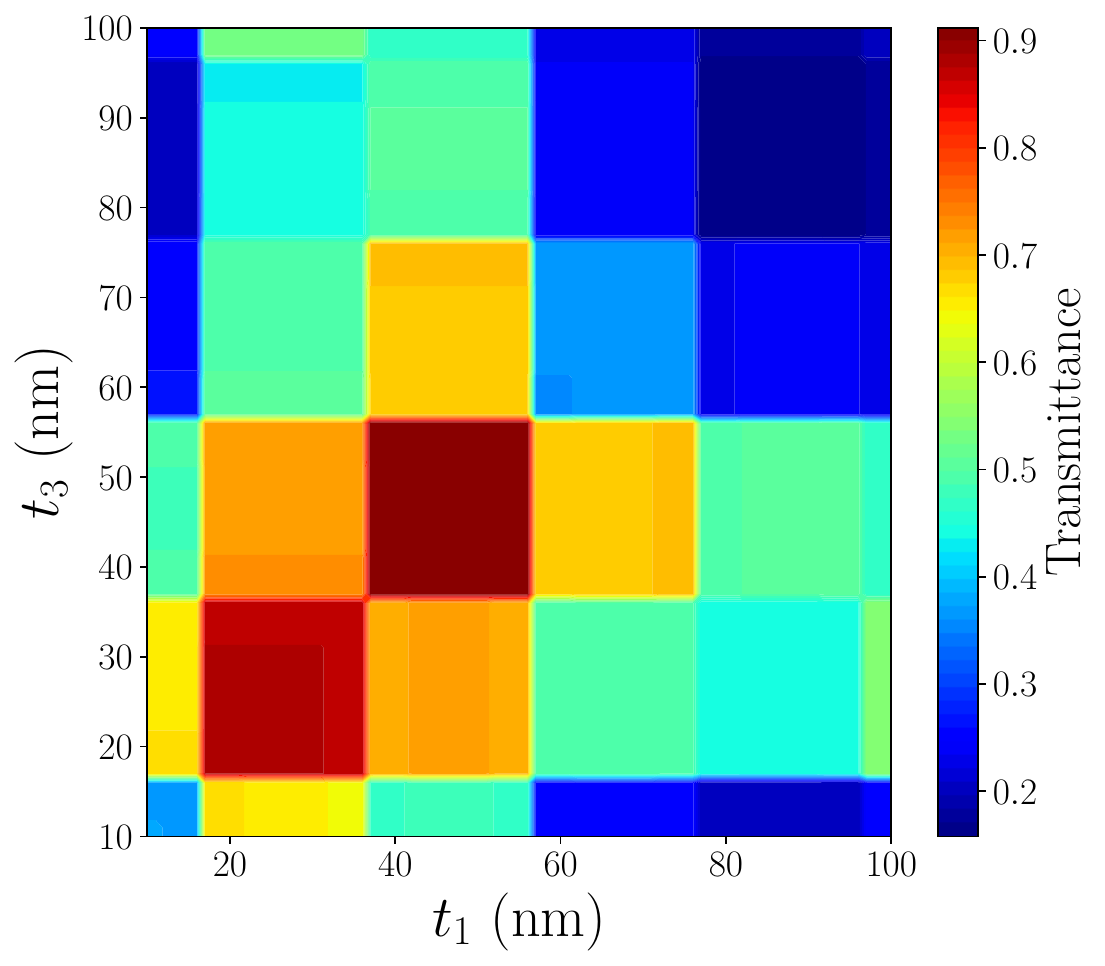}
        \includegraphics[width=0.31\textwidth]{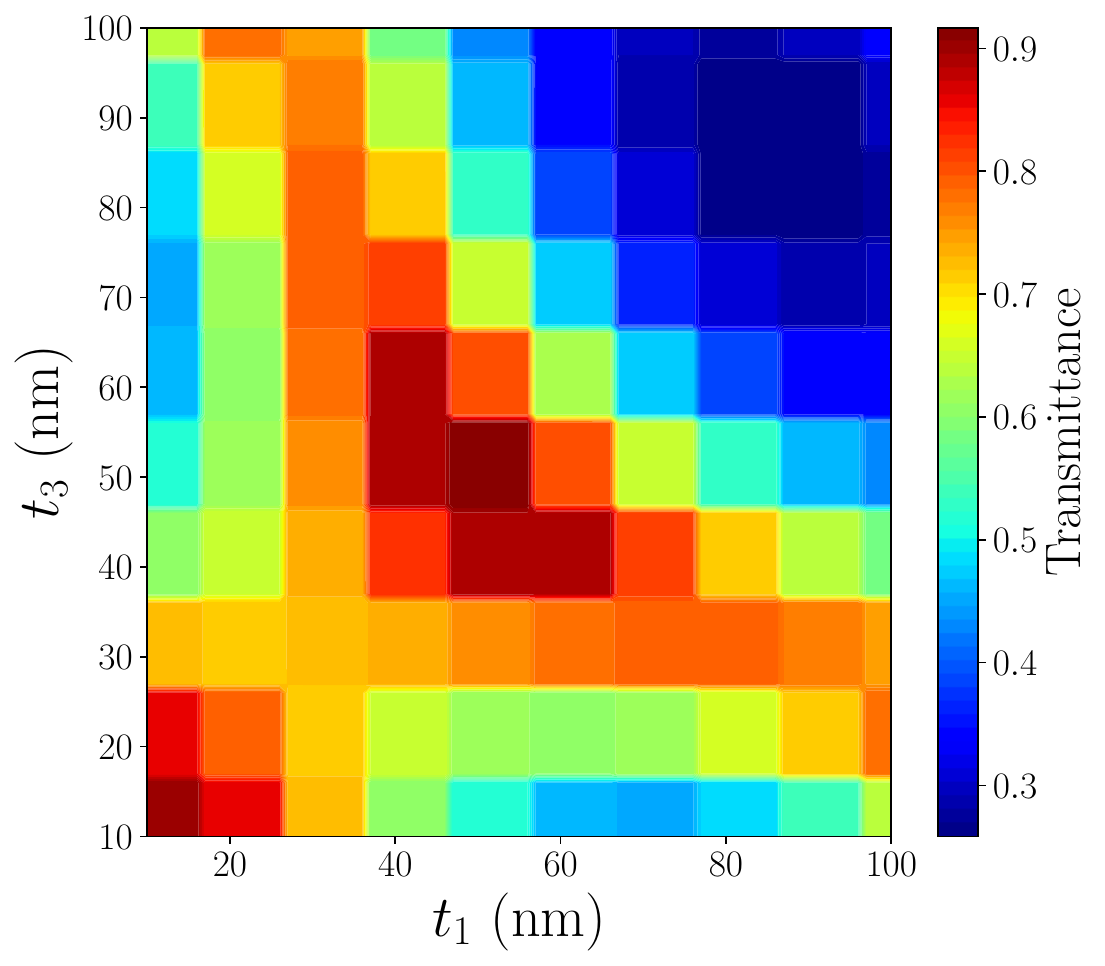}
        \includegraphics[width=0.31\textwidth]{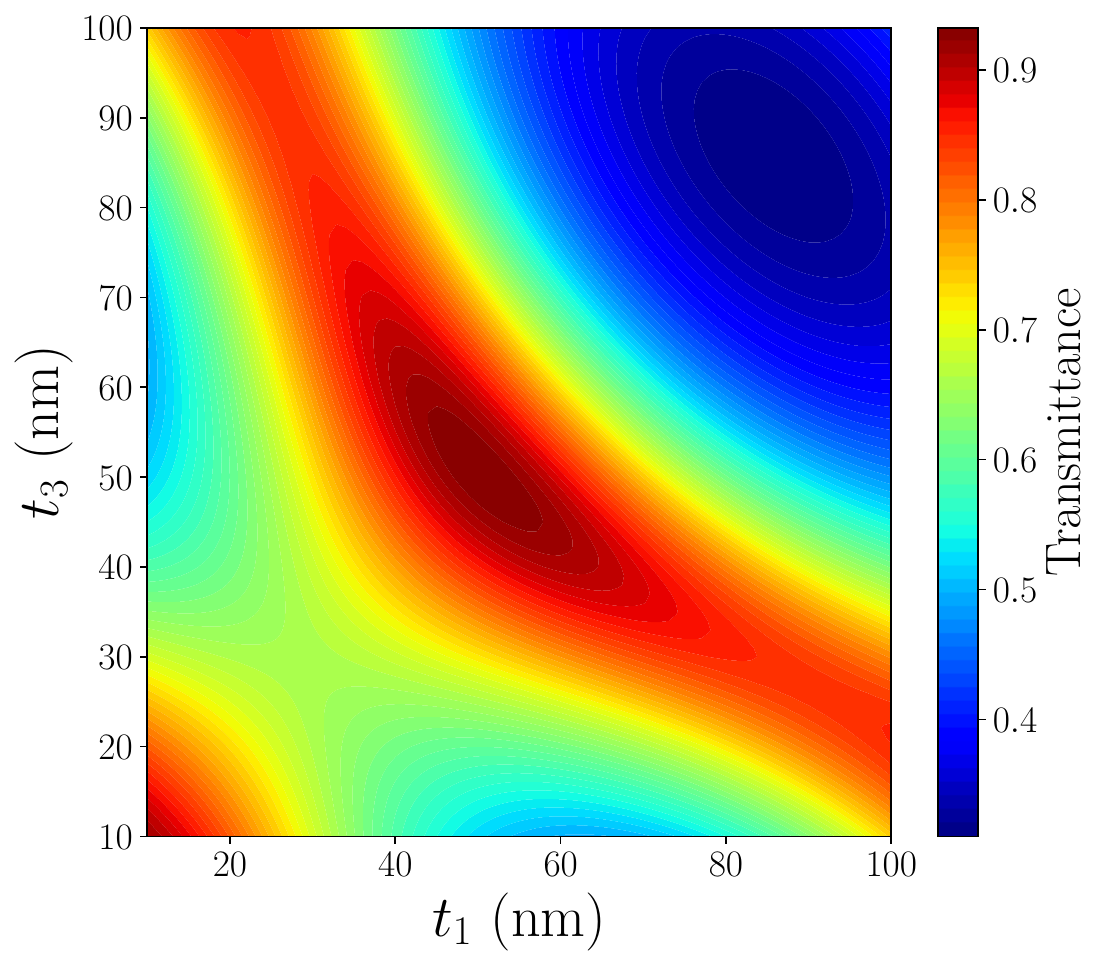}
        \caption{$t_2 = 7$}
    \end{subfigure}
    \begin{subfigure}[b]{\textwidth}
        \centering
        \includegraphics[width=0.31\textwidth]{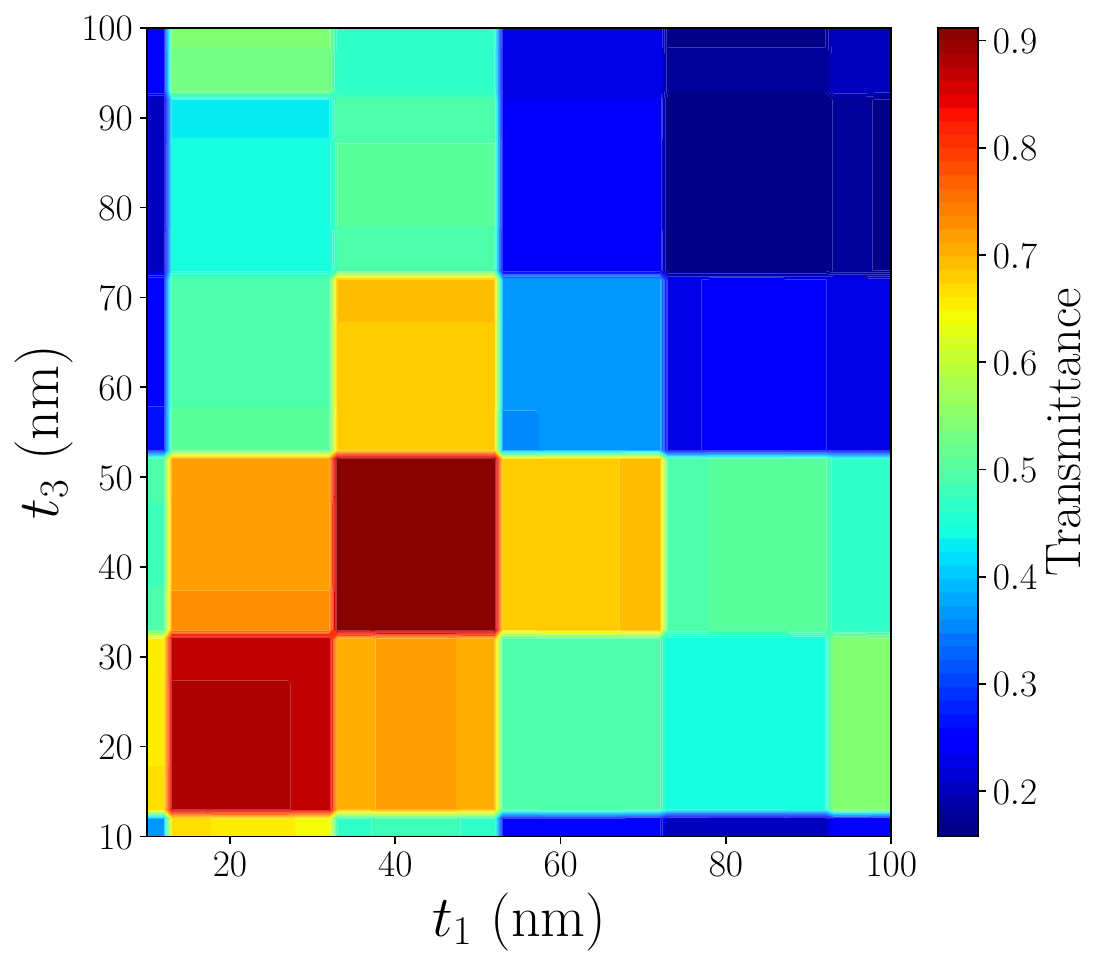}
        \includegraphics[width=0.31\textwidth]{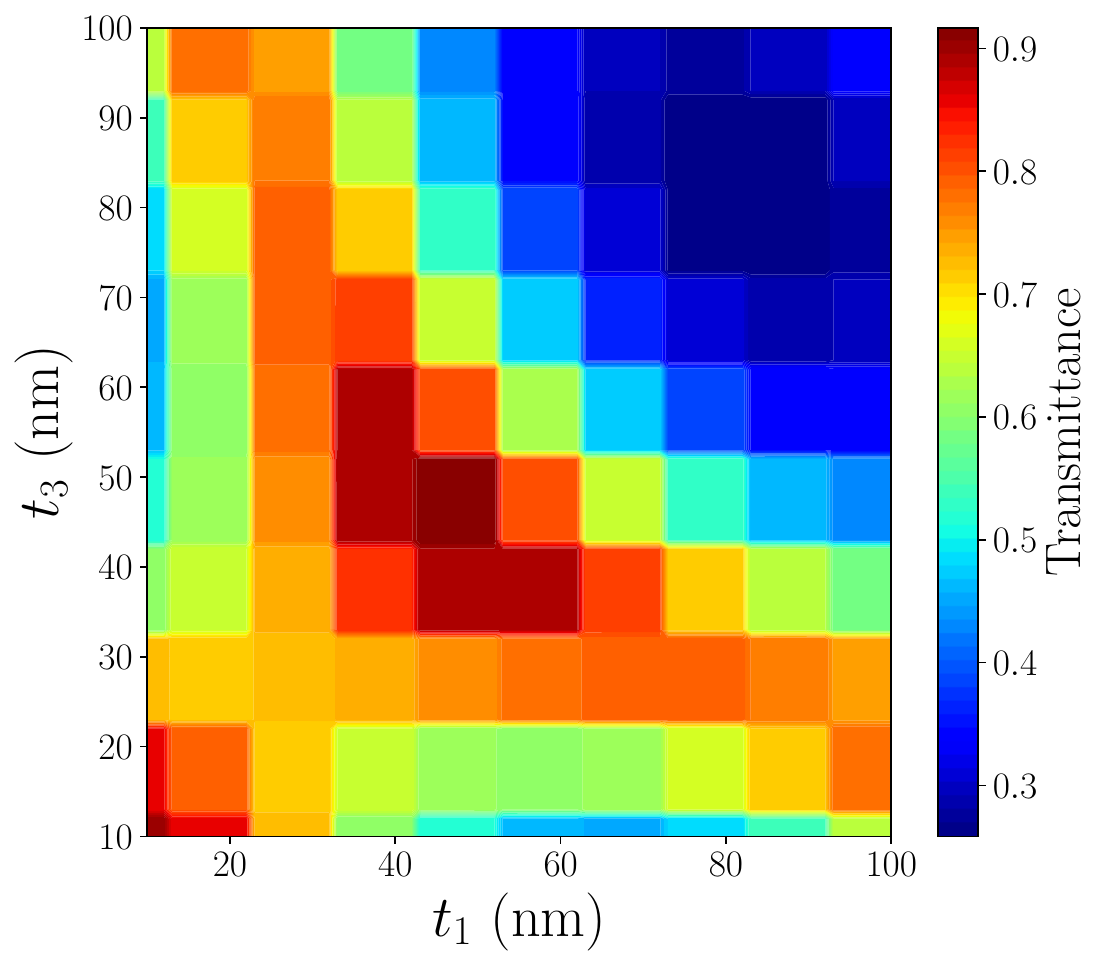}
        \includegraphics[width=0.31\textwidth]{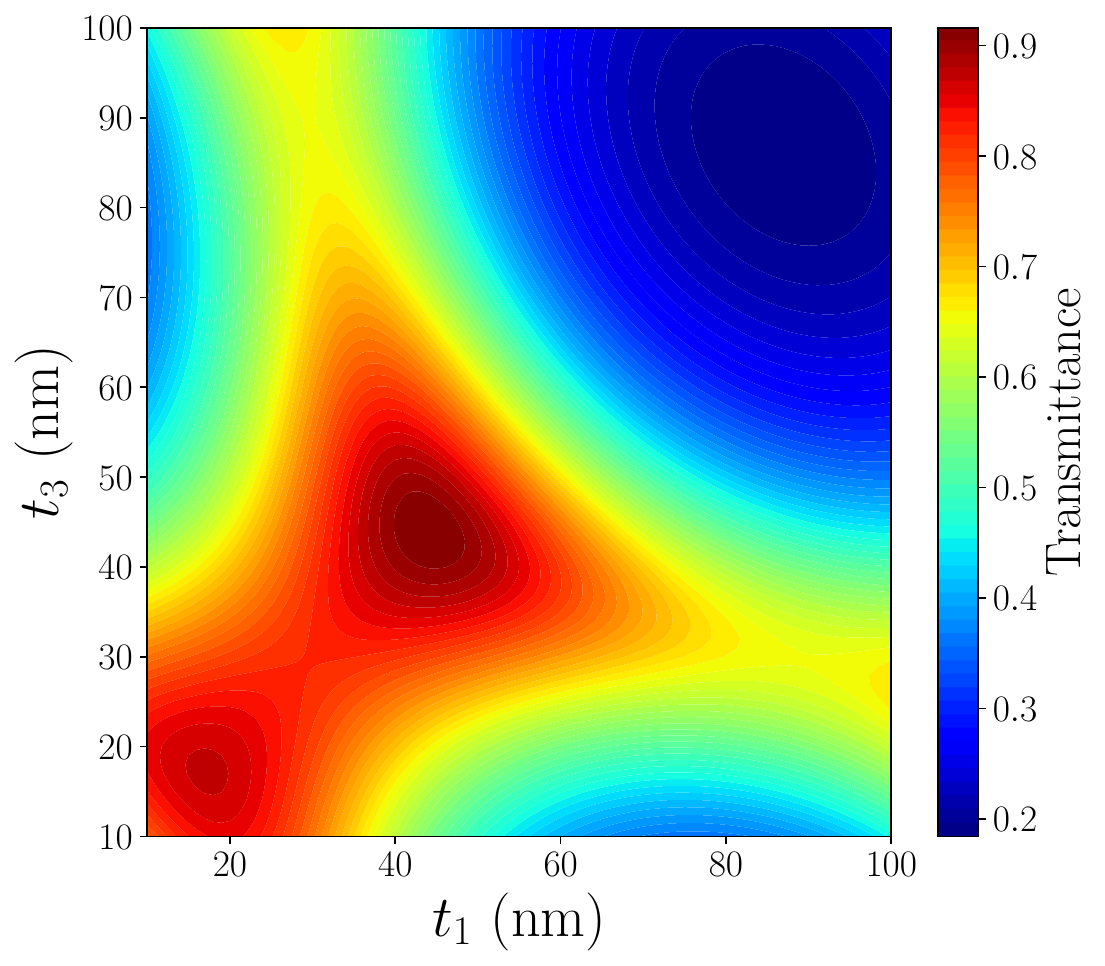}
        \caption{$t_2 = 15$}
    \end{subfigure}
    \caption{Visualization of the transmittance of the three-layer film made of \ce{TiO2}/\ce{Ag}/\ce{TiO2} for three different fidelity levels, i.e., low fidelity (shown in left panels), medium fidelity (shown in center panels), and high fidelity (shown in right panels).}
    \label{fig:threelayers_tio2_ag_tio2}
\end{figure}

\begin{figure}[t!]
    \centering
    \begin{subfigure}[b]{\textwidth}
        \centering
        \includegraphics[width=0.31\textwidth]{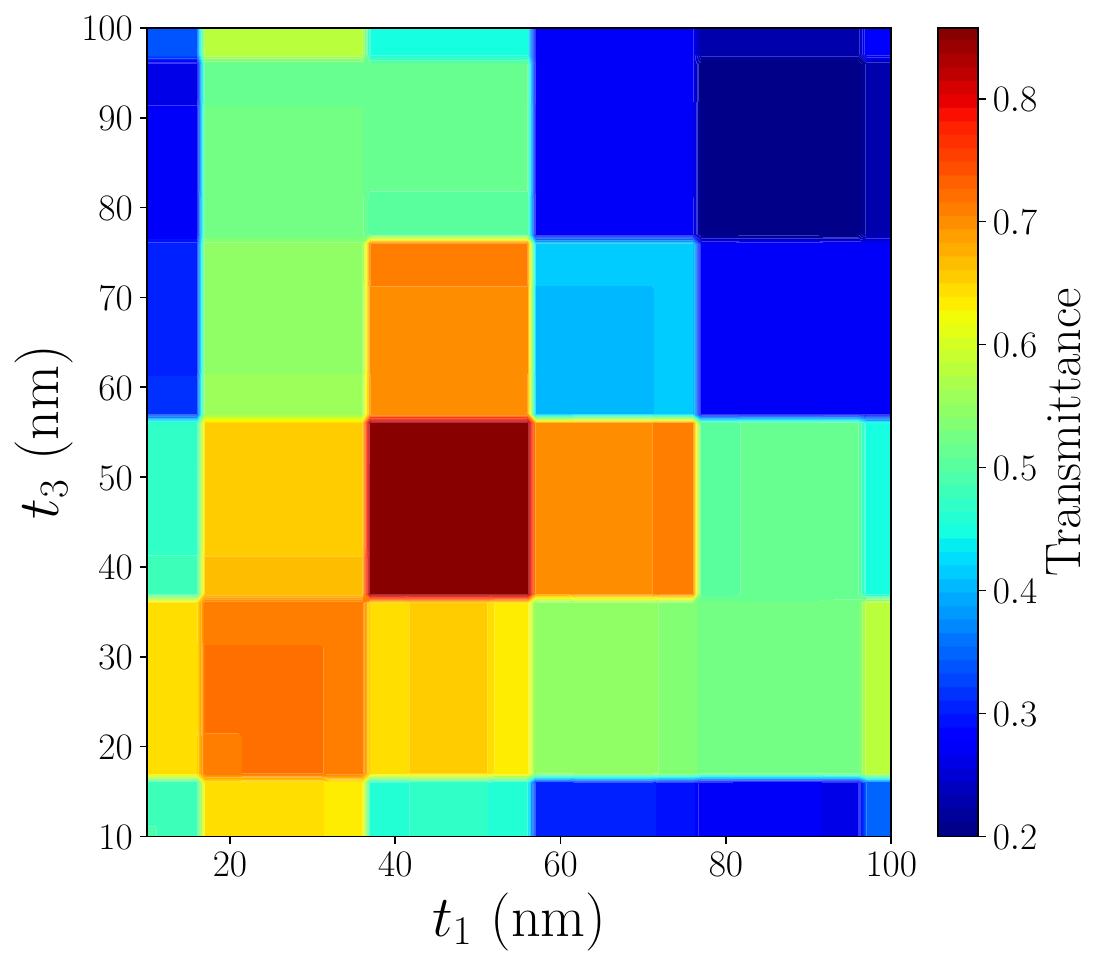}
        \includegraphics[width=0.31\textwidth]{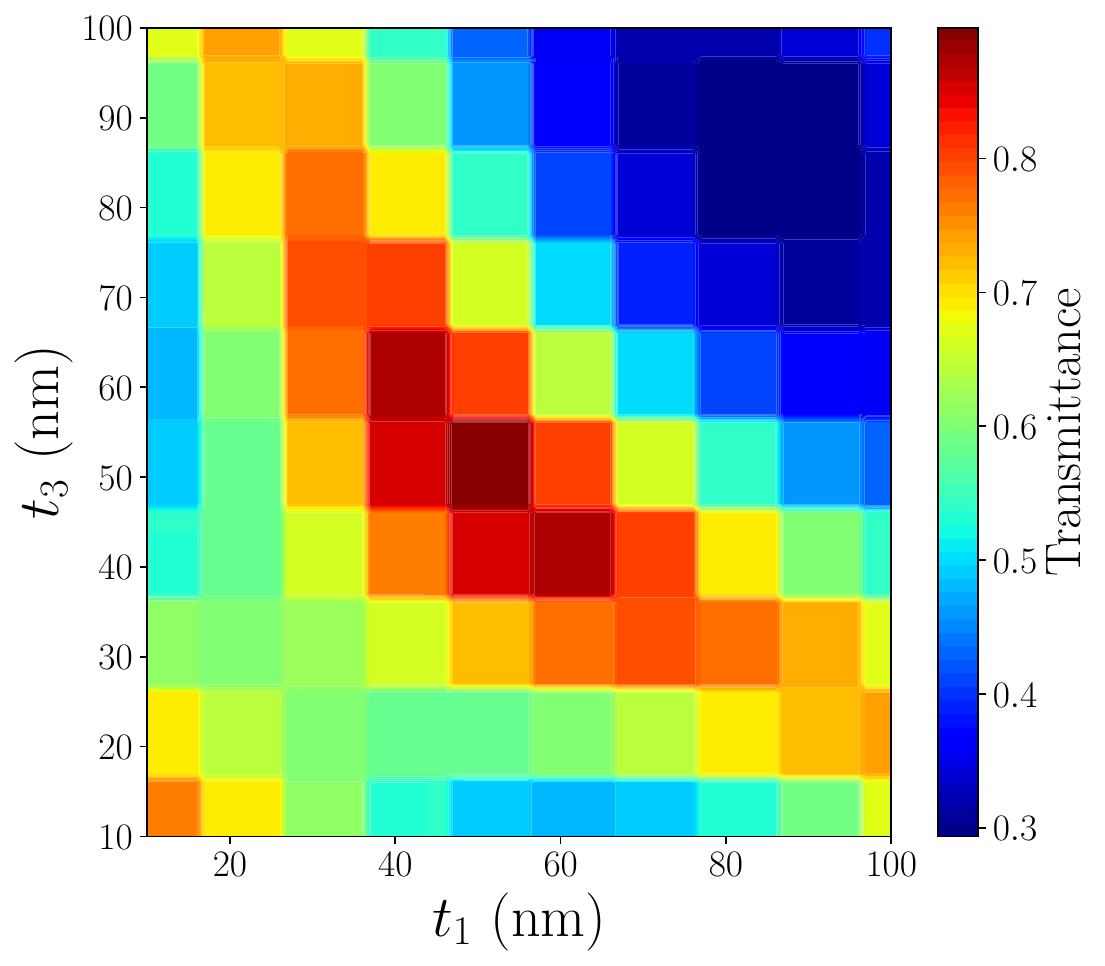}
        \includegraphics[width=0.31\textwidth]{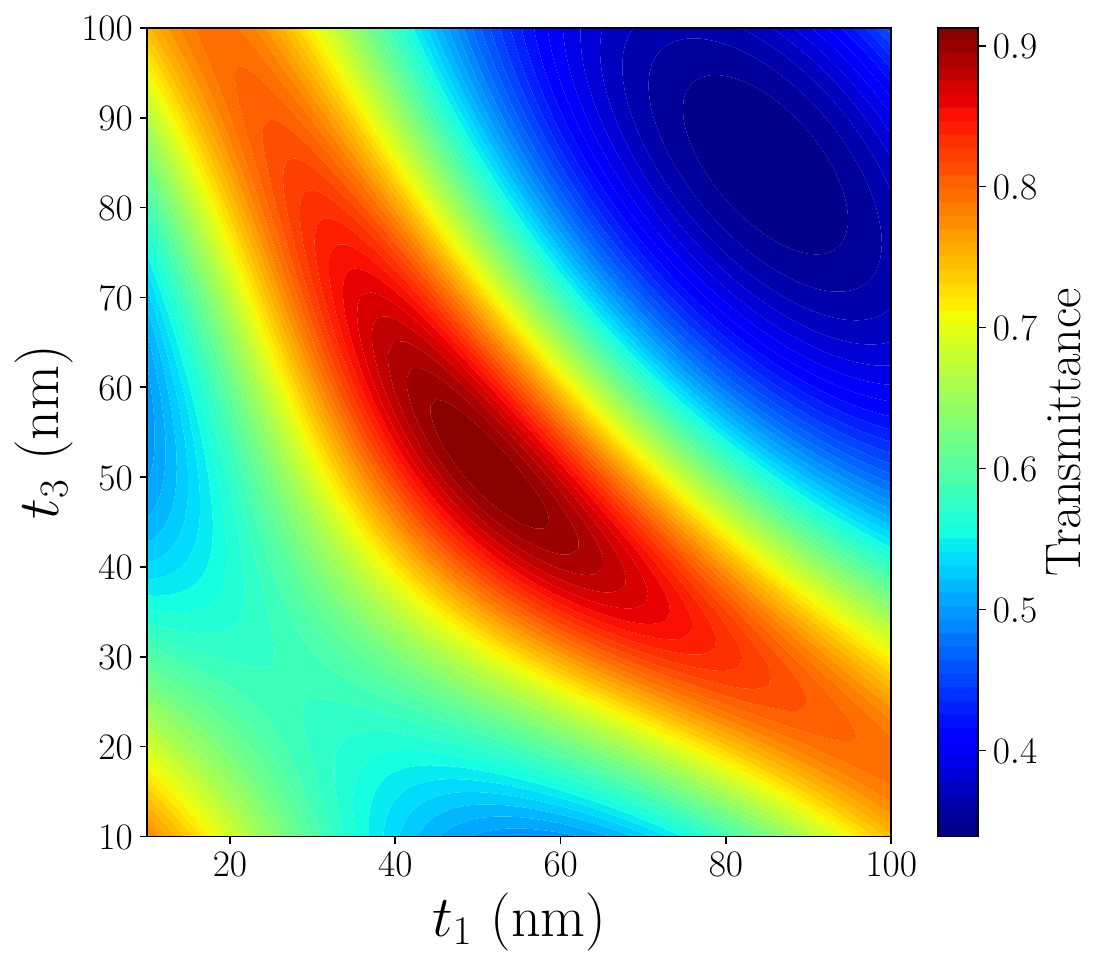}
        \caption{$t_2 = 7$}
    \end{subfigure}
    \begin{subfigure}[b]{\textwidth}
        \centering
        \includegraphics[width=0.31\textwidth]{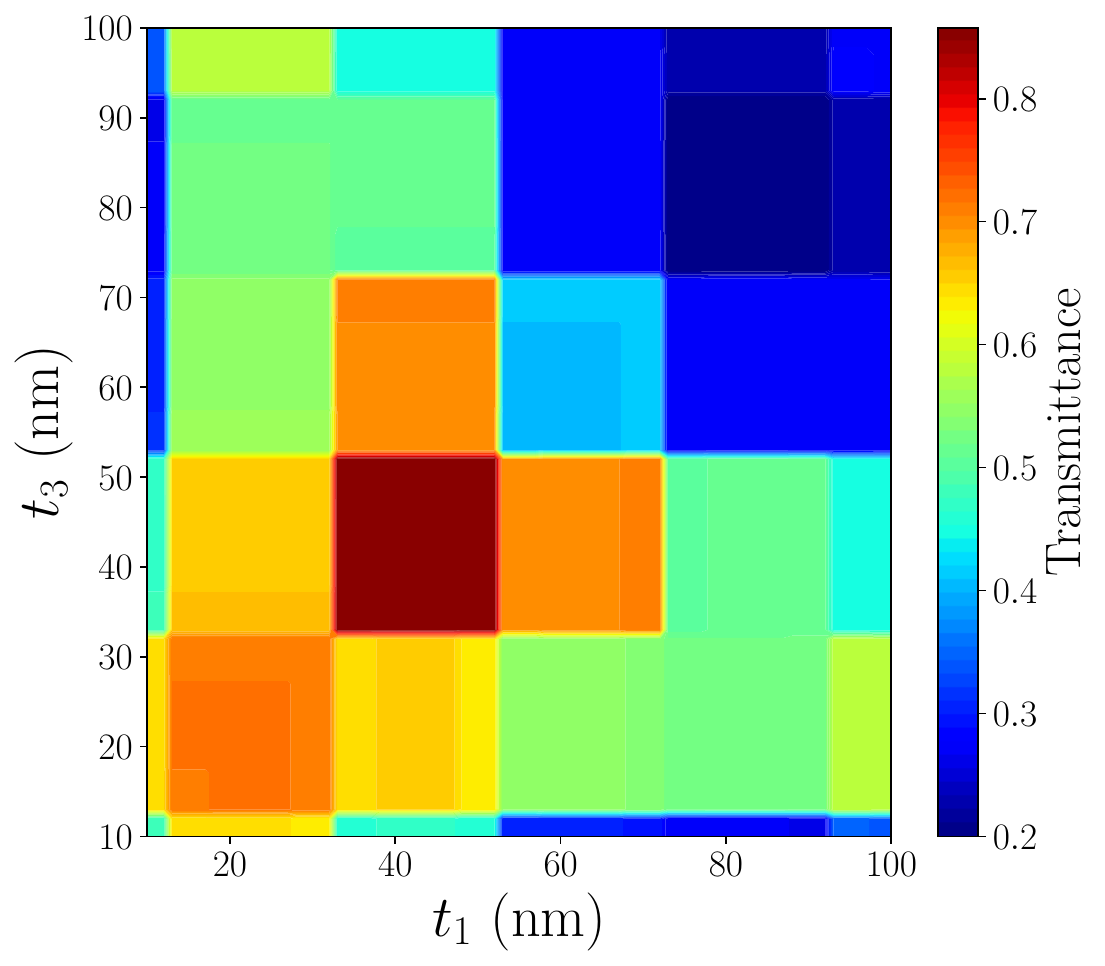}
        \includegraphics[width=0.31\textwidth]{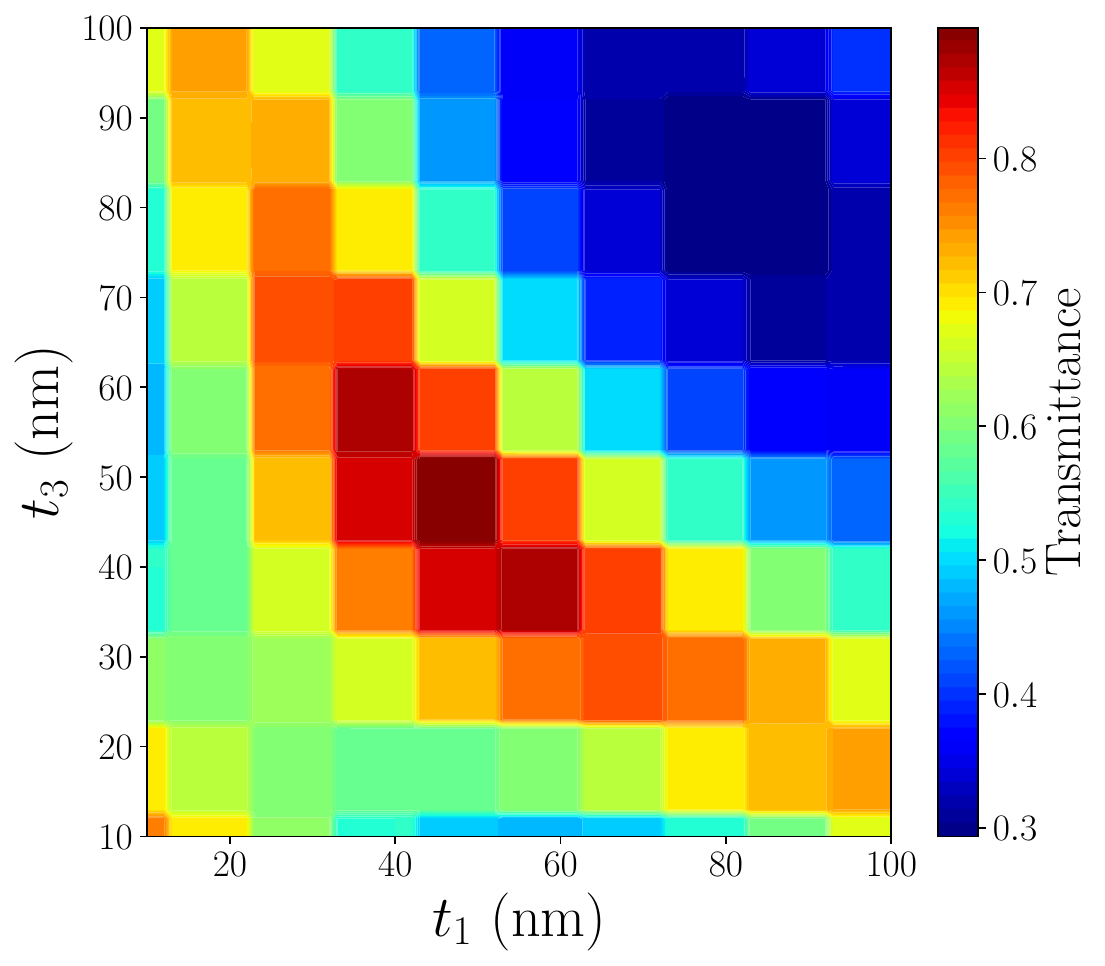}
        \includegraphics[width=0.31\textwidth]{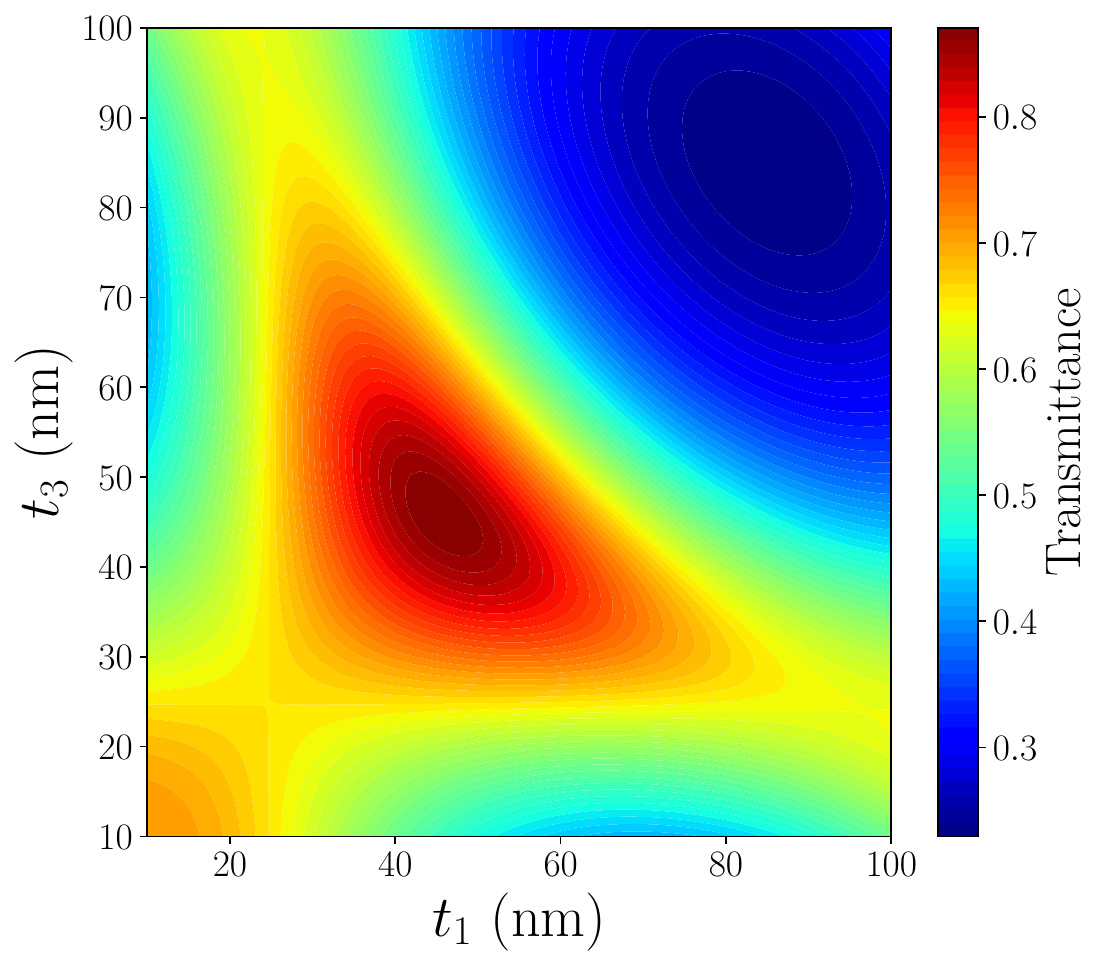}
        \caption{$t_2 = 15$}
    \end{subfigure}
    \caption{Visualization of the transmittance of the three-layer film made of \ce{TiO2}/\ce{Au}/\ce{TiO2} for three different fidelity levels, i.e., low fidelity (shown in left panels), medium fidelity (shown in center panels), and high fidelity (shown in right panels).}
    \label{fig:threelayers_tio2_au_tio2}
\end{figure}

\begin{figure}[t]
    \centering
    \begin{subfigure}[b]{\textwidth}
        \centering
        \includegraphics[width=0.31\textwidth]{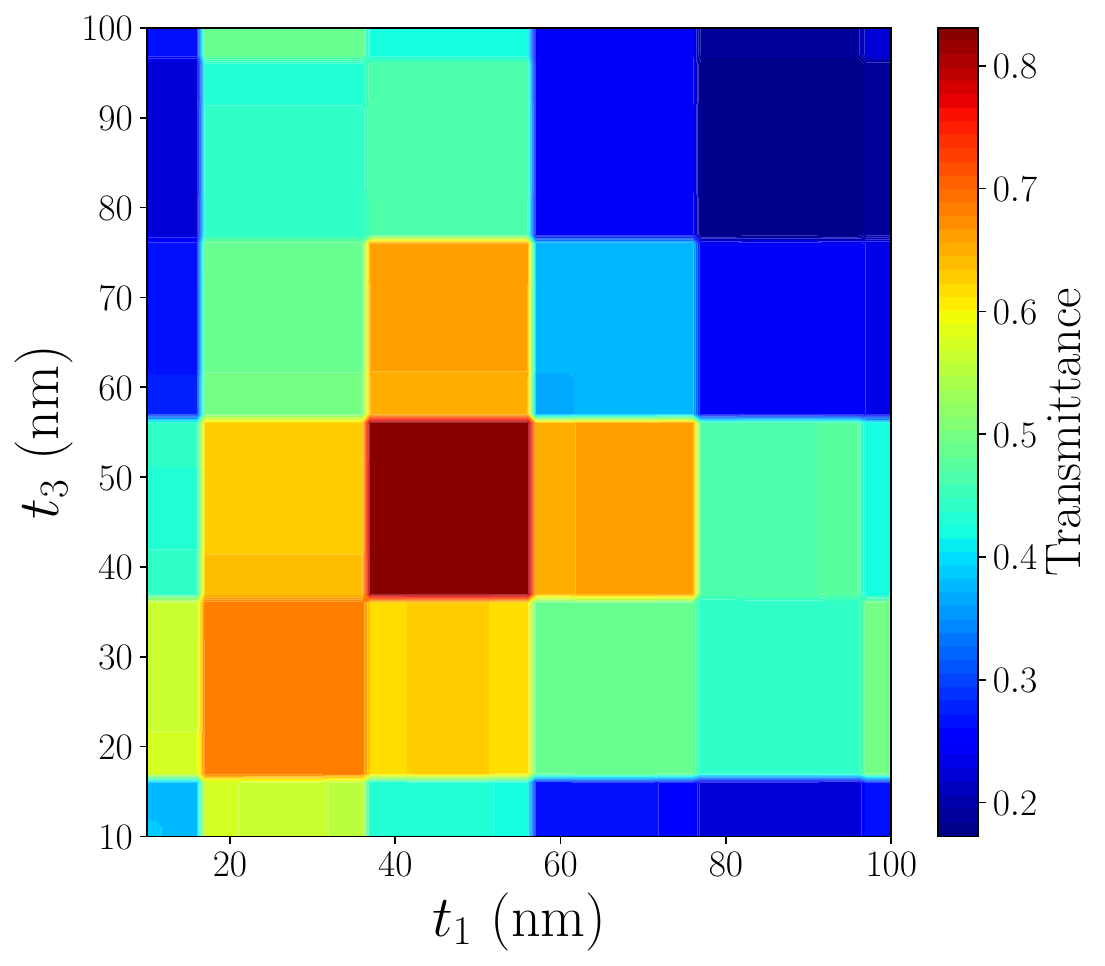}
        \includegraphics[width=0.31\textwidth]{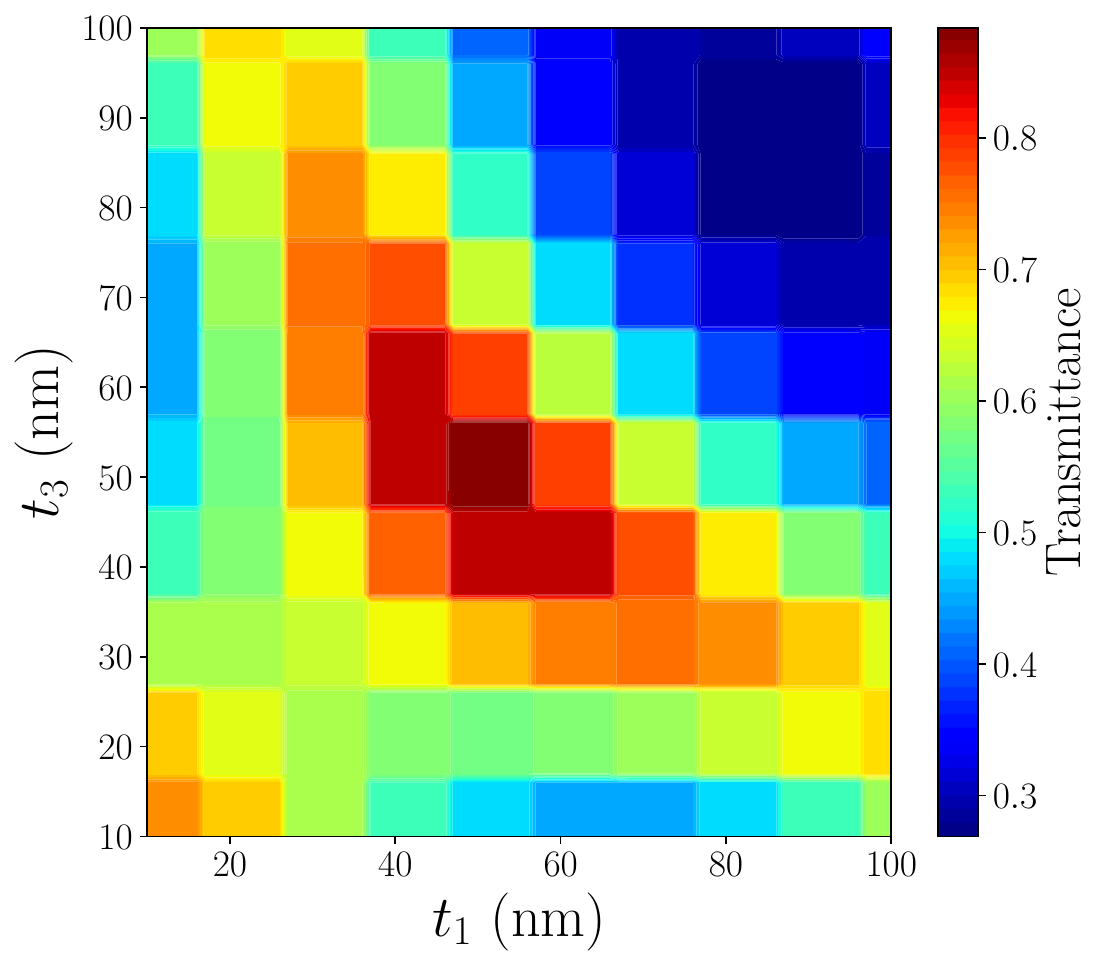}
        \includegraphics[width=0.31\textwidth]{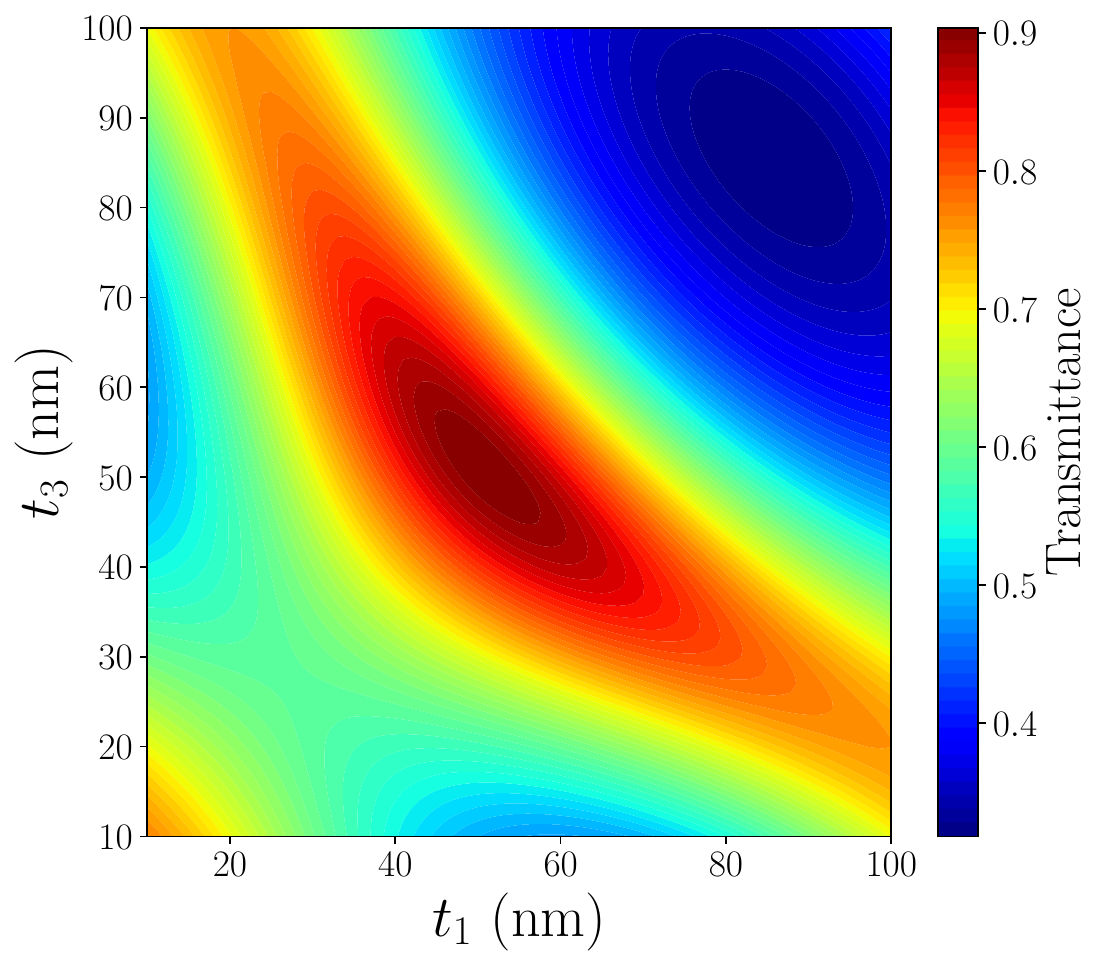}
        \caption{$t_2 = 7$}
    \end{subfigure}
    \begin{subfigure}[b]{\textwidth}
        \centering
        \includegraphics[width=0.31\textwidth]{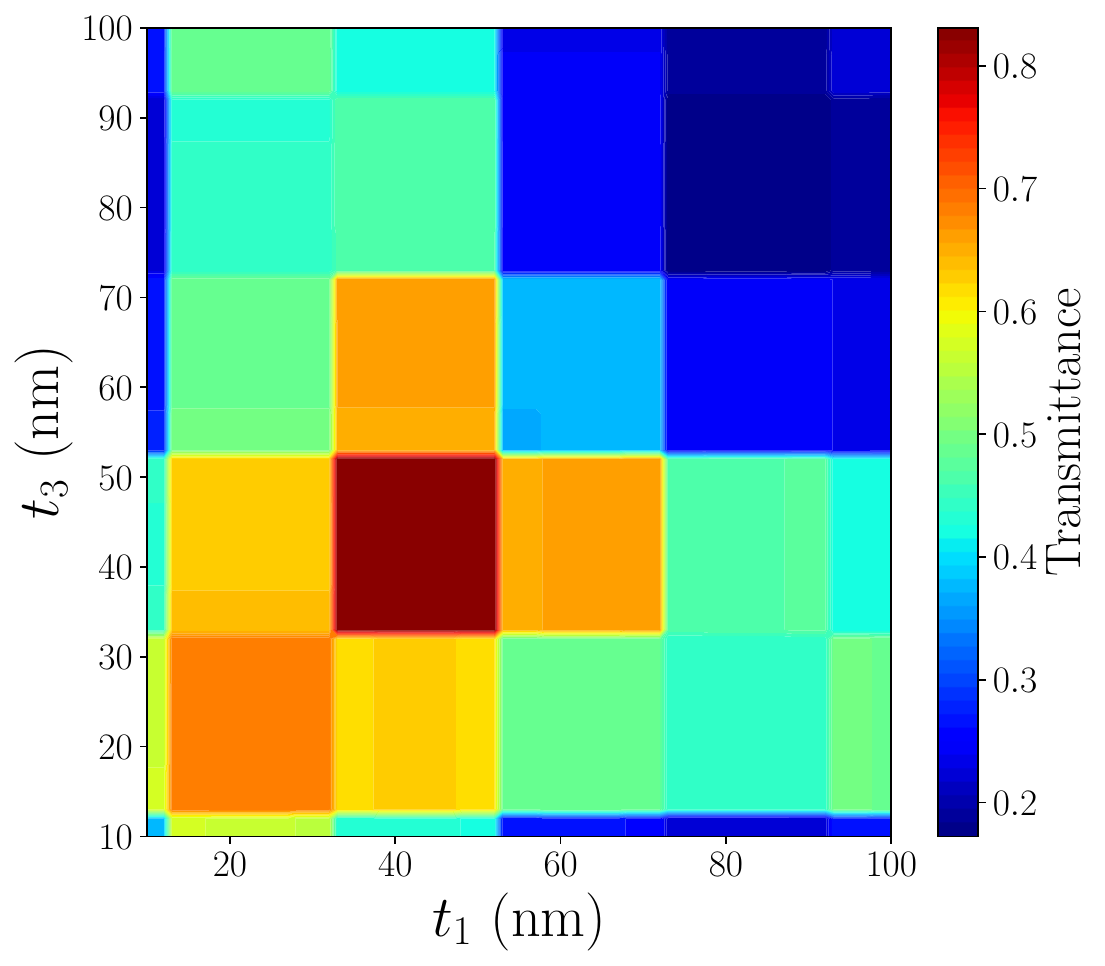}
        \includegraphics[width=0.31\textwidth]{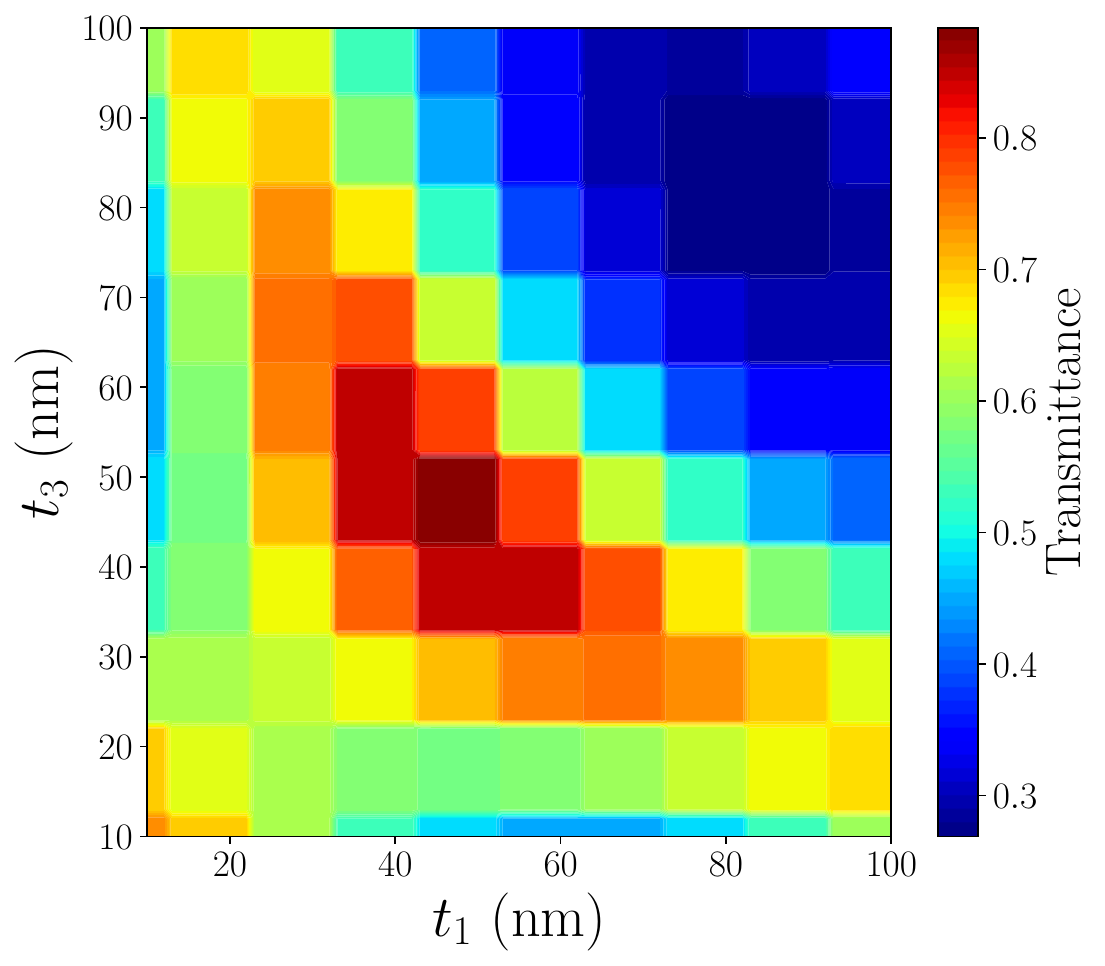}
        \includegraphics[width=0.31\textwidth]{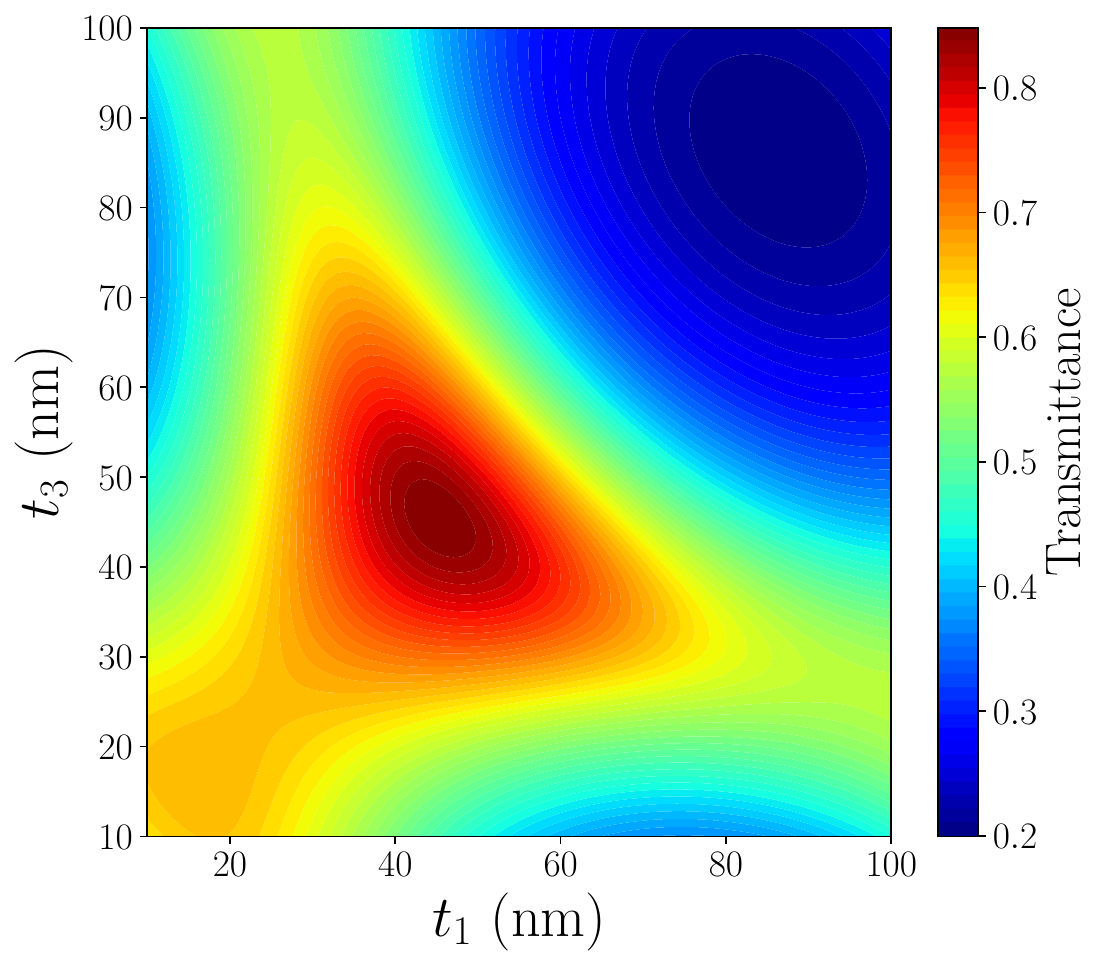}
        \caption{$t_2 = 15$}
    \end{subfigure}
    \caption{Visualization of the transmittance of the three-layer film made of \ce{TiO2}/\ce{Cu}/\ce{TiO2} for three different fidelity levels, i.e., low fidelity (shown in left panels), medium fidelity (shown in center panels), and high fidelity (shown in right panels).}
    \label{fig:threelayers_tio2_cu_tio2}
\end{figure}

\begin{figure}[t!]
    \centering
    \begin{subfigure}[b]{\textwidth}
        \centering
        \includegraphics[width=0.31\textwidth]{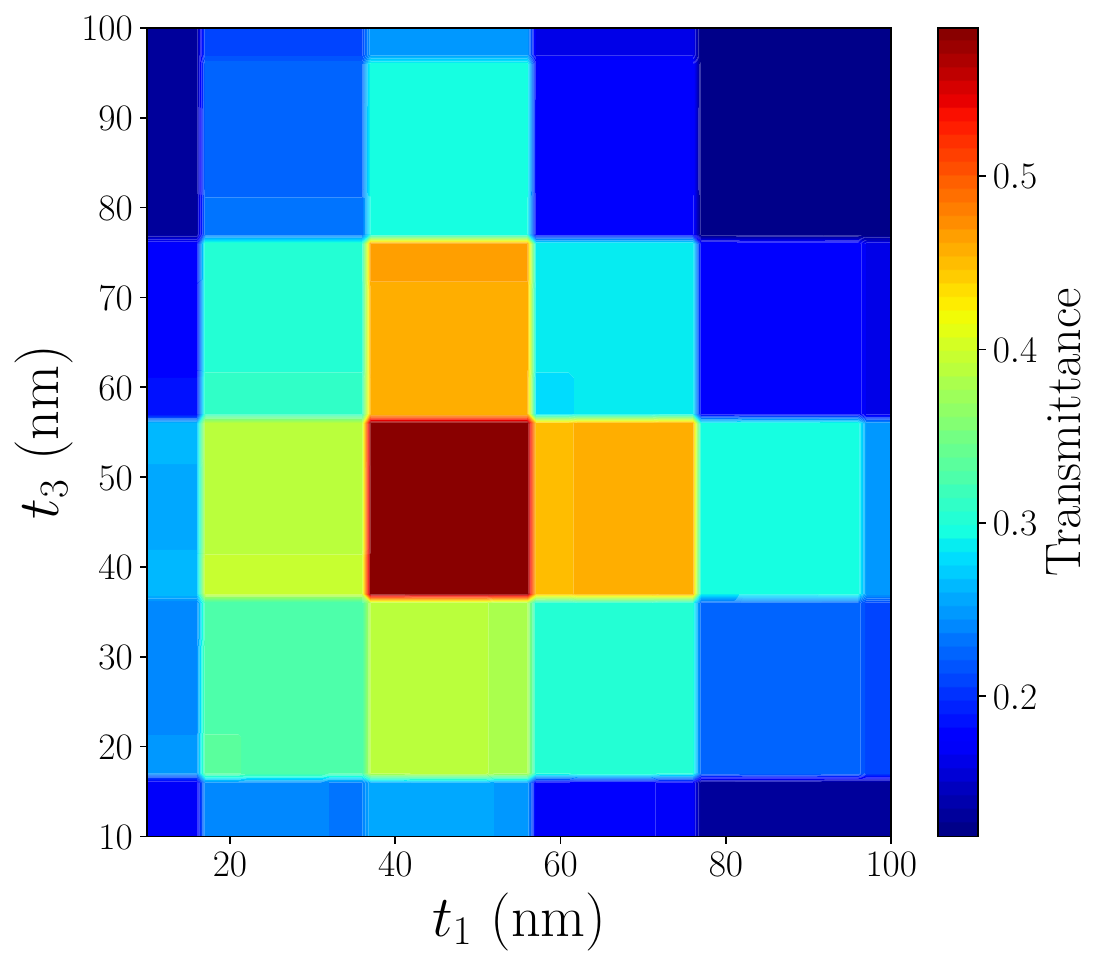}
        \includegraphics[width=0.31\textwidth]{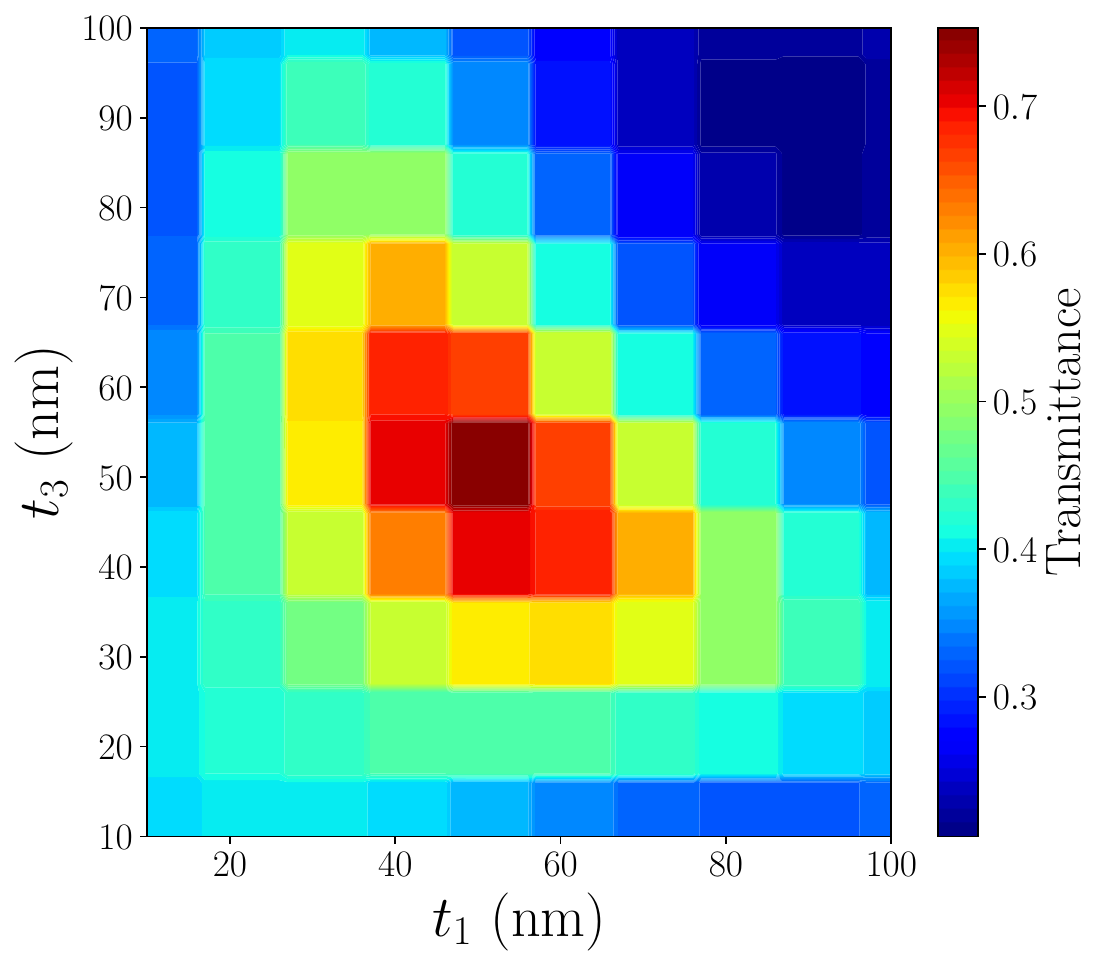}
        \includegraphics[width=0.31\textwidth]{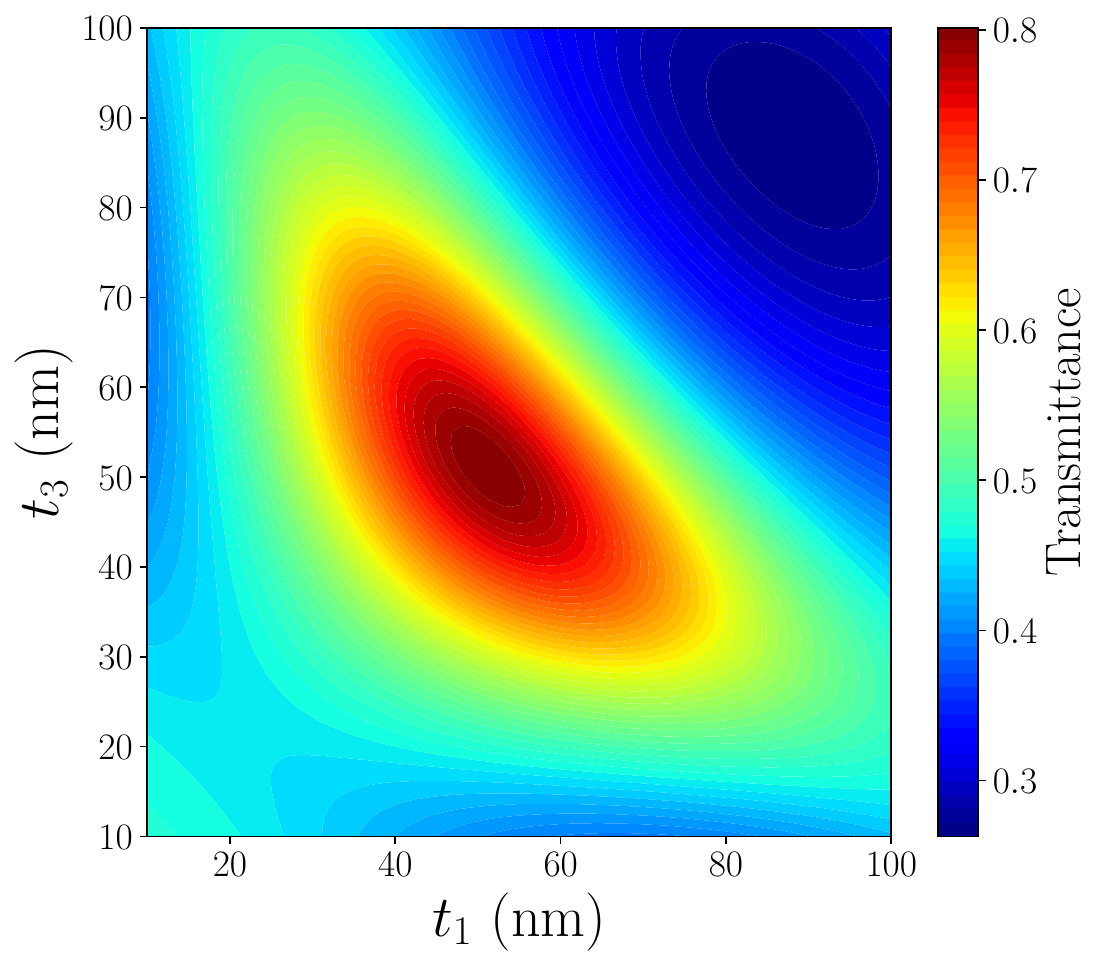}
        \caption{$t_2 = 7$}
    \end{subfigure}
    \begin{subfigure}[b]{\textwidth}
        \centering
        \includegraphics[width=0.31\textwidth]{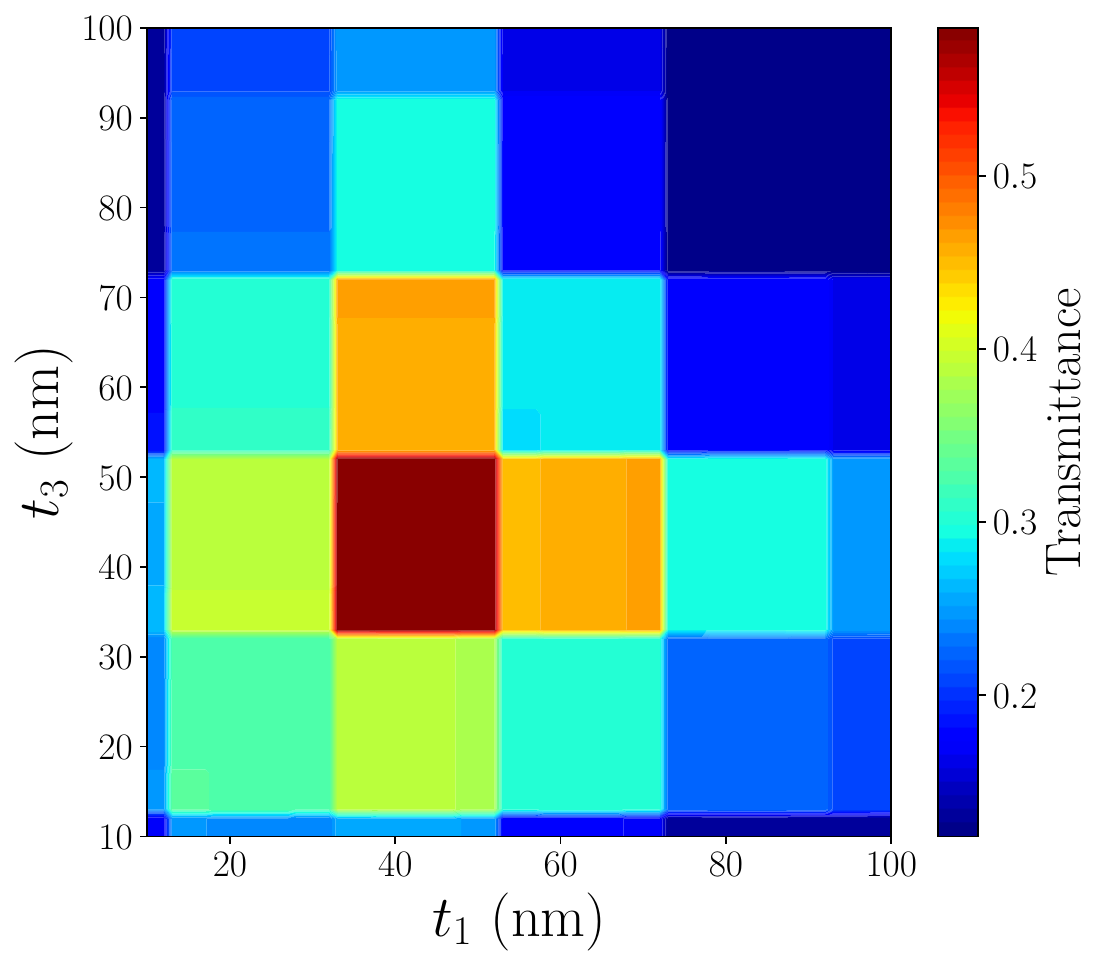}
        \includegraphics[width=0.31\textwidth]{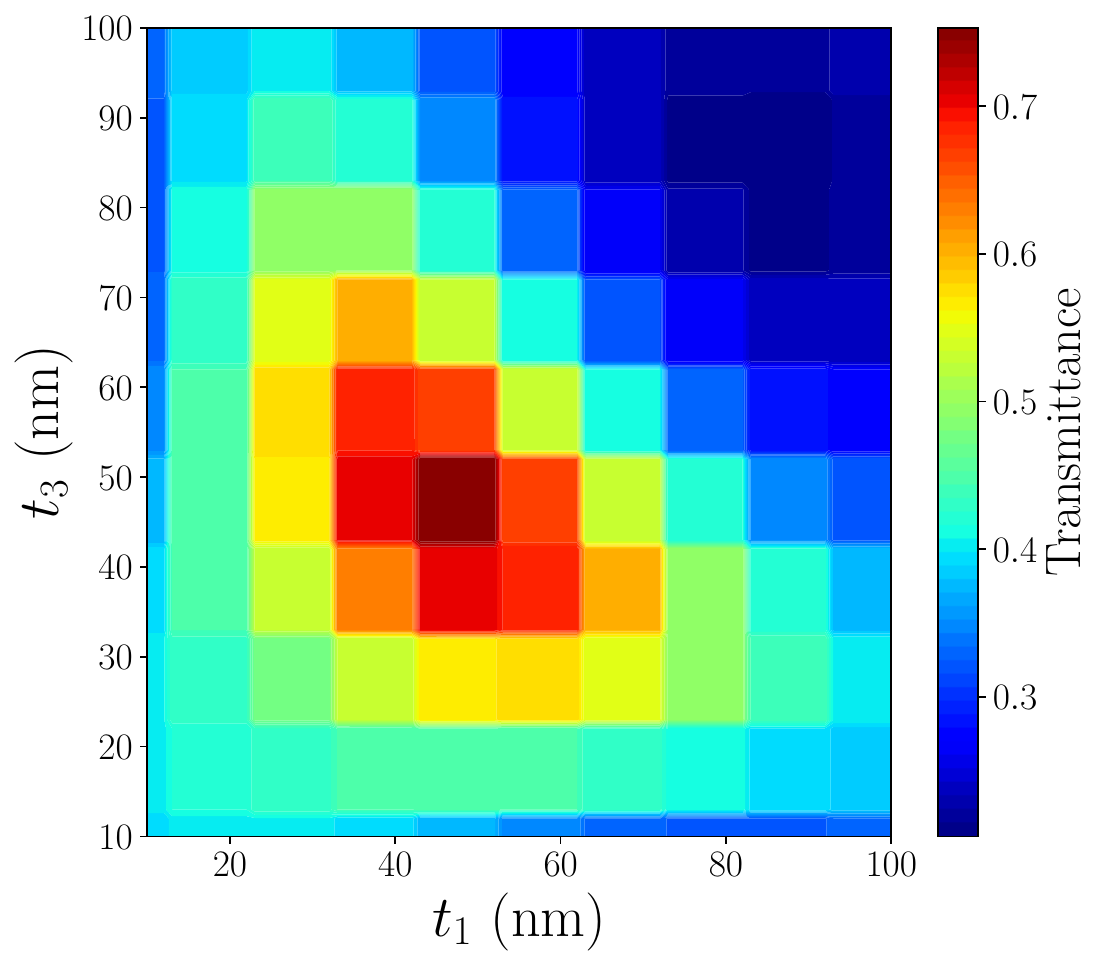}
        \includegraphics[width=0.31\textwidth]{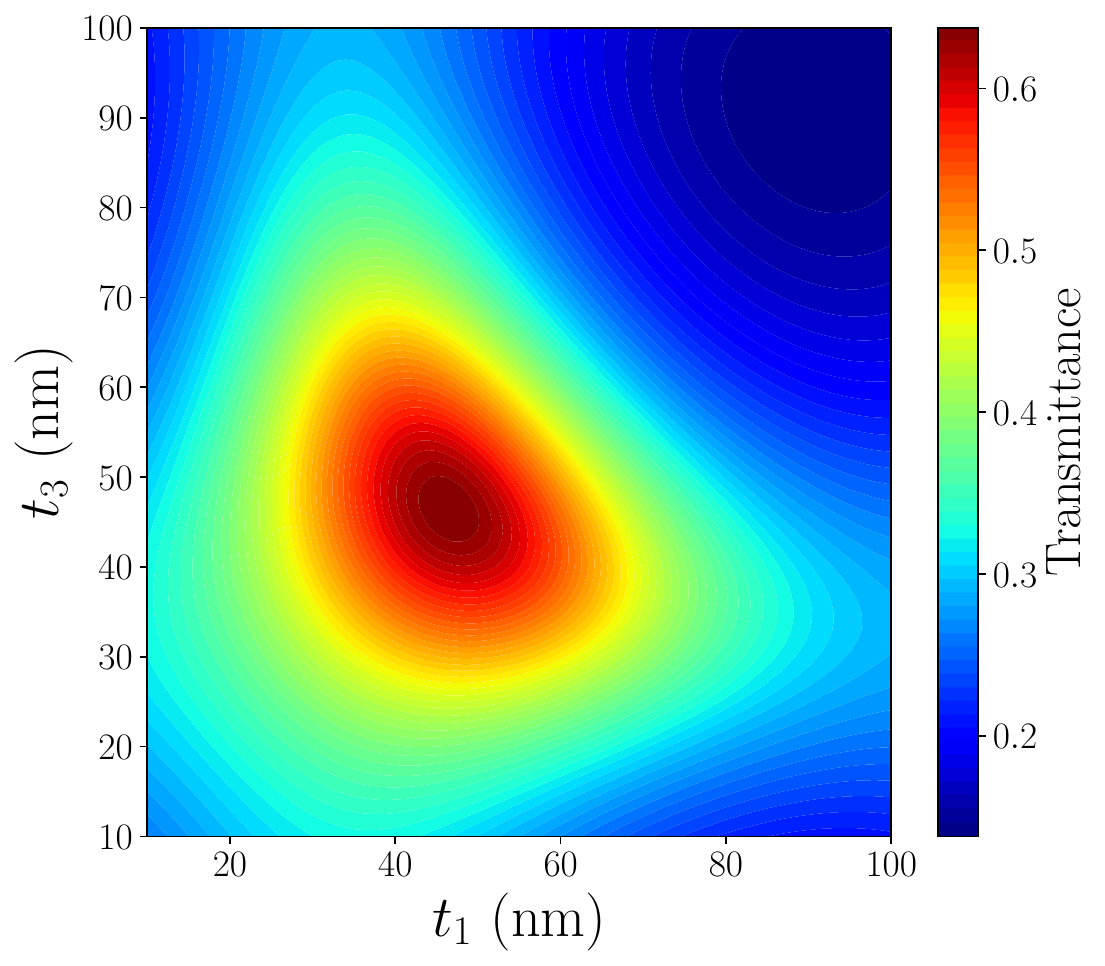}
        \caption{$t_2 = 15$}
    \end{subfigure}
    \caption{Visualization of the transmittance of the three-layer film made of \ce{TiO2}/\ce{Ni}/\ce{TiO2} for three different fidelity levels, i.e., low fidelity (shown in left panels), medium fidelity (shown in center panels), and high fidelity (shown in right panels).}
    \label{fig:threelayers_tio2_ni_tio2}
\end{figure}

\begin{figure}[t]
    \centering
    \begin{subfigure}[b]{\textwidth}
        \centering
        \includegraphics[width=0.31\textwidth]{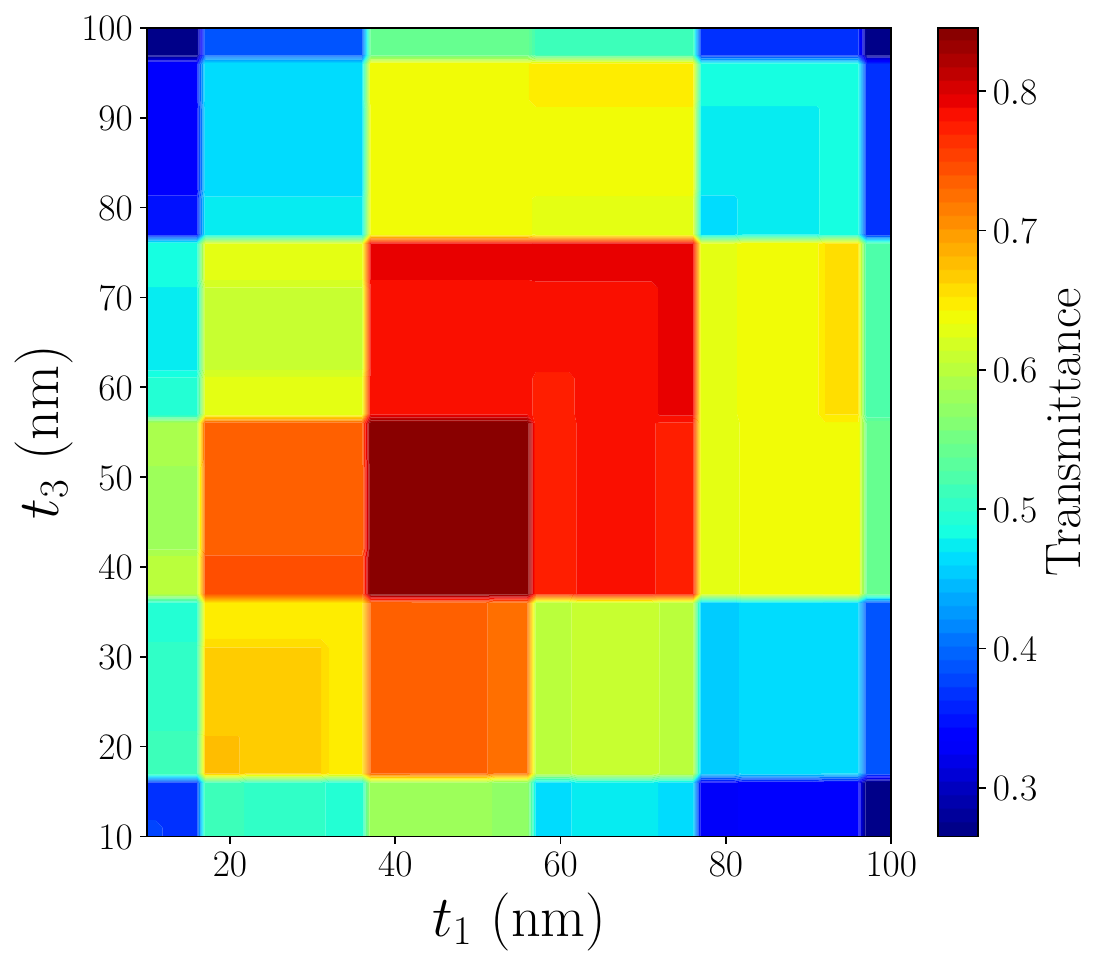}
        \includegraphics[width=0.31\textwidth]{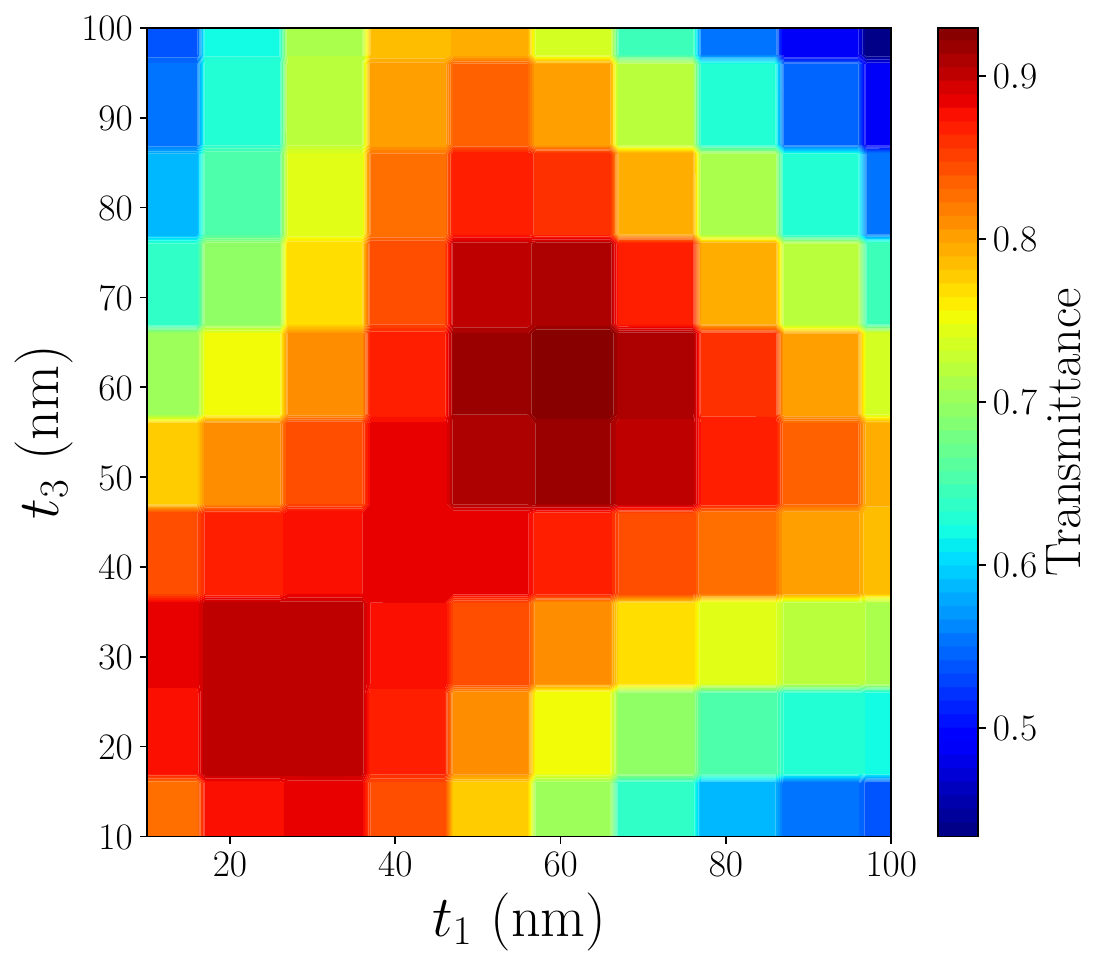}
        \includegraphics[width=0.31\textwidth]{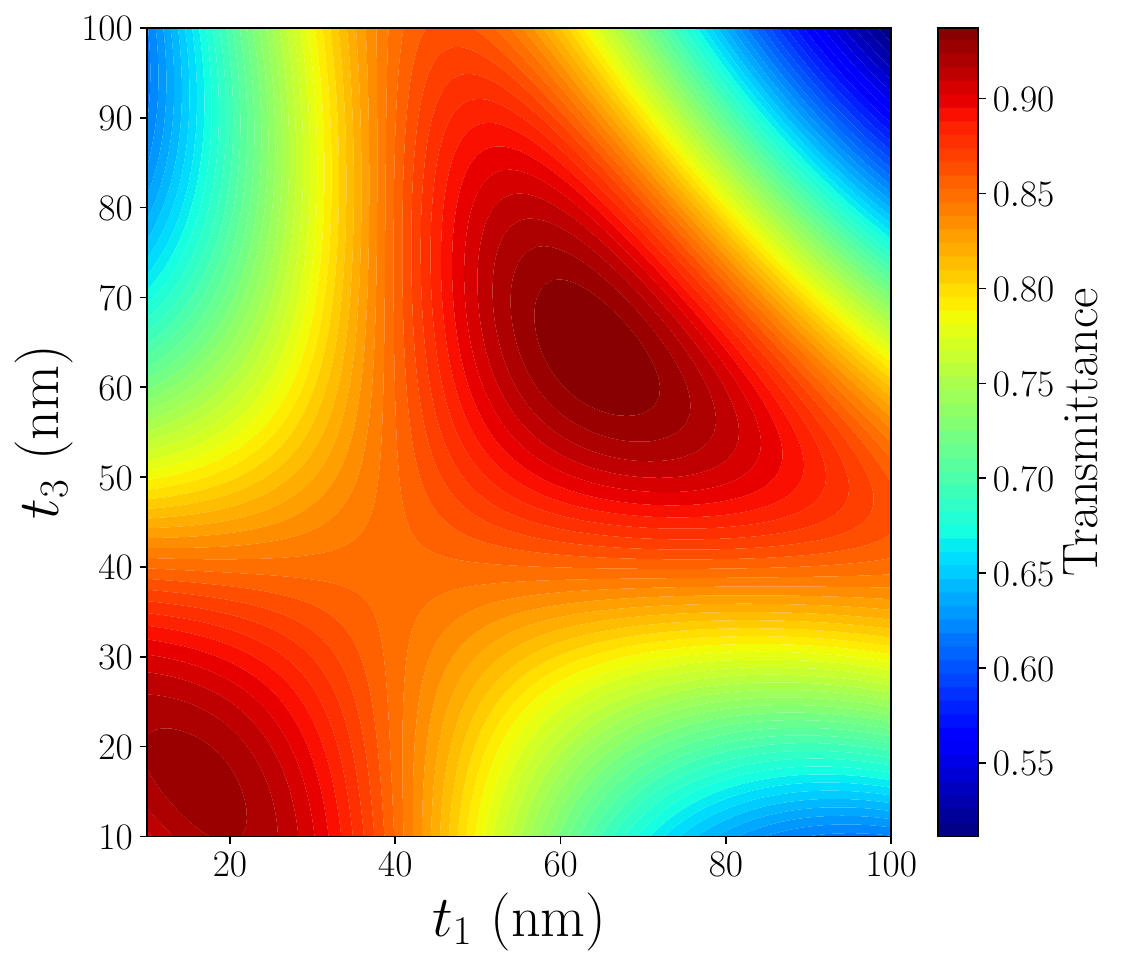}
        \caption{$t_2 = 7$}
    \end{subfigure}
    \begin{subfigure}[b]{\textwidth}
        \centering
        \includegraphics[width=0.31\textwidth]{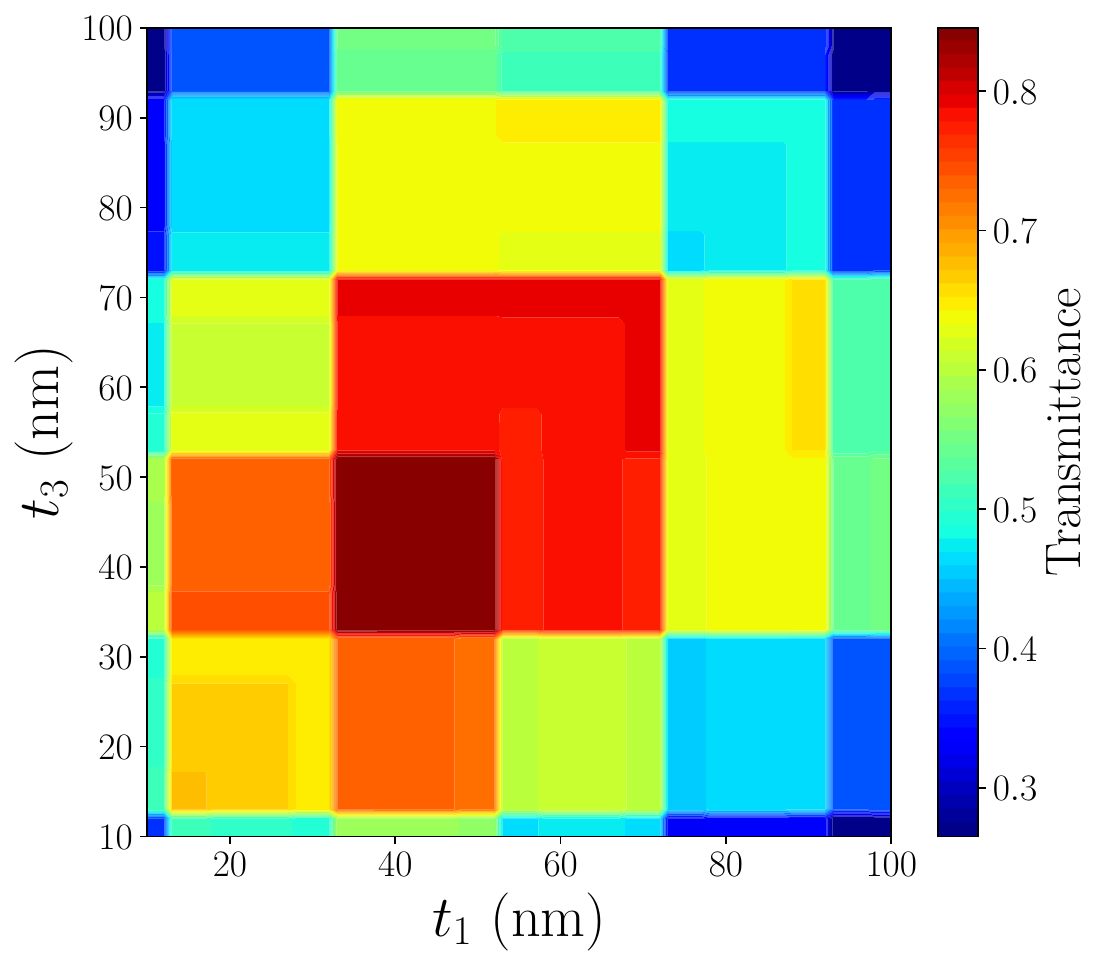}
        \includegraphics[width=0.31\textwidth]{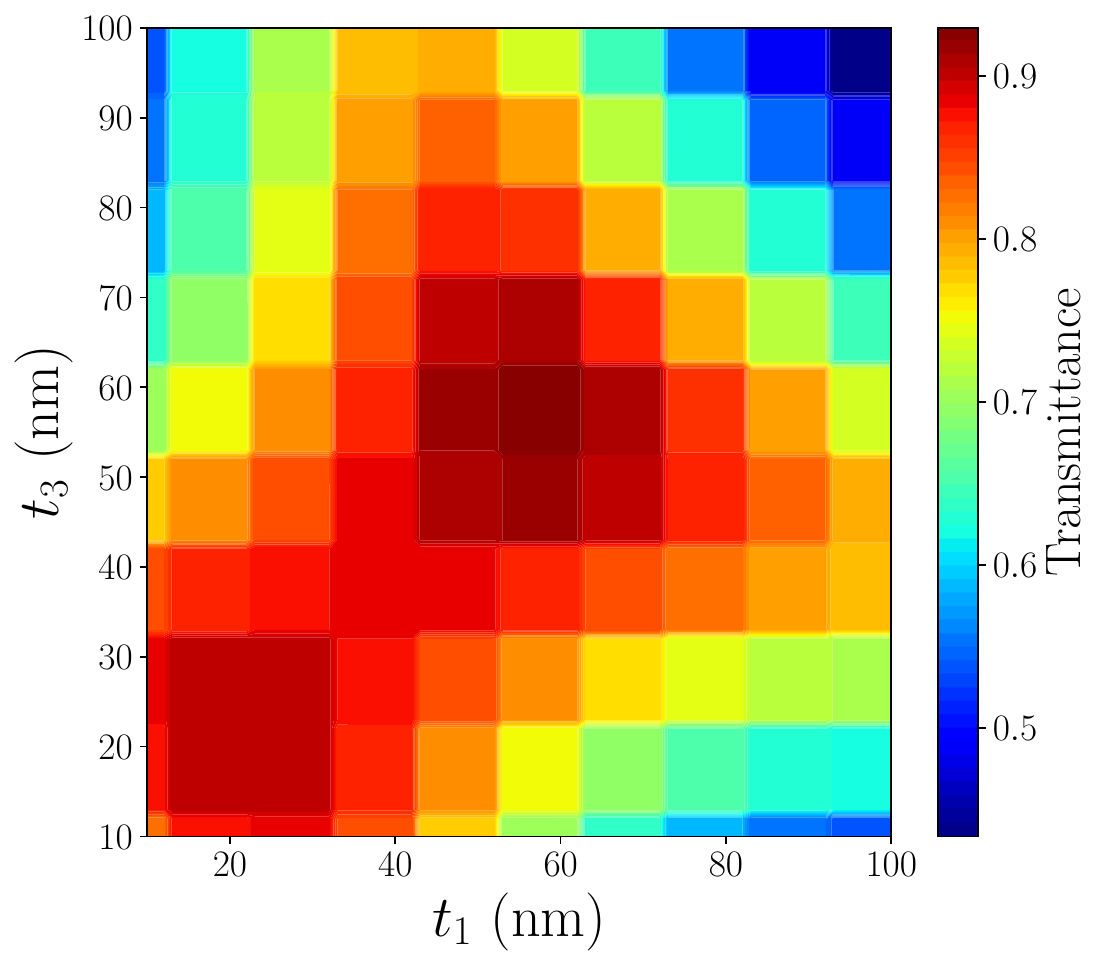}
        \includegraphics[width=0.31\textwidth]{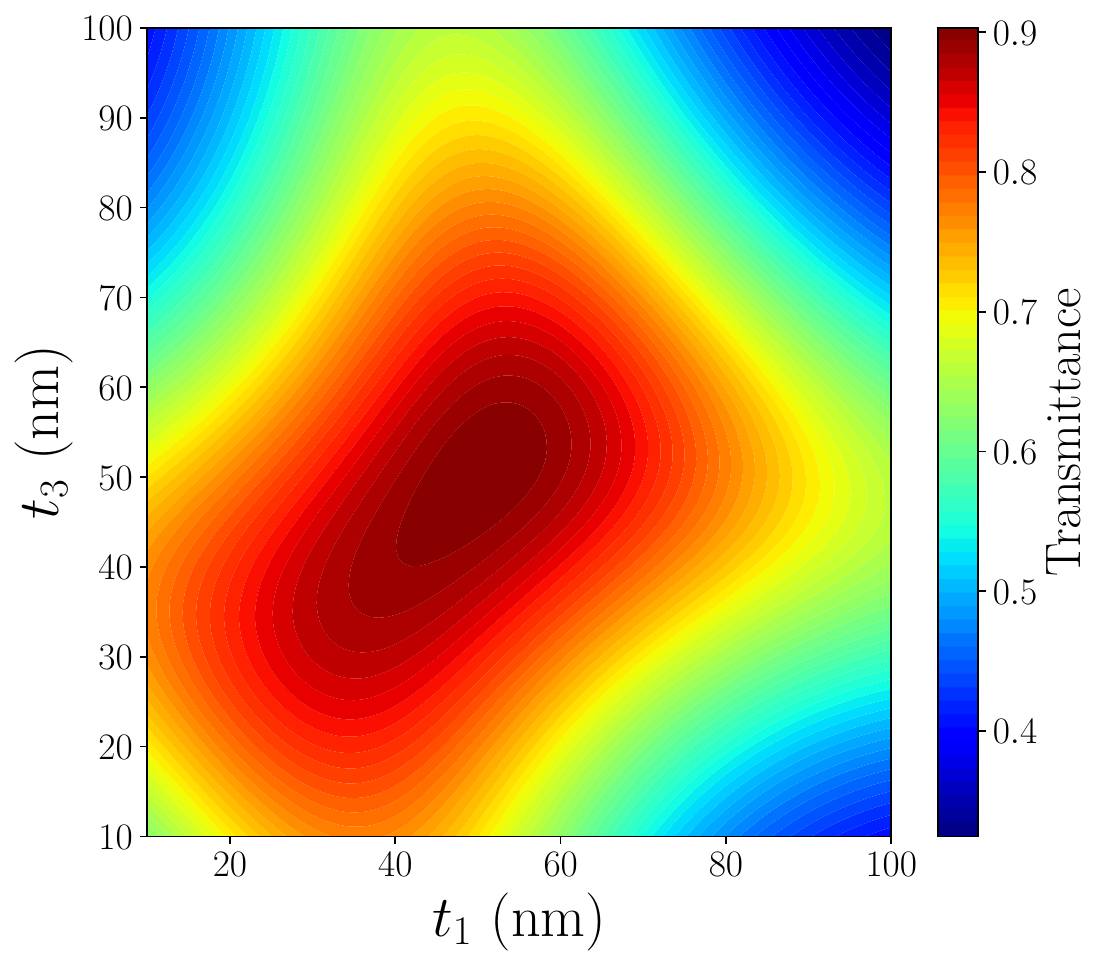}
        \caption{$t_2 = 15$}
    \end{subfigure}
    \caption{Visualization of the transmittance of the three-layer film made of \ce{AZO}/\ce{Ag}/\ce{AZO} for three different fidelity levels, i.e., low fidelity (shown in left panels), medium fidelity (shown in center panels), and high fidelity (shown in right panels).}
    \label{fig:threelayers_azo_ag_azo}
\end{figure}

\begin{figure}[t]
    \centering
    \begin{subfigure}[b]{\textwidth}
        \centering
        \includegraphics[width=0.31\textwidth]{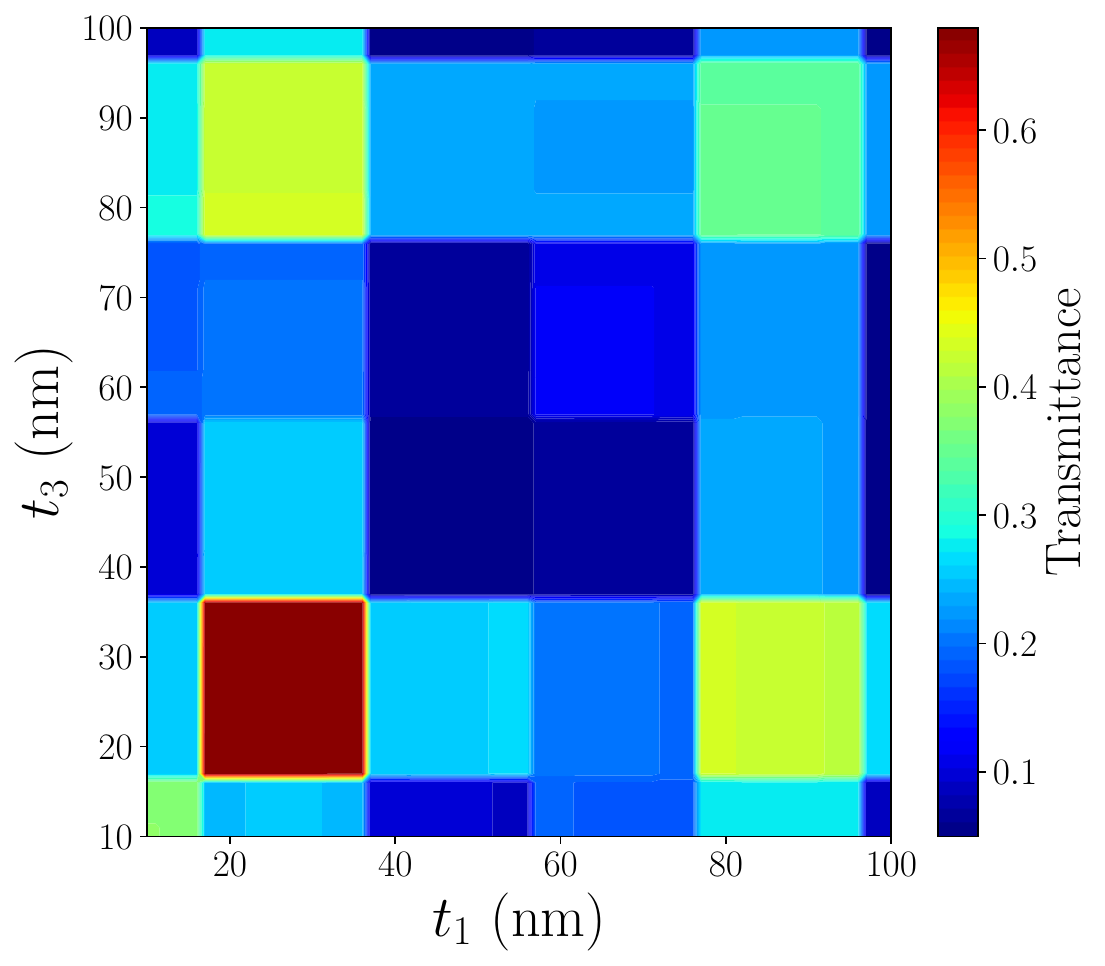}
        \includegraphics[width=0.31\textwidth]{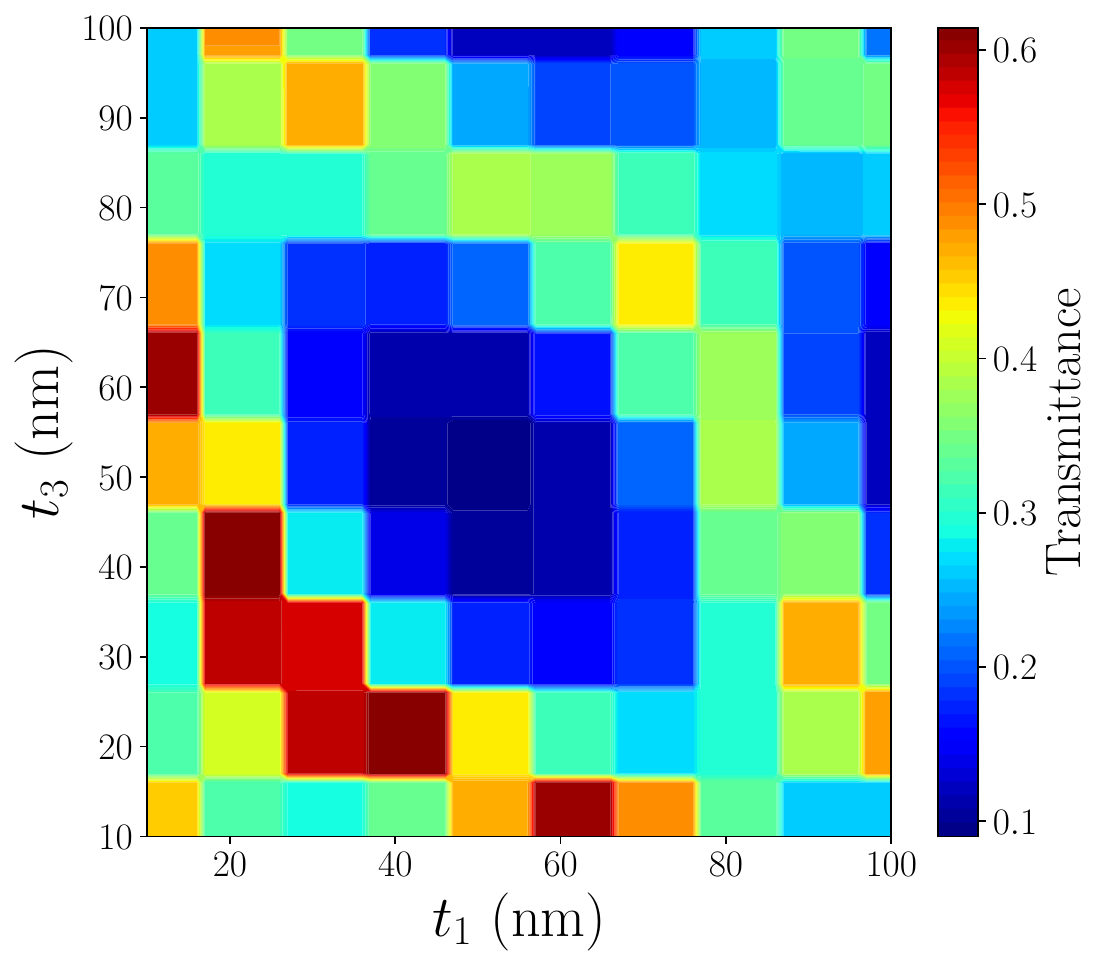}
        \includegraphics[width=0.31\textwidth]{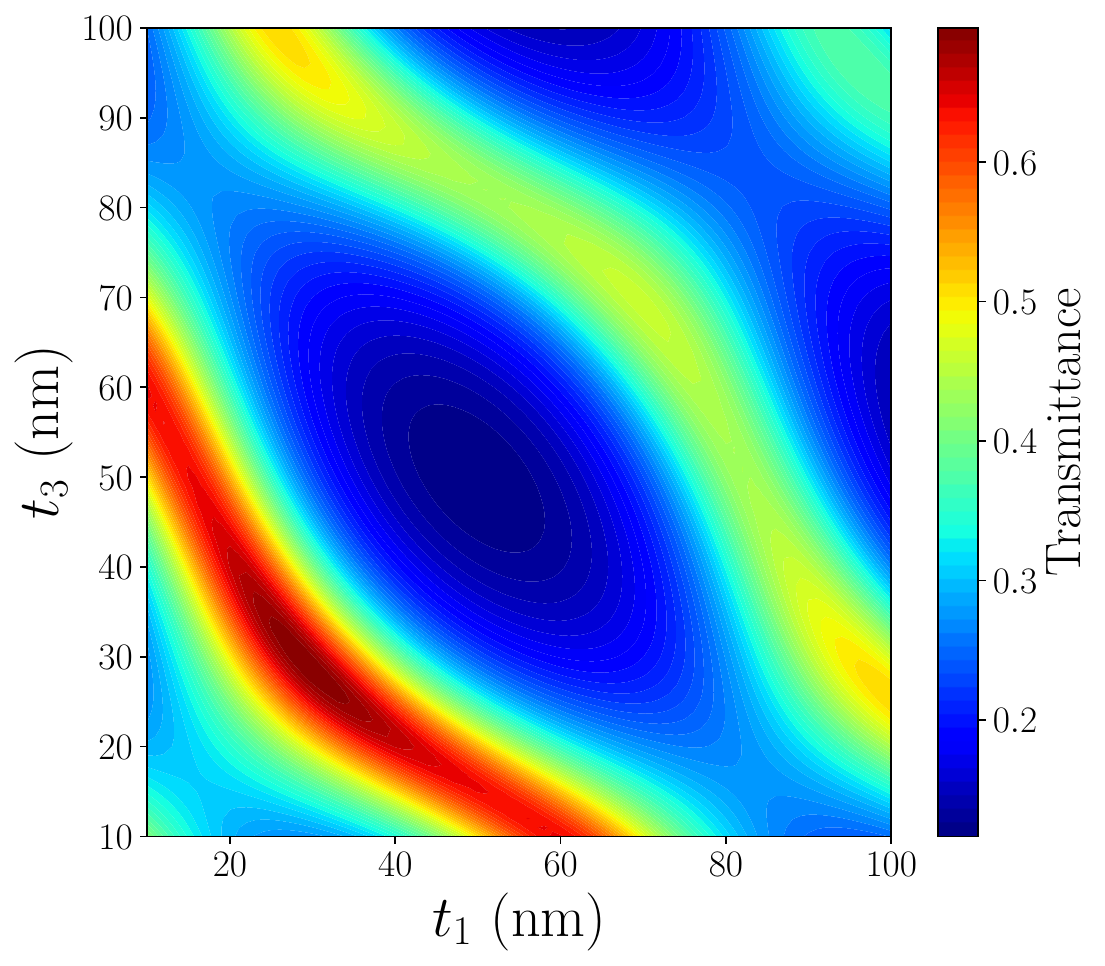}
        \caption{$t_2 = 7$}
    \end{subfigure}
    \begin{subfigure}[b]{\textwidth}
        \centering
        \includegraphics[width=0.31\textwidth]{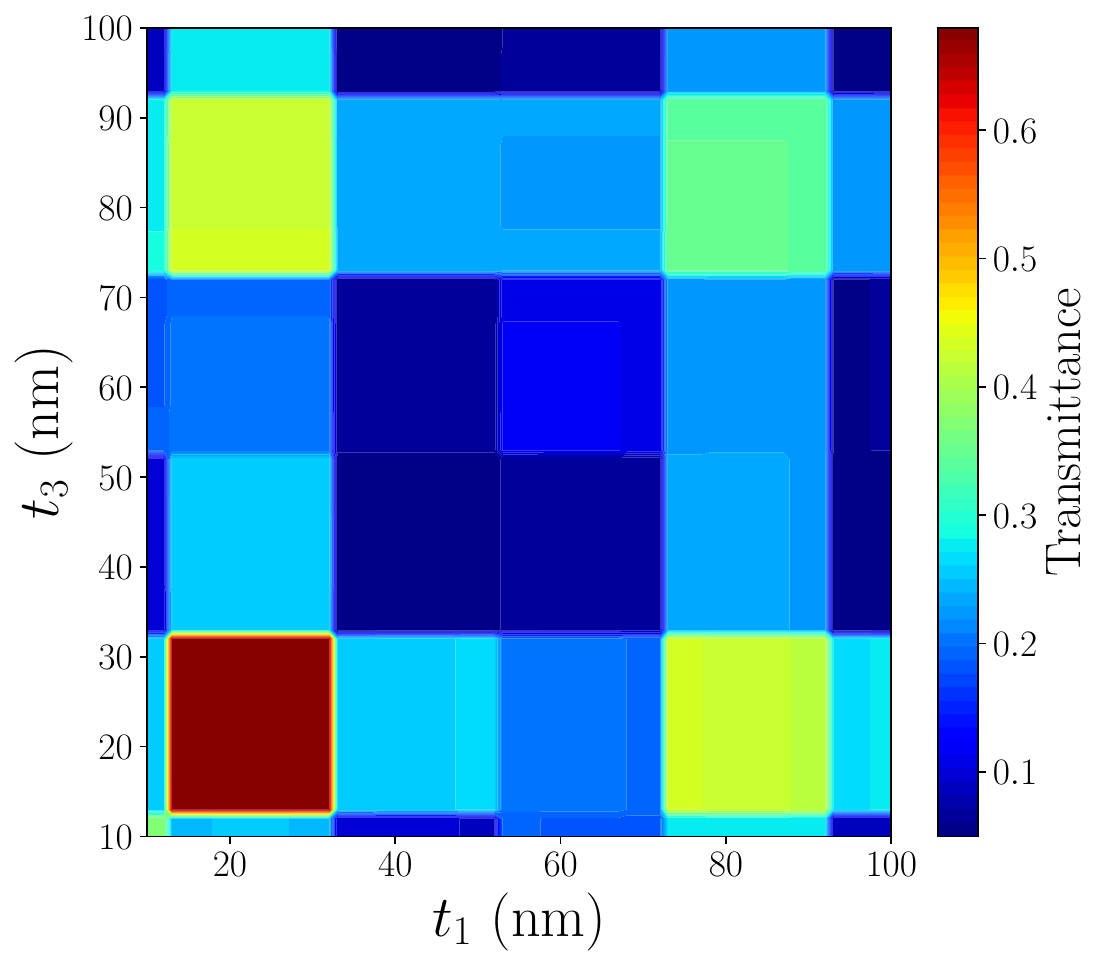}
        \includegraphics[width=0.31\textwidth]{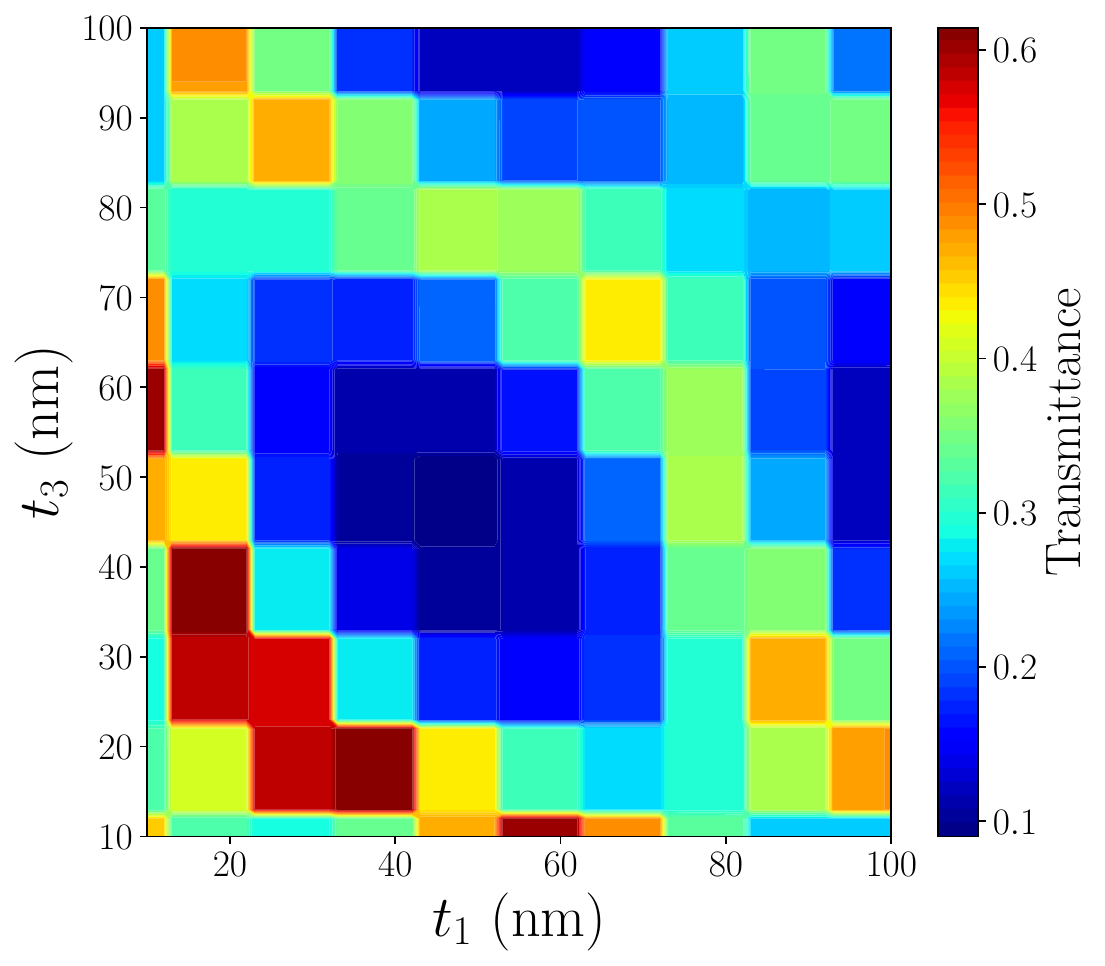}
        \includegraphics[width=0.31\textwidth]{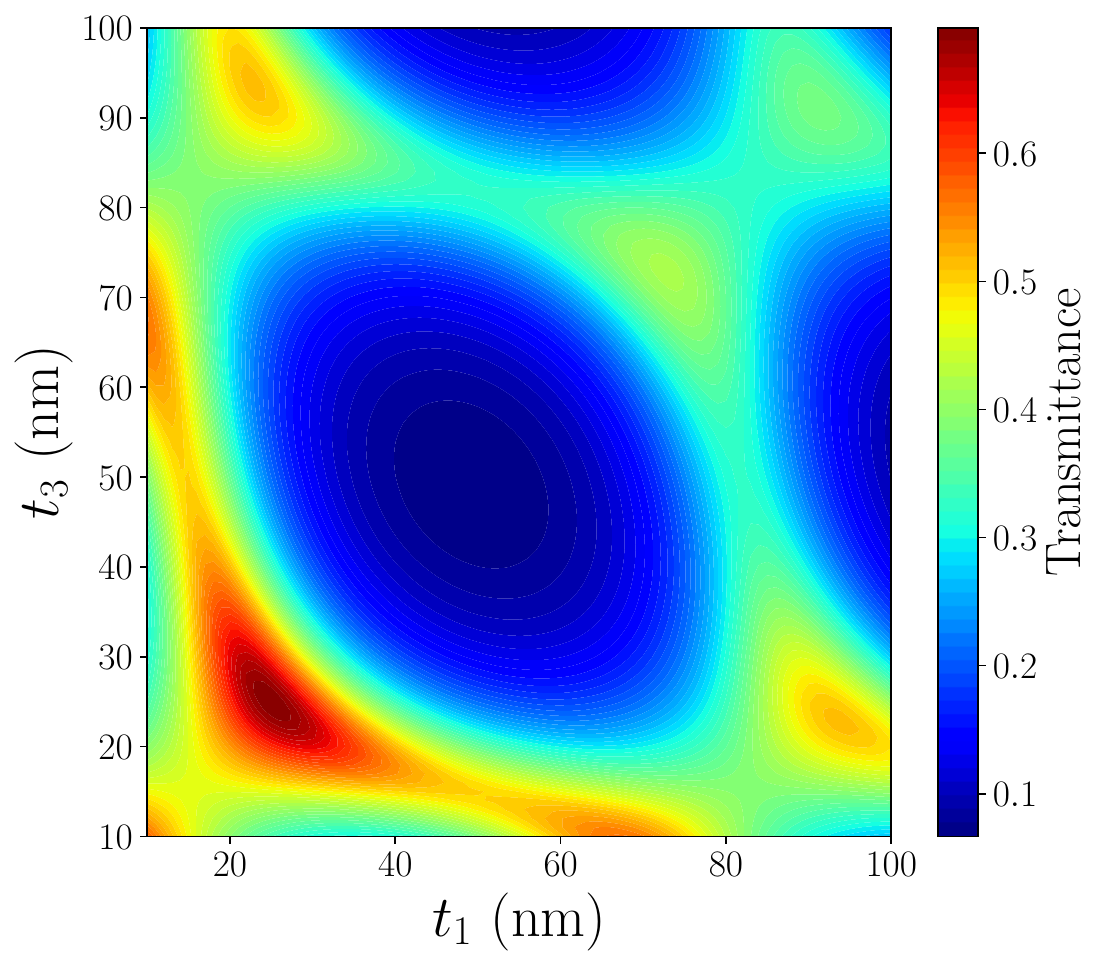}
        \caption{$t_2 = 15$}
    \end{subfigure}
    \caption{Visualization of the transmittance of the three-layer film made of \ce{cSi}/\ce{Ag}/\ce{cSi} for three different fidelity levels, i.e., low fidelity (shown in left panels), medium fidelity (shown in center panels), and high fidelity (shown in right panels).}
    \label{fig:threelayers_csi_ag_csi}
\end{figure}

\begin{figure}[t]
    \centering
    \begin{subfigure}[b]{\textwidth}
        \centering
        \includegraphics[width=0.31\textwidth]{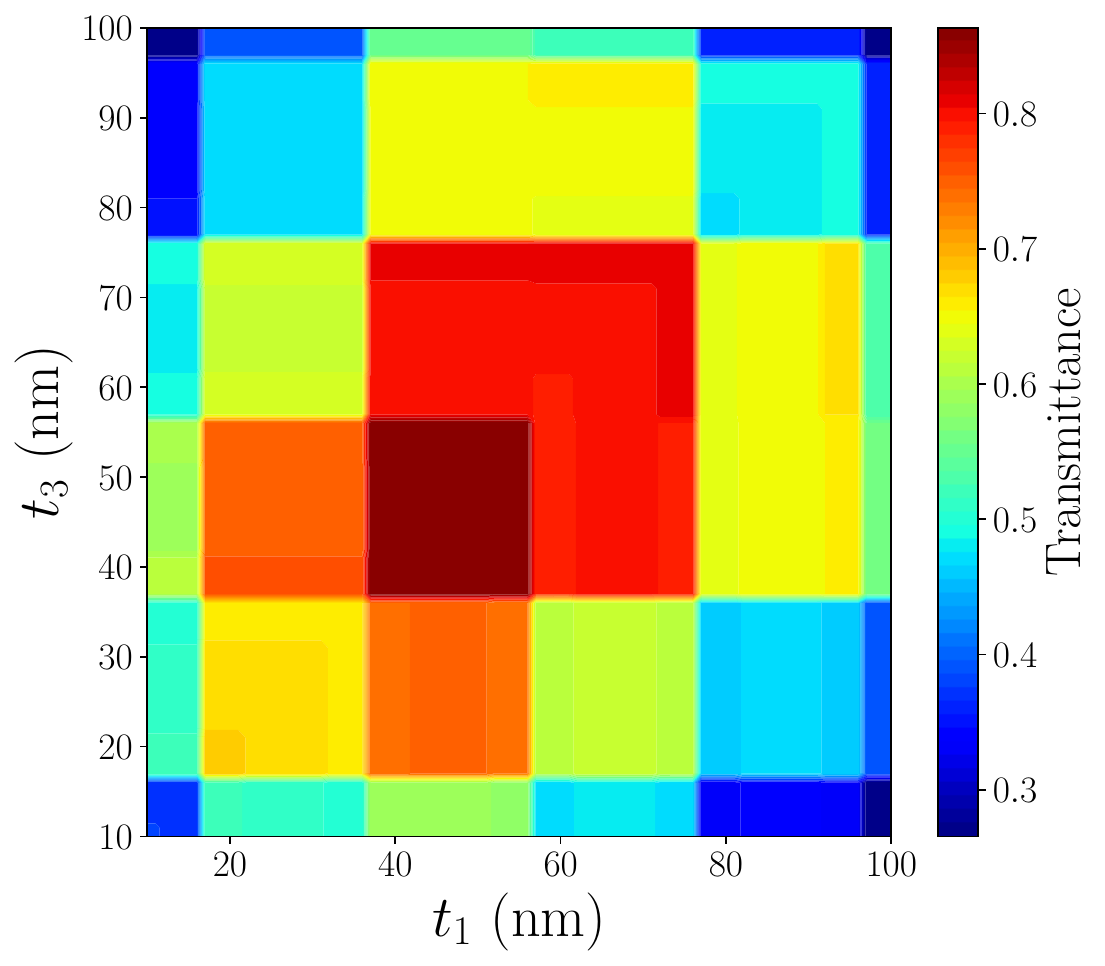}
        \includegraphics[width=0.31\textwidth]{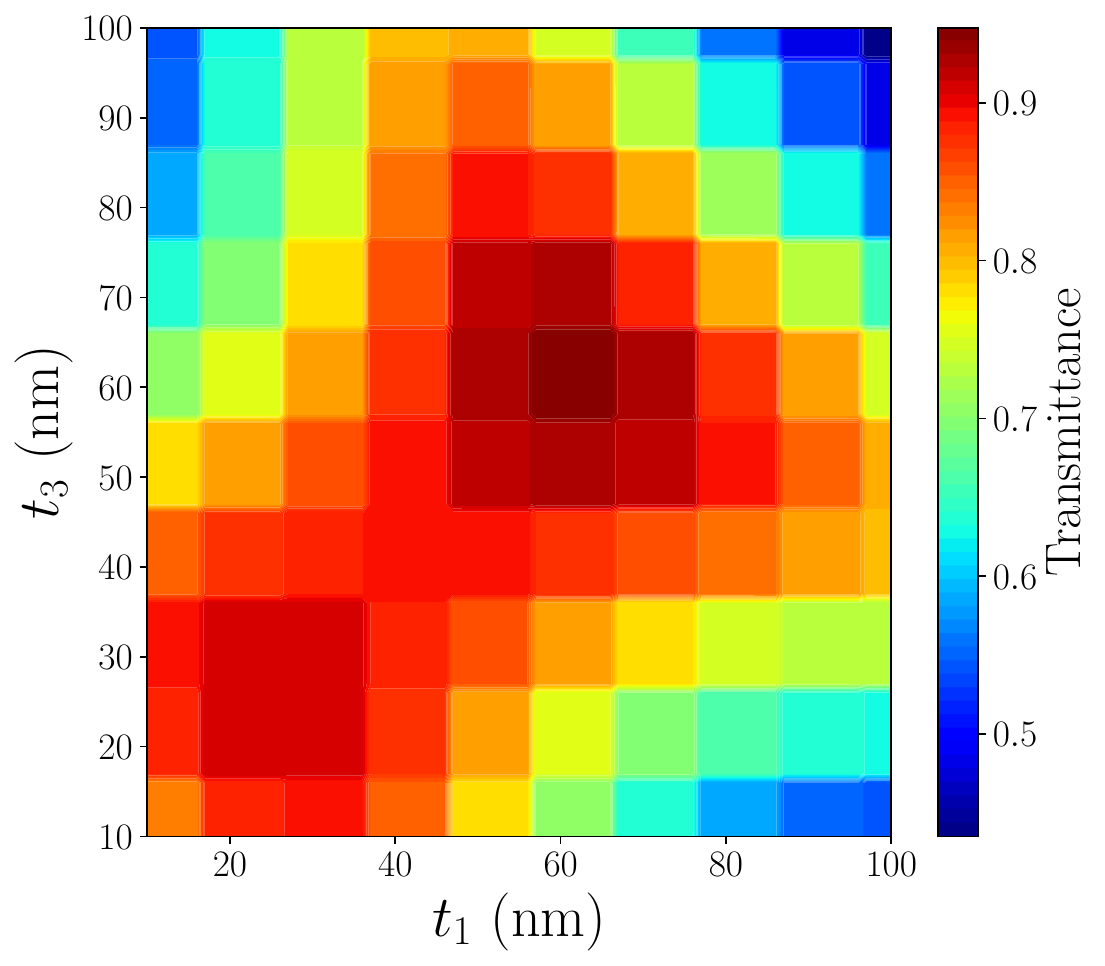}
        \includegraphics[width=0.31\textwidth]{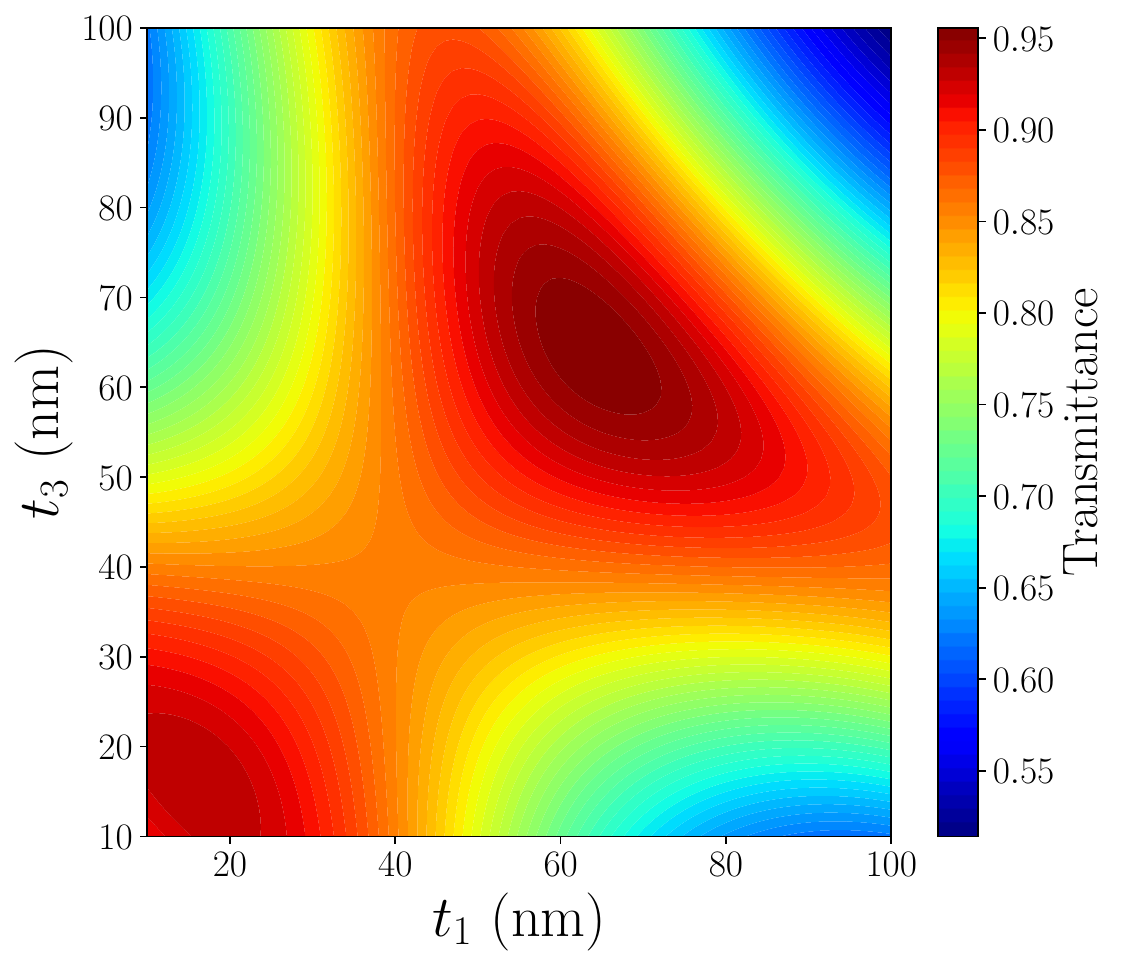}
        \caption{$t_2 = 7$}
    \end{subfigure}
    \begin{subfigure}[b]{\textwidth}
        \centering
        \includegraphics[width=0.31\textwidth]{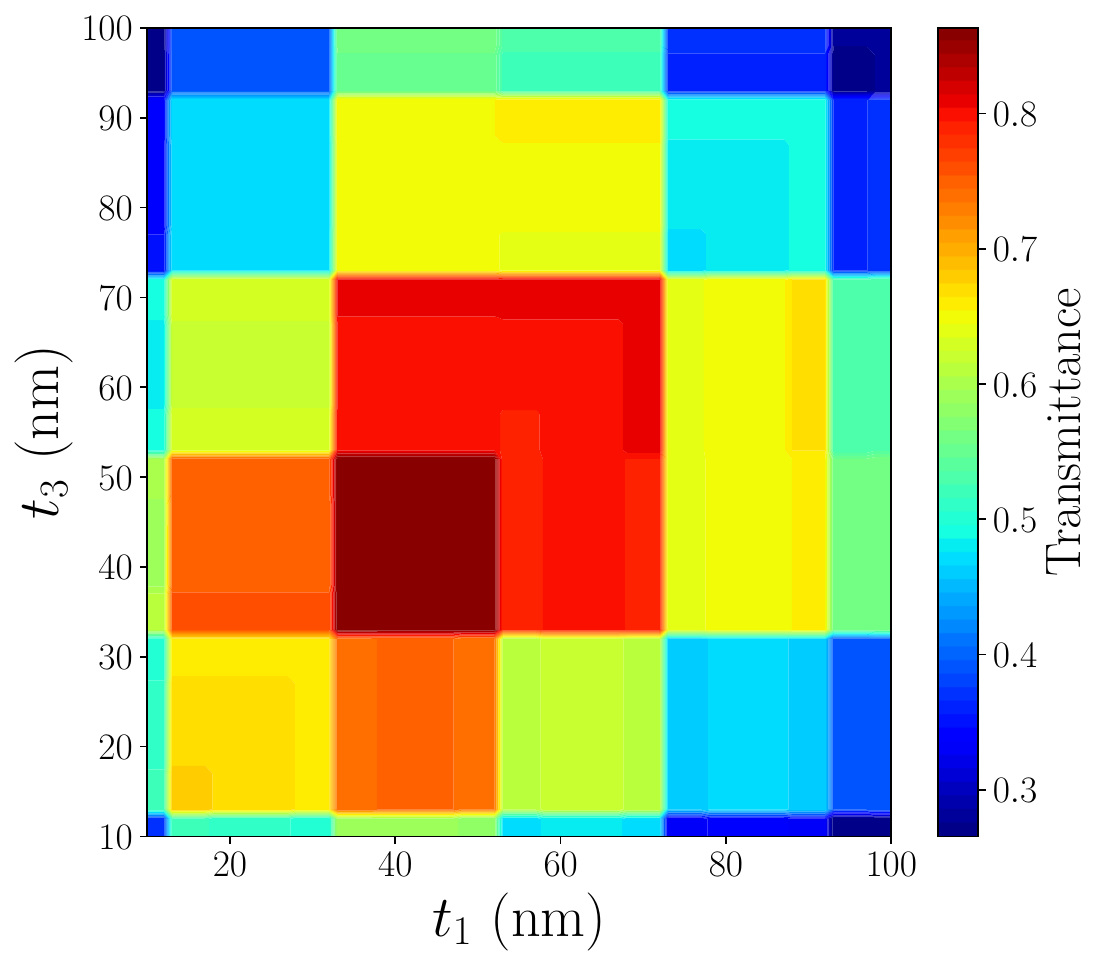}
        \includegraphics[width=0.31\textwidth]{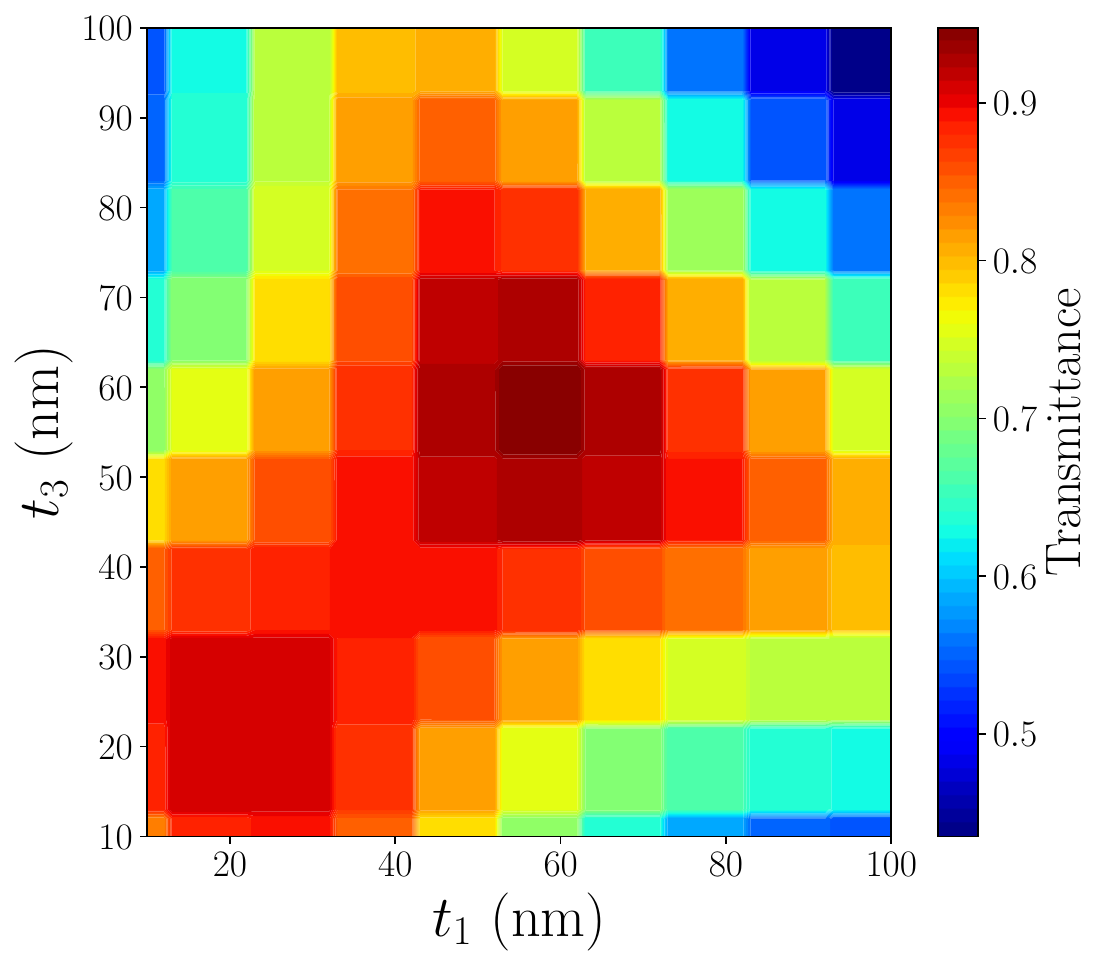}
        \includegraphics[width=0.31\textwidth]{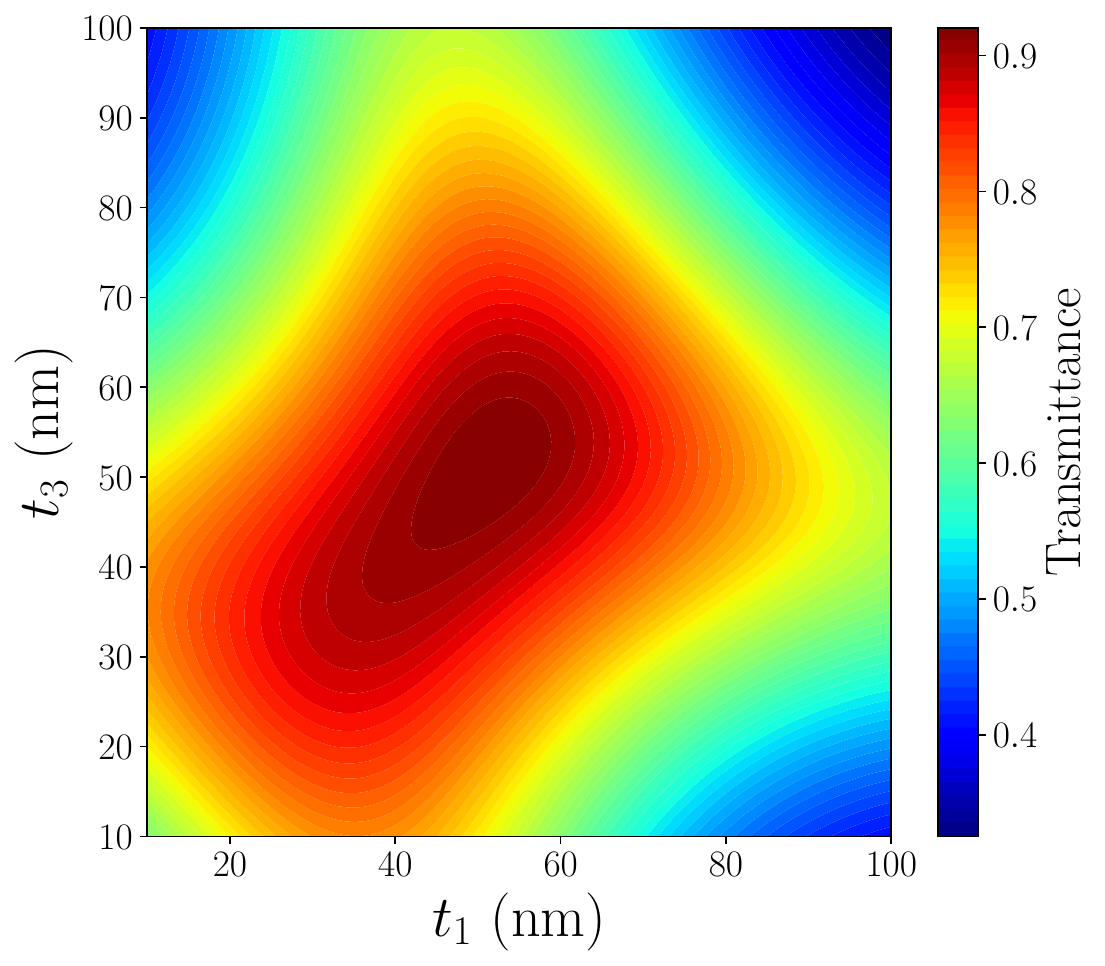}
        \caption{$t_2 = 15$}
    \end{subfigure}
    \caption{Visualization of the transmittance of the three-layer film made of \ce{ITO}/\ce{Ag}/\ce{ITO} for three different fidelity levels, i.e., low fidelity (shown in left panels), medium fidelity (shown in center panels), and high fidelity (shown in right panels).}
    \label{fig:threelayers_ito_ag_ito}
\end{figure}

\begin{figure}[t]
    \centering
    \begin{subfigure}[b]{\textwidth}
        \centering
        \includegraphics[width=0.31\textwidth]{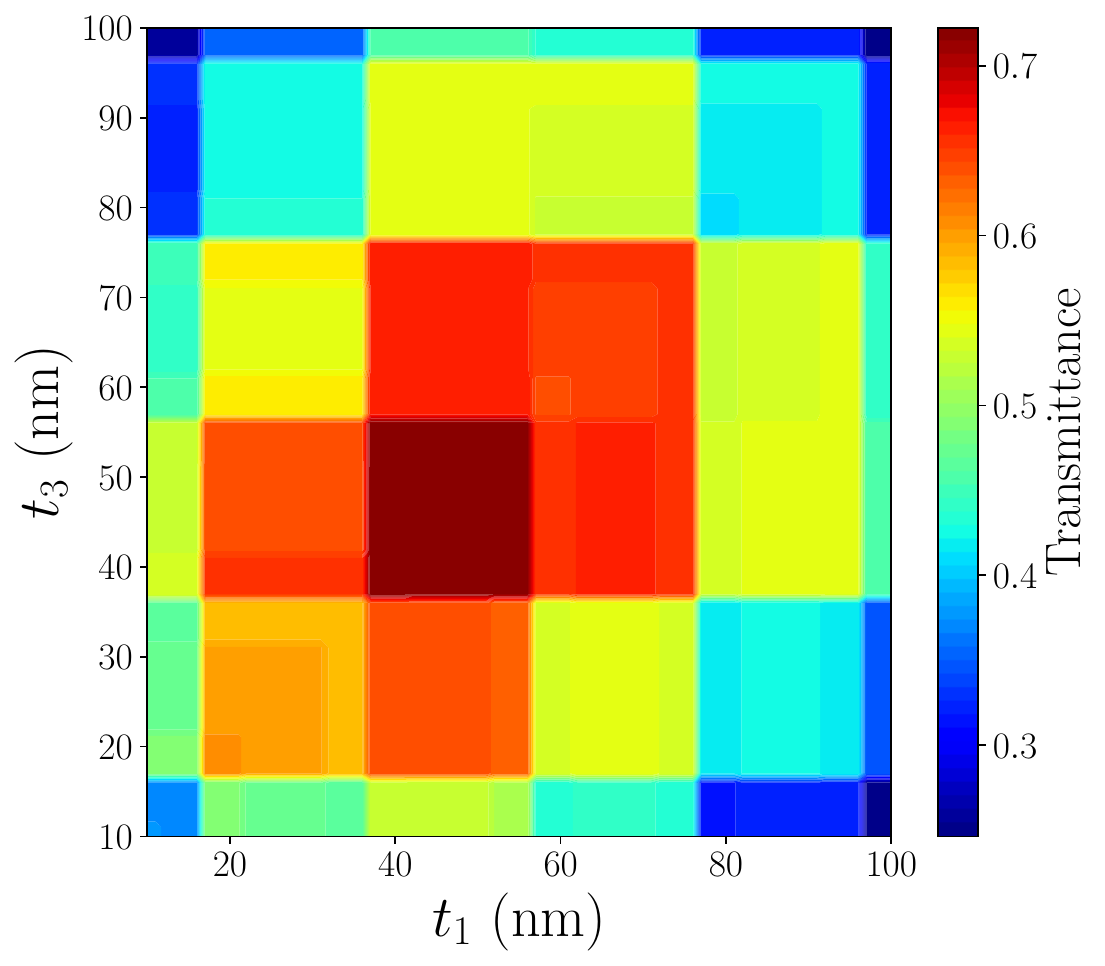}
        \includegraphics[width=0.31\textwidth]{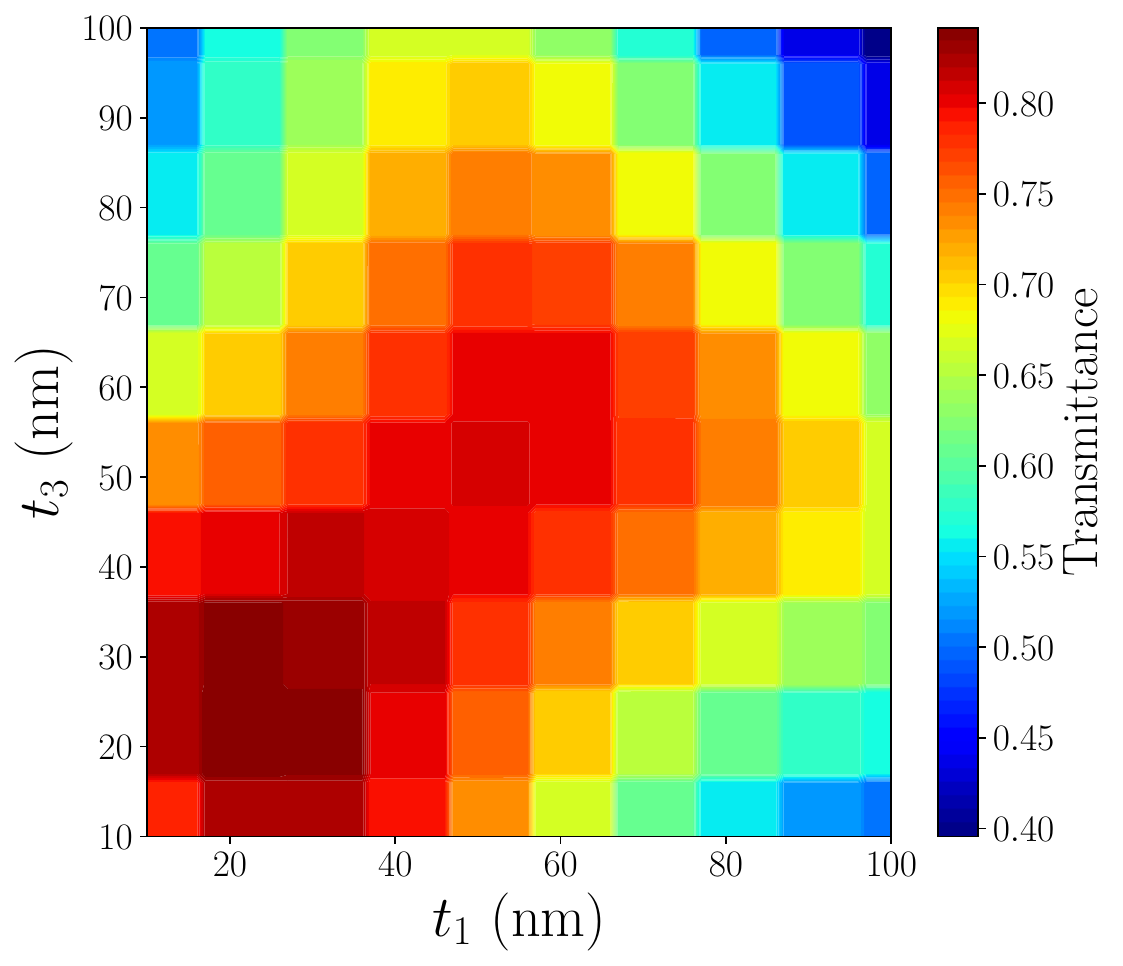}
        \includegraphics[width=0.31\textwidth]{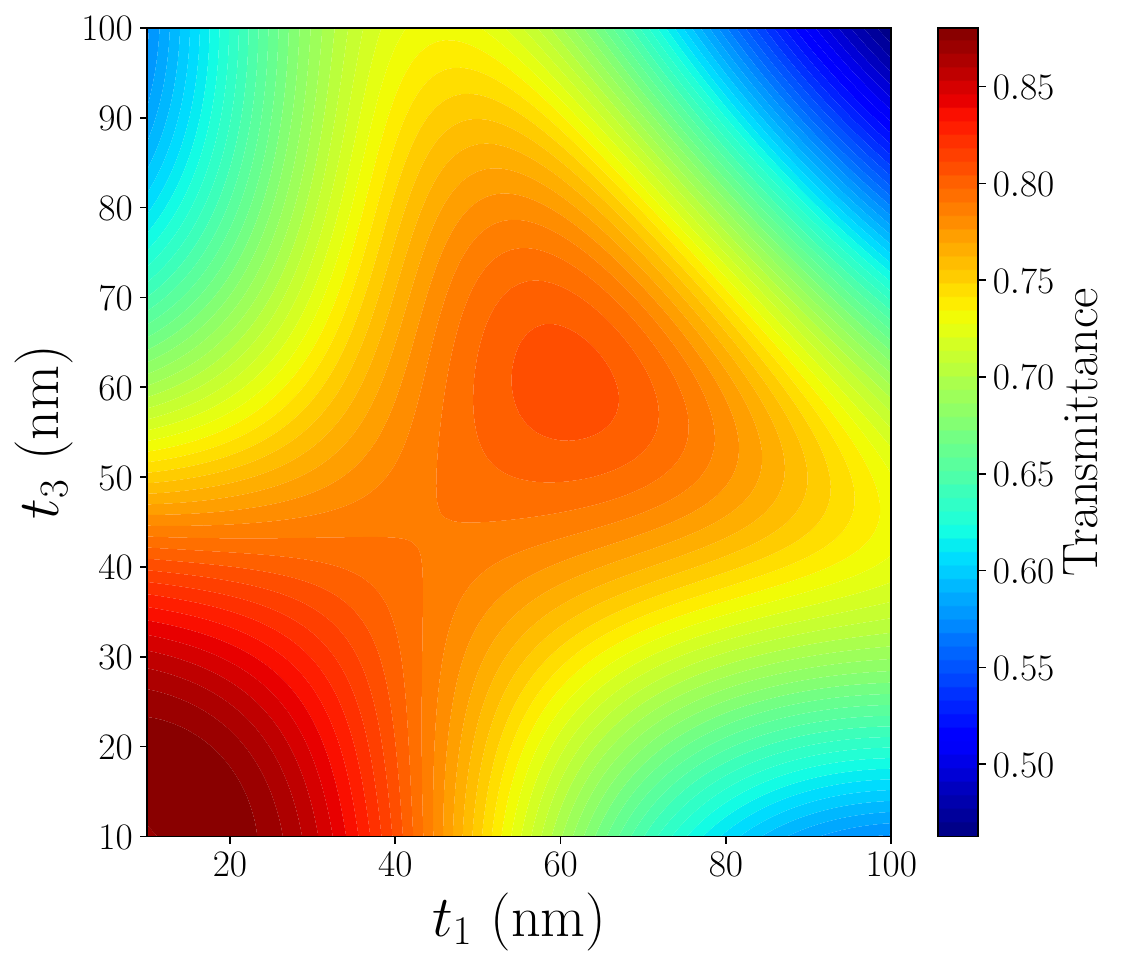}
        \caption{$t_2 = 7$}
    \end{subfigure}
    \begin{subfigure}[b]{\textwidth}
        \centering
        \includegraphics[width=0.31\textwidth]{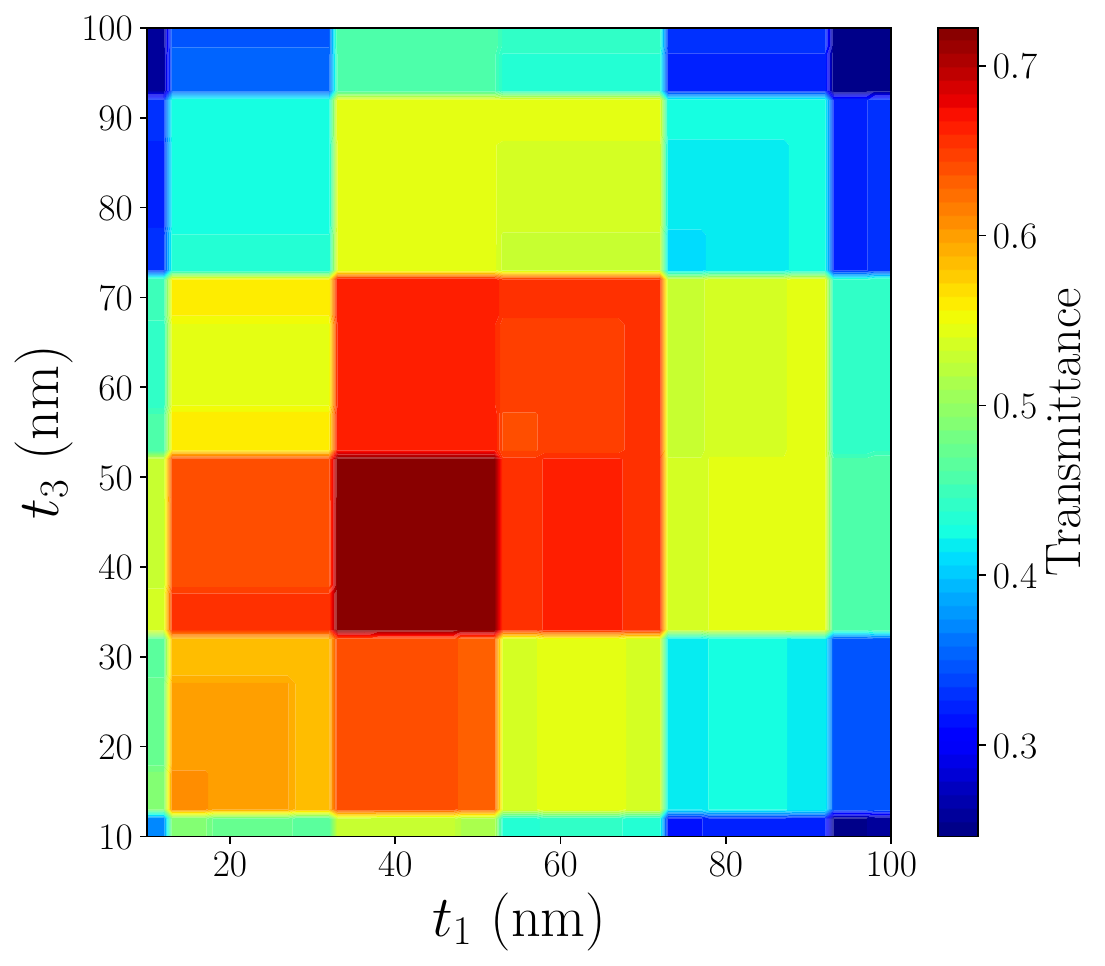}
        \includegraphics[width=0.31\textwidth]{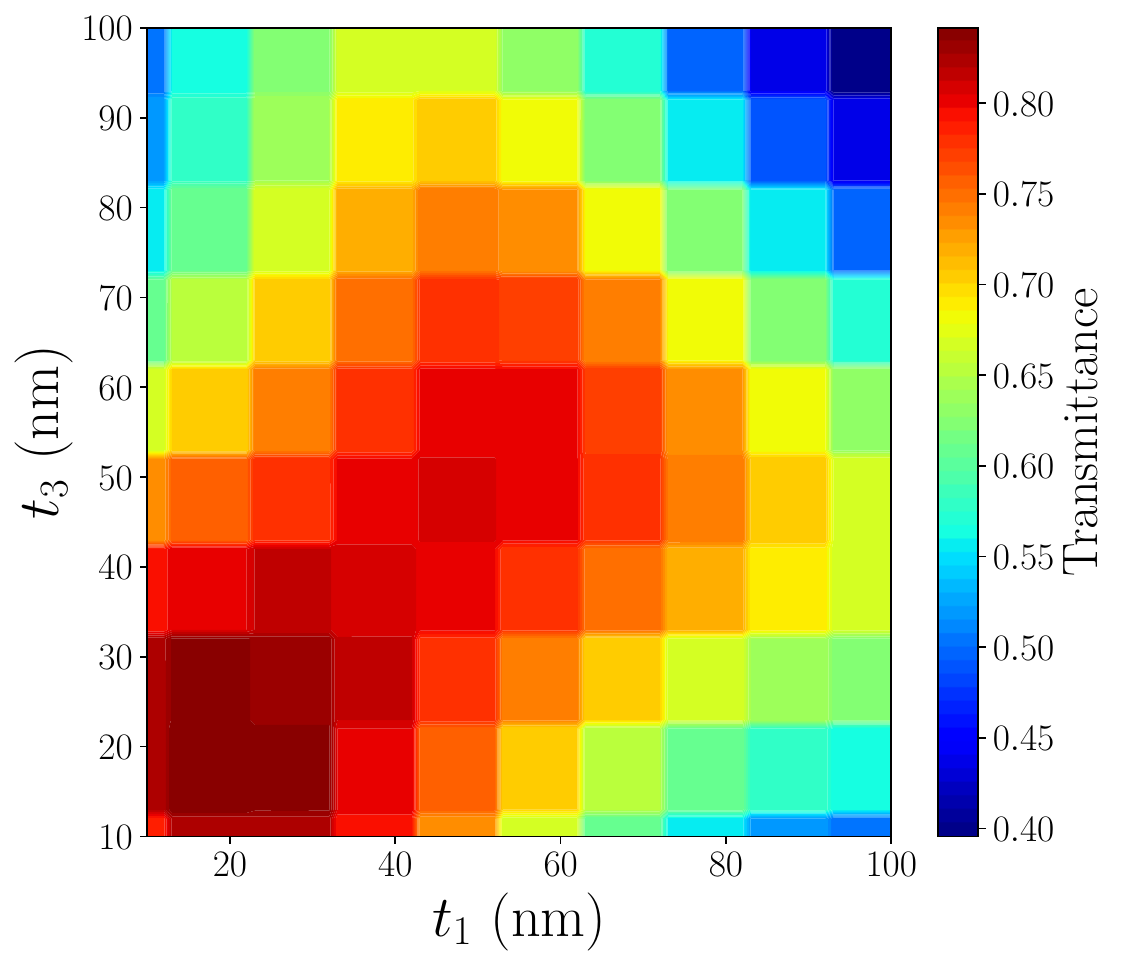}
        \includegraphics[width=0.31\textwidth]{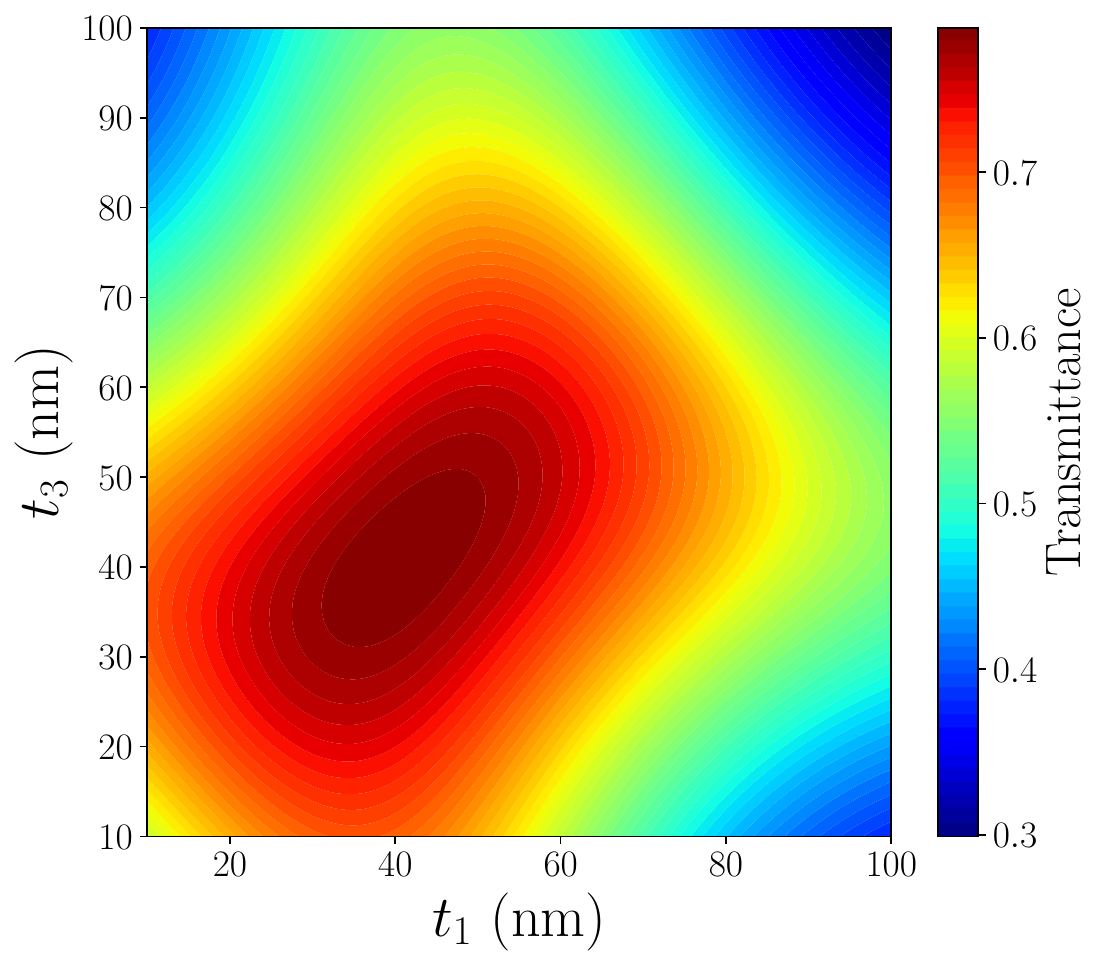}
        \caption{$t_2 = 15$}
    \end{subfigure}
    \caption{Visualization of the transmittance of the three-layer film made of \ce{ZnO}/\ce{Ag}/\ce{ZnO} for three different fidelity levels, i.e., low fidelity (shown in left panels), medium fidelity (shown in center panels), and high fidelity (shown in right panels).}
    \label{fig:threelayers_zno_ag_zno}
\end{figure}

\begin{figure}[t]
    \centering
    \begin{subfigure}[b]{0.31\textwidth}
        \centering
        \includegraphics[width=\textwidth]{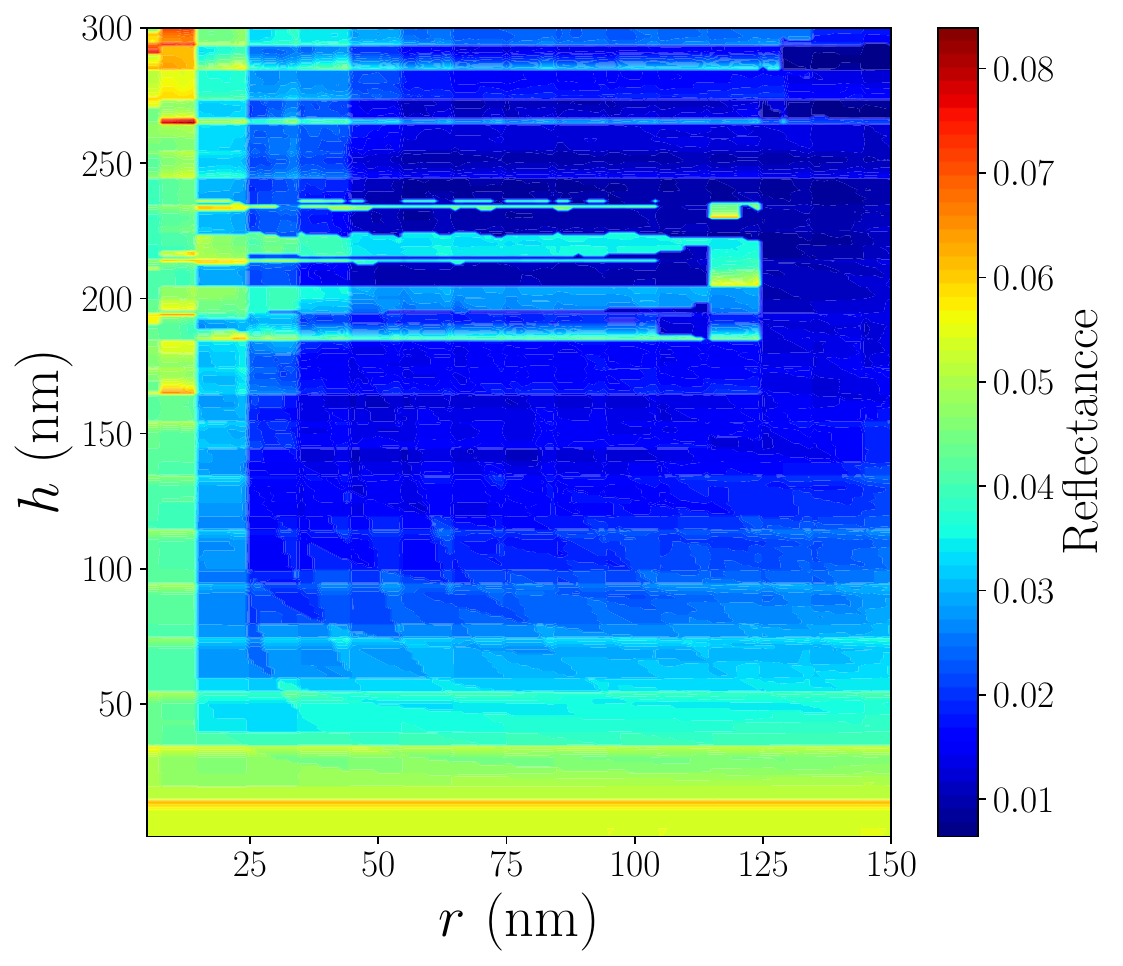}
        \caption{Low fidelity}
    \end{subfigure}
    \begin{subfigure}[b]{0.31\textwidth}
        \centering
        \includegraphics[width=\textwidth]{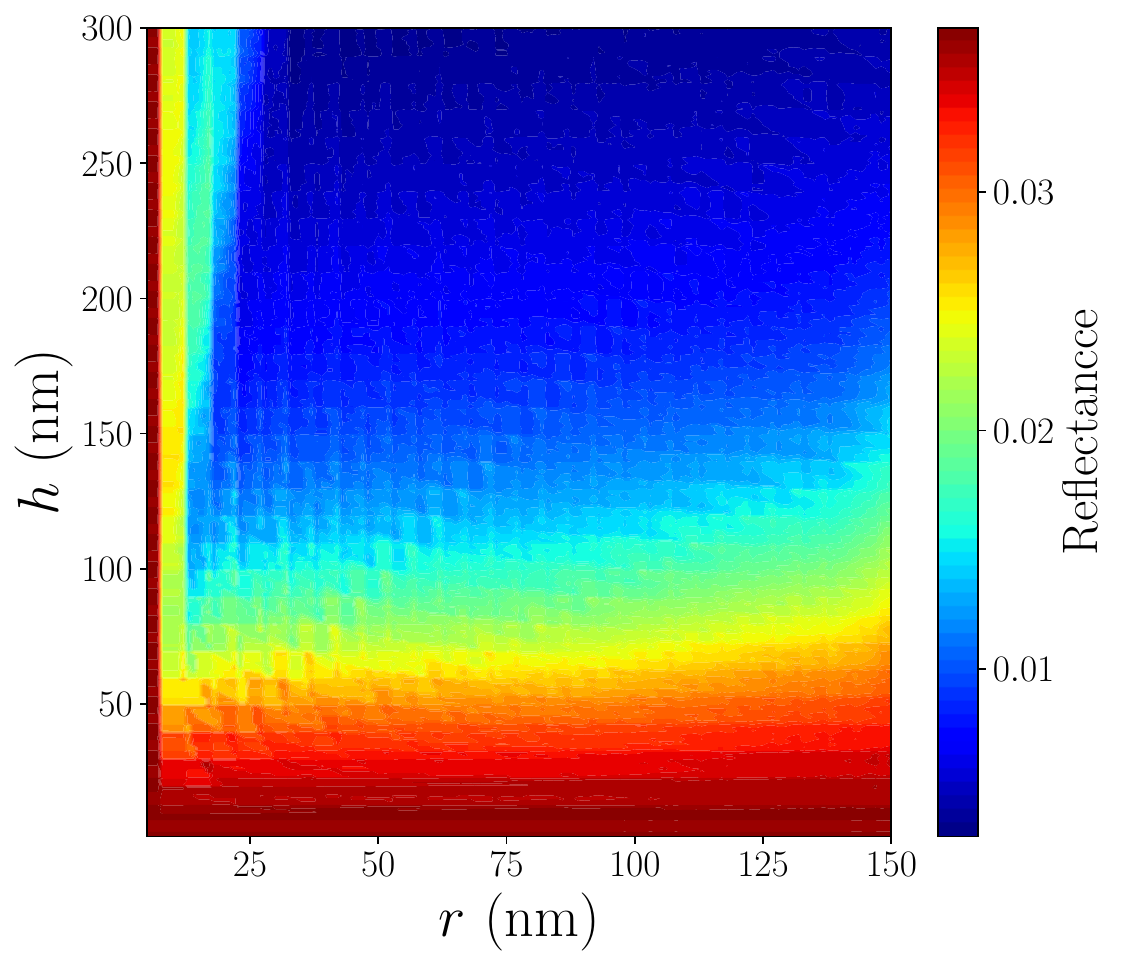}
        \caption{Medium fidelity}
    \end{subfigure}
    \begin{subfigure}[b]{0.31\textwidth}
        \centering
        \includegraphics[width=\textwidth]{figures/dataset_nanocones2d_fusedsilica_fusedsilica_0.1_reflectance_0.pdf}
        \caption{High fidelity}
    \end{subfigure}
    \caption{Visualization of the reflectance of the anti-reflective nanocones made of fused silica for three different fidelity levels.}
    \label{fig:nanocones_all}
\end{figure}

\begin{figure}[t]
    \centering
    \begin{subfigure}[b]{\textwidth}
        \centering
        \includegraphics[width=0.31\textwidth]{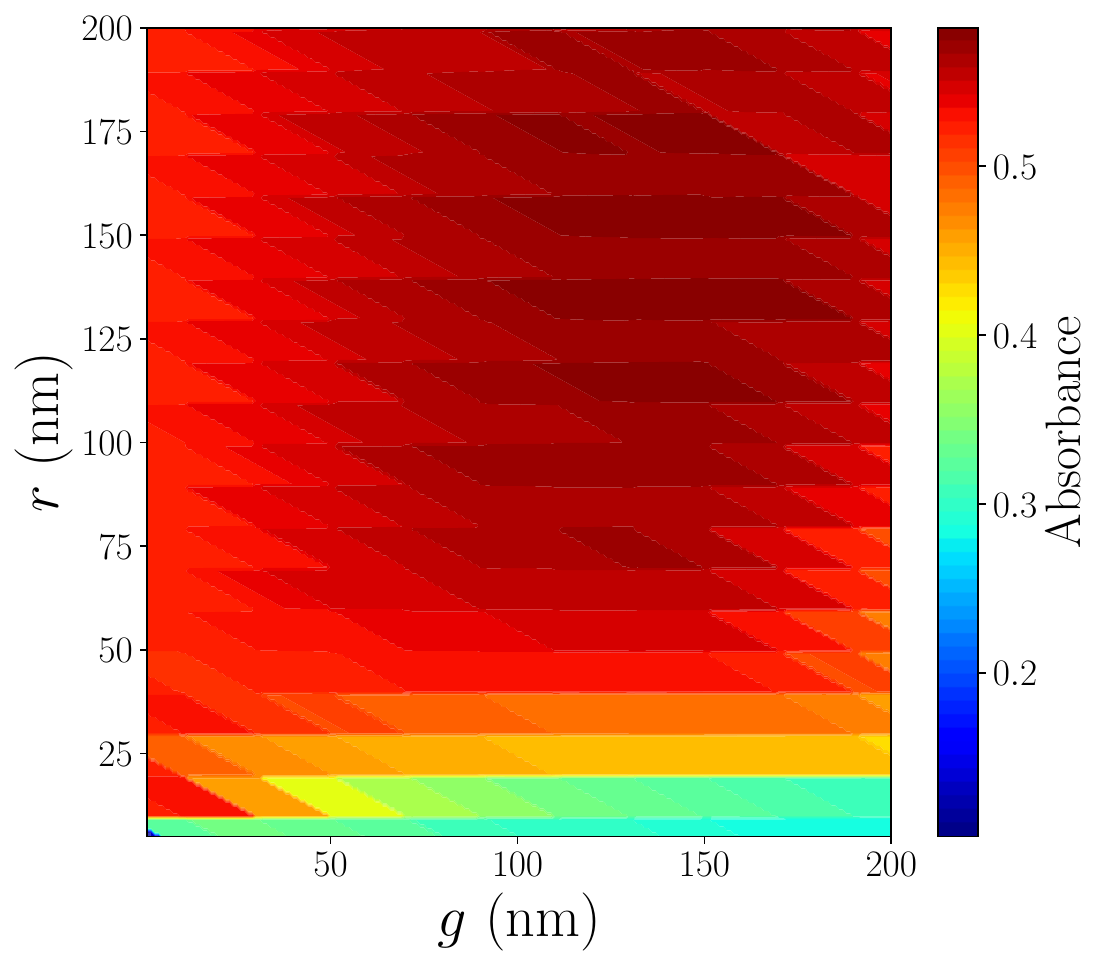}
        \includegraphics[width=0.31\textwidth]{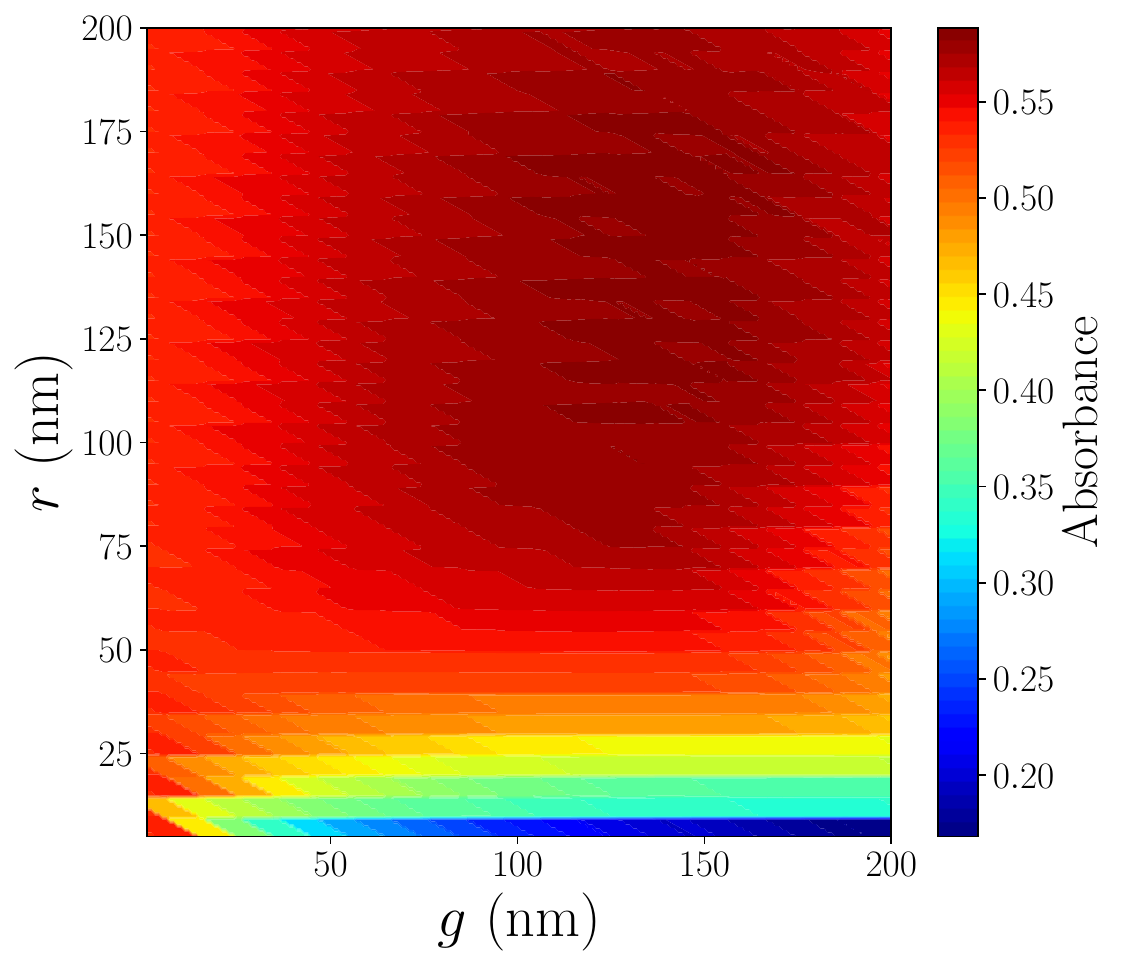}
        \includegraphics[width=0.31\textwidth]{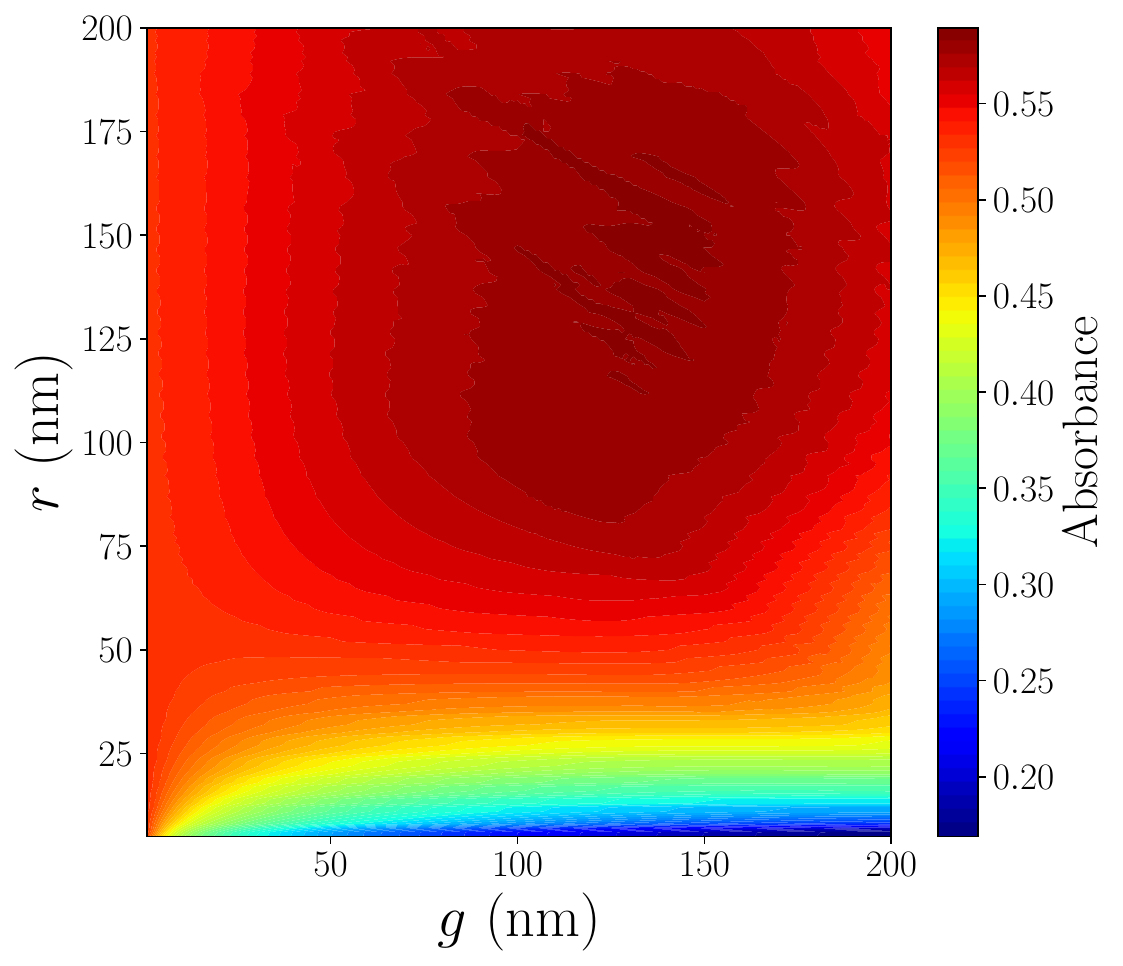}
        \caption{\ce{CH3NH3PbI3}}
    \end{subfigure}
    \begin{subfigure}[b]{\textwidth}
        \centering
        \includegraphics[width=0.31\textwidth]{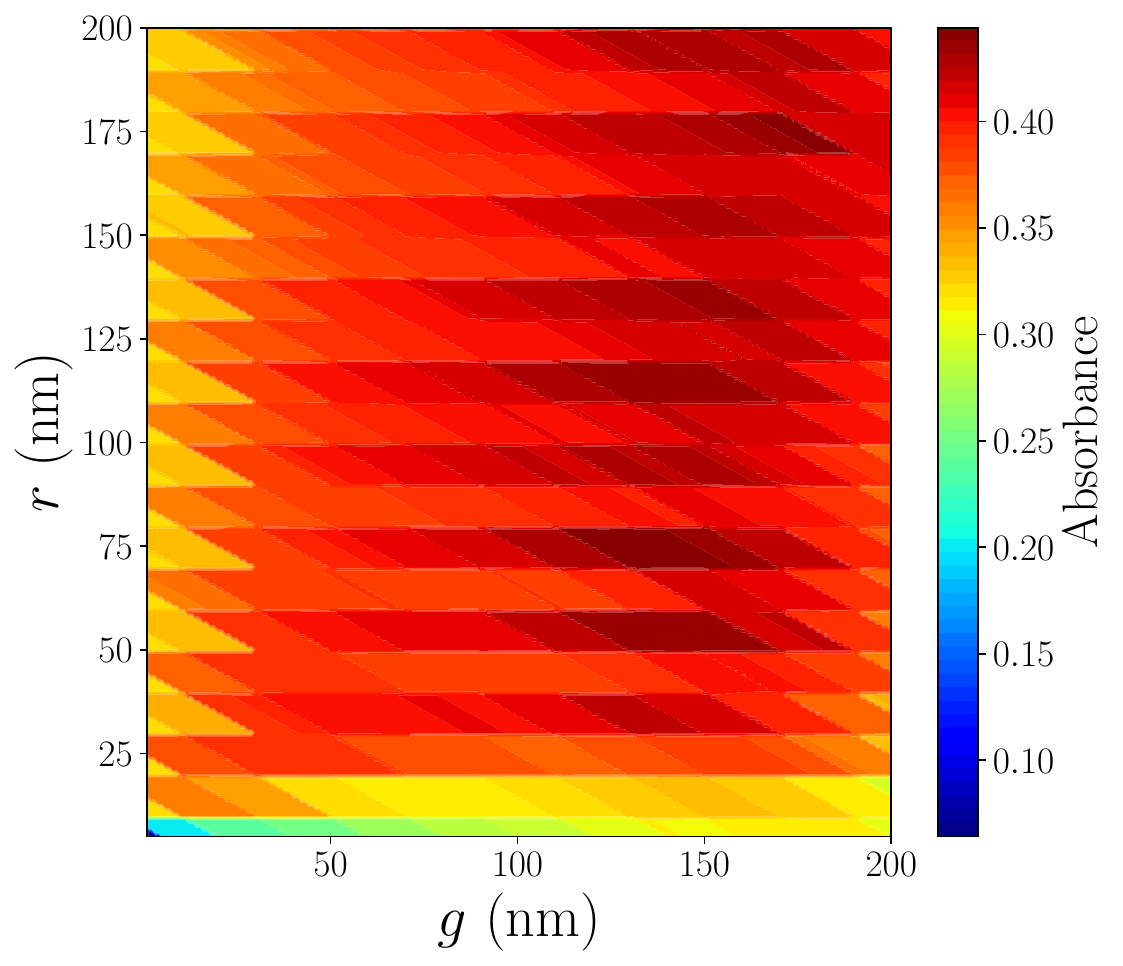}
        \includegraphics[width=0.31\textwidth]{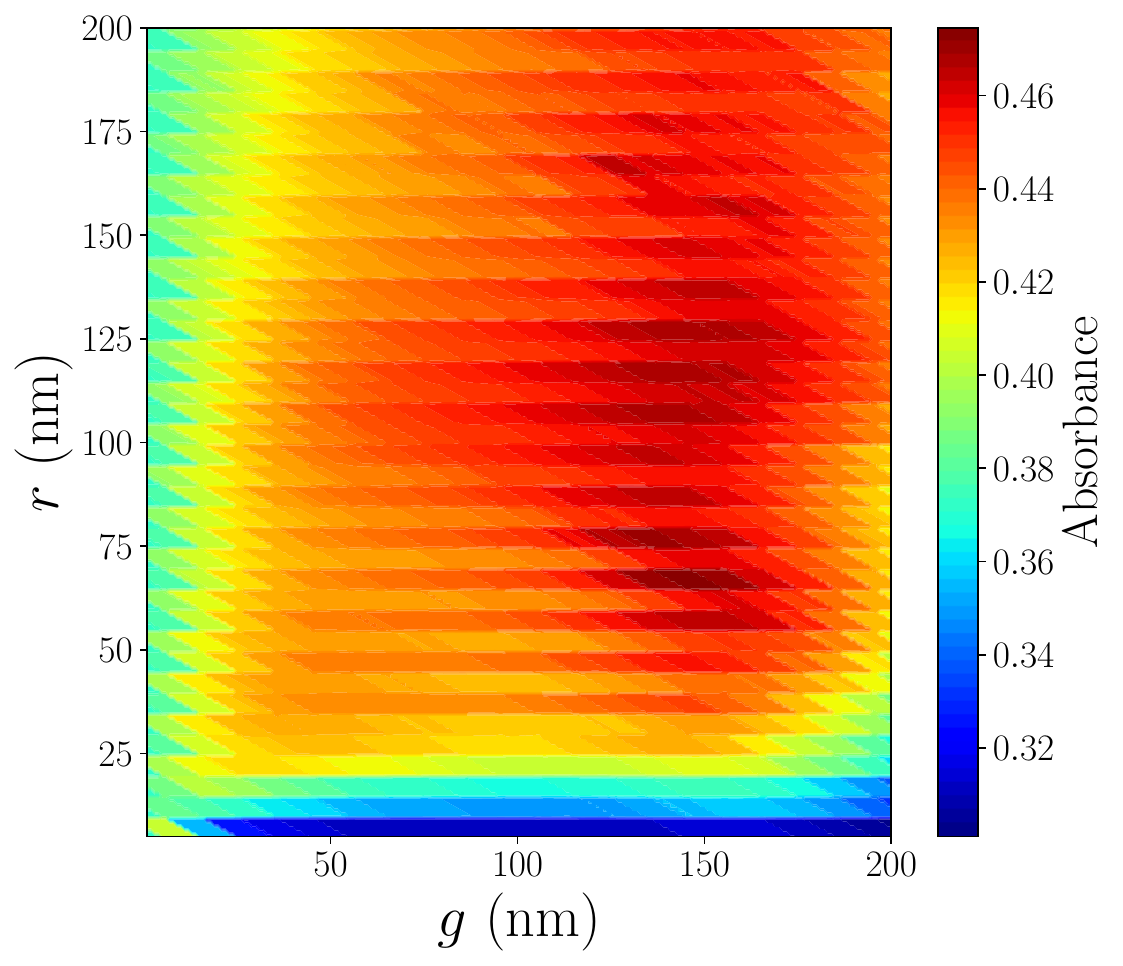}
        \includegraphics[width=0.31\textwidth]{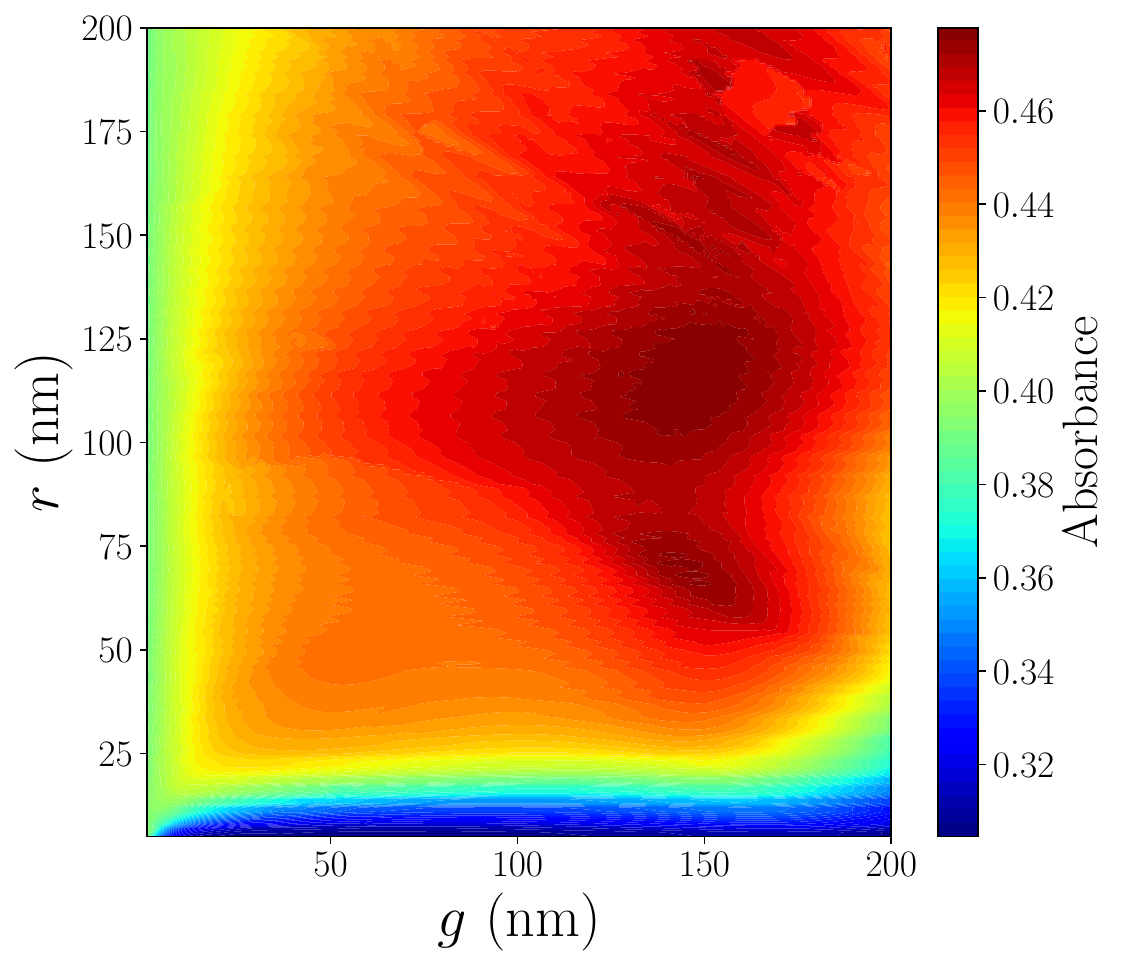}
        \caption{\ce{GaAs}}
    \end{subfigure}
    \caption{Visualization of the absorbance of the vertical nanowires made of \ce{CH3NH3PbI3} or \ce{GaAs} for three different fidelity levels, i.e., low fidelity (shown in left panels), medium fidelity (shown in center panels), and high fidelity (shown in right panels).}
    \label{fig:nanowires_all}
\end{figure}

\begin{figure}[t]
    \centering
    \begin{subfigure}[b]{\textwidth}
        \centering
        \includegraphics[width=0.31\textwidth]{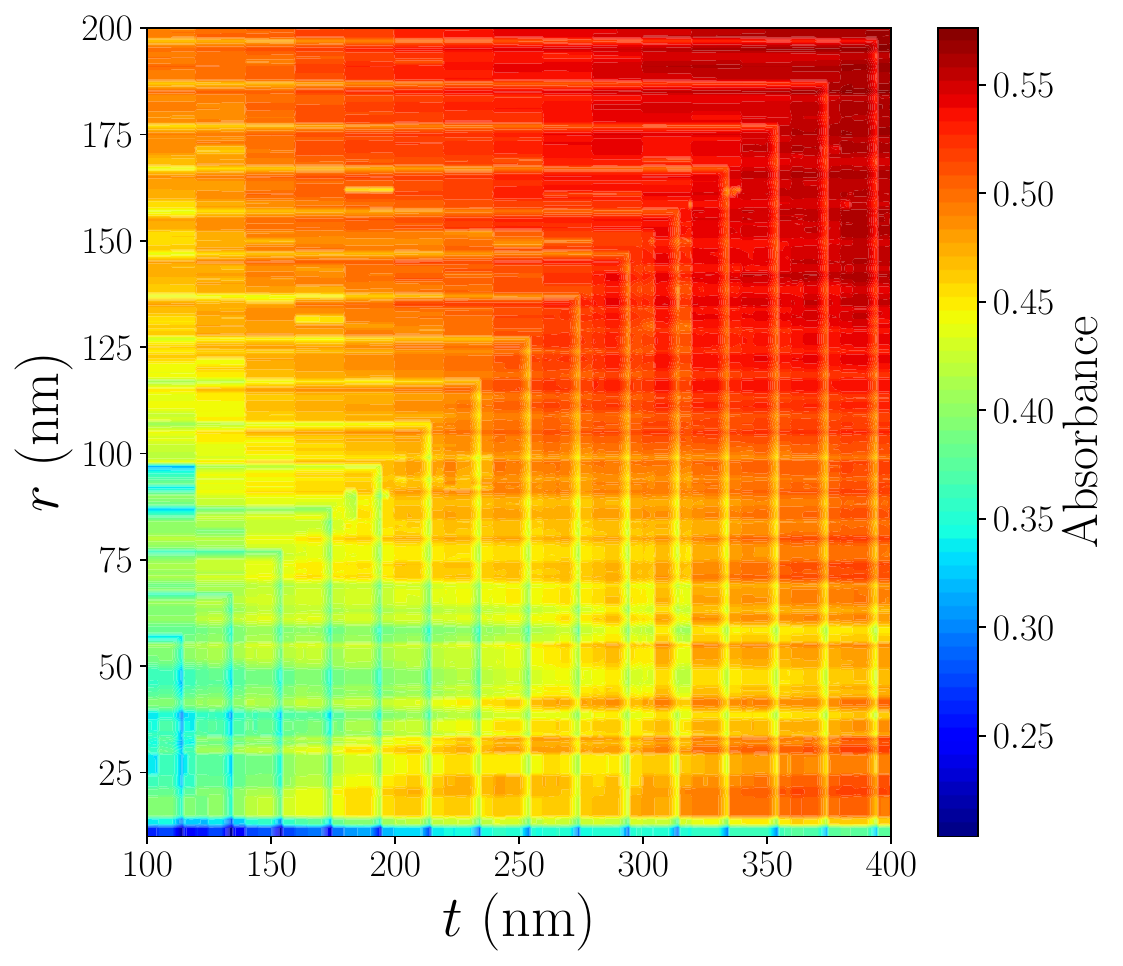}
        \includegraphics[width=0.31\textwidth]{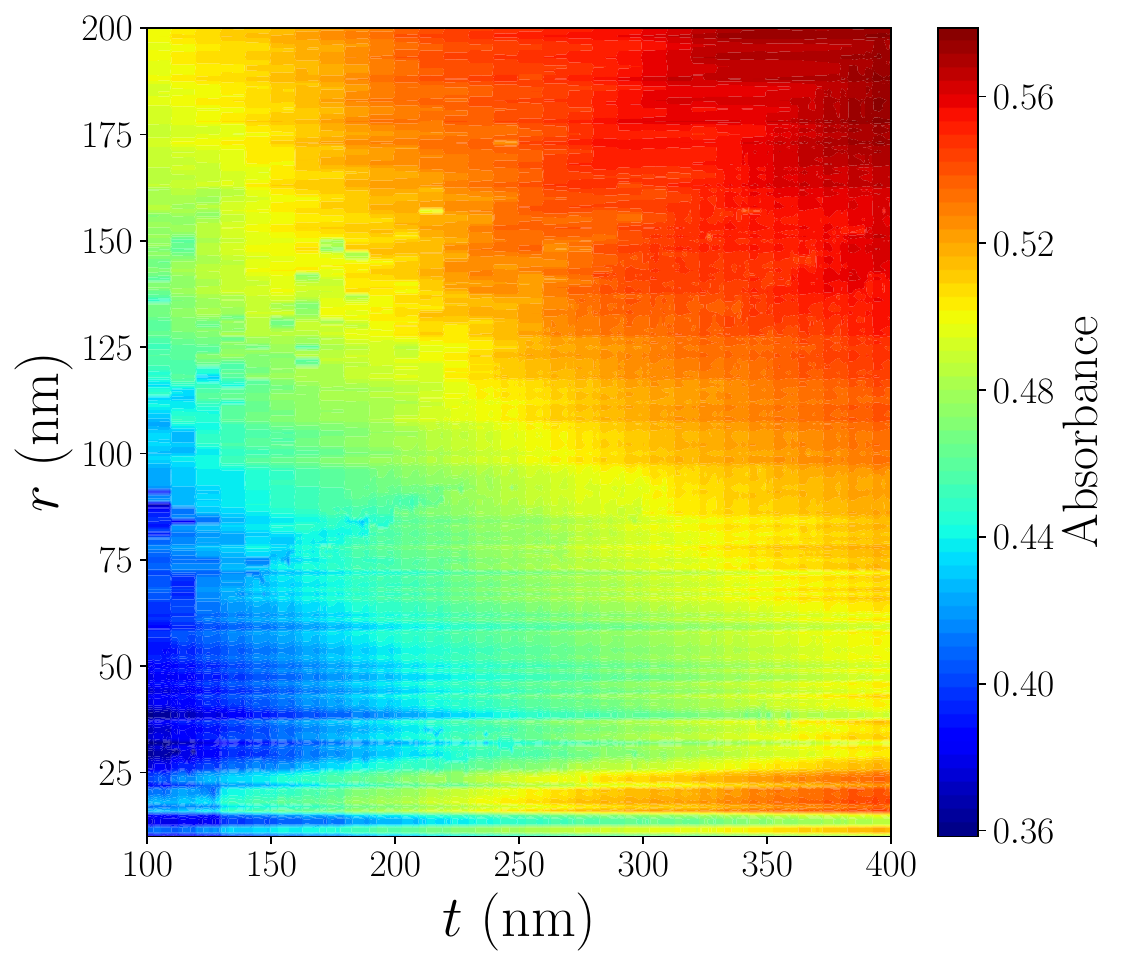}
        \includegraphics[width=0.31\textwidth]{figures/dataset_nanospheres2d_cSi_TiO2_0.1_absorbance_0.pdf}
        \caption{\ce{cSi}}
    \end{subfigure}
    \begin{subfigure}[b]{\textwidth}
        \centering
        \includegraphics[width=0.31\textwidth]{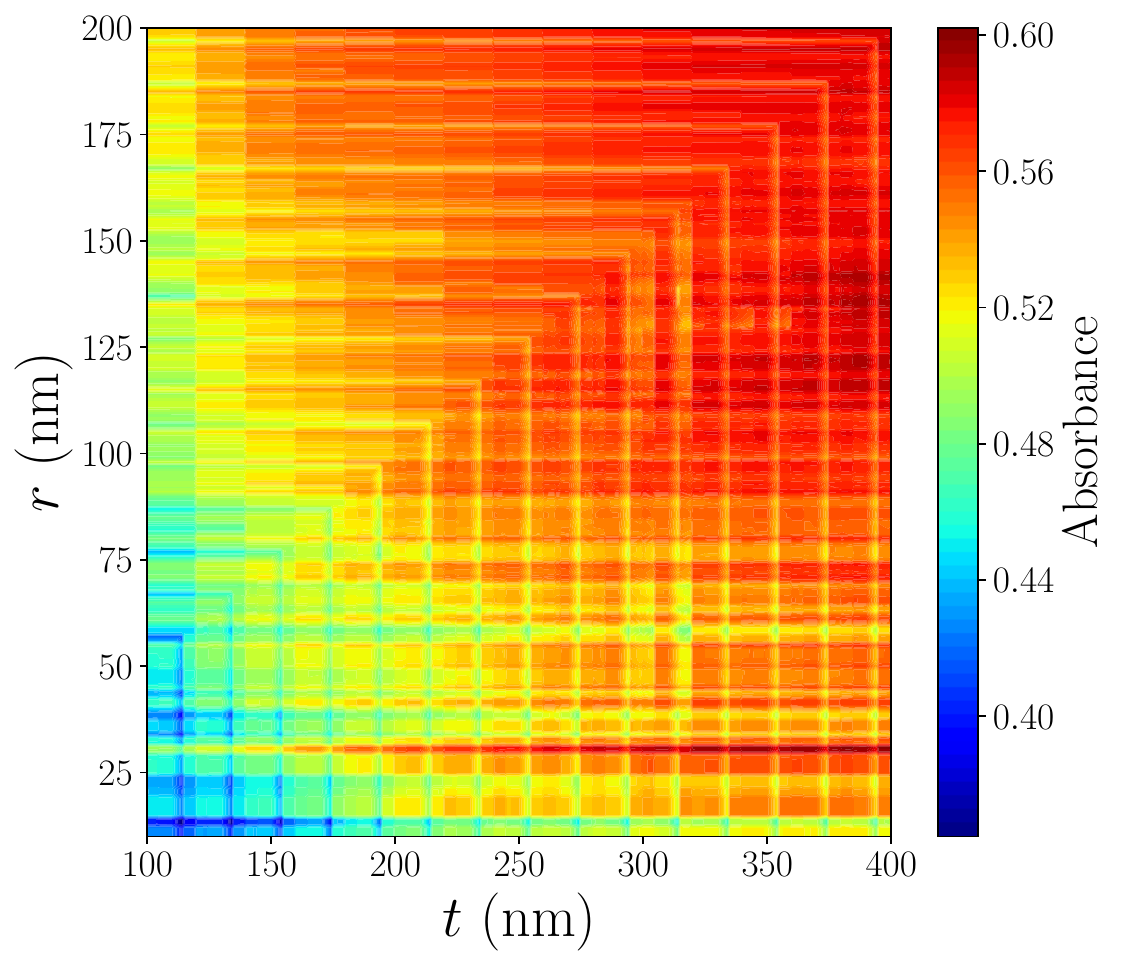}
        \includegraphics[width=0.31\textwidth]{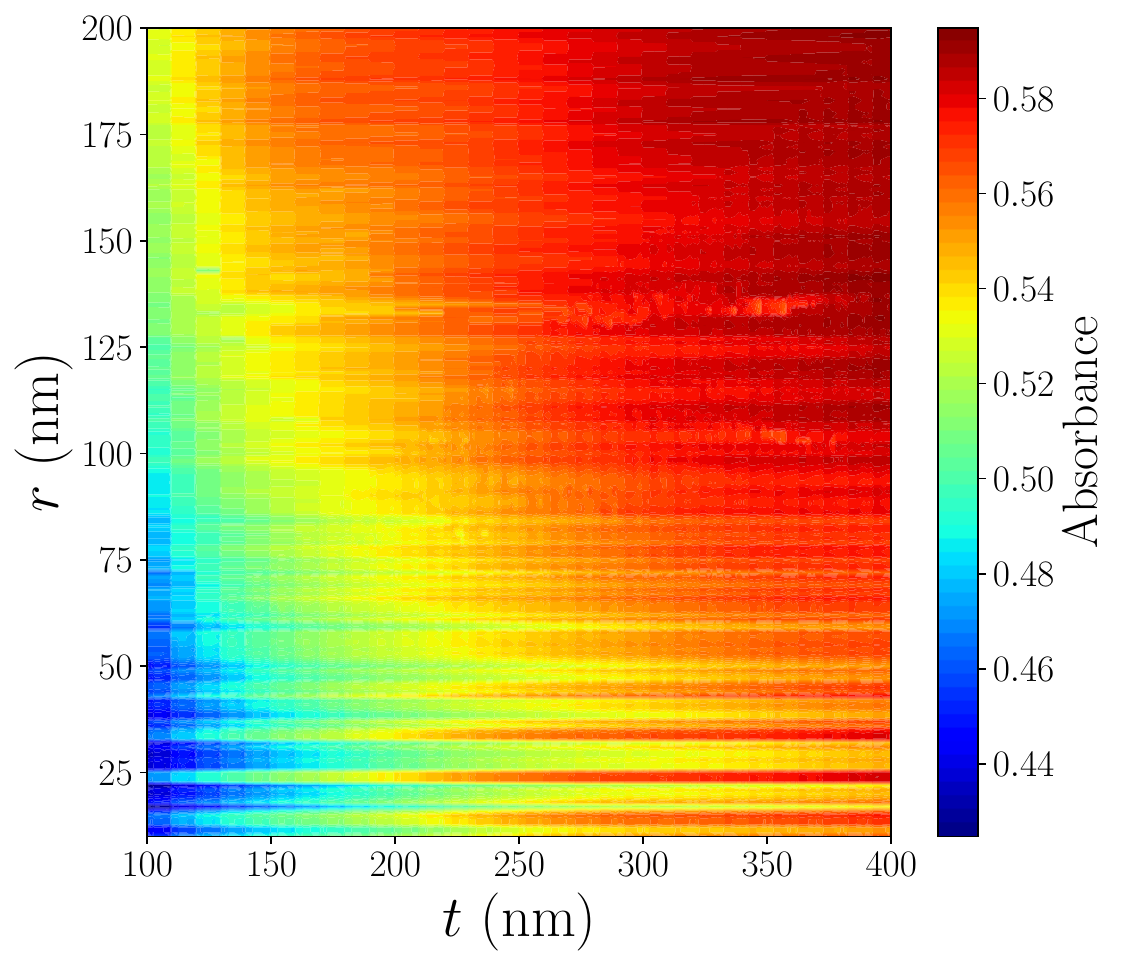}
        \includegraphics[width=0.31\textwidth]{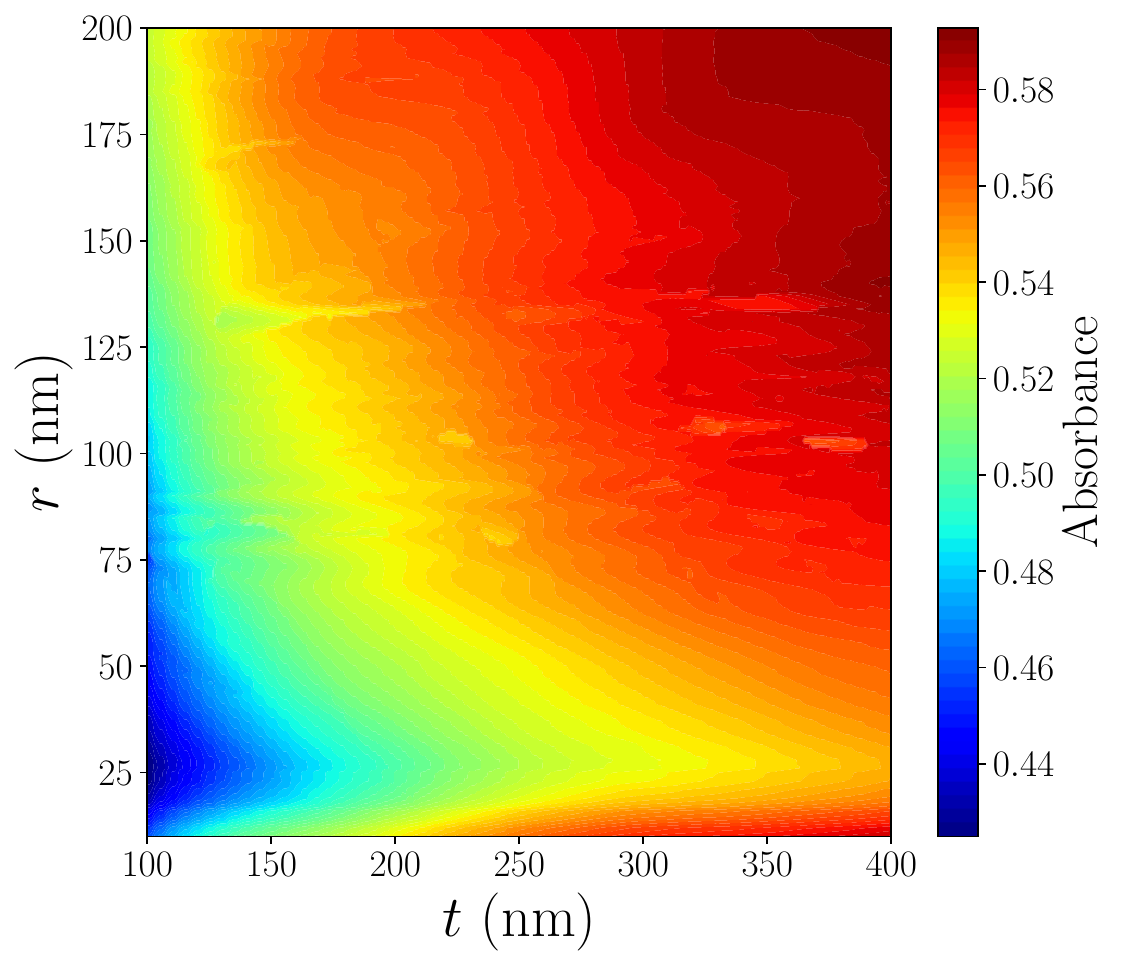}
        \caption{\ce{CH3NH3PbI3}}
    \end{subfigure}
    \begin{subfigure}[b]{\textwidth}
        \centering
        \includegraphics[width=0.31\textwidth]{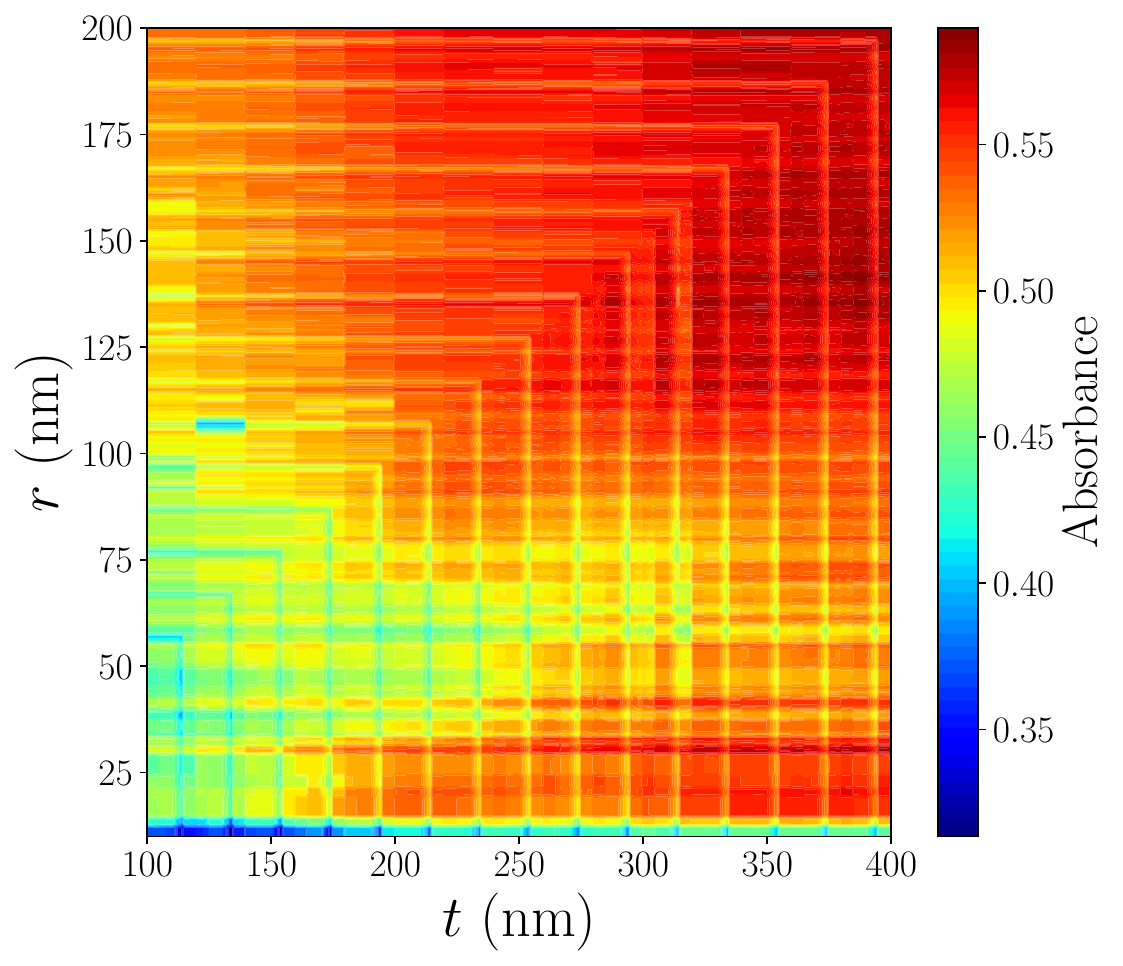}
        \includegraphics[width=0.31\textwidth]{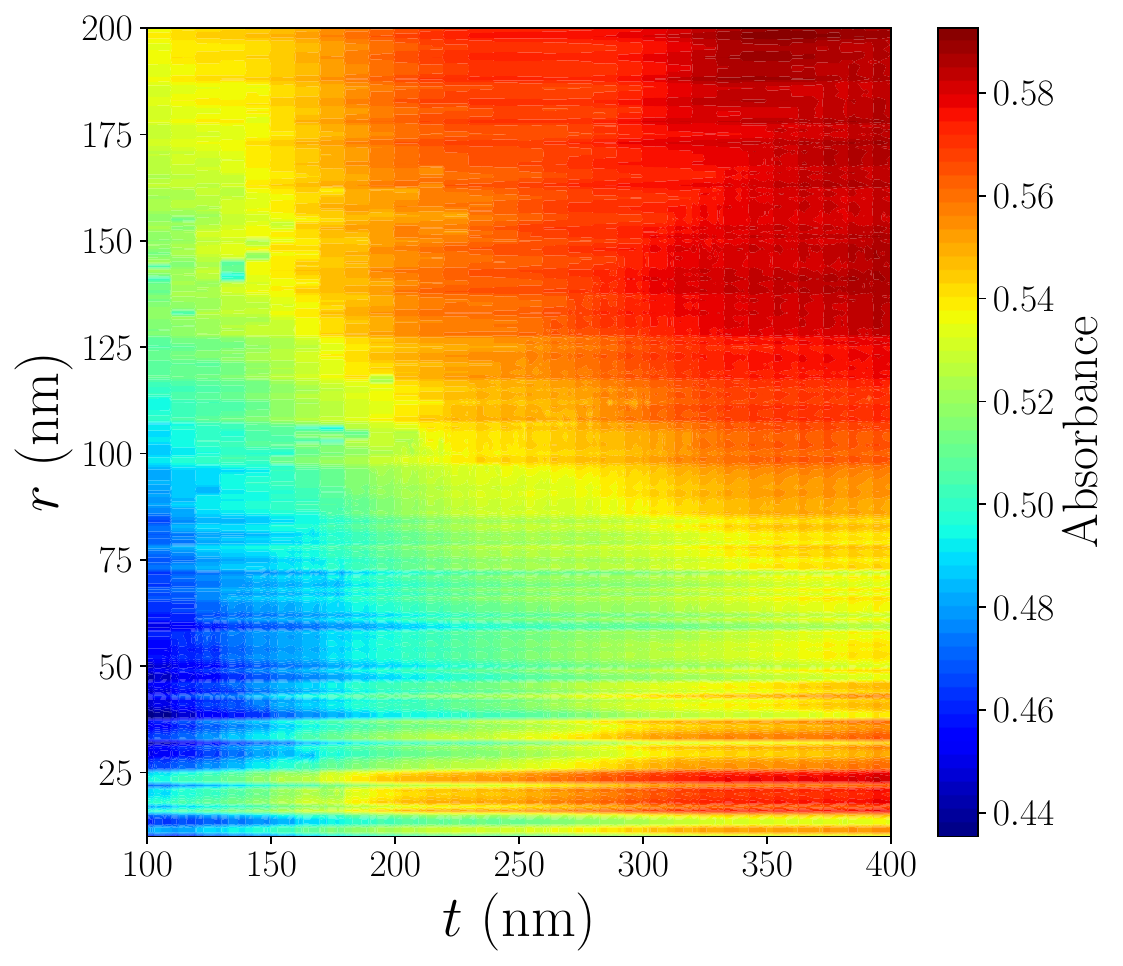}
        \includegraphics[width=0.31\textwidth]{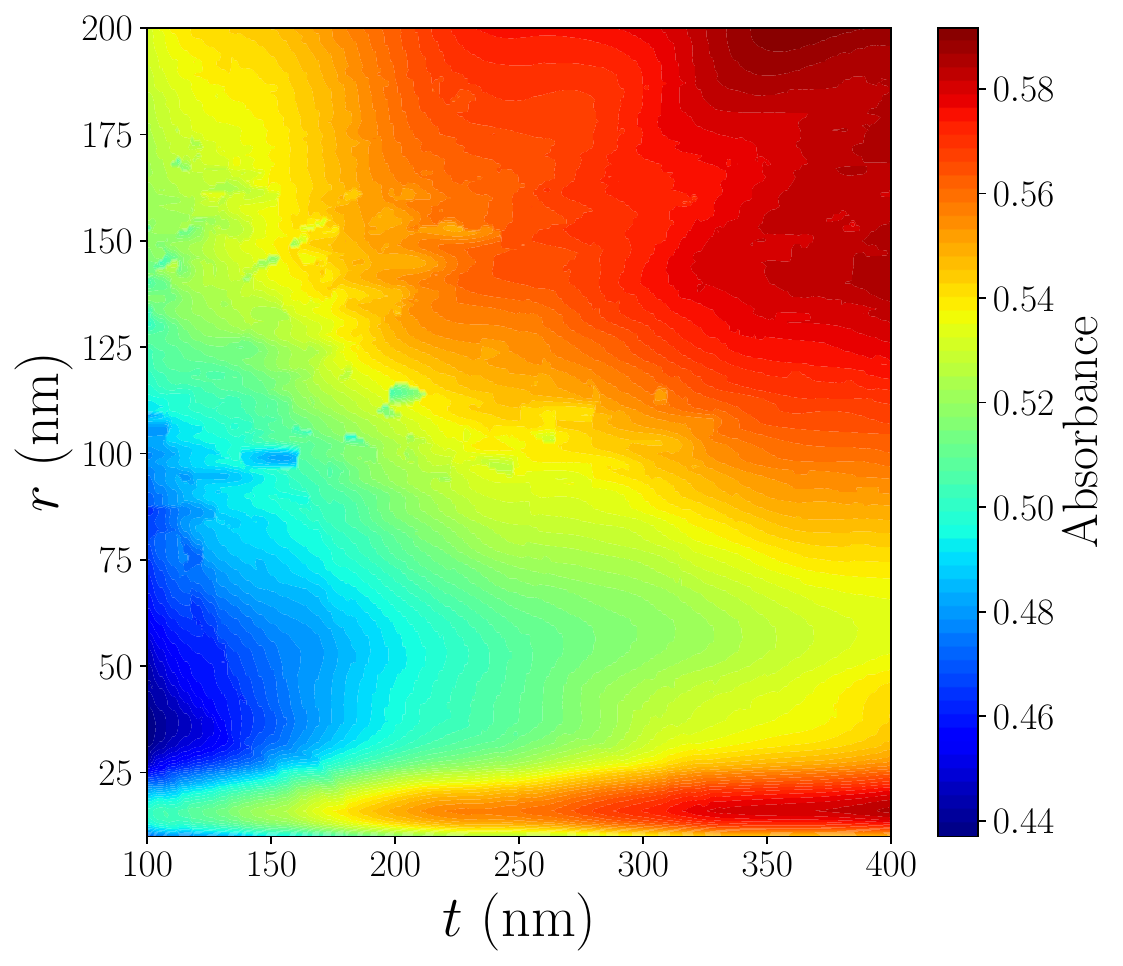}
        \caption{\ce{GaAs}}
    \end{subfigure}
    \caption{Visualization of the absorbance of the close-packed \ce{TiO2} nanospheres on top of a thin film made of \ce{cSi}, \ce{CH3NH3PbI3}, or \ce{GaAs} for three different fidelity levels, i.e., low fidelity (shown in left panels), medium fidelity (shown in center panels), and high fidelity (shown in right panels).}
    \label{fig:nanospheres_all}
\end{figure}

\begin{figure}[t]
    \centering
    \begin{subfigure}[b]{\textwidth}
        \centering
        \includegraphics[width=0.31\textwidth]{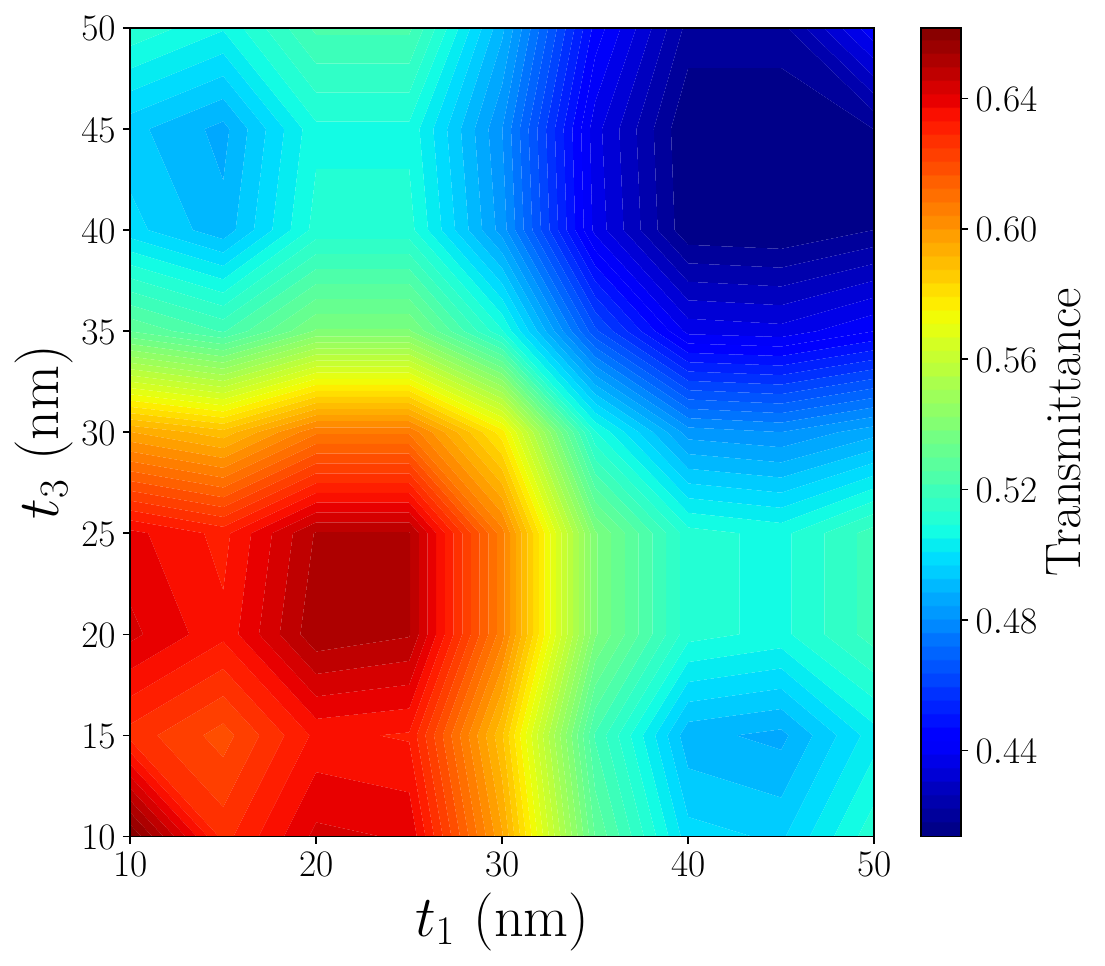}
        \includegraphics[width=0.31\textwidth]{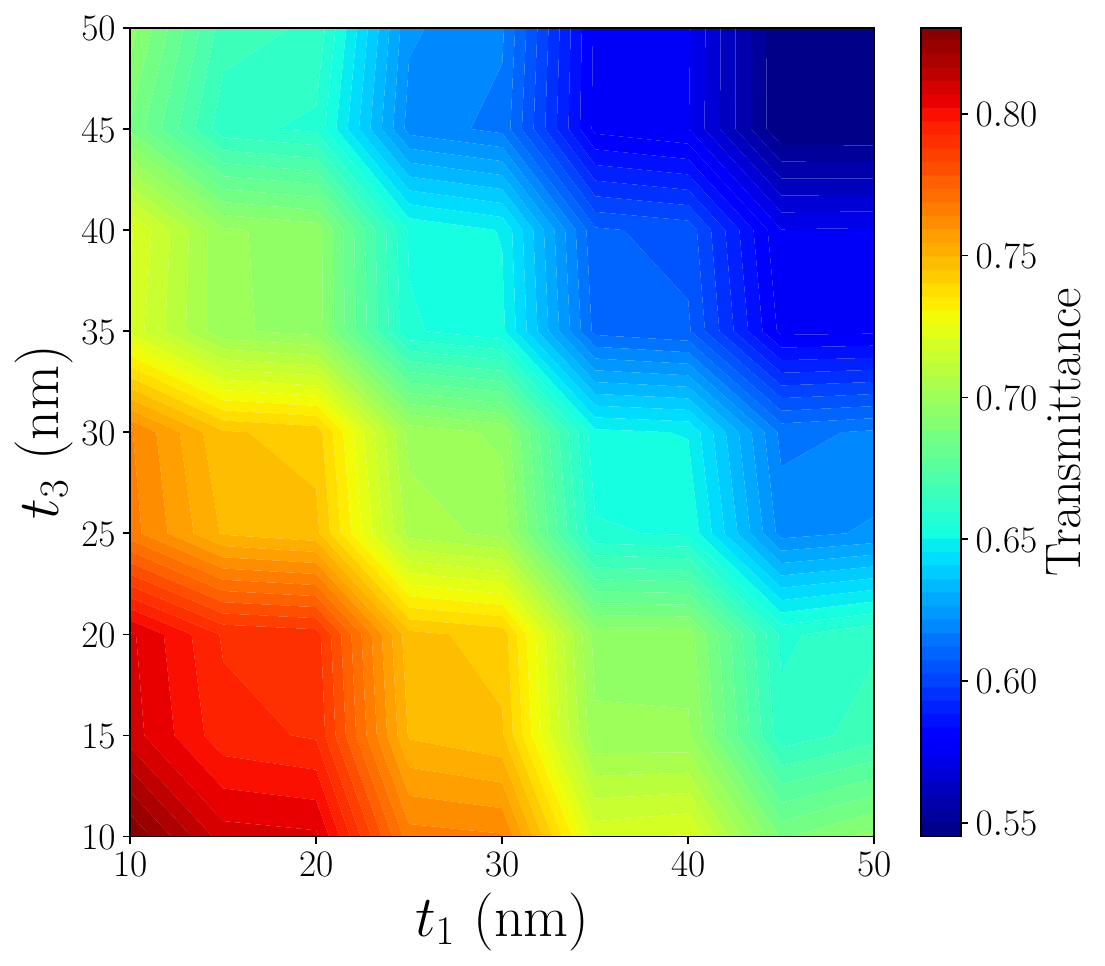}
        \includegraphics[width=0.31\textwidth]{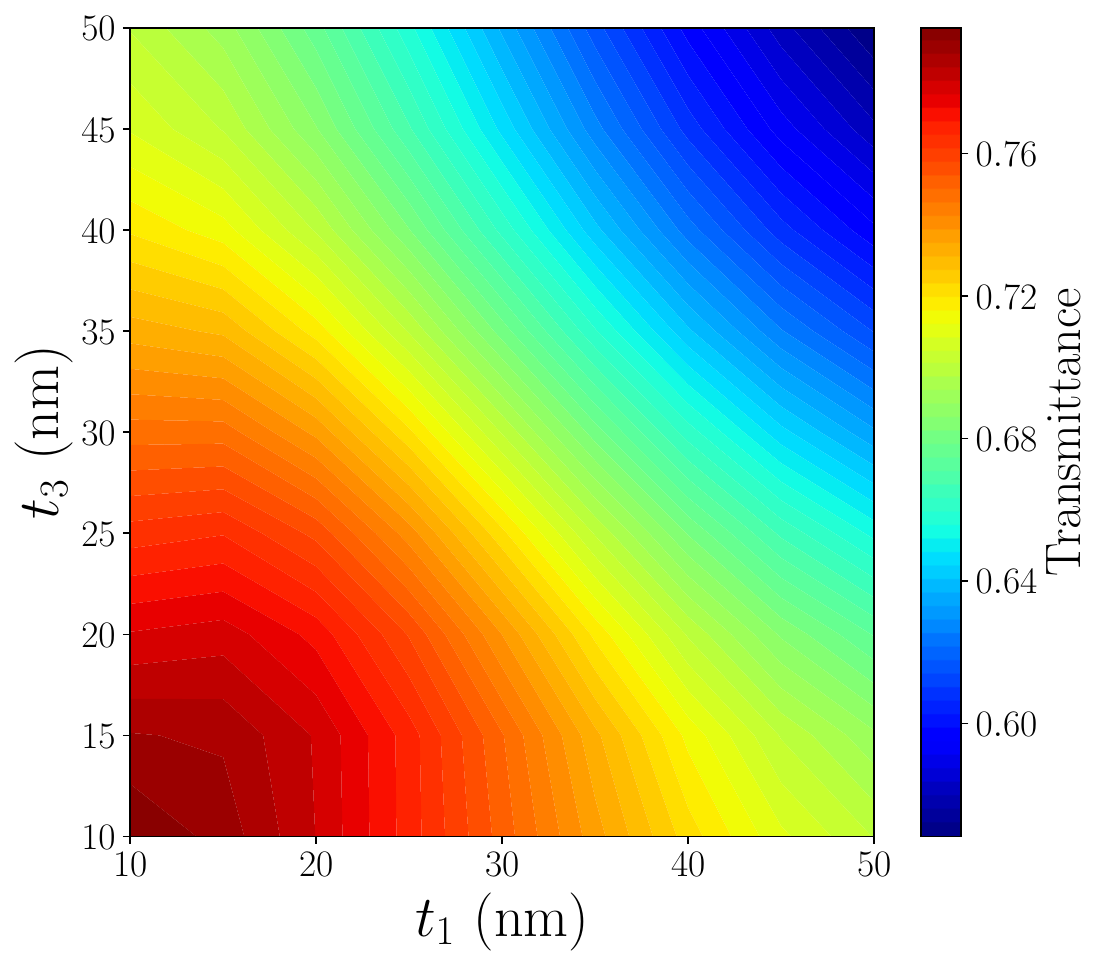}
        \caption{$t_2 = 8, r_1 = r_2 = 20, h_1 = h_2 = 50$}
    \end{subfigure}
    \begin{subfigure}[b]{\textwidth}
        \centering
        \includegraphics[width=0.31\textwidth]{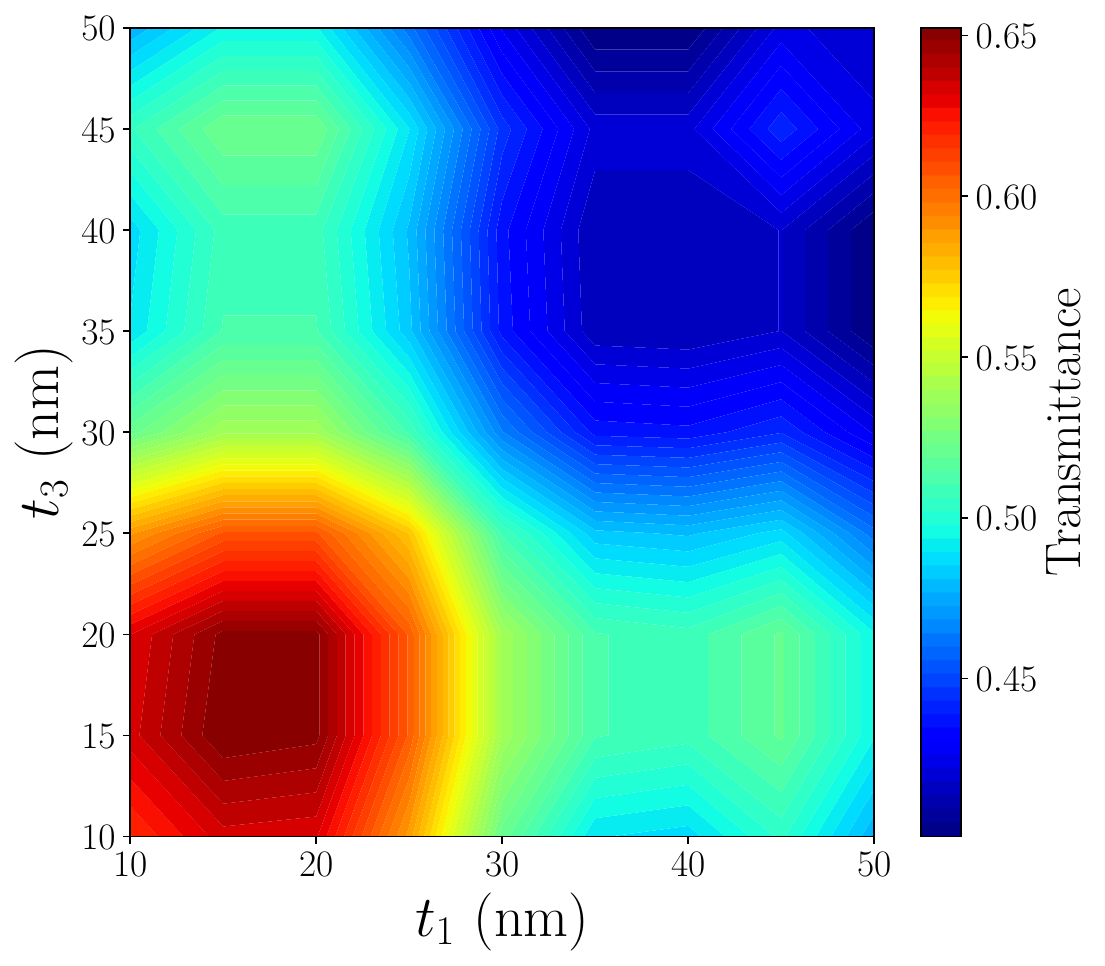}
        \includegraphics[width=0.31\textwidth]{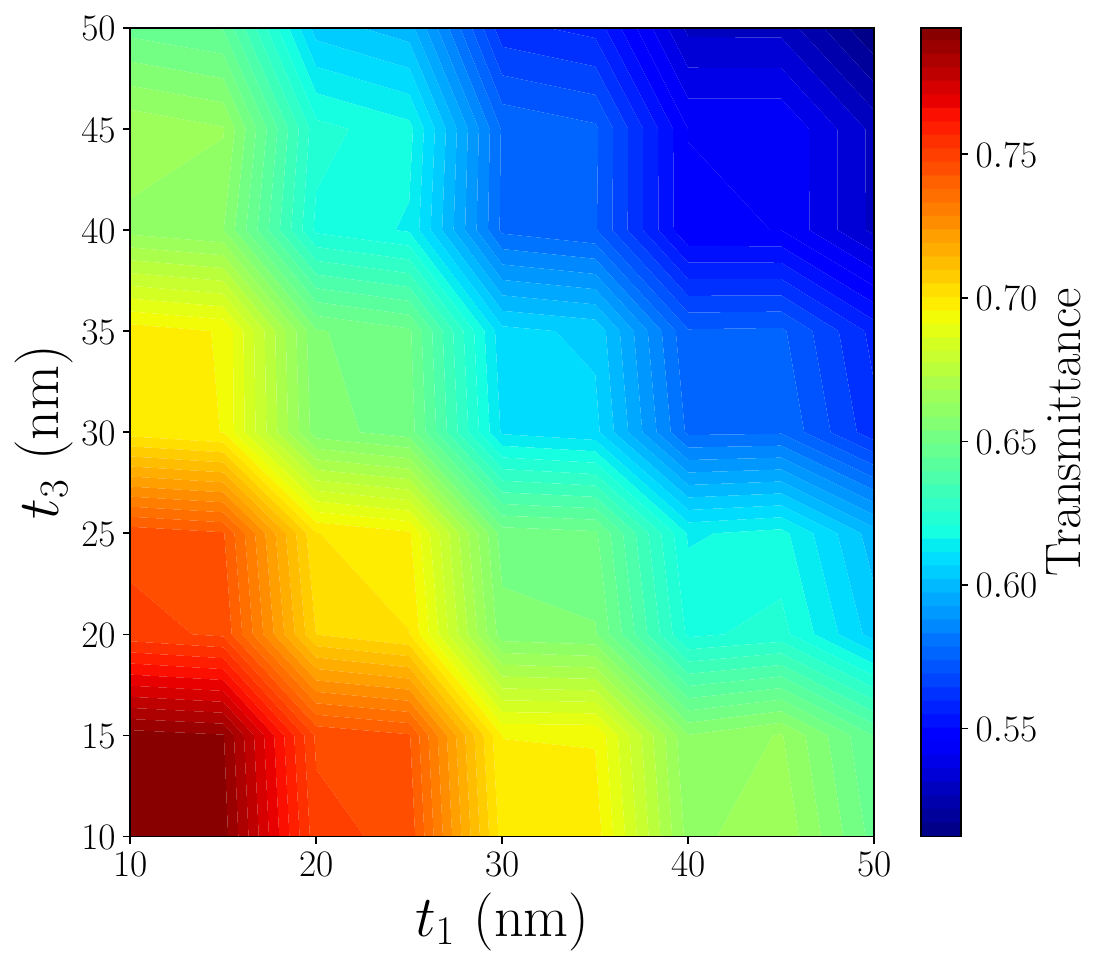}
        \includegraphics[width=0.31\textwidth]{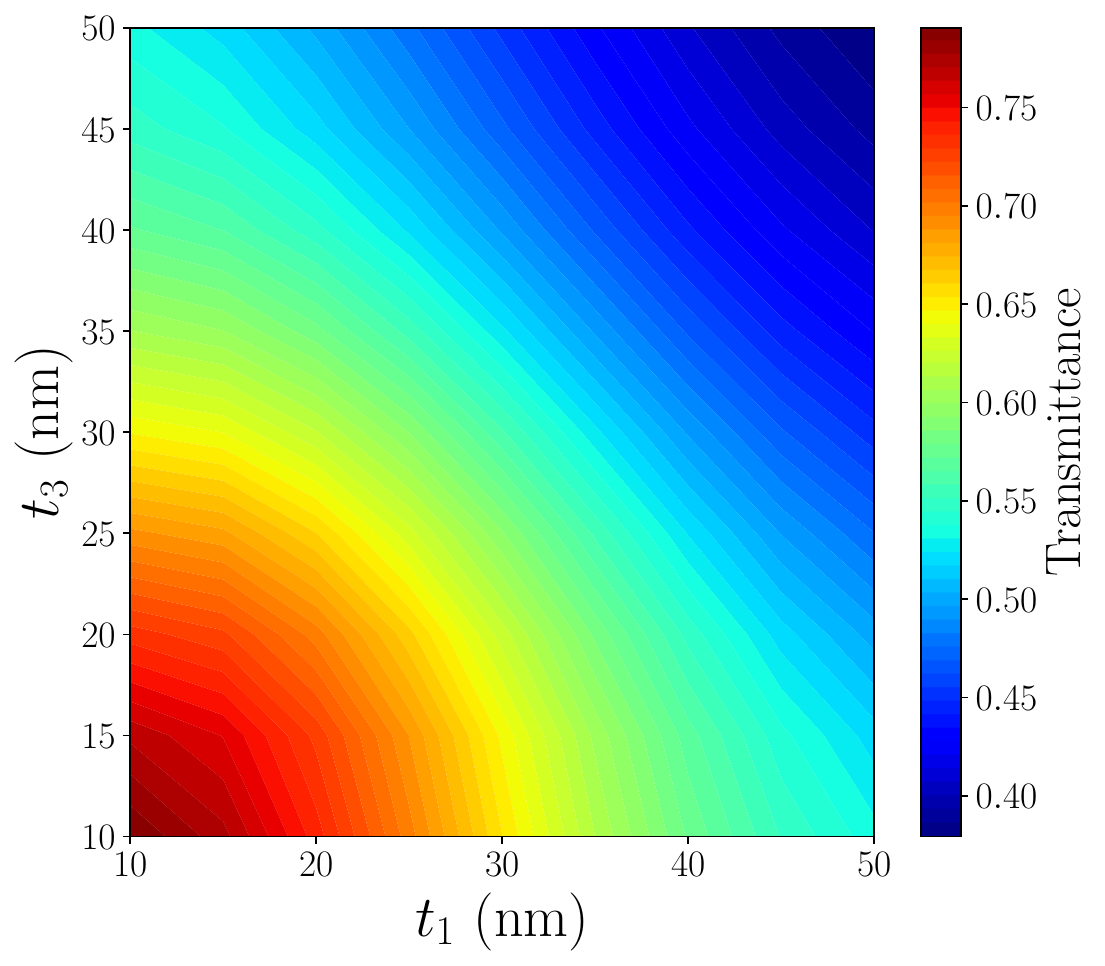}
        \caption{$t_2 = 18, r_1 = r_2 = 20, h_1 = h_2 = 50$}
    \end{subfigure}
    \caption{Visualization of the transmittance of the three-layer film with double-sided nanocones made of \ce{TiO2}/\ce{Ag}/\ce{TiO2}/\ce{TiO2}/\ce{TiO2} for three different fidelity levels, i.e., low fidelity (shown in left panels), medium fidelity (shown in center panels), and high fidelity (shown in right panels).}
    \label{fig:doublenanocones_tio2_ag_tio2}
\end{figure}

\begin{figure}[t]
    \centering
    \begin{subfigure}[b]{\textwidth}
        \centering
        \includegraphics[width=0.31\textwidth]{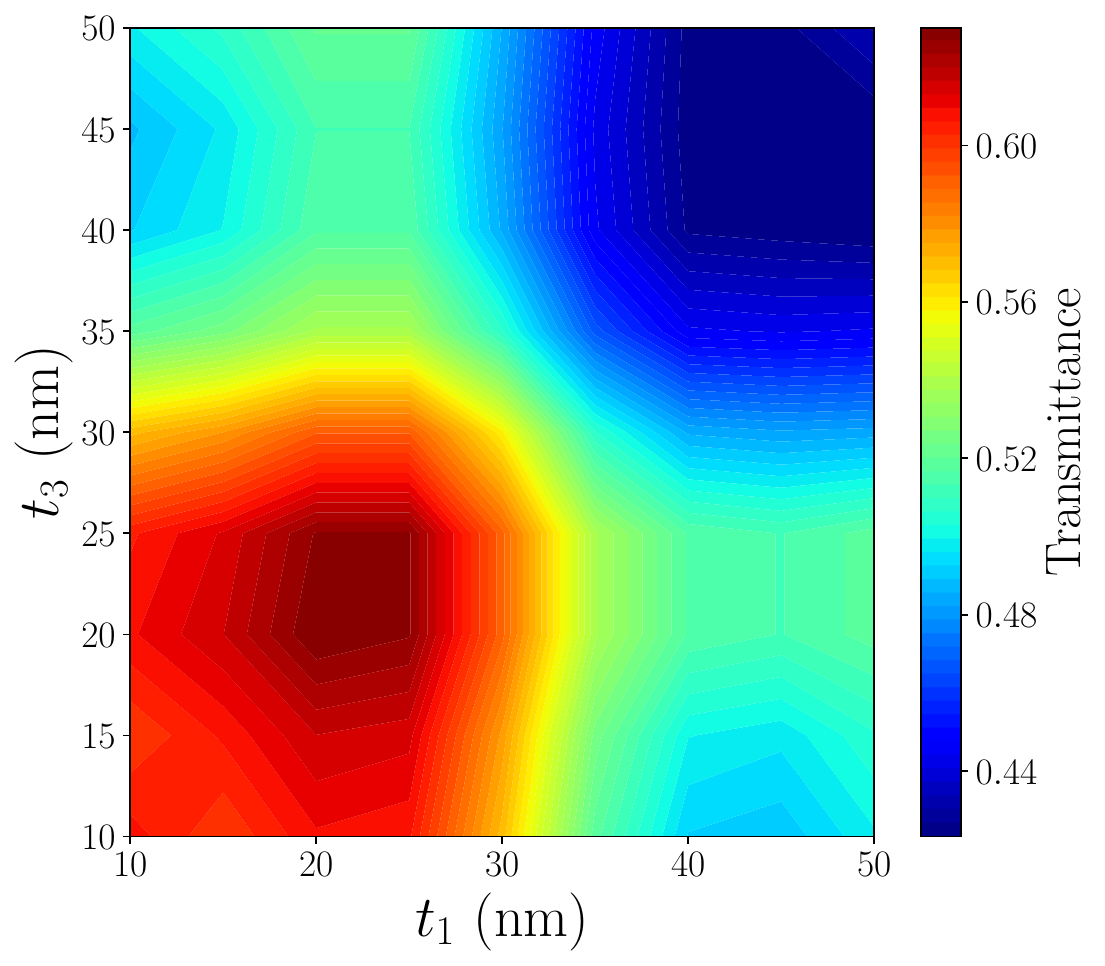}
        \includegraphics[width=0.31\textwidth]{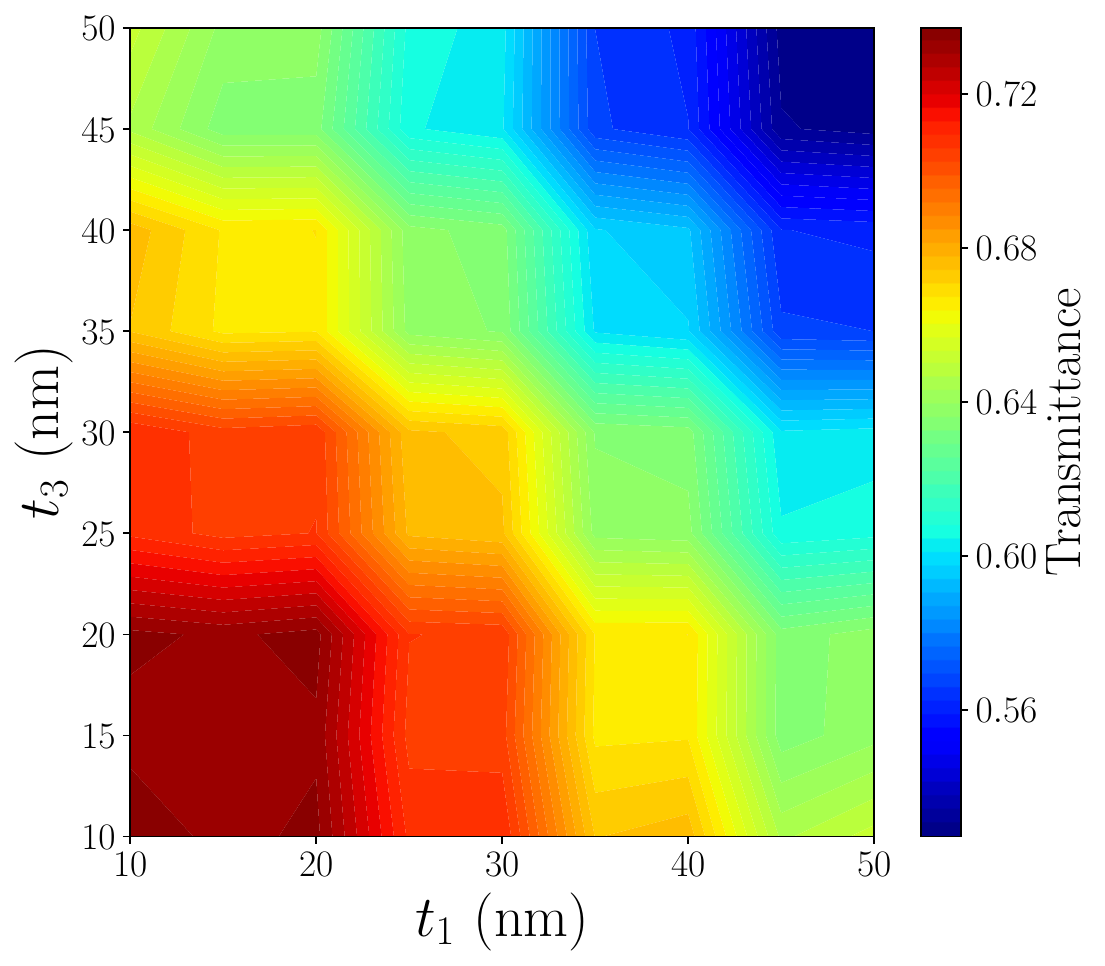}
        \includegraphics[width=0.31\textwidth]{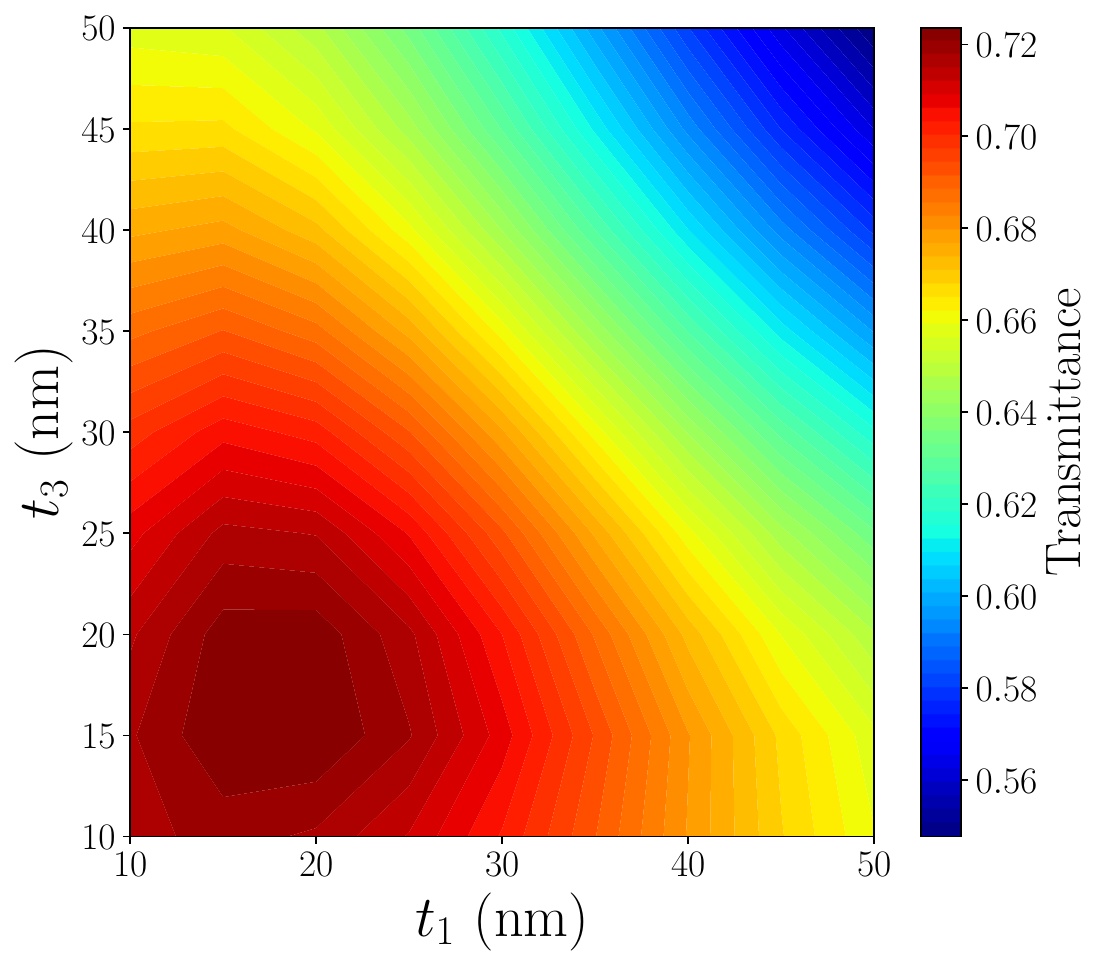}
        \caption{$t_2 = 8, r_1 = r_2 = 20, h_1 = h_2 = 50$}
    \end{subfigure}
    \begin{subfigure}[b]{\textwidth}
        \centering
        \includegraphics[width=0.31\textwidth]{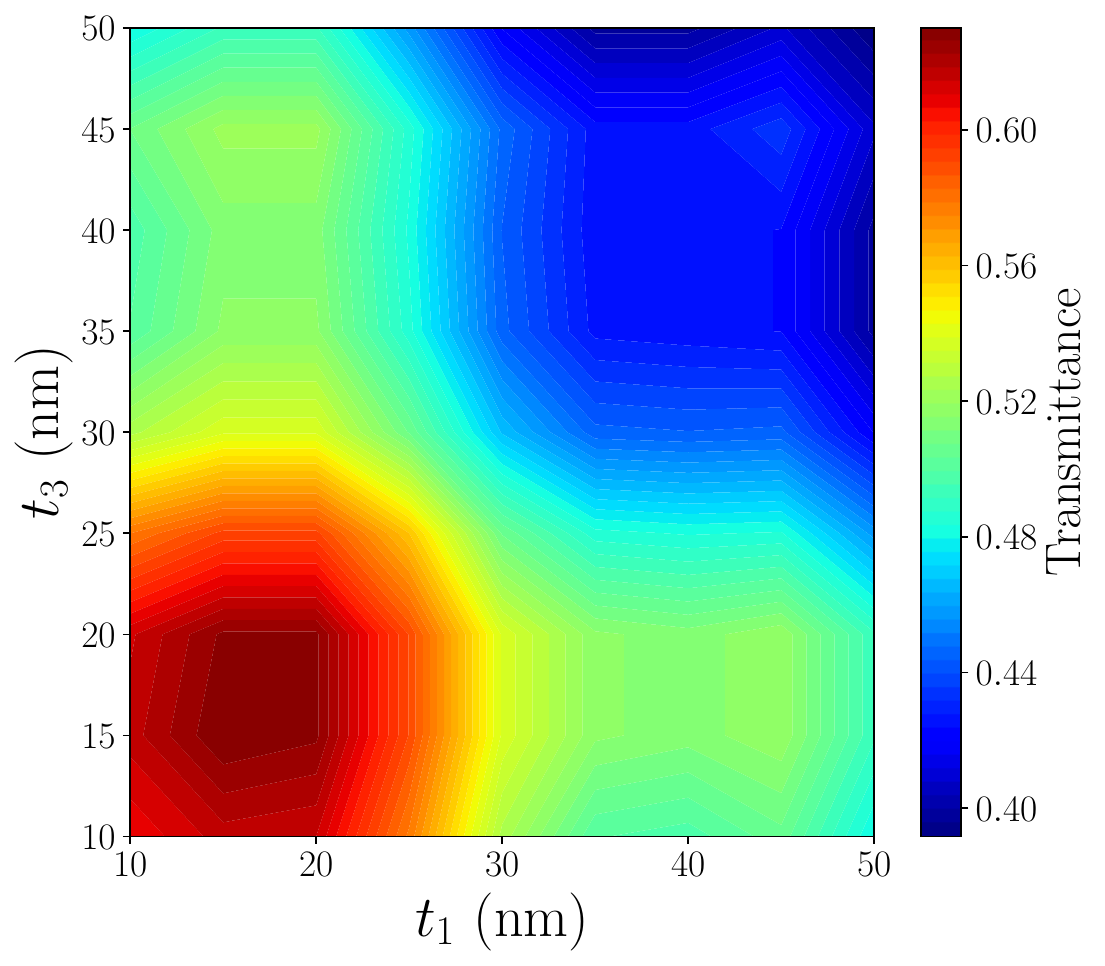}
        \includegraphics[width=0.31\textwidth]{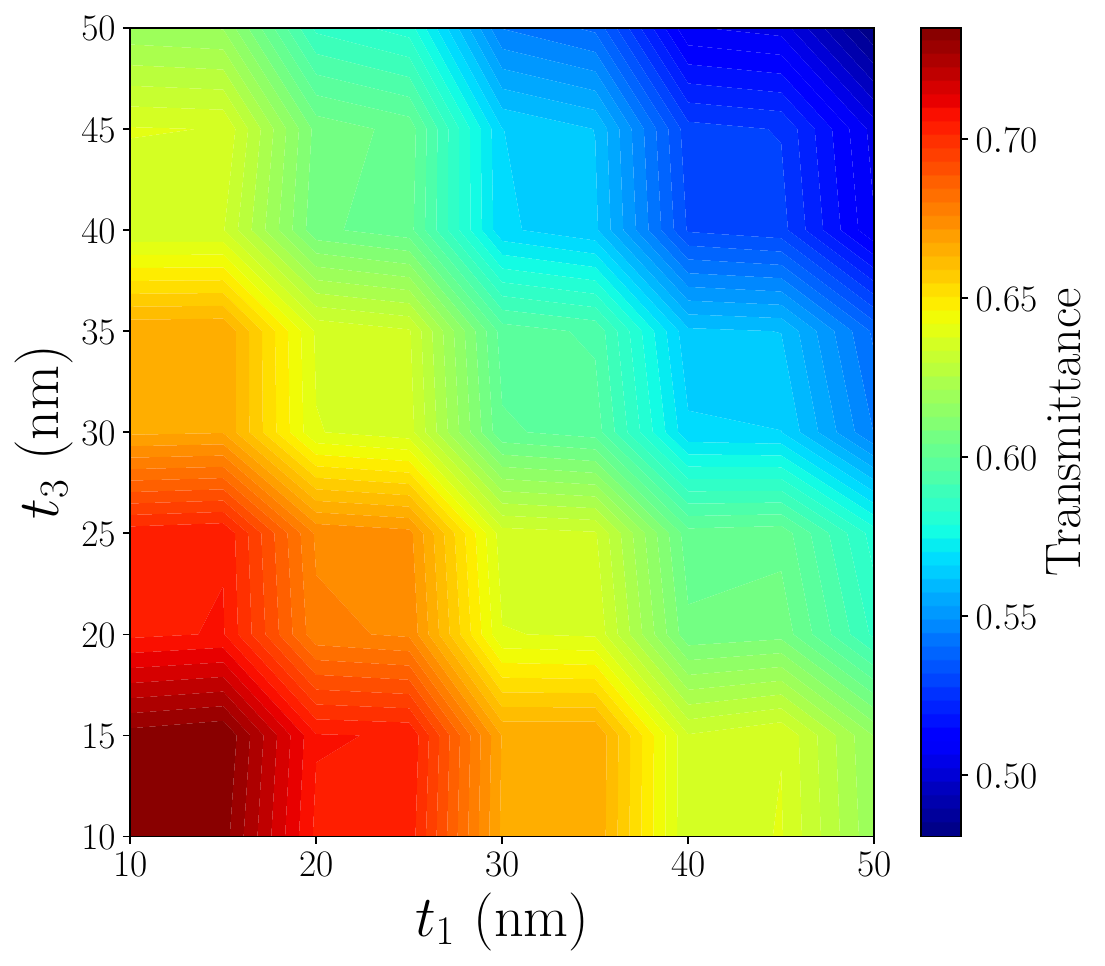}
        \includegraphics[width=0.31\textwidth]{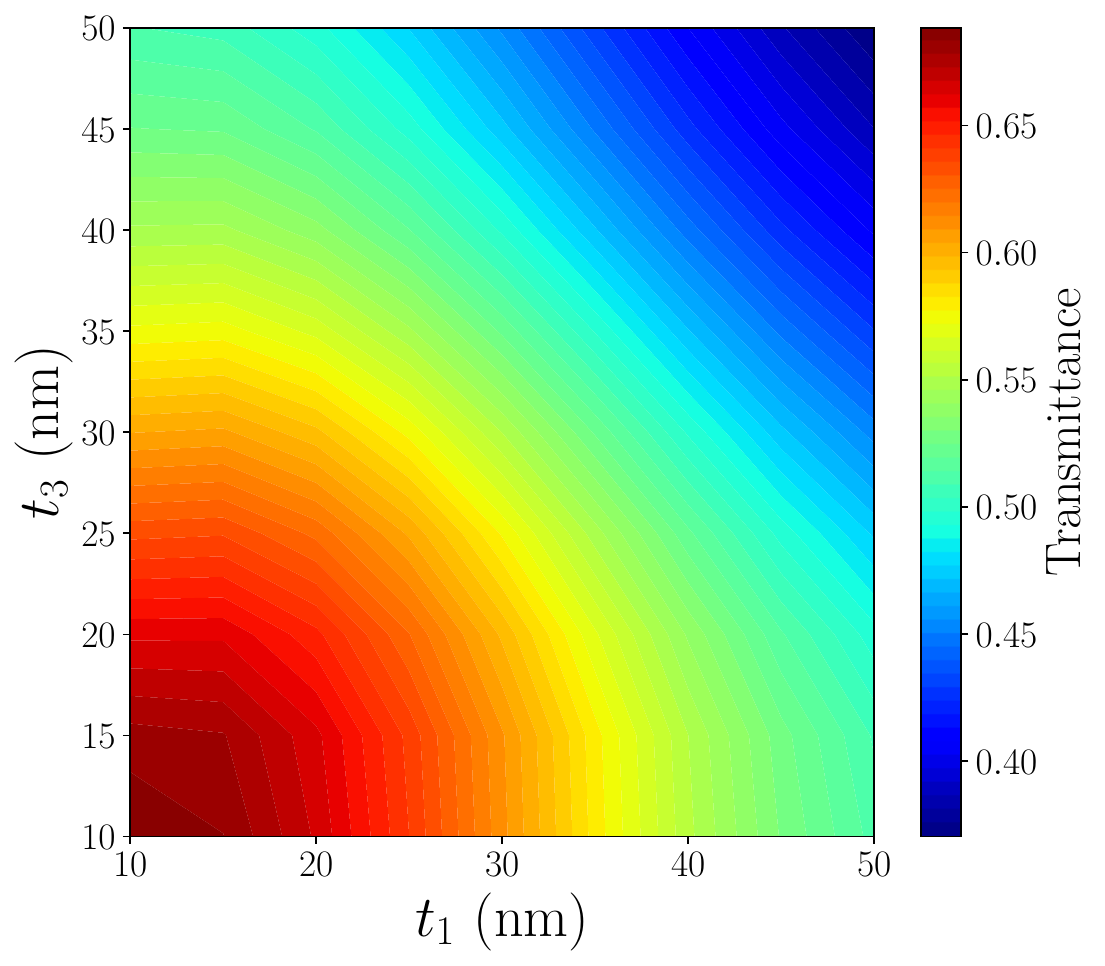}
        \caption{$t_2 = 18, r_1 = r_2 = 20, h_1 = h_2 = 50$}
    \end{subfigure}
    \caption{Visualization of the transmittance of the three-layer film with double-sided nanocones made of \ce{TiO2}/\ce{Au}/\ce{TiO2}/\ce{TiO2}/\ce{TiO2} for three different fidelity levels, i.e., low fidelity (shown in left panels), medium fidelity (shown in center panels), and high fidelity (shown in right panels).}
    \label{fig:doublenanocones_tio2_au_tio2}
\end{figure}

\begin{figure}[t]
    \centering
    \begin{subfigure}[b]{\textwidth}
        \centering
        \includegraphics[width=0.31\textwidth]{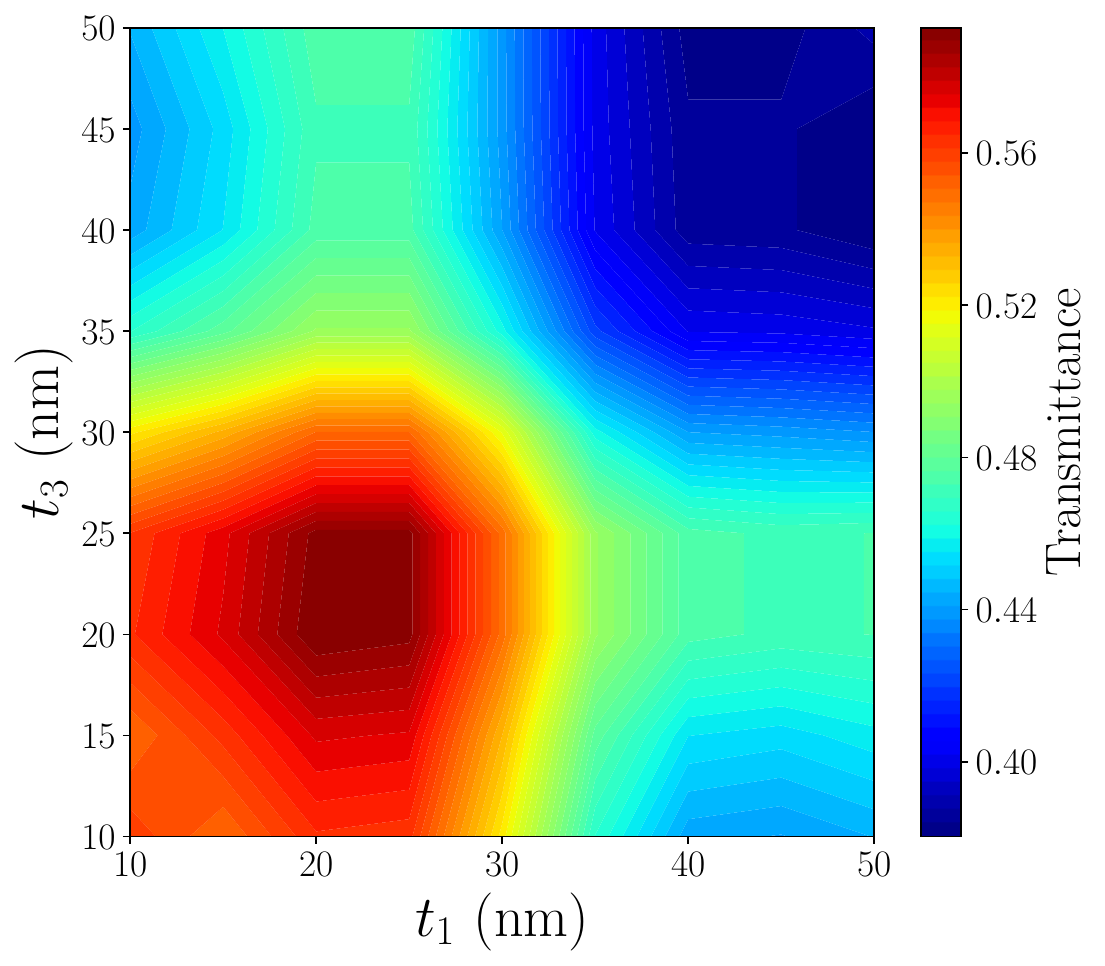}
        \includegraphics[width=0.31\textwidth]{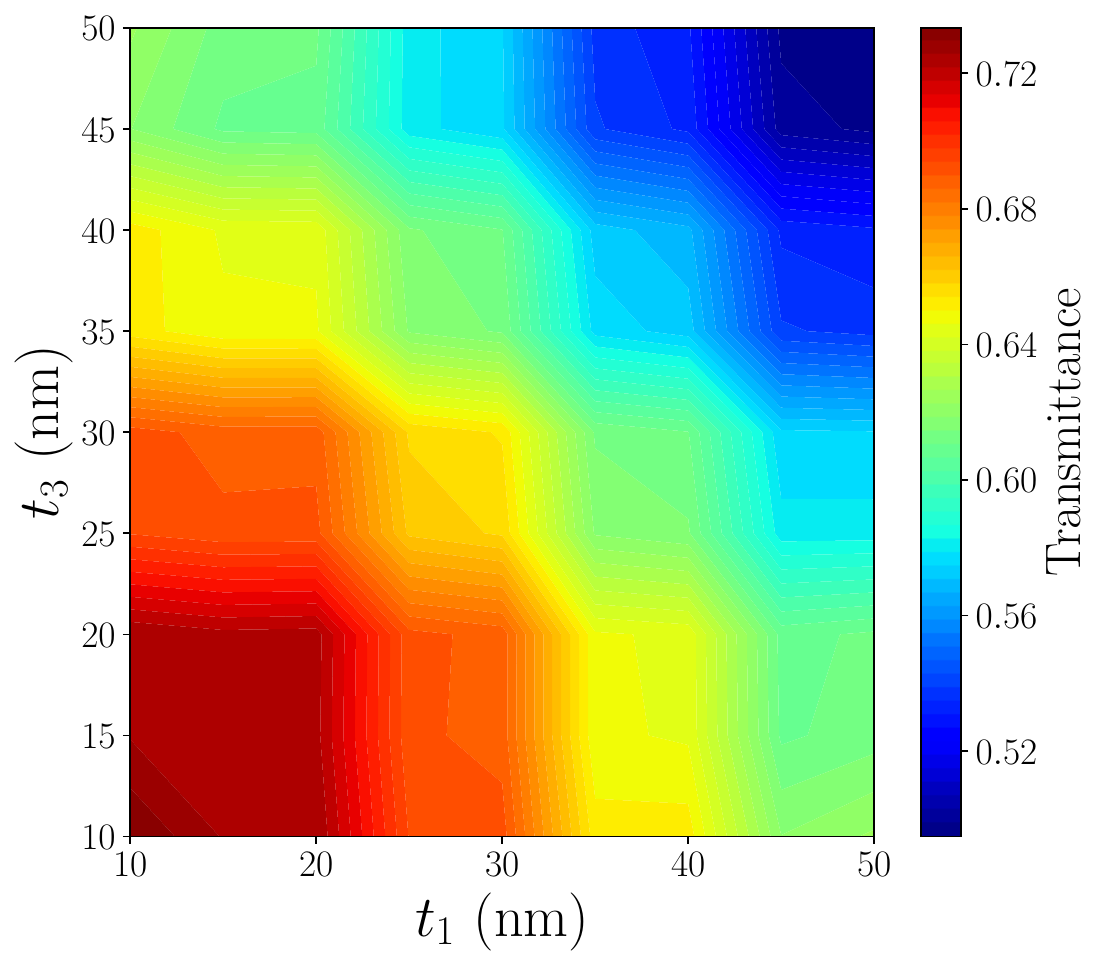}
        \includegraphics[width=0.31\textwidth]{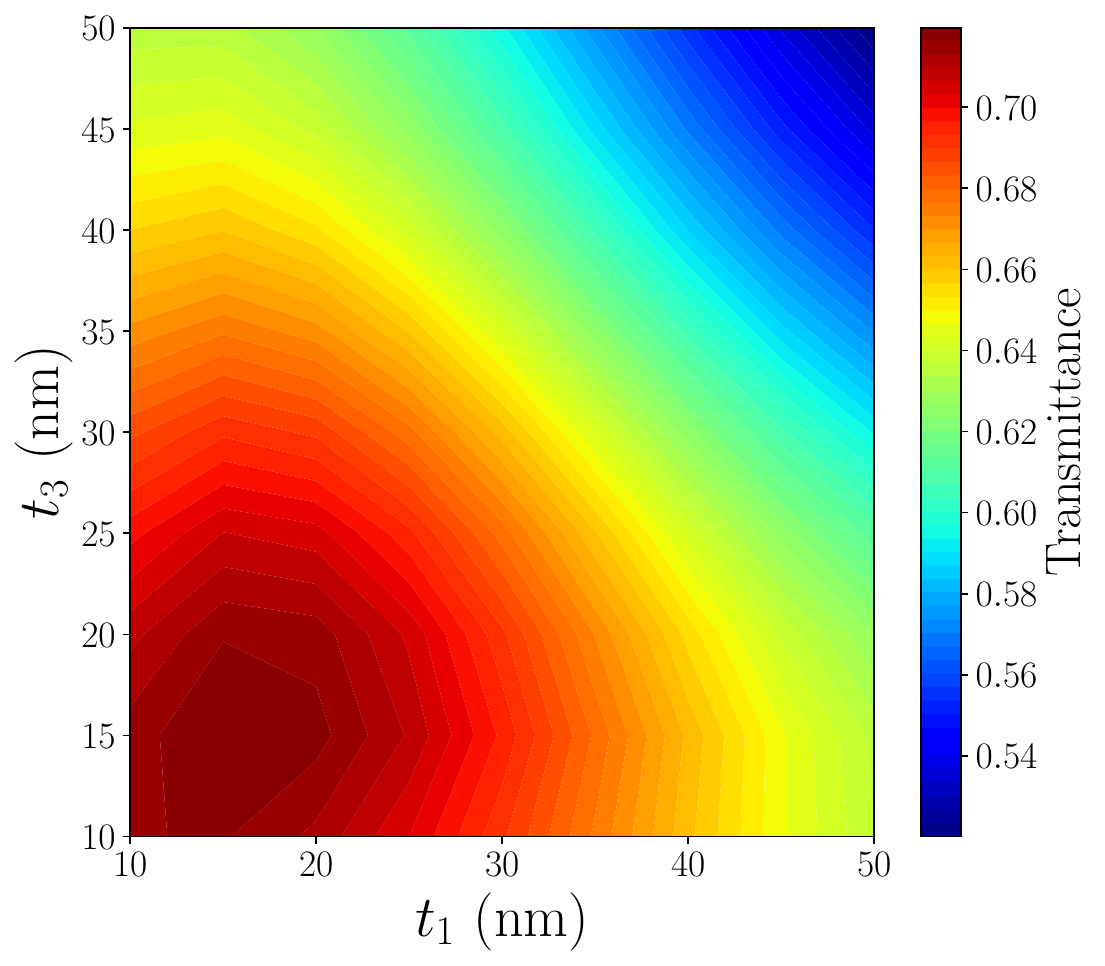}
        \caption{$t_2 = 8, r_1 = r_2 = 20, h_1 = h_2 = 50$}
    \end{subfigure}
    \begin{subfigure}[b]{\textwidth}
        \centering
        \includegraphics[width=0.31\textwidth]{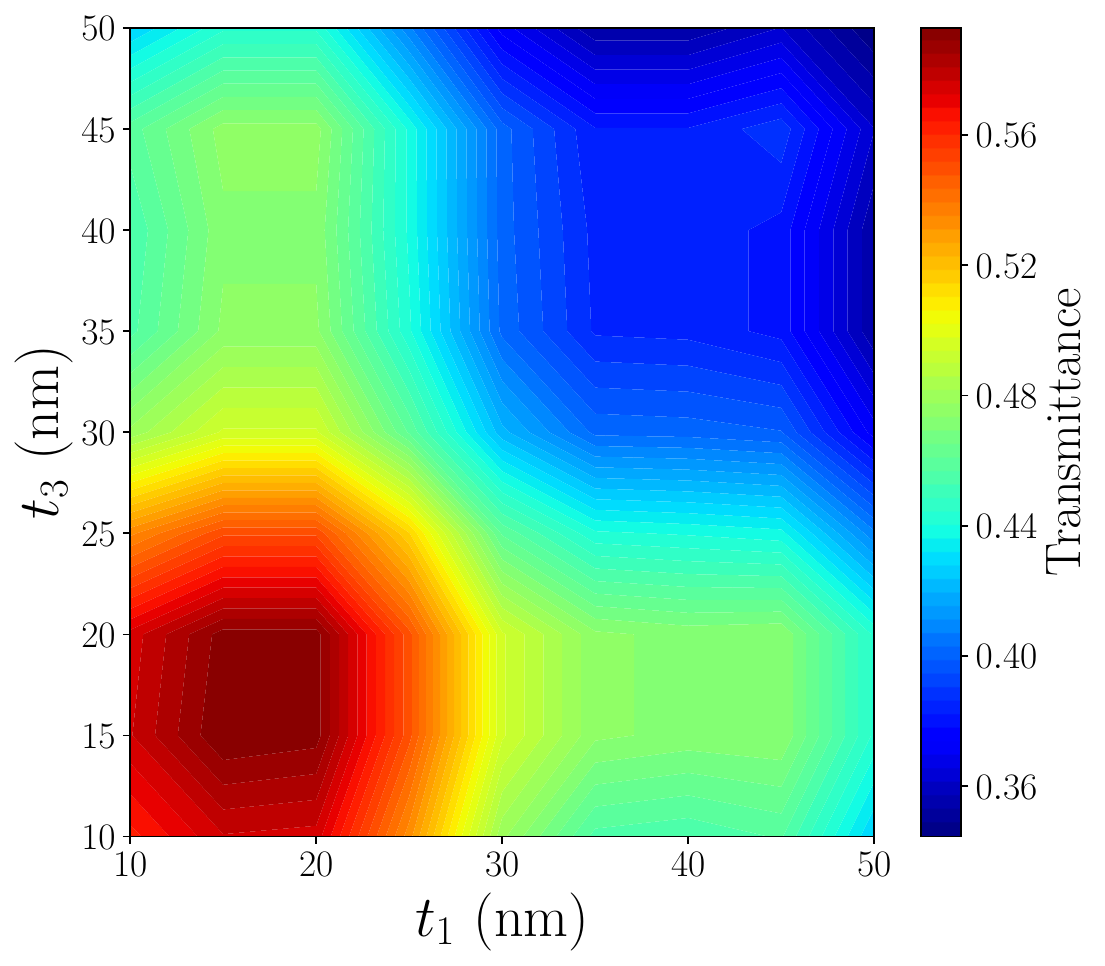}
        \includegraphics[width=0.31\textwidth]{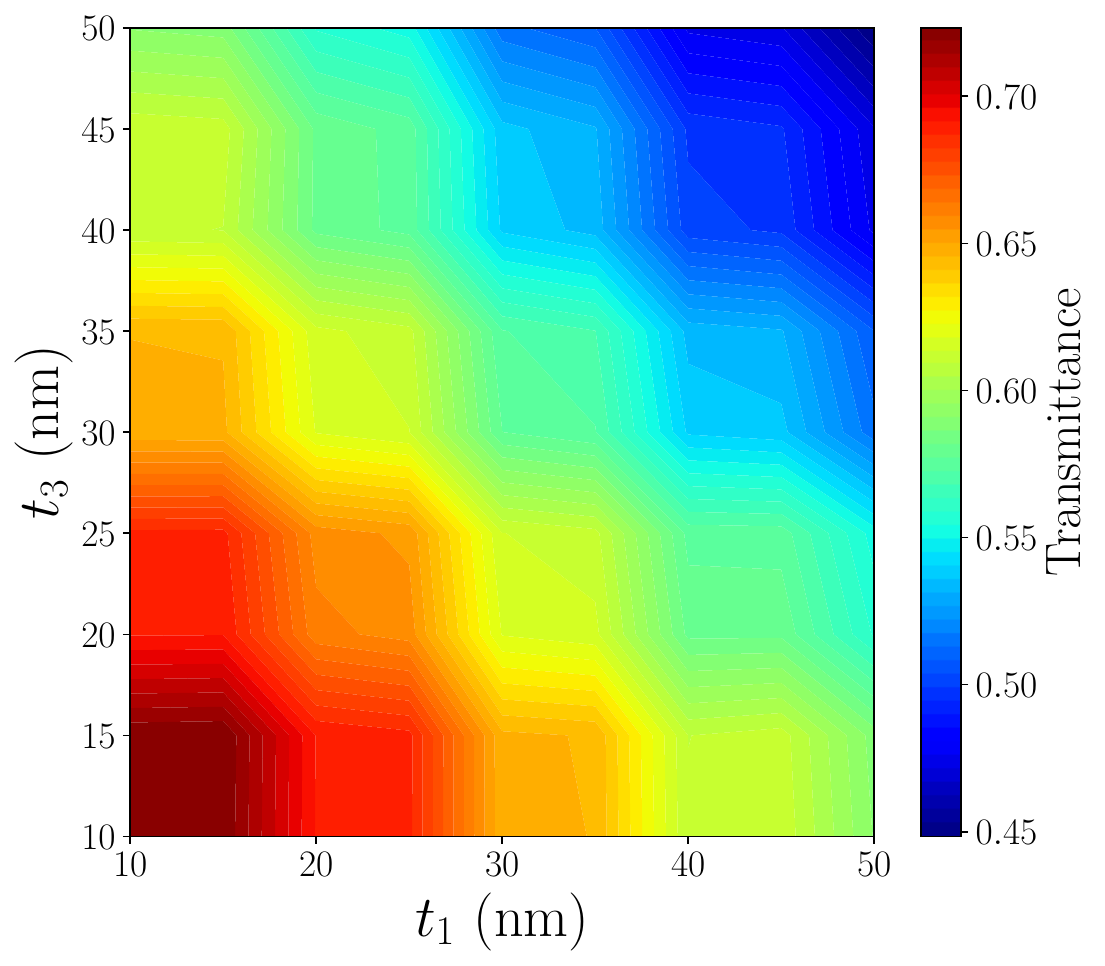}
        \includegraphics[width=0.31\textwidth]{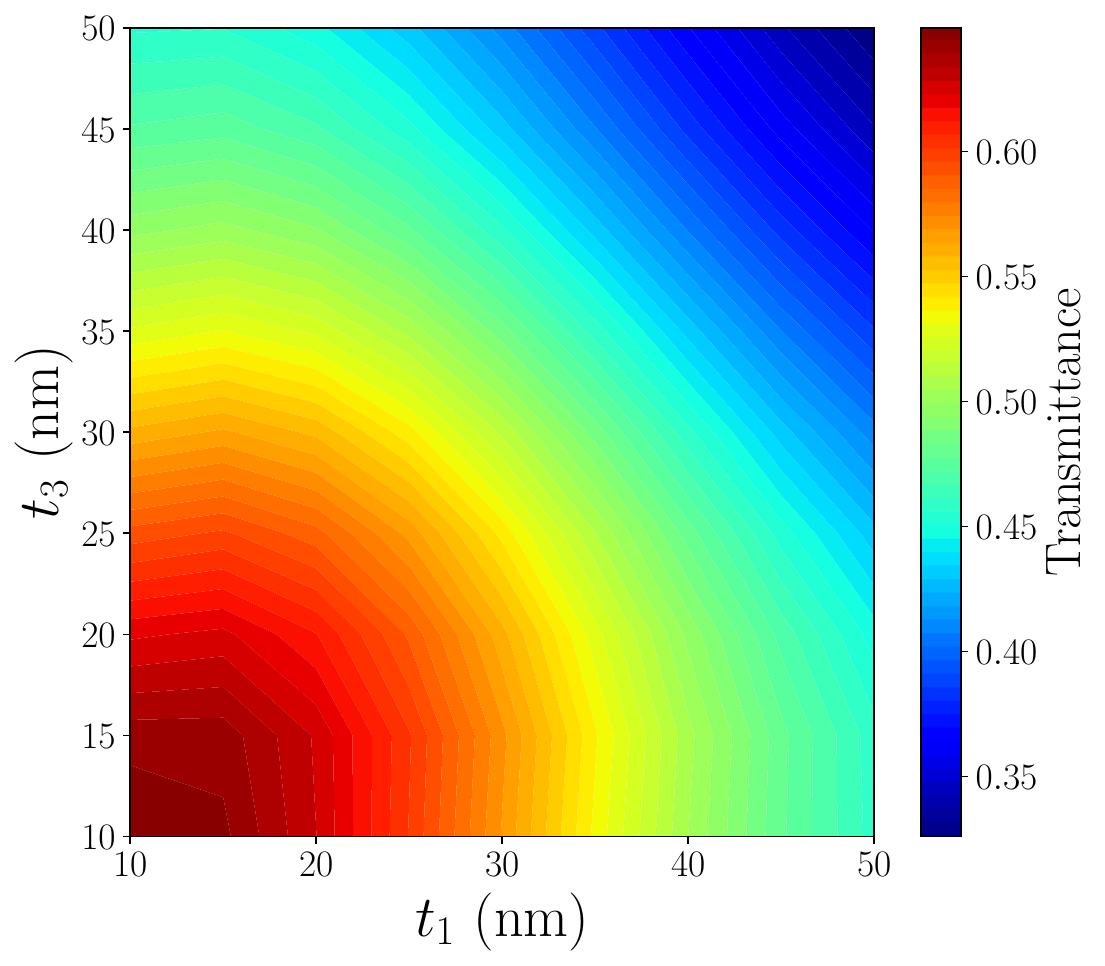}
        \caption{$t_2 = 18, r_1 = r_2 = 20, h_1 = h_2 = 50$}
    \end{subfigure}
    \caption{Visualization of the transmittance of the three-layer film with double-sided nanocones made of \ce{TiO2}/\ce{Cu}/\ce{TiO2}/\ce{TiO2}/\ce{TiO2} for three different fidelity levels, i.e., low fidelity (shown in left panels), medium fidelity (shown in center panels), and high fidelity (shown in right panels).}
    \label{fig:doublenanocones_tio2_cu_tio2}
\end{figure}

\begin{figure}[t]
    \centering
    \begin{subfigure}[b]{\textwidth}
        \centering
        \includegraphics[width=0.31\textwidth]{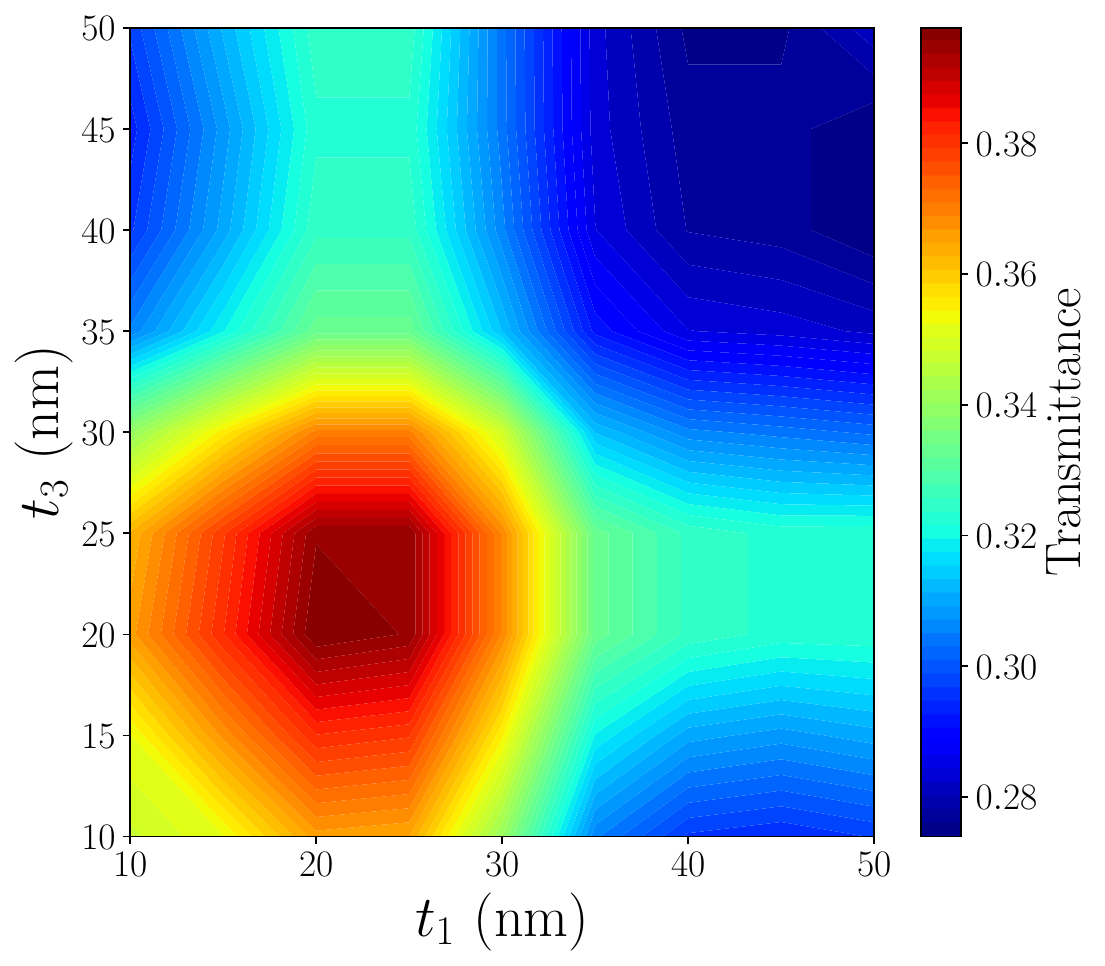}
        \includegraphics[width=0.31\textwidth]{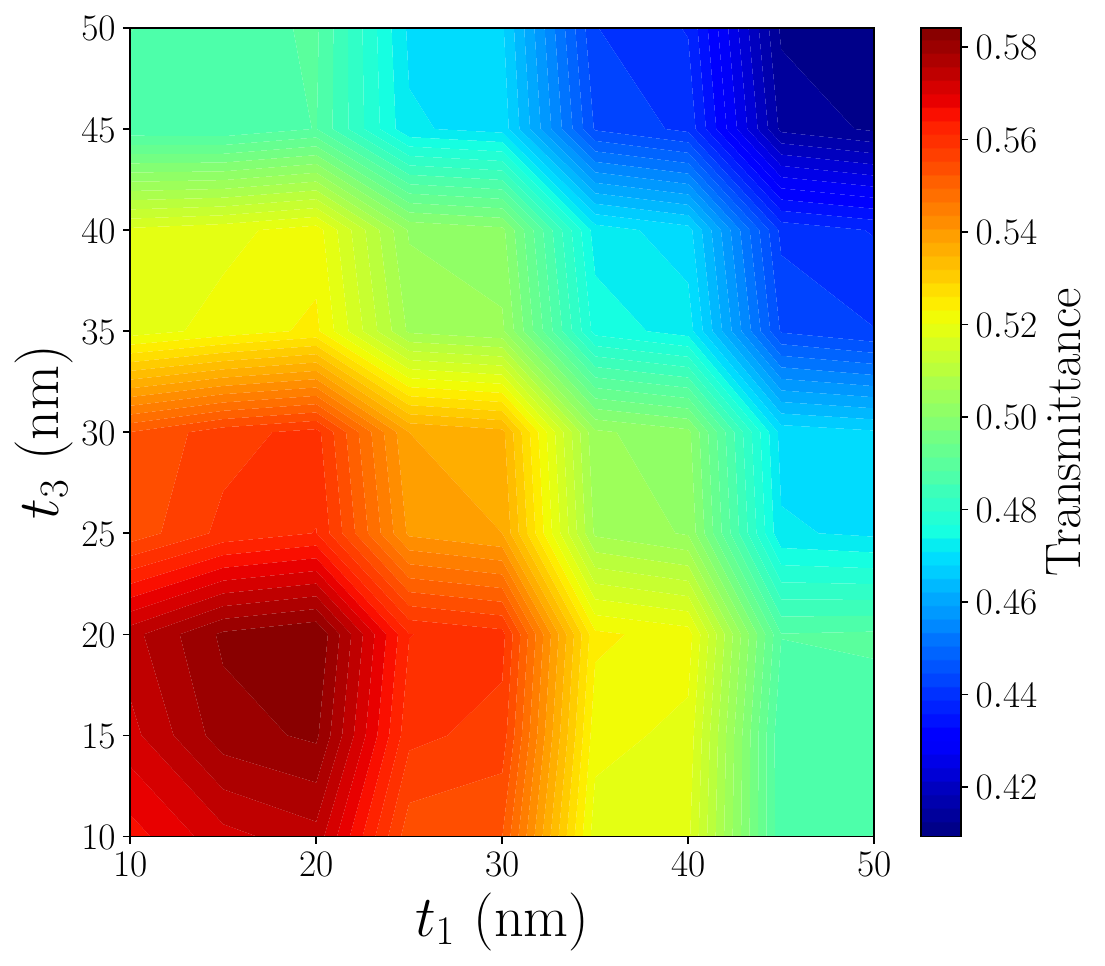}
        \includegraphics[width=0.31\textwidth]{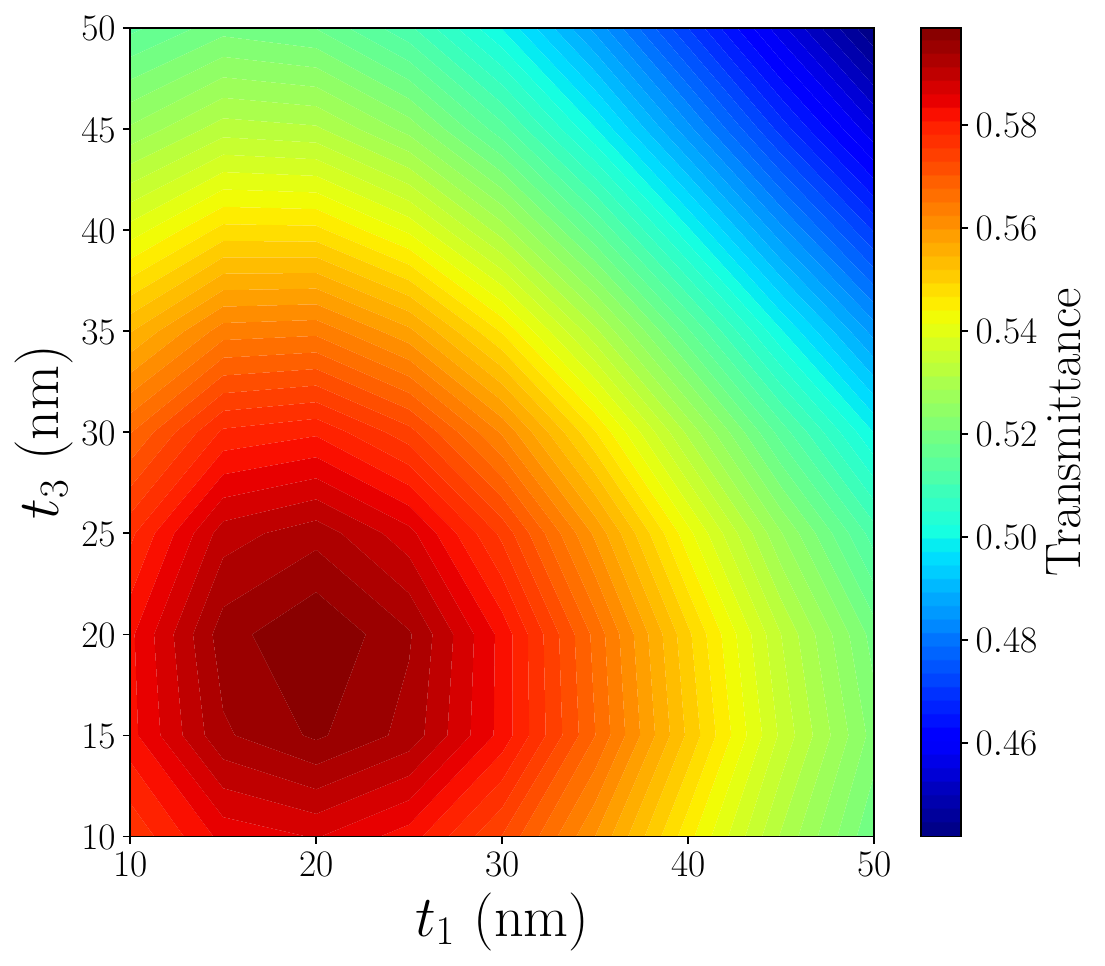}
        \caption{$t_2 = 8, r_1 = r_2 = 20, h_1 = h_2 = 50$}
    \end{subfigure}
    \begin{subfigure}[b]{\textwidth}
        \centering
        \includegraphics[width=0.31\textwidth]{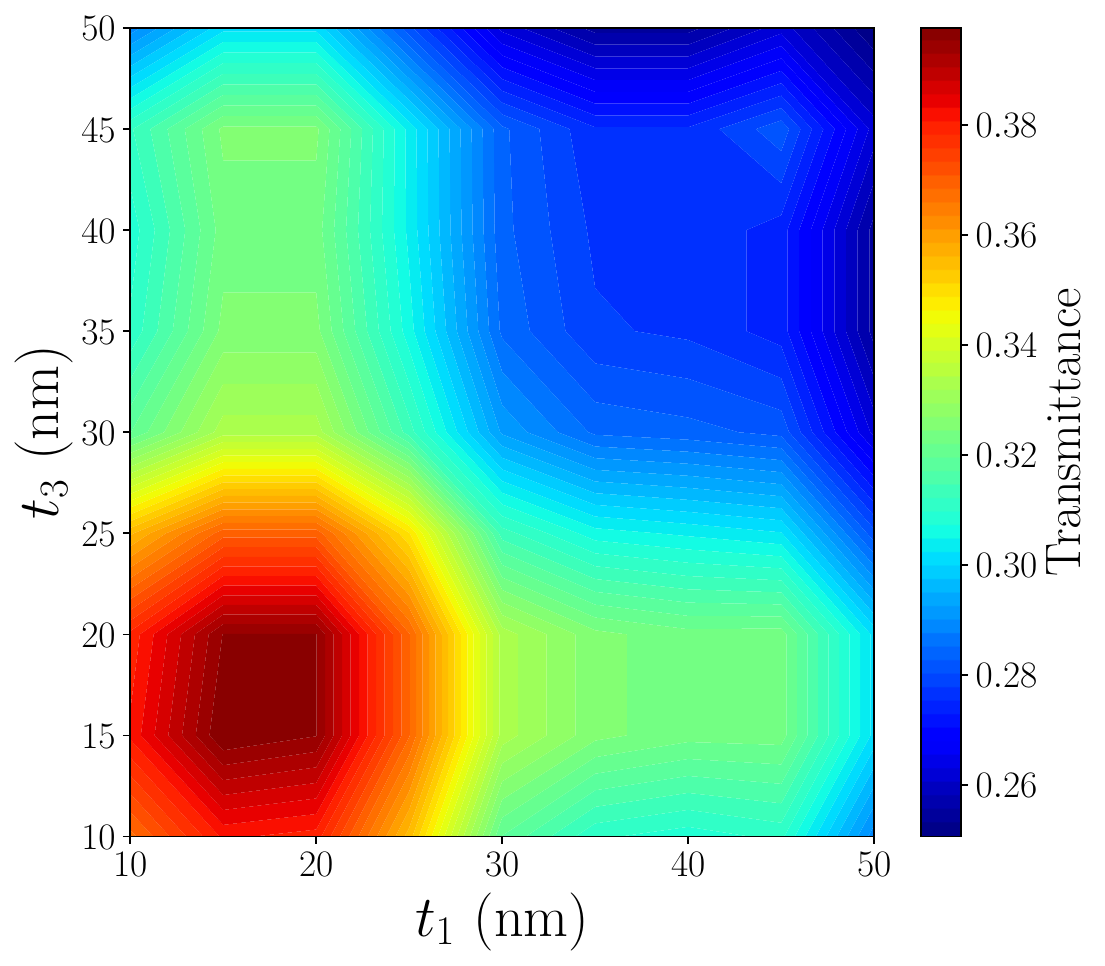}
        \includegraphics[width=0.31\textwidth]{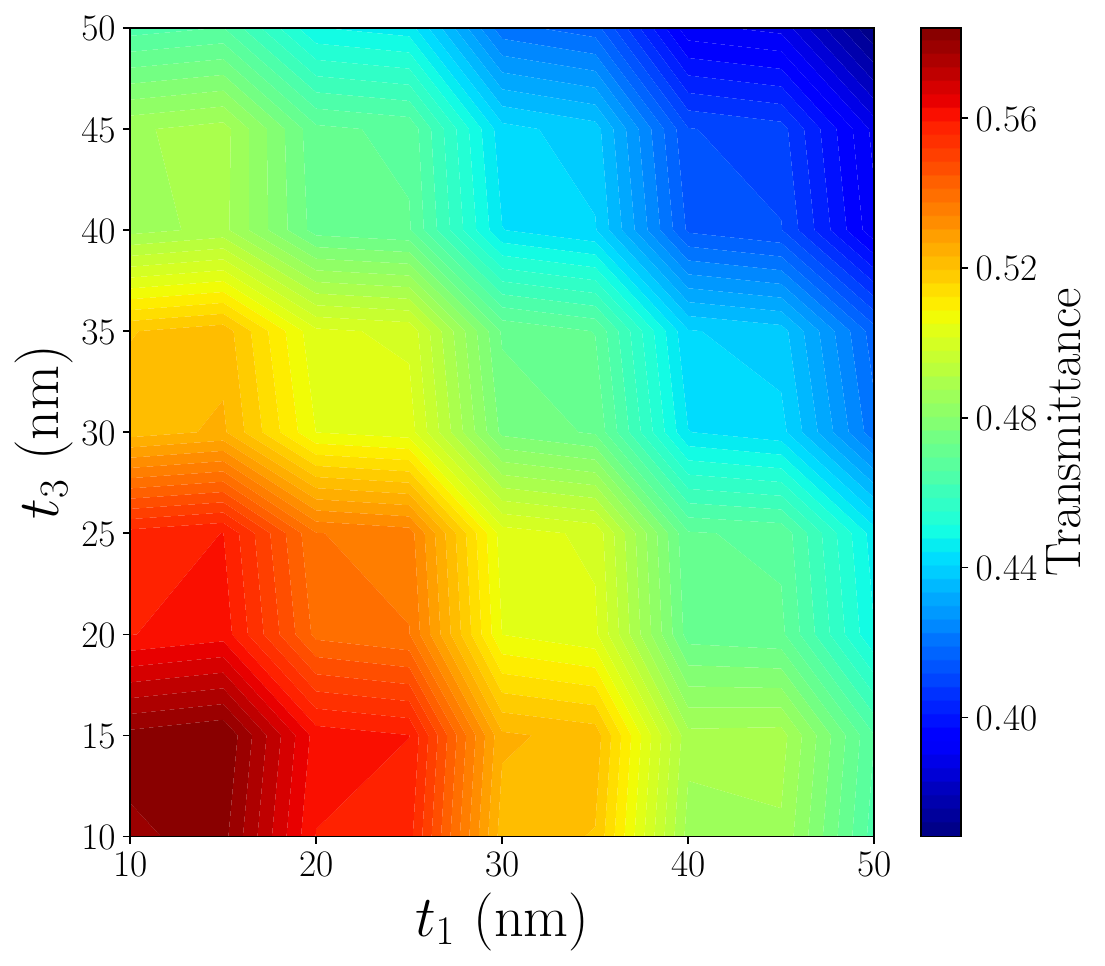}
        \includegraphics[width=0.31\textwidth]{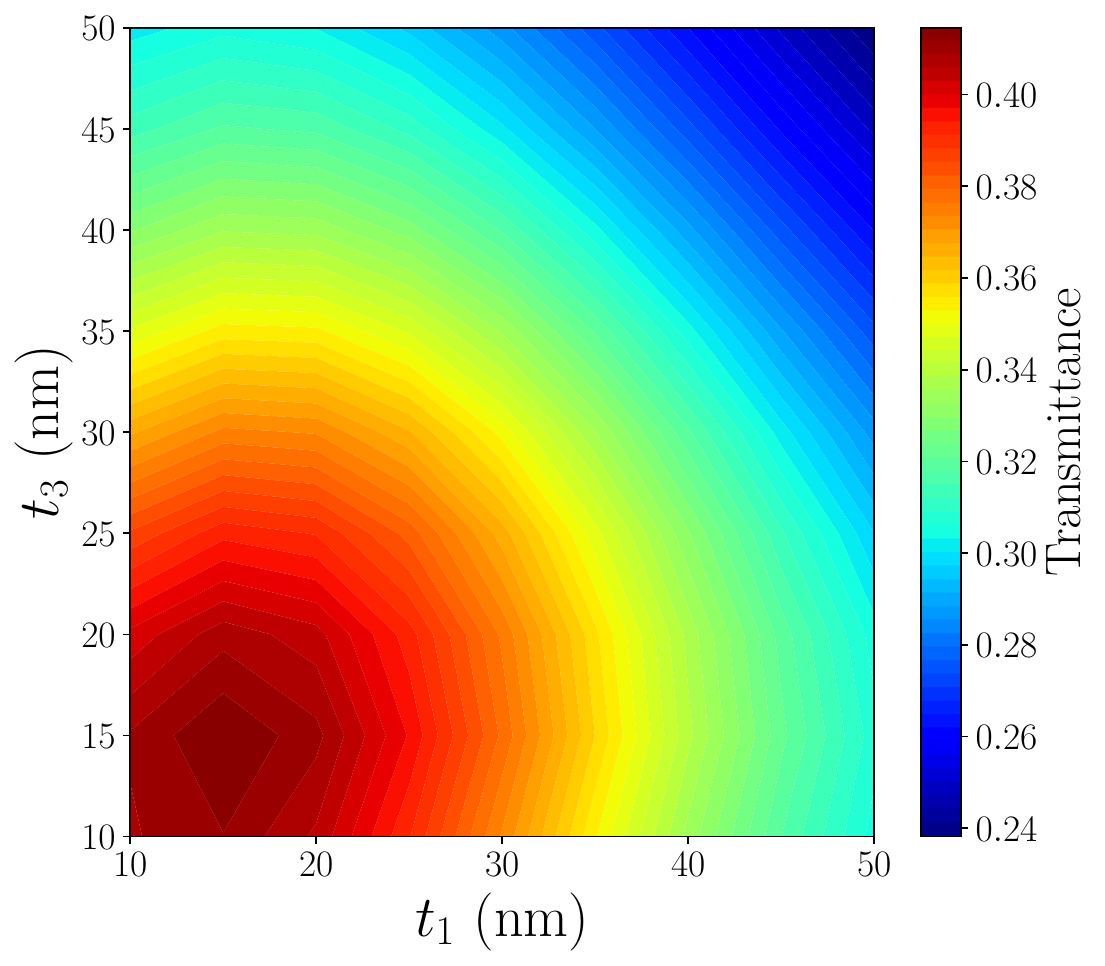}
        \caption{$t_2 = 18, r_1 = r_2 = 20, h_1 = h_2 = 50$}
    \end{subfigure}
    \caption{Visualization of the transmittance of the three-layer film with double-sided nanocones made of \ce{TiO2}/\ce{Ni}/\ce{TiO2}/\ce{TiO2}/\ce{TiO2} for three different fidelity levels, i.e., low fidelity (shown in left panels), medium fidelity (shown in center panels), and high fidelity (shown in right panels).}
    \label{fig:doublenanocones_tio2_ni_tio2}
\end{figure}

\begin{figure}[t]
    \centering
    \begin{subfigure}[b]{\textwidth}
        \centering
        \includegraphics[width=0.31\textwidth]{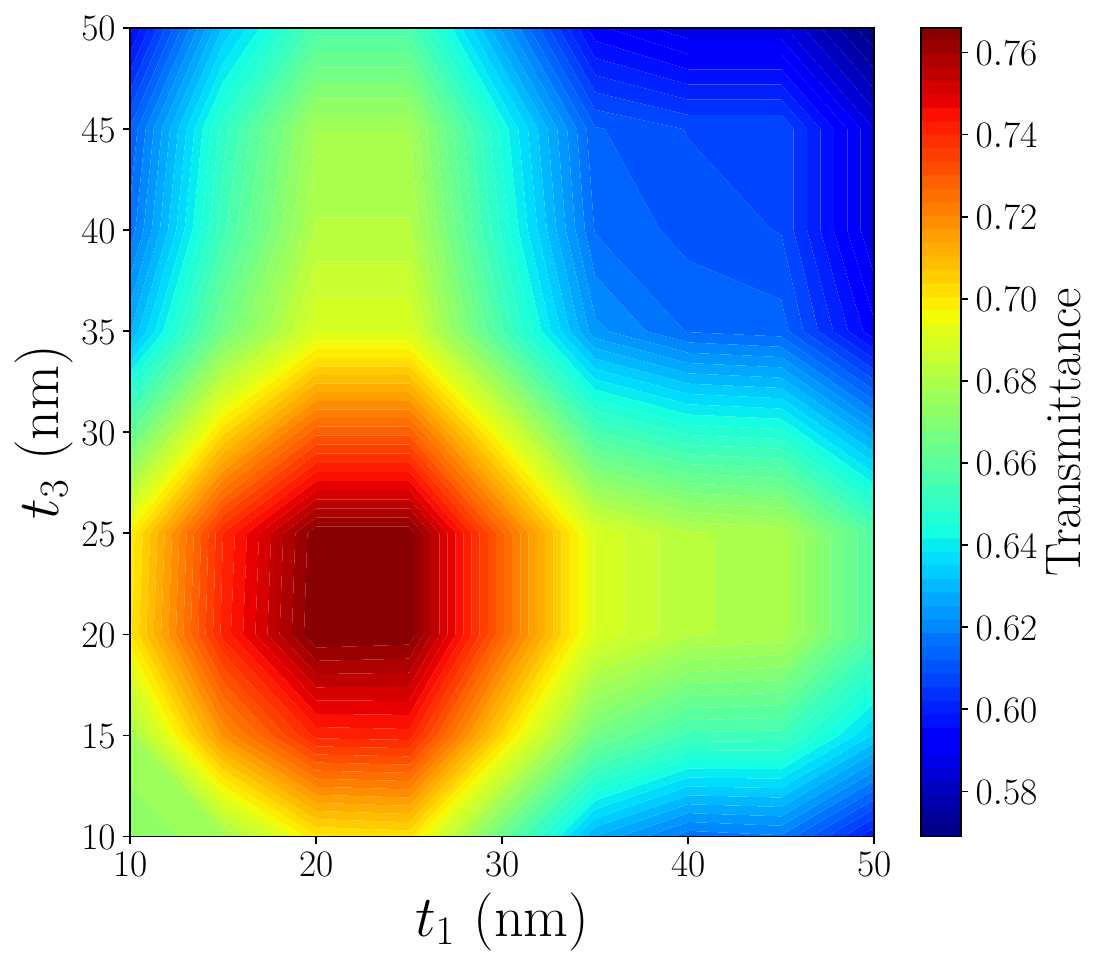}
        \includegraphics[width=0.31\textwidth]{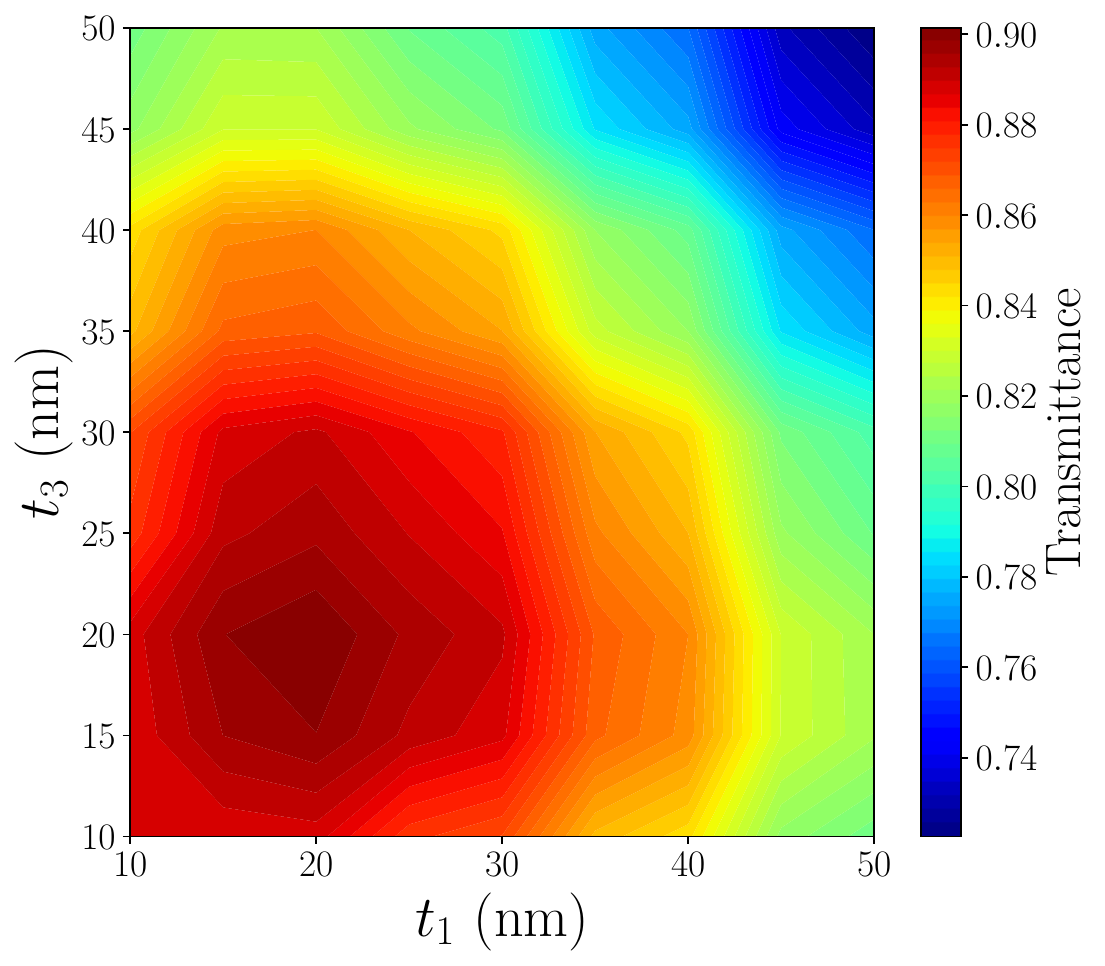}
        \includegraphics[width=0.31\textwidth]{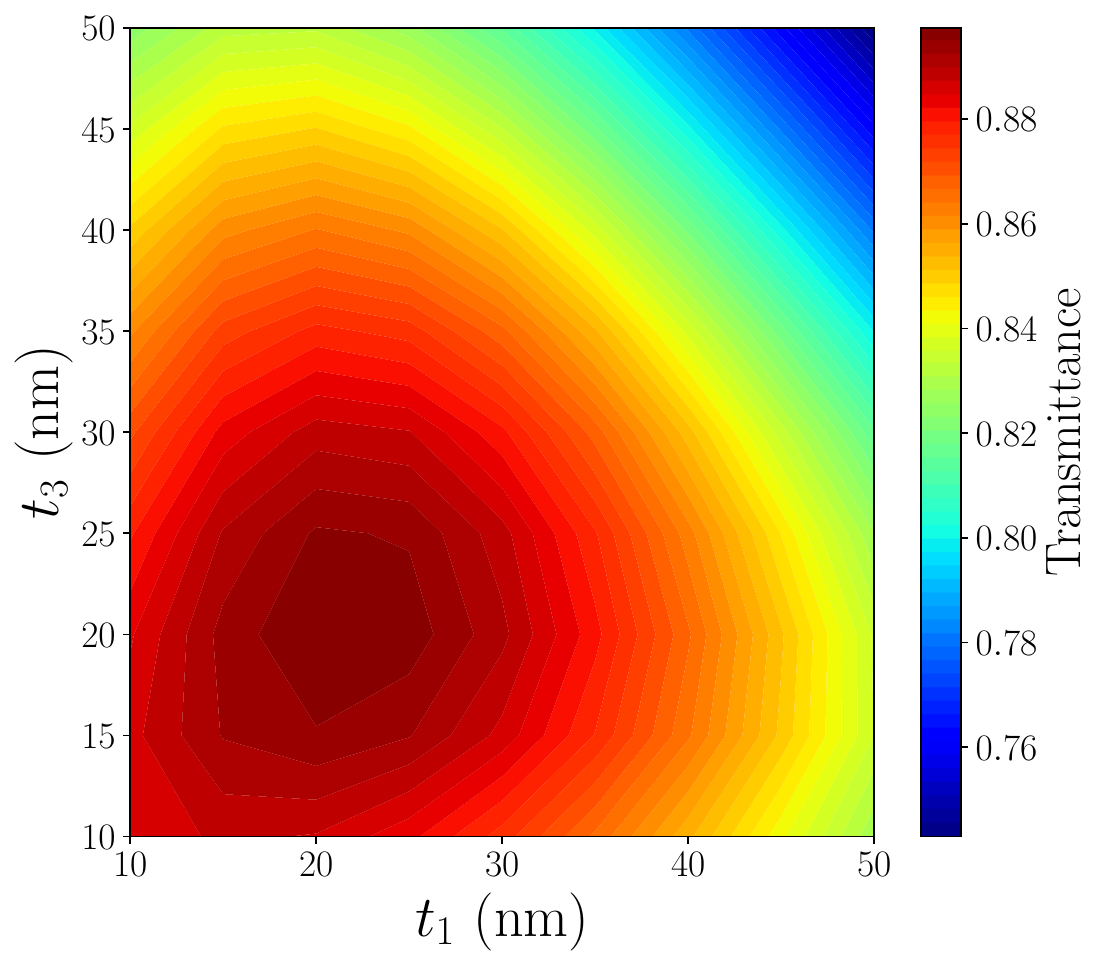}
        \caption{$t_2 = 8, r_1 = r_2 = 20, h_1 = h_2 = 50$}
    \end{subfigure}
    \begin{subfigure}[b]{\textwidth}
        \centering
        \includegraphics[width=0.31\textwidth]{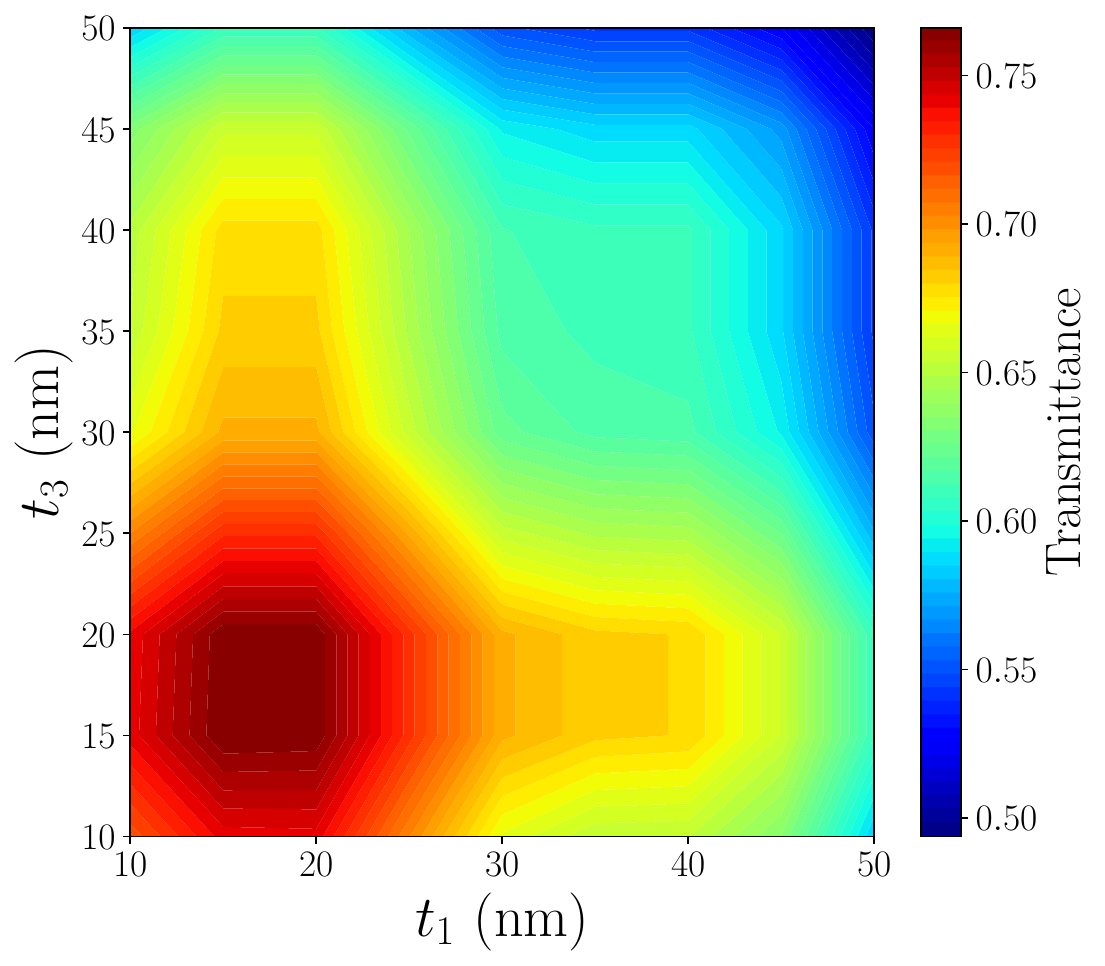}
        \includegraphics[width=0.31\textwidth]{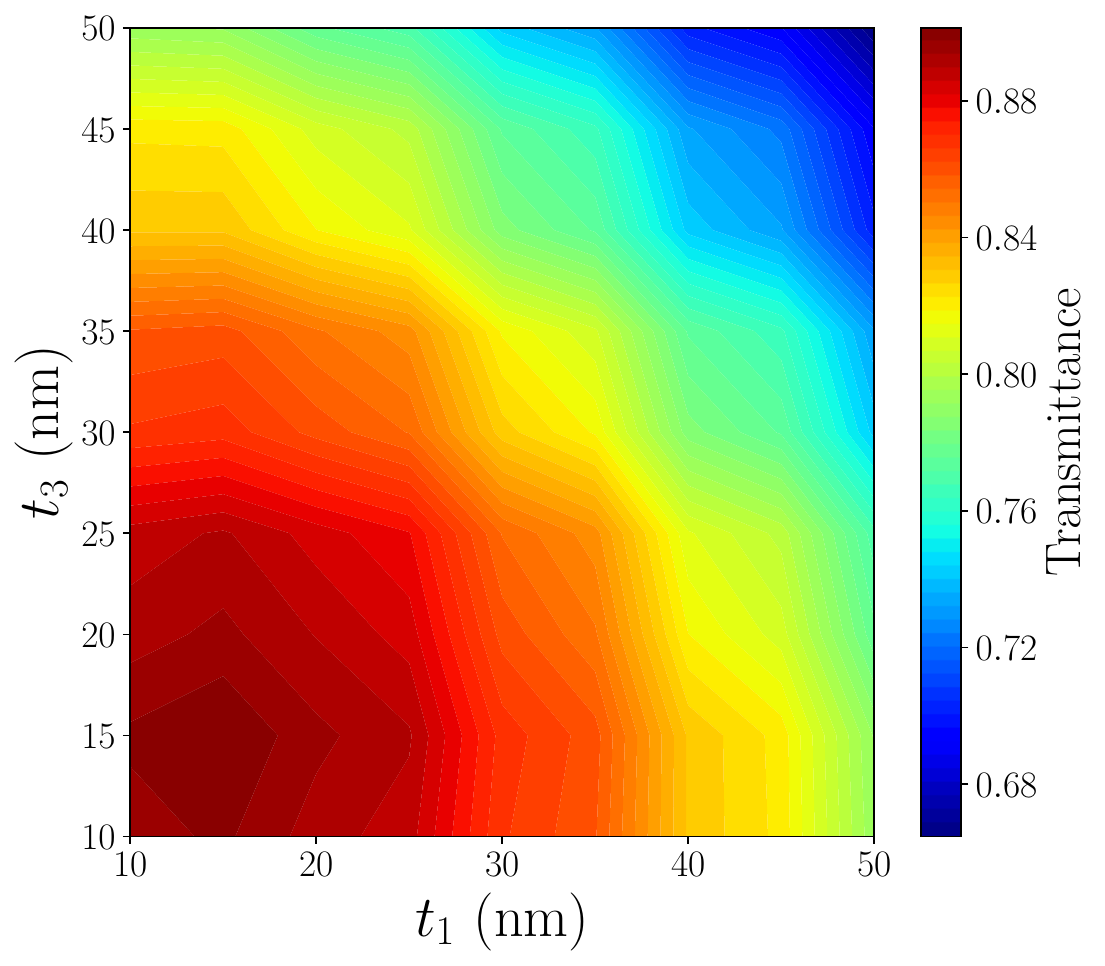}
        \includegraphics[width=0.31\textwidth]{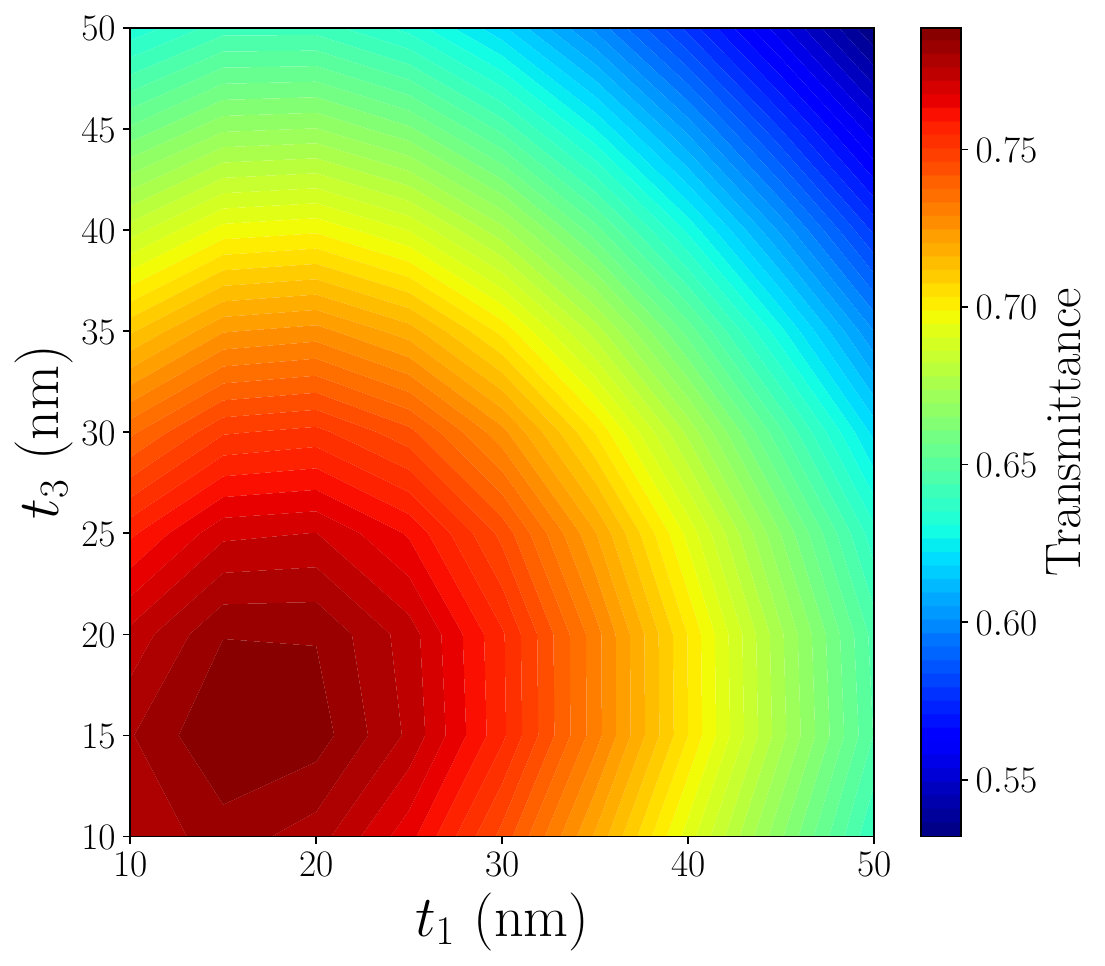}
        \caption{$t_2 = 18, r_1 = r_2 = 20, h_1 = h_2 = 50$}
    \end{subfigure}
    \caption{Visualization of the transmittance of the three-layer film with double-sided nanocones made of \ce{AZO}/\ce{Ag}/\ce{AZO}/\ce{AZO}/\ce{AZO} for three different fidelity levels, i.e., low fidelity (shown in left panels), medium fidelity (shown in center panels), and high fidelity (shown in right panels).}
    \label{fig:doublenanocones_azo_ag_azo}
\end{figure}

\begin{figure}[t]
    \centering
    \begin{subfigure}[b]{\textwidth}
        \centering
        \includegraphics[width=0.31\textwidth]{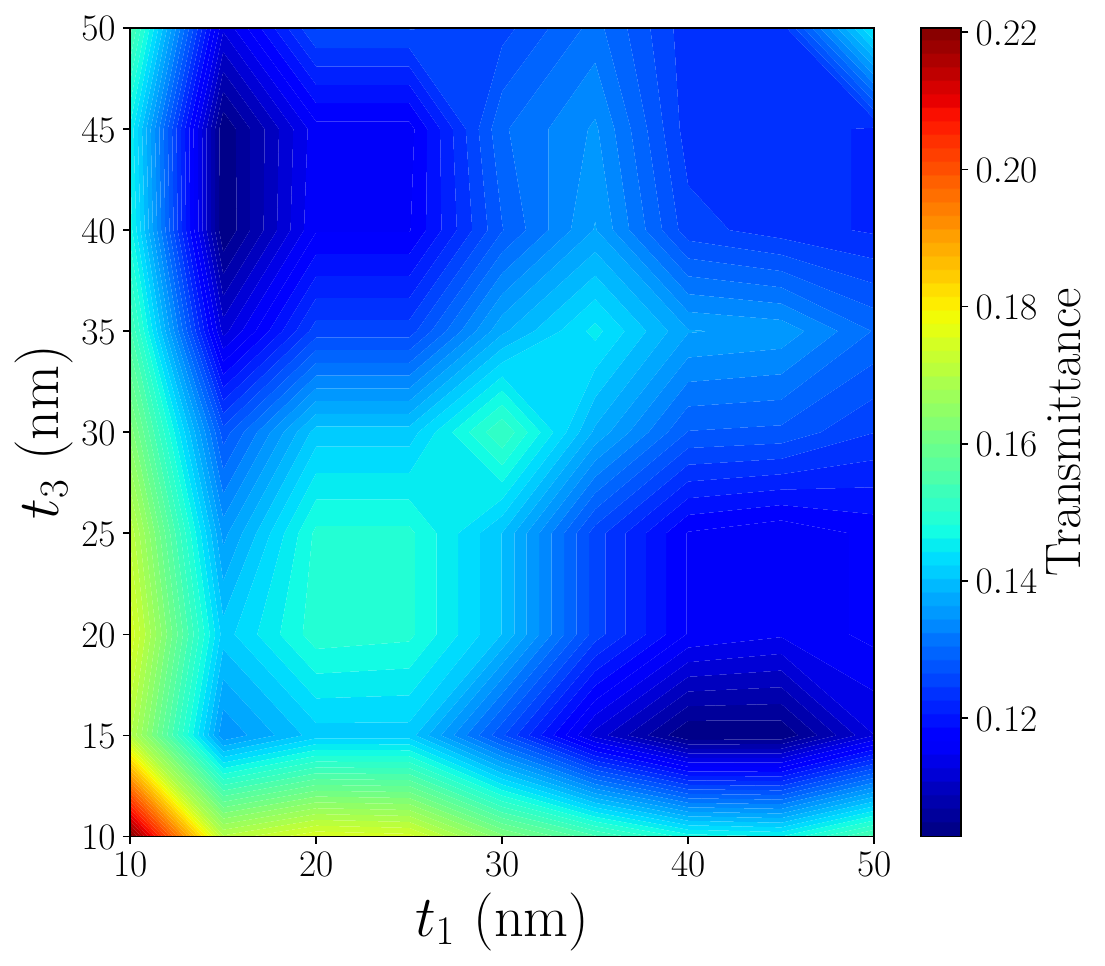}
        \includegraphics[width=0.31\textwidth]{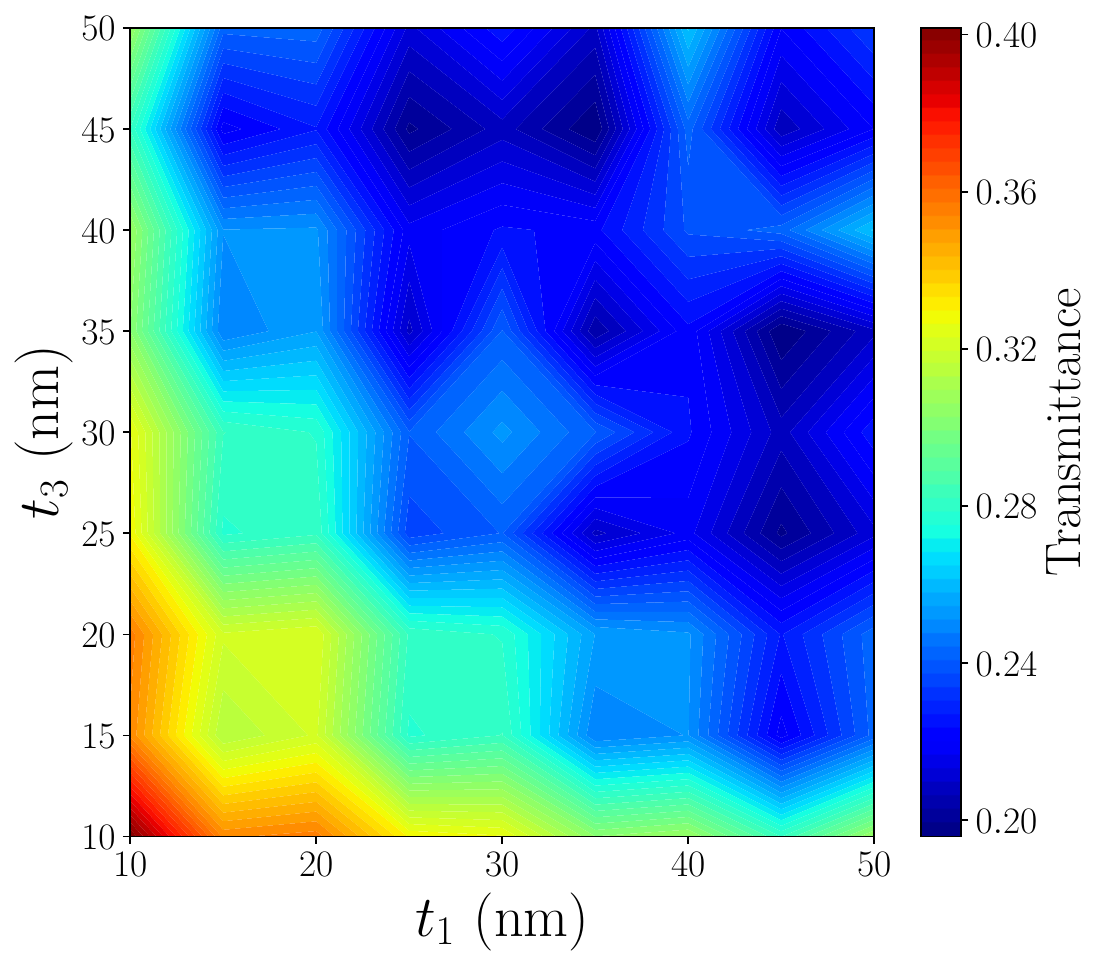}
        \includegraphics[width=0.31\textwidth]{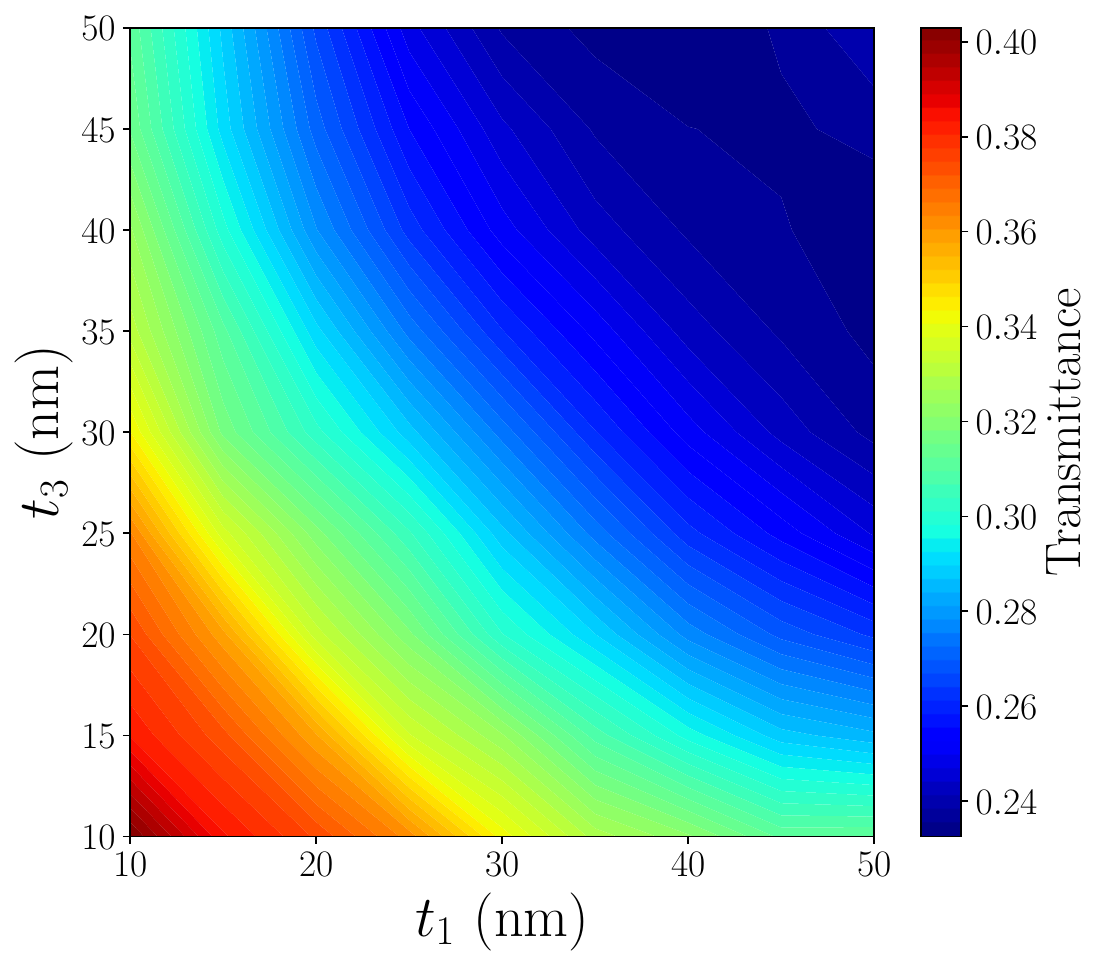}
        \caption{$t_2 = 8, r_1 = r_2 = 20, h_1 = h_2 = 50$}
    \end{subfigure}
    \begin{subfigure}[b]{\textwidth}
        \centering
        \includegraphics[width=0.31\textwidth]{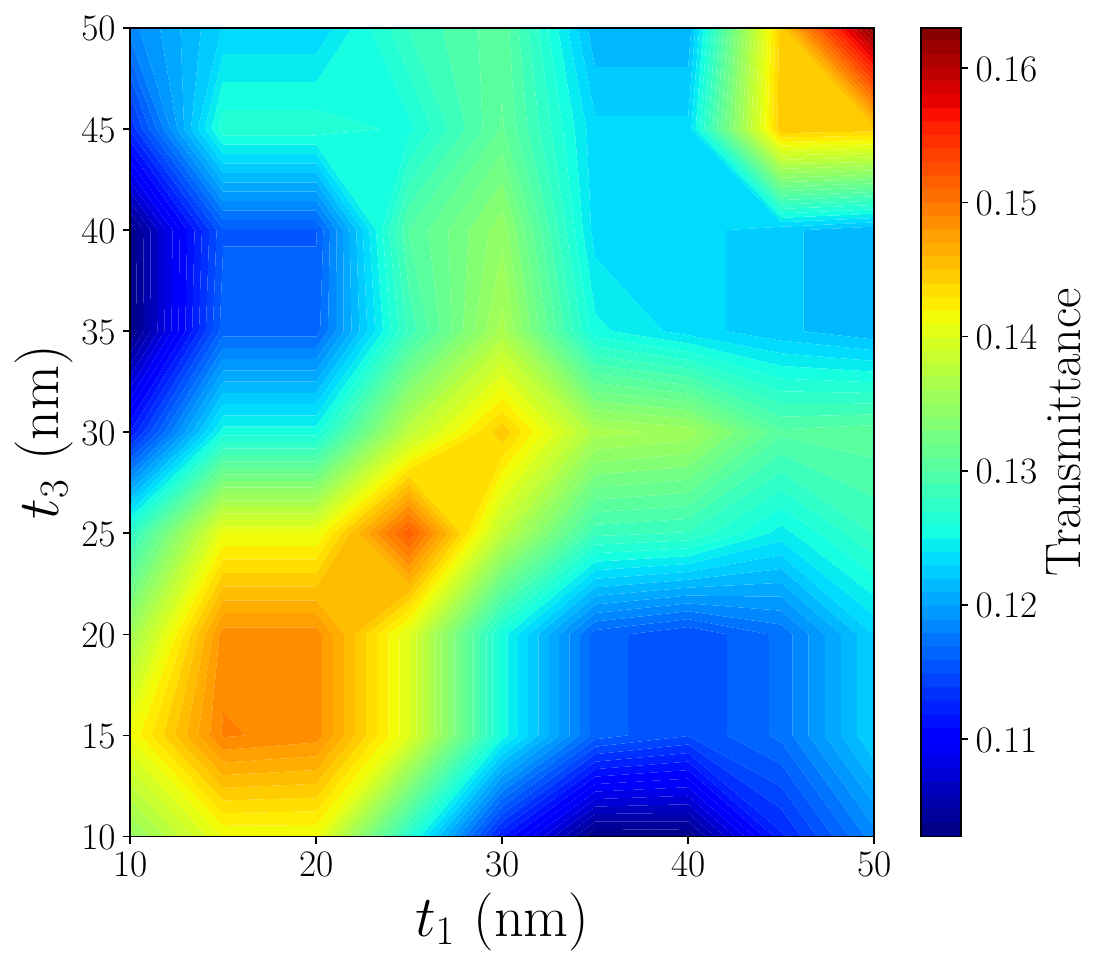}
        \includegraphics[width=0.31\textwidth]{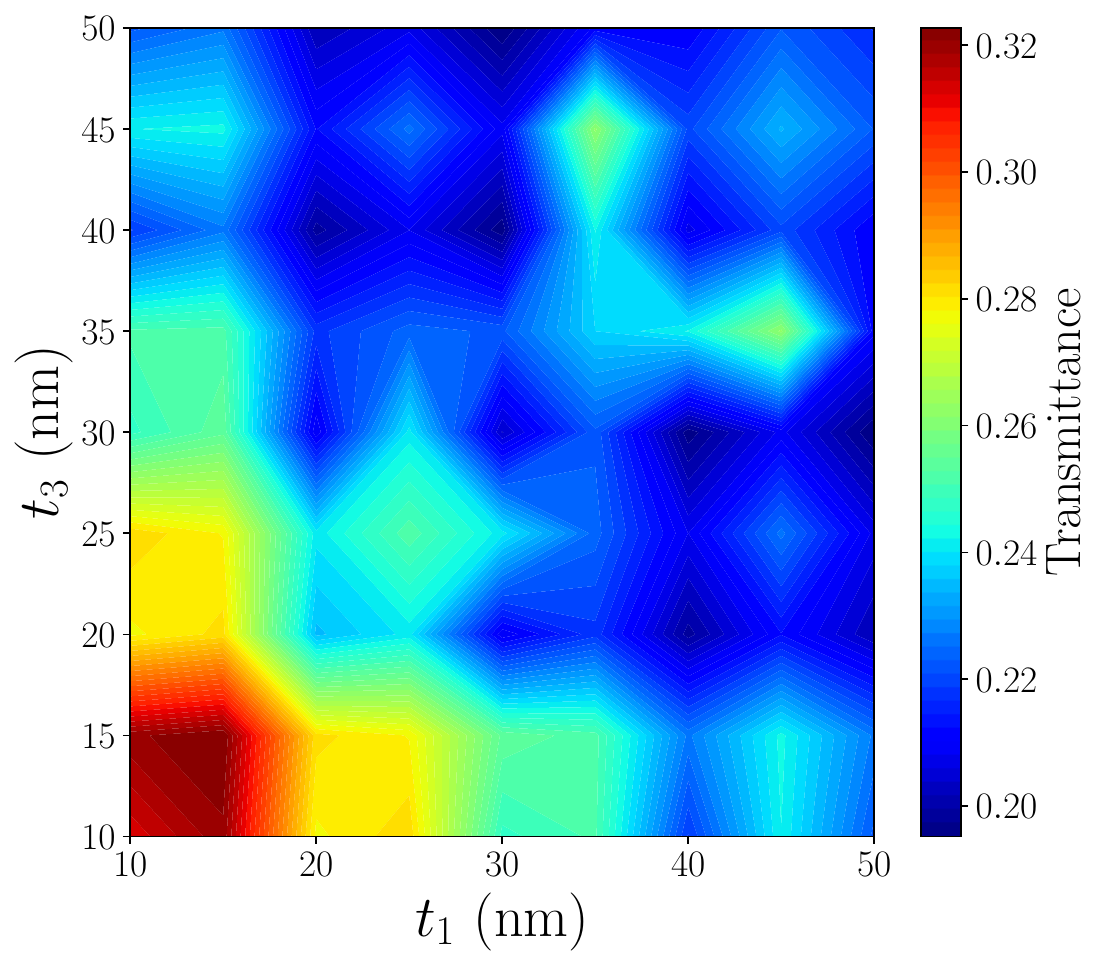}
        \includegraphics[width=0.31\textwidth]{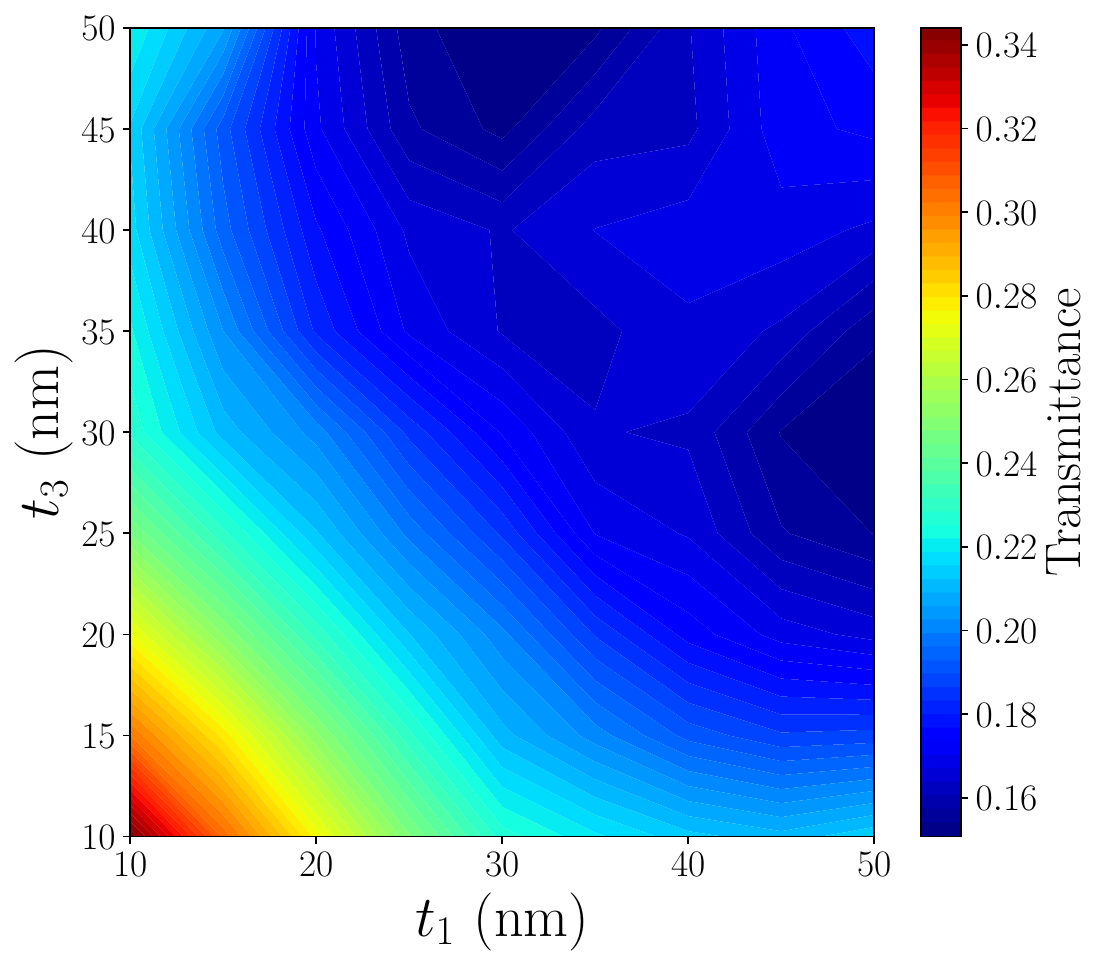}
        \caption{$t_2 = 18, r_1 = r_2 = 20, h_1 = h_2 = 50$}
    \end{subfigure}
    \caption{Visualization of the transmittance of the three-layer film with double-sided nanocones made of \ce{cSi}/\ce{Ag}/\ce{cSi}/\ce{cSi}/\ce{cSi} for three different fidelity levels, i.e., low fidelity (shown in left panels), medium fidelity (shown in center panels), and high fidelity (shown in right panels).}
    \label{fig:doublenanocones_csi_ag_csi}
\end{figure}

\begin{figure}[t]
    \centering
    \begin{subfigure}[b]{\textwidth}
        \centering
        \includegraphics[width=0.31\textwidth]{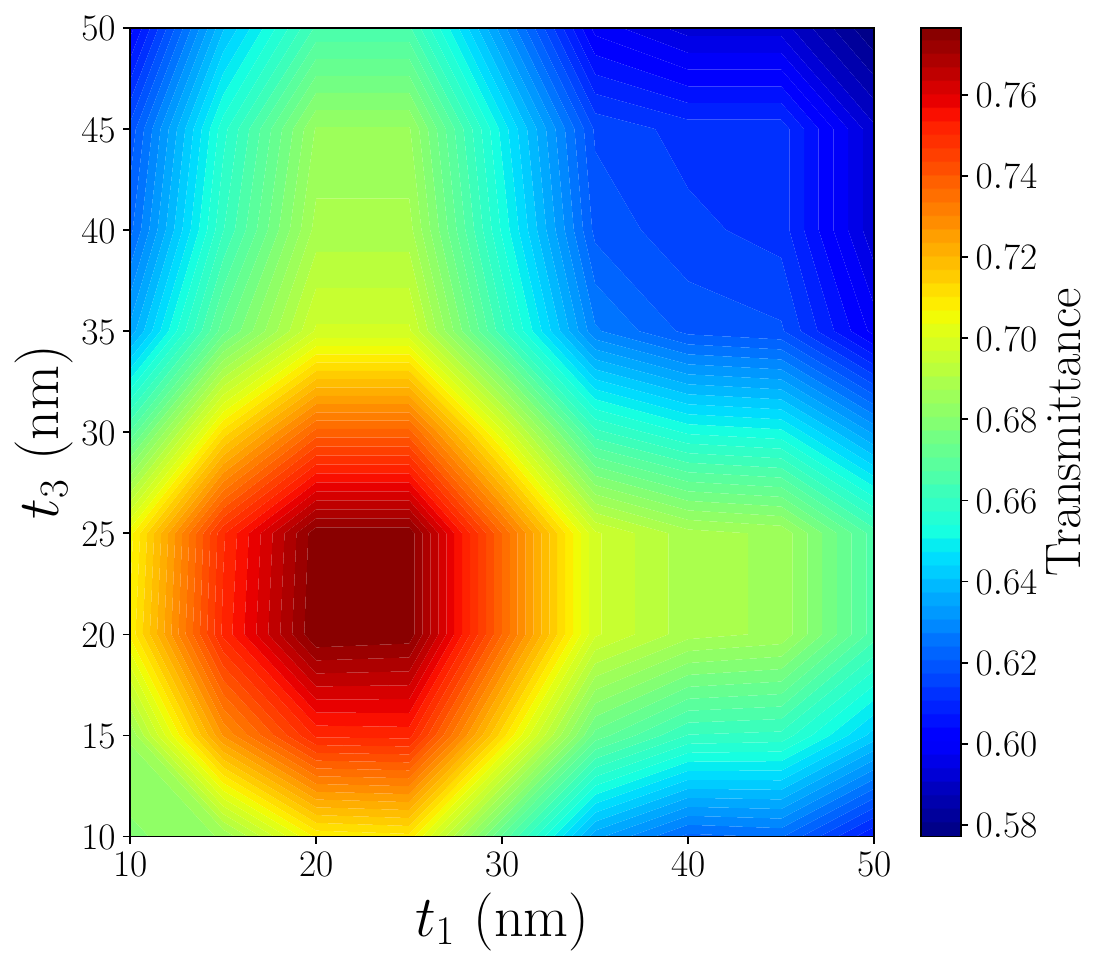}
        \includegraphics[width=0.31\textwidth]{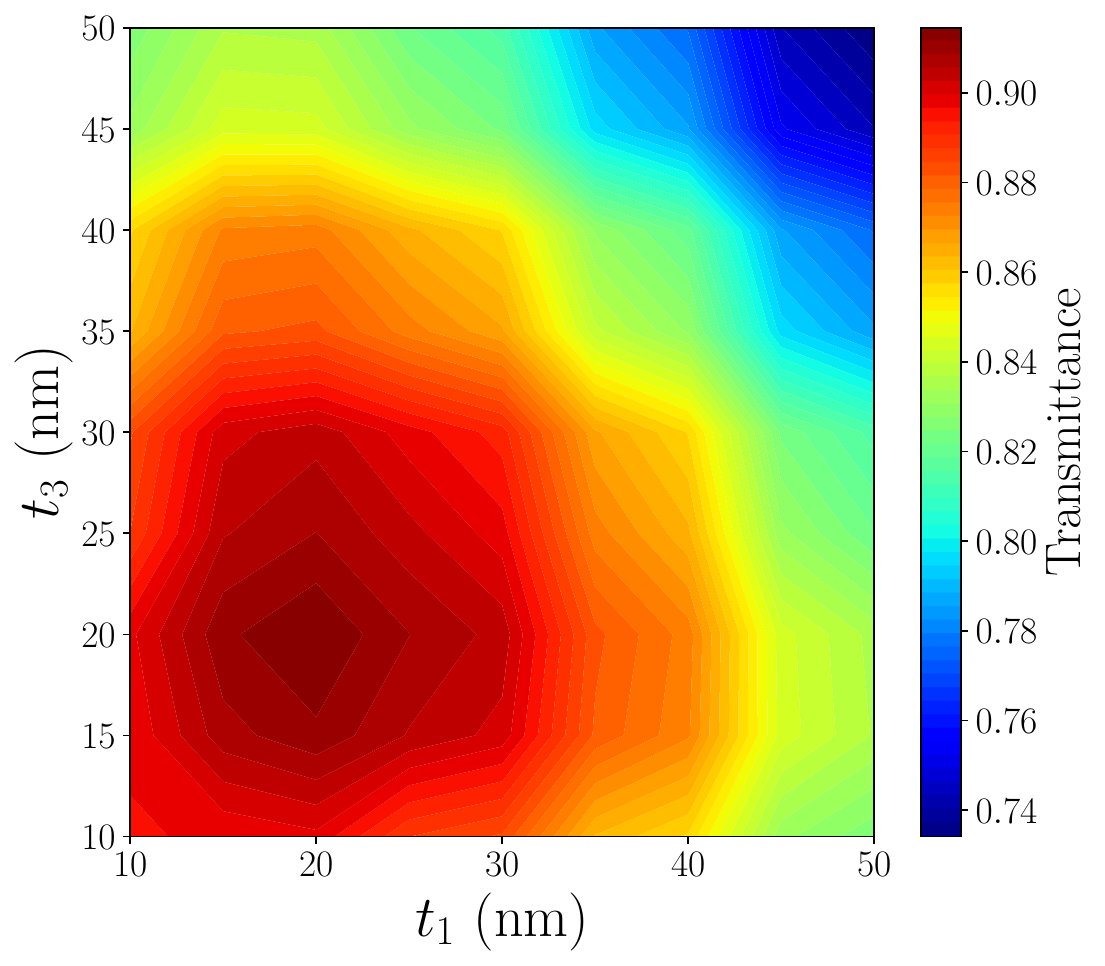}
        \includegraphics[width=0.31\textwidth]{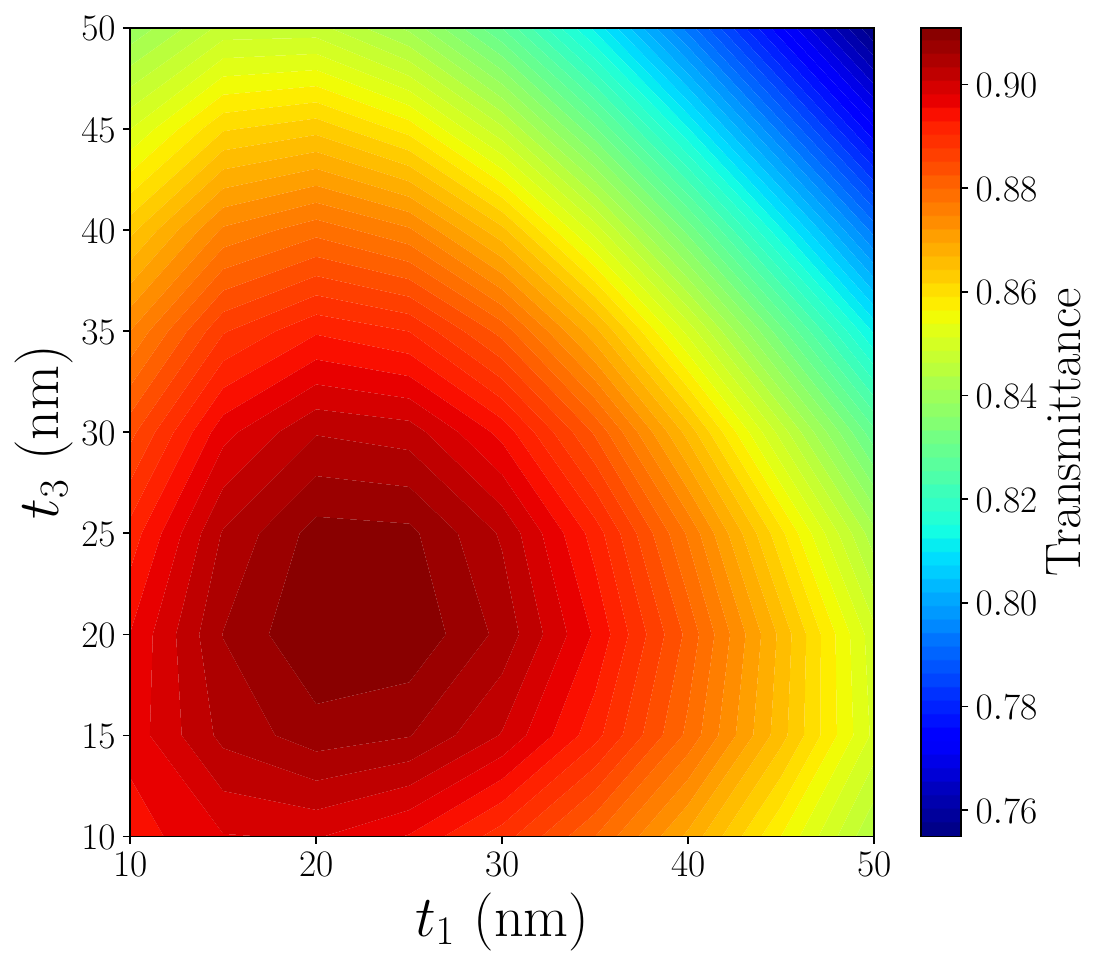}
        \caption{$t_2 = 8, r_1 = r_2 = 20, h_1 = h_2 = 50$}
    \end{subfigure}
    \begin{subfigure}[b]{\textwidth}
        \centering
        \includegraphics[width=0.31\textwidth]{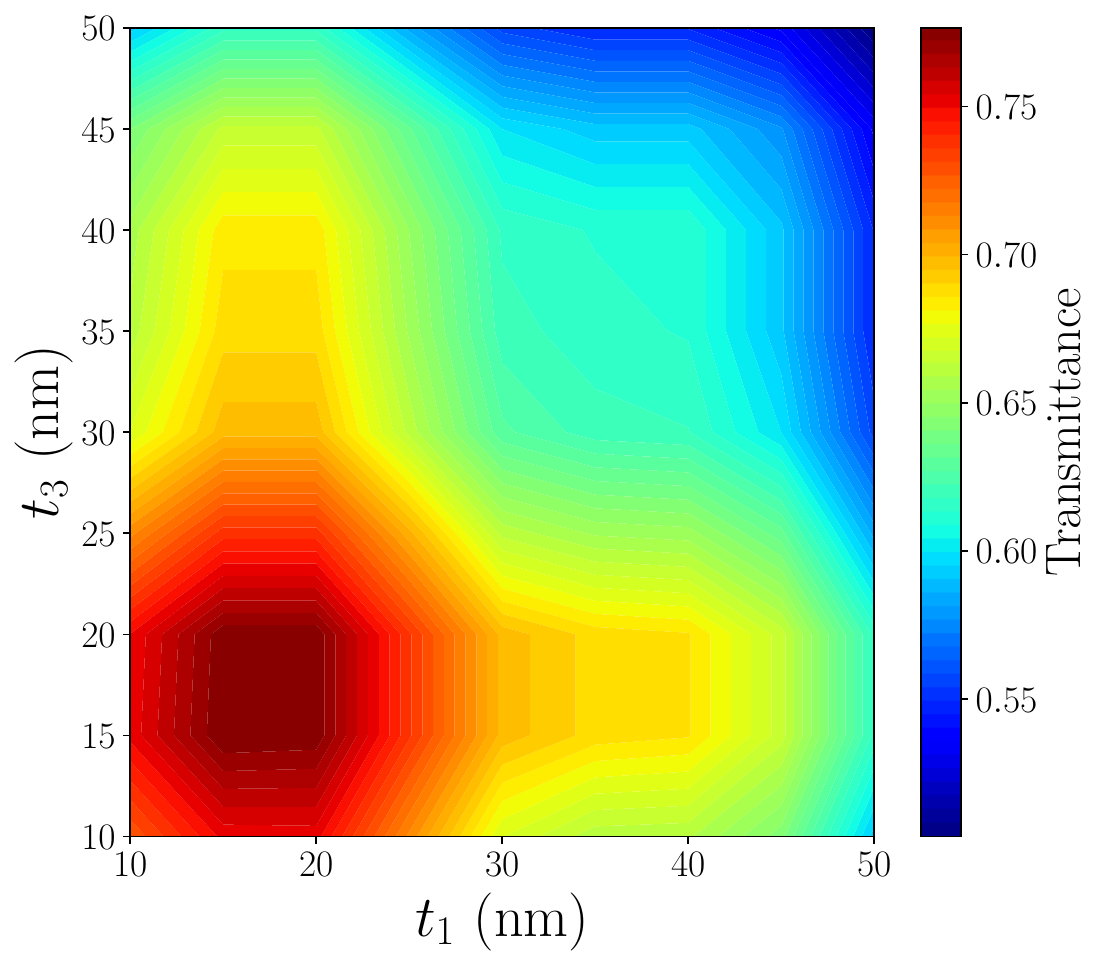}
        \includegraphics[width=0.31\textwidth]{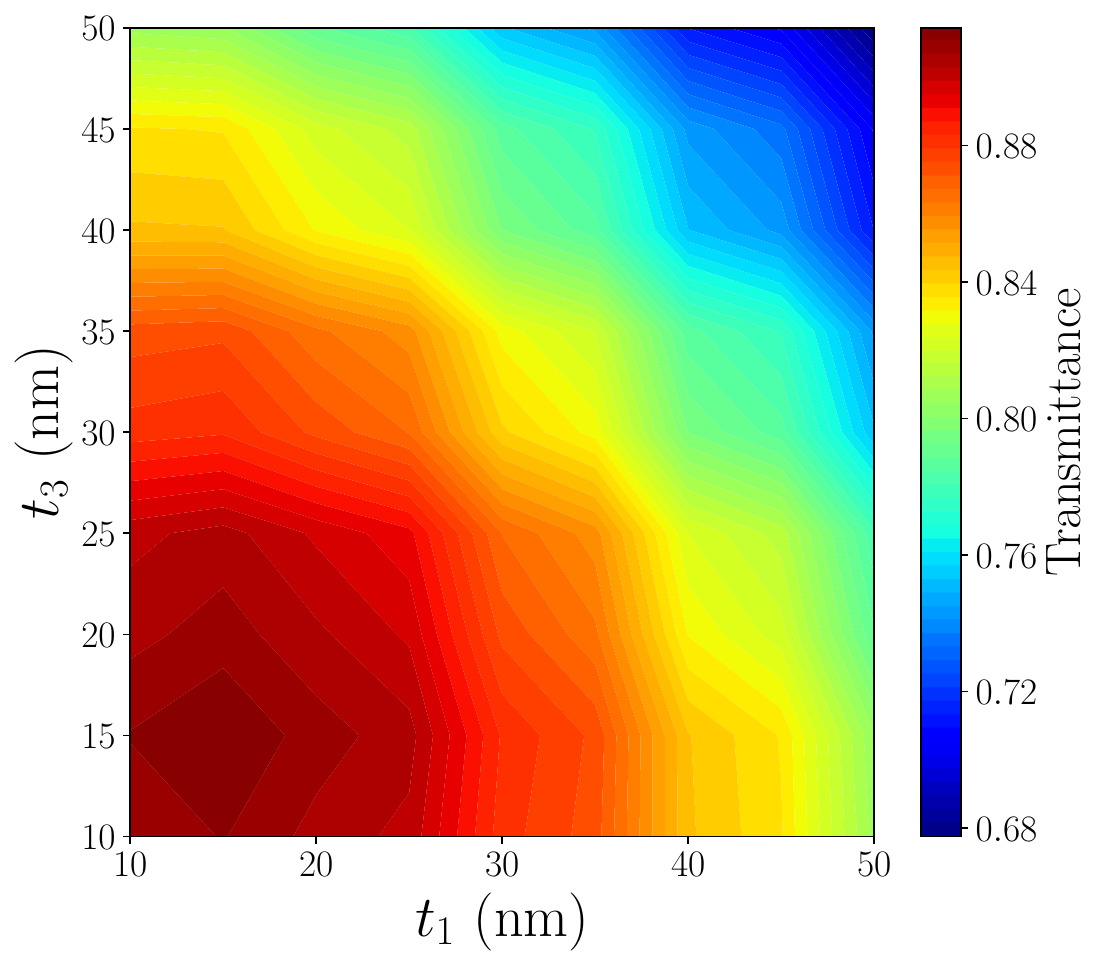}
        \includegraphics[width=0.31\textwidth]{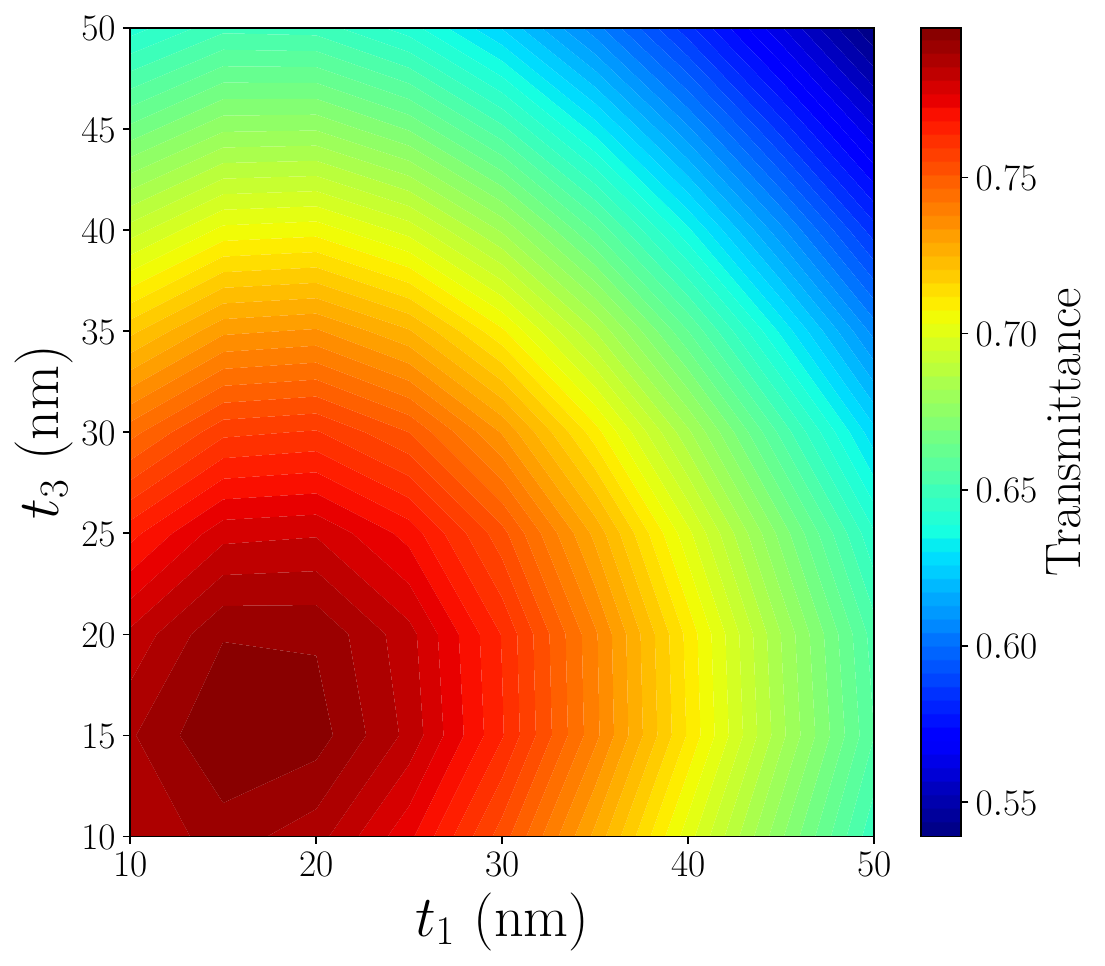}
        \caption{$t_2 = 18, r_1 = r_2 = 20, h_1 = h_2 = 50$}
    \end{subfigure}
    \caption{Visualization of the transmittance of the three-layer film with double-sided nanocones made of \ce{ITO}/\ce{Ag}/\ce{ITO}/\ce{ITO}/\ce{ITO} for three different fidelity levels, i.e., low fidelity (shown in left panels), medium fidelity (shown in center panels), and high fidelity (shown in right panels).}
    \label{fig:doublenanocones_ito_ag_ito}
\end{figure}

\begin{figure}[t]
    \centering
    \begin{subfigure}[b]{\textwidth}
        \centering
        \includegraphics[width=0.31\textwidth]{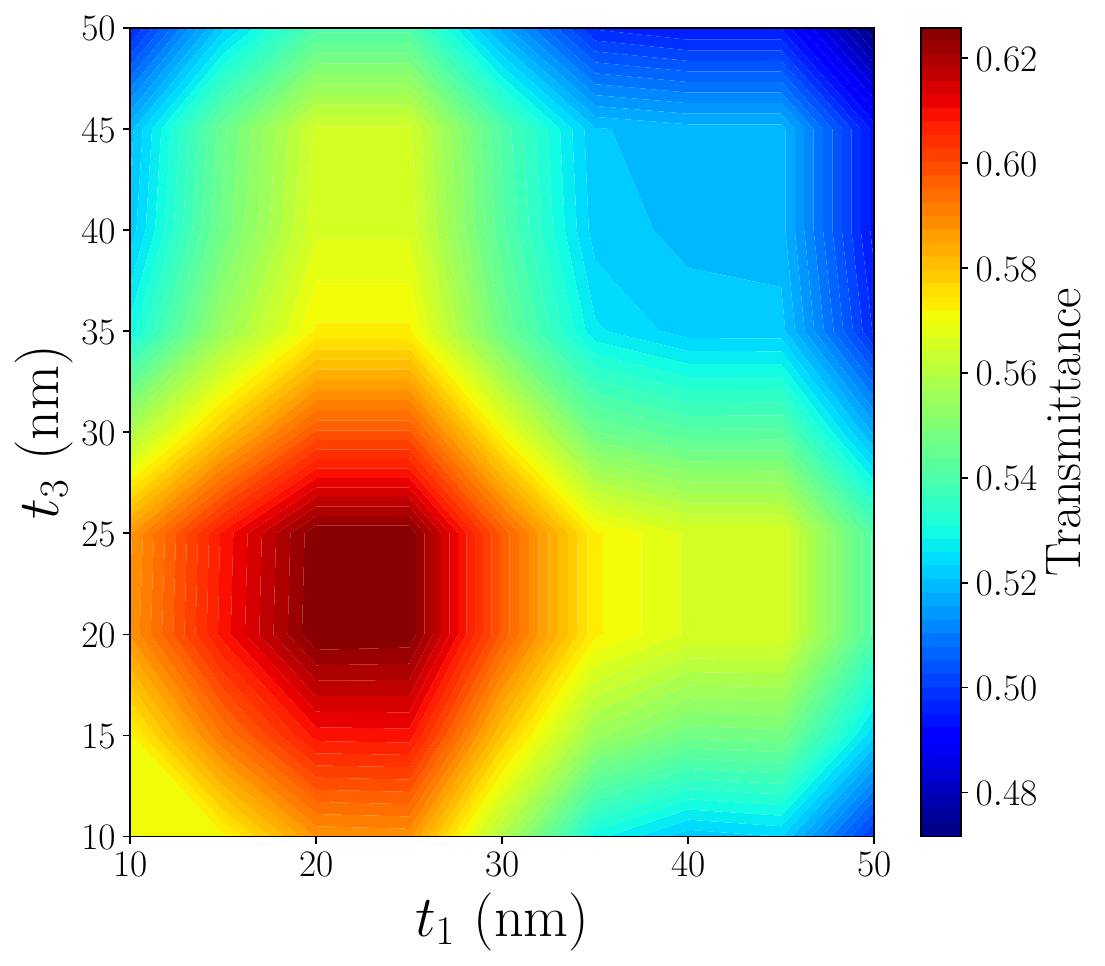}
        \includegraphics[width=0.31\textwidth]{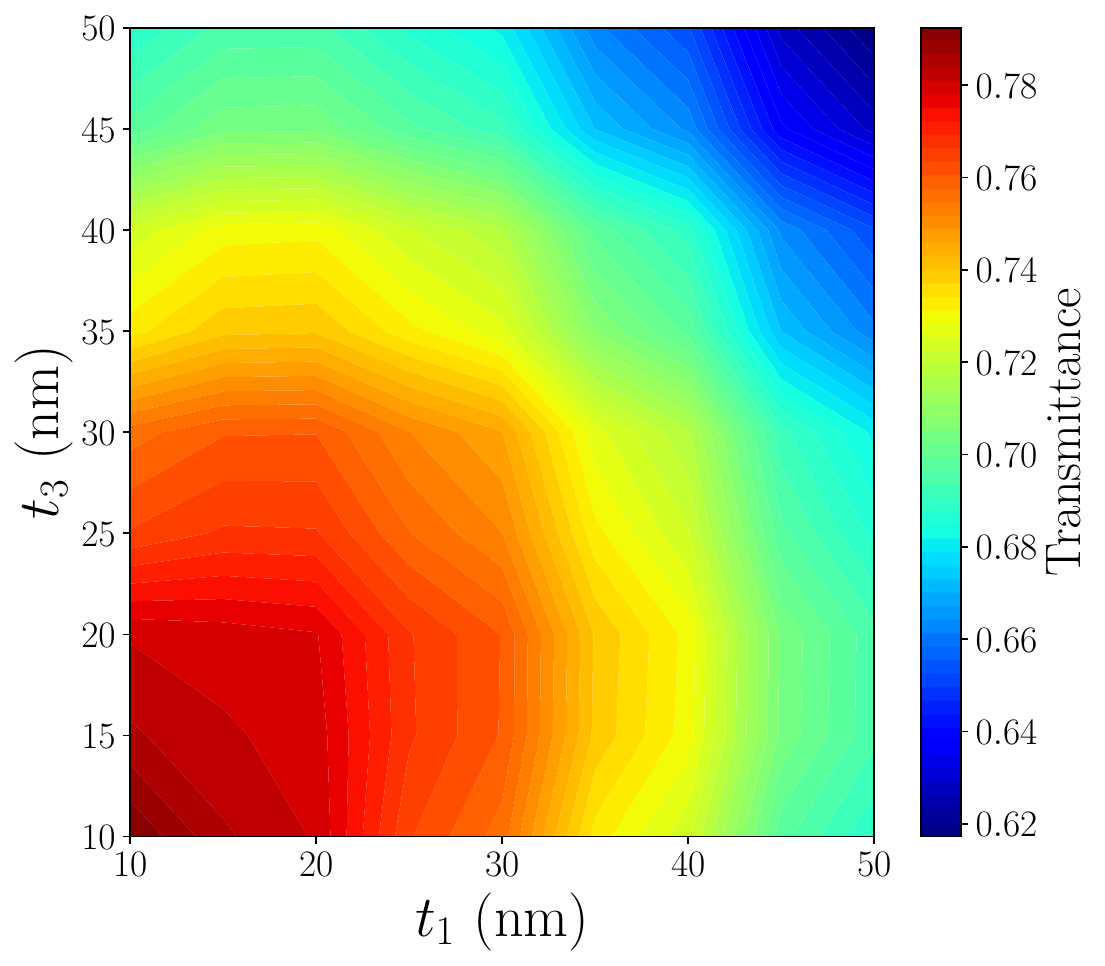}
        \includegraphics[width=0.31\textwidth]{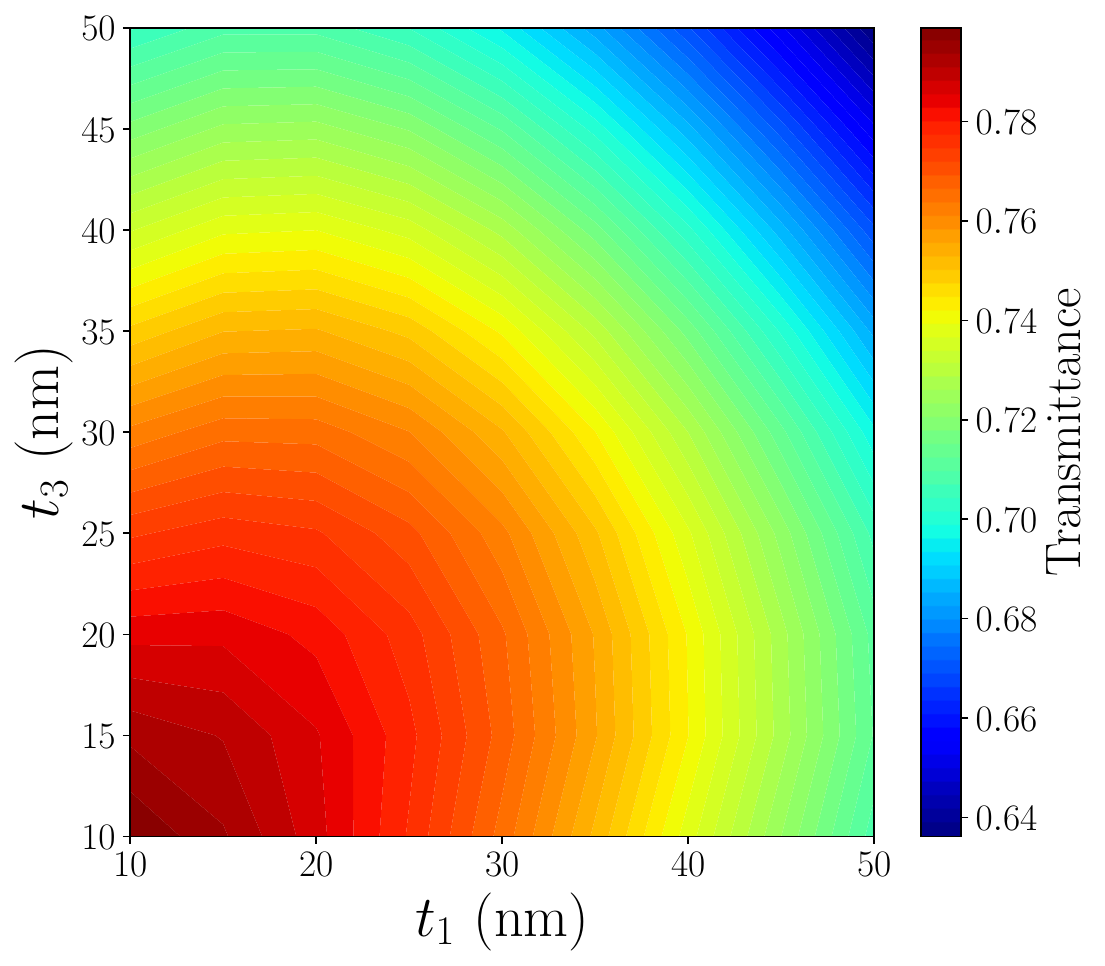}
        \caption{$t_2 = 8, r_1 = r_2 = 20, h_1 = h_2 = 50$}
    \end{subfigure}
    \begin{subfigure}[b]{\textwidth}
        \centering
        \includegraphics[width=0.31\textwidth]{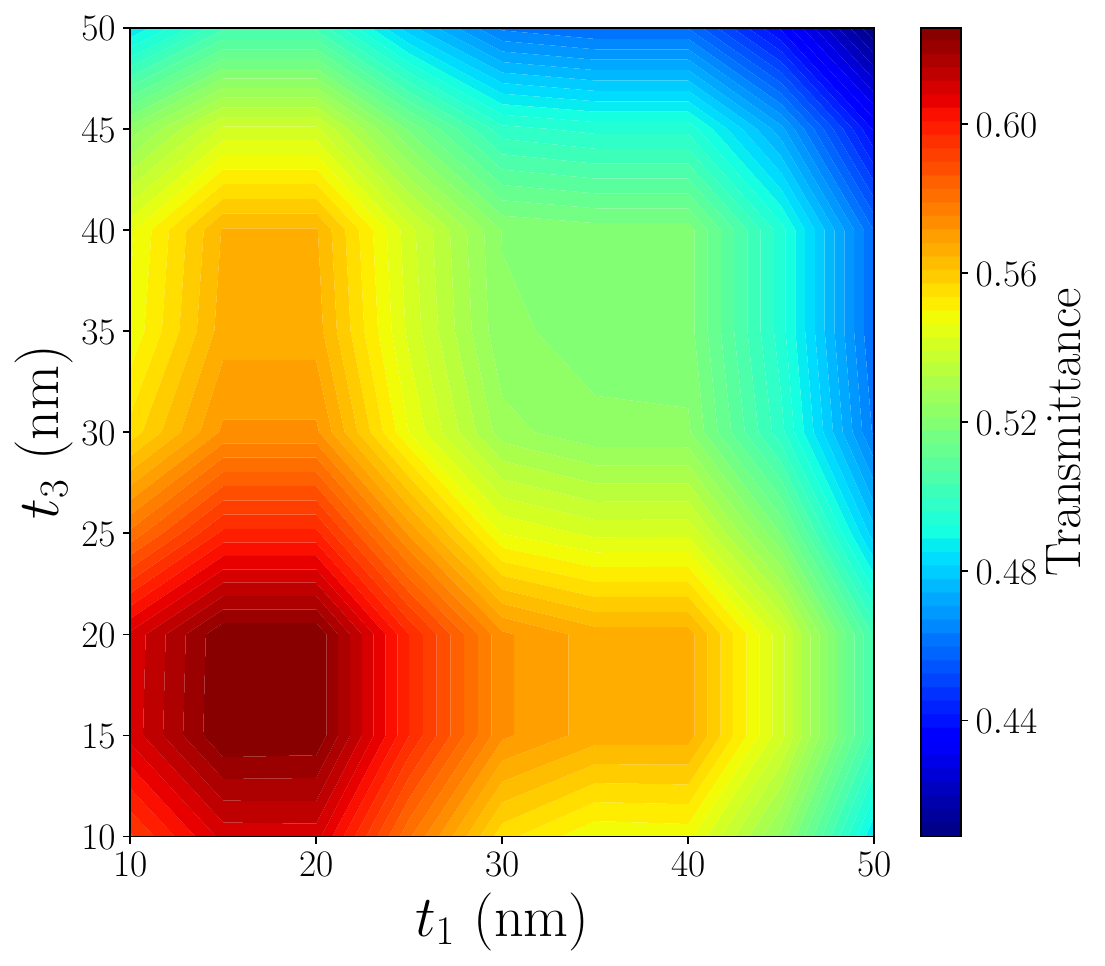}
        \includegraphics[width=0.31\textwidth]{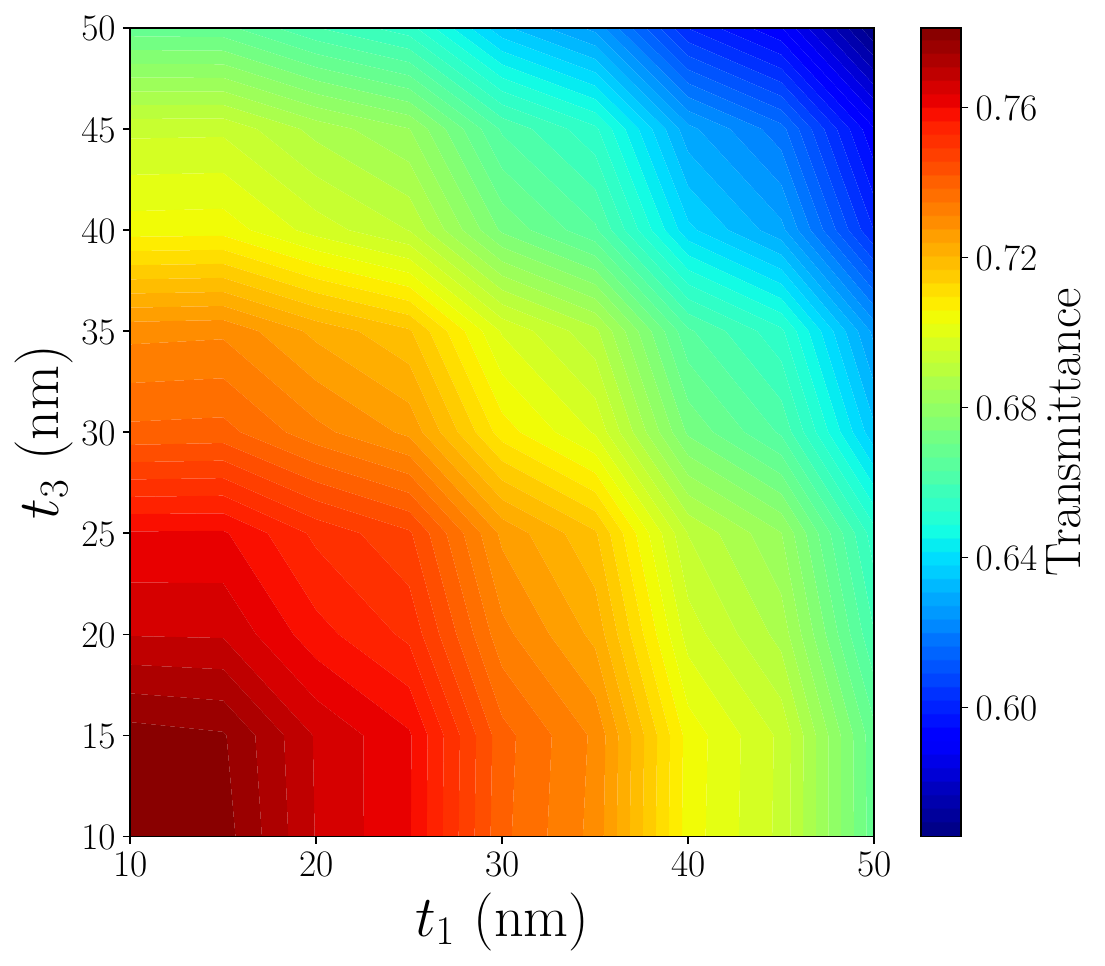}
        \includegraphics[width=0.31\textwidth]{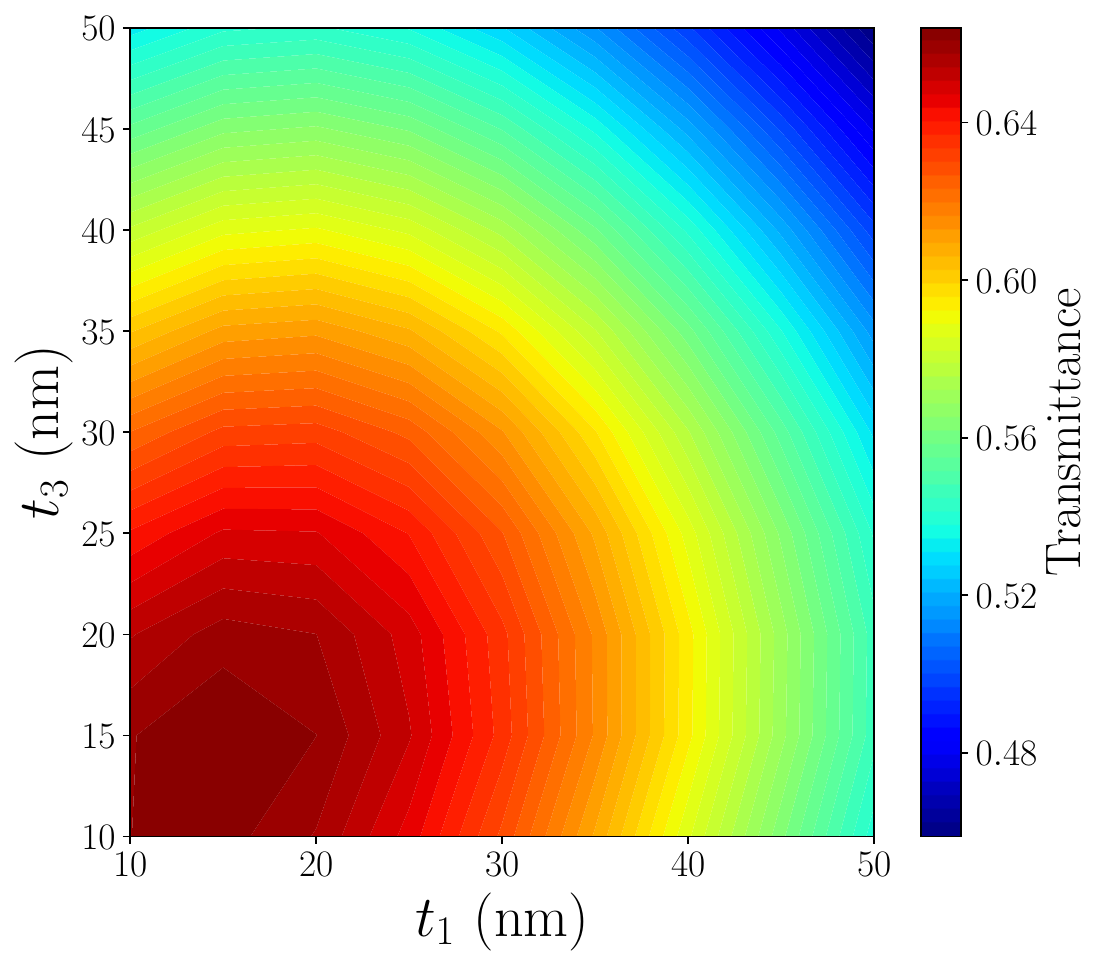}
        \caption{$t_2 = 18, r_1 = r_2 = 20, h_1 = h_2 = 50$}
    \end{subfigure}
    \caption{Visualization of the transmittance of the three-layer film with double-sided nanocones made of \ce{ZnO}/\ce{Ag}/\ce{ZnO}/\ce{ZnO}/\ce{ZnO} for three different fidelity levels, i.e., low fidelity (shown in left panels), medium fidelity (shown in center panels), and high fidelity (shown in right panels).}
    \label{fig:doublenanocones_zno_ag_zno}
\end{figure}

Along with the visualizations presented in~\secref{sec:visualization},
additional examples pertaining to various nanophotonic structures are provided in this section.
For the three-layer film structure, we display simulation results with different material compositions in Figures~\ref{fig:threelayers_tio2_ag_tio2} through~\ref{fig:threelayers_zno_ag_zno}.
Specifically, these illustrate variations in the nanophotonic structure when different materials are used in its composition.
Next, we present simulation results for the anti-reflective nanocones in~\figref{fig:nanocones_all},
showcasing variations in this particular structure type.
This is followed by simulation results for the vertical nanowires,
exhibited in~\figref{fig:nanowires_all},
and the close-packed nanospheres, shown in~\figref{fig:nanospheres_all}.
Furthermore,
we provide the visualization of simulations results for the three-layer film with double-sided nanocones in Figures~\ref{fig:doublenanocones_tio2_ag_tio2} through~\ref{fig:doublenanocones_zno_ag_zno}.
Similar to other visualization,
these demonstrate how the results change with different material compositions.
It is worth noting that for nanophotonic structures characterized by more than two parameters, we vary two of the parameters while keeping the others constant.
This approach allows us to effectively represent their parameter spaces in two-dimensional plots.

\section{Results on Elapsed Time}
\label{sec:results_elapsed_time}

\begin{table}[ht]
    \caption{Results on elapsed time for diverse nanophotonic structures and fidelity levels. Mean and standard deviation are reported. All values are in seconds.}
    \label{tab:elapsed_time}
    \begin{center}
    \small
    \setlength{\tabcolsep}{2pt}
    \begin{tabular}{lccc}
        \toprule
        \textbf{Structure} & \textbf{Low Fidelity} & \textbf{Medium Fidelity} & \textbf{High Fidelity}  \\
        \midrule
        Three-layer film, \ce{TiO2}/\ce{Ag}/\ce{TiO2} & 0.5927 $\pm$ 0.0810 & 1.1277 $\pm$ 0.1373 & 54.4757 $\pm$ 23.1198 \\
        Anti-reflective nanocones, fused silica & 0.5999 $\pm$ 0.1357 & 0.6762 $\pm$ 0.1583 & 60.6909 $\pm$ 54.9886 \\
        Vertical nanowires, \ce{cSi} & 1.0668 $\pm$ 0.2669 & 1.5037 $\pm$ 0.4536 & 377.0777 $\pm$ 215.7626 \\
        Close-packed nanospheres, \ce{cSi}/\ce{TiO2} & 1.5695 $\pm$ 3.2443 & 2.3399 $\pm$ 4.8280 & 673.3744 $\pm$ 848.0346 \\
        Film with nanocones, \ce{TiO2}/\ce{Ag}/\ce{TiO2}/\ce{TiO2}/\ce{TiO2} & 2.2194 $\pm$ 0.7021 & 2.3046 $\pm$ 0.7497 & 115.0109 $\pm$ 57.6097 \\
        \bottomrule
    \end{tabular}
    \end{center}
\end{table}

We demonstrate results on elapsed time for the nanophotonic structures we study and three fidelity levels in~\tabref{tab:elapsed_time}.
Since we run a large number of simulations through a job scheduler,
i.e., the Slurm workload manager,
the respective simulations can be run on distinct machines; see~\secref{sec:details_experiments}.
Although we assign the same number of threads for each job,
the time elapsed for simulations might vary depending on the status of machines.
Nevertheless,
assuming that all simulation jobs are fairly distributed to the machines and machine states are also similar on average,
the results on elapsed time are provided for analysis.
As expected,
simulations with low fidelity is faster than ones with medium fidelity and high fidelity,
and simulations with high fidelity is slower than the others.
Moreover,
the time differences between low fidelity and medium fidelity are marginal
compared to the differences between medium fidelity and high fidelity
or between low fidelity and high fidelity.

\section{Details of Regression Models}
\label{sec:details_regression_models}

To establish a regression model for predicting a target optical property,
we begin by partitioning our dataset into subsets designated for training, validation, and testing.
Specifically, these disjoint subsets are 70\%, 10\%, and 20\% of the entire dataset, respectively.

The regression model for the surrogate model mode is a multi-layer perceptron, detailed as follows:
\begin{itemize}
\item[] First layer: (the number of parameters, 128, batch normalization, ReLU);
\item[] Second layer: (128, 64, batch normalization, ReLU);
\item[] Third layer: (64, 1, --, Logistic),
\end{itemize}
where $(x, y, f, g)$ implies (the number of input dimensions, the number of output dimensions, a normalization technique, an activation function).
Training of this network is performed using the Adam optimizer~\citep{KingmaDP2015iclr} and the mean squared error,
configured with a learning rate of 0.001 and a batch size of 64.
The network architecture and its associated components including batch normalization~\citep{IoffeS2015icml}
are implemented using PyTorch~\citep{PaszkeA2019neurips}.
To enhance the training process and select an optimal model,
we employ early stopping based on the validation dataset's performance.
We set the maximum number of training epochs as 200 and use the average of the last five validation losses as a threshold to determine when to stop training early.
This approach helps us to finalize our model by ensuring that it generalizes well without overfitting.

Upon completion of the training process,
the final model's performance on the test dataset is evaluated in terms of mean squared errors,
yielding the following results for various nanophotonic structures:
\begin{itemize}
\item Three-layer film made of \ce{TiO2}/\ce{Ag}/\ce{TiO2}: mean squared error of 0.000549;
\item Anti-reflective nanocones made of fused silica: mean squared error of 0.000004;
\item Vertical nanowires made of \ce{cSi}: mean squared error of 0.000058;
\item Close-packed nanospheres made of \ce{cSi}/\ce{TiO2}: mean squared error of 0.000061;
\item Three-layer film with double-sided nanocones made of \ce{TiO2}/\ce{Ag}/\ce{TiO2}/\ce{TiO2}/\ce{TiO2}: mean squared error of 0.000020.
\end{itemize}

These results demonstrate the models' ability to reliably estimate the target optical properties across different structural configurations.

\section{Details of Experiments}
\label{sec:details_experiments}

In the main text,
we delineate the comparative analysis of various optimization techniques applied to parametric structure optimization.
The optimization algorithms used in this work can be enumerated as follows:
\begin{itemize}
    \item Random search: a technique of selecting points from a uniform distribution;
    \item Powell's method~\citep{PowellMJD1964tcj}: a method to find a local solution without an assumption on differentiability;
    \item Py-BOBYQA~\citep{CartisC2019acmtoms}: a derivative-free approach for local optimization;
    \item DIRECT~\citep{JonesDR1993jota}: a global optimization method without the Lipschitz constant;
    \item Differential evolution~\citep{StornR1997jgo}: a metaheuristic approach that iteratively improves solution candidates;
    \item Bayesian optimization~\citep{GarnettR2023book}: a probabilistic model-based global optimization strategy for black-box functions.
\end{itemize}

For the implementation of random search,
we employ the NumPy's uniform distributions~\citep{HarrisCR2020nature},
where NumPy is under the BSD license.
Powell's method, DIRECT, and differential evolution are executed via their respective implementations in SciPy~\citep{VirtanenP2020nm}, also BSD licensed.
Py-BOBYQA is run using the implementation obtained from its public repository,\footnote{Accessible here: \href{https://github.com/numericalalgorithmsgroup/pybobyqa}{https://github.com/numericalalgorithmsgroup/pybobyqa}.} which operates under the GNU General Public License.
For Bayesian optimization, we utilize BayesO~\citep{KimJ2023joss}, which is available under the MIT license.
A variety of commercial CPUs,
e.g., AMD EPYC 9374F, AMD EPYC 7302, and Intel Xeon Gold 6126,
are used as the computational resources deployed for dataset creation, benchmarking, and structure optimization.

\section{Experiments for the Combinatorial System with Material Blocks}
\label{sec:combinatorial_material_blocks}

\begin{figure}[ht]
    \centering
    \includegraphics[width=0.31\textwidth]{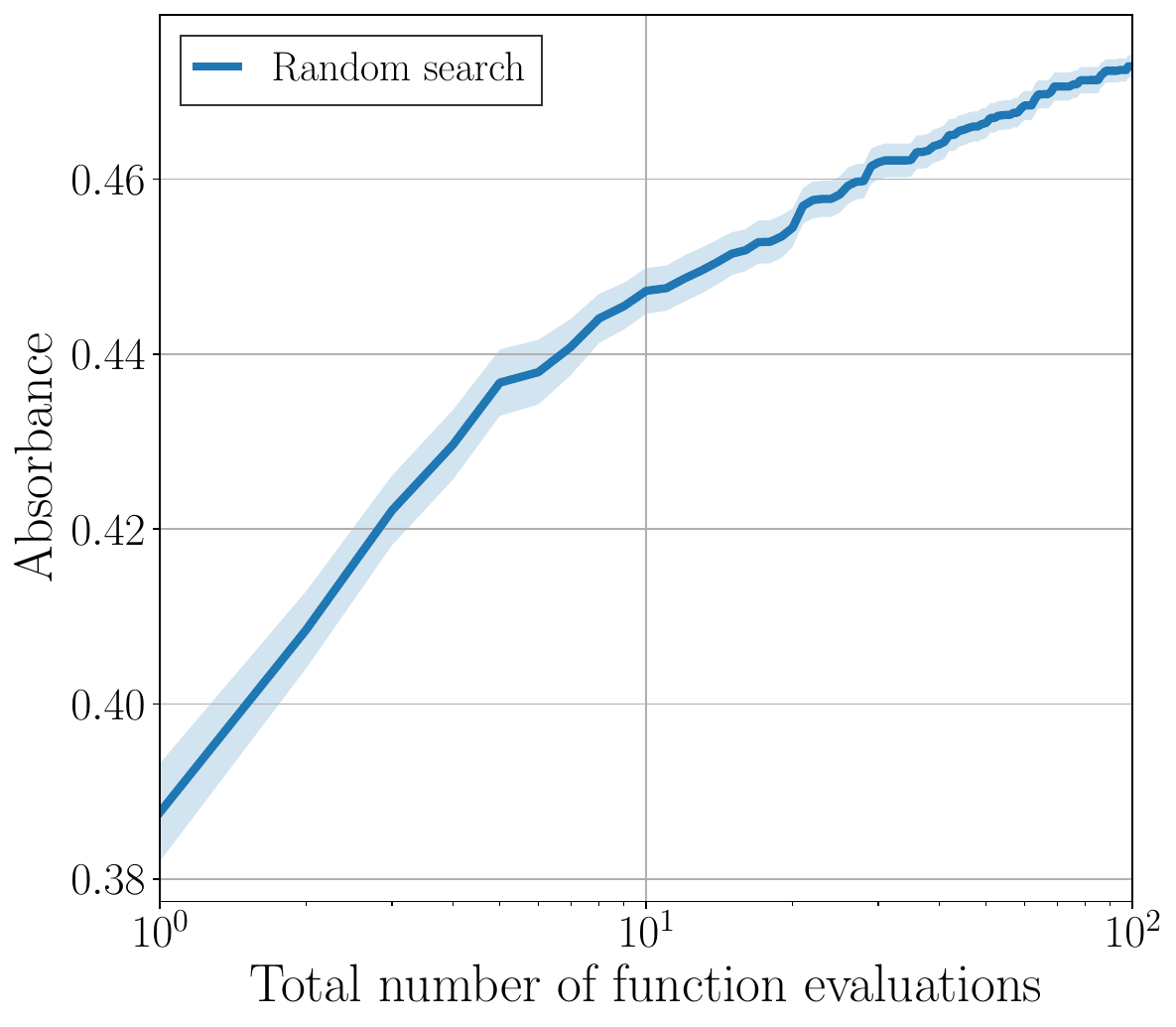}
    \caption{Results of structure optimization for the combinatorial system with material blocks using random search. Similar to~\figref{fig:optimization_experiments}, it repeats each experiment 50 times. The sample mean and the standard error of the sample mean are plotted.}
    \label{fig:optimization_results_combinatorial}
\end{figure}

For the combinatorial system with material blocks,
we test random search with uniform distributions
in the process of selecting materials for all the material blocks in the structure.
The objective of this problem is absorbance maximization
for the AM1.5 global solar spectrum.
To simplify the task,
we dismiss band gaps for particular materials for these experiments.
Since we have twelve options for materials,
the number of possible structural configurations,
i.e., in this case 12$^{\textrm{80}}$, is enormous.
This benchmark can be utilized to compare various combinatorial optimization algorithms,
and the development and utilization of more advanced combinatorial optimization techniques are left for future work.

\section{Future Directions}
\label{sec:future_directions}

Topology optimization via automatic differentiation has been studied in the field of photonics~\citep{HammondAM2021oe,SchubertMF2022acsp}.
Currently, we do not include support for such gradient-based methods.
However, the FDTD simulation software including Meep~\citep{OskooiAF2010cpc} and Ansys Lumerical has added features for efficient gradient computation (with roughly twice the cost of a single function evaluation)~\citep{HammondAM2022oe}.
Incorporating this into our benchmarks is a natural topic for future work.

In addition, we implement a user-friendly process for the integration of additional structures,
supporting the involvement of a broad background of researchers and practitioners.
We aim to constantly develop and maintain this work,
and will strive to actively encourage the contributions of other researchers and practitioners.

\section{Limitations}
\label{sec:limitations}

The simulations we conduct are subject to potential instabilities and require precise settings.
As a result, there are occasions when the combined values of reflectance, absorbance, and transmittance might exceed 1.
It is important to note, however, that the results presented in this work do not yield any negative values in the reflectance, absorbance, and transmittance measurements we are concerned with.
That said, it is within the realm of possibility for simulations to produce negative results.

Additionally, in contexts of continuous optimization with a surrogate model, the regression models we employ are not entirely devoid of errors,
which can stem from both incorrect assumptions made by the model,
i.e., inductive bias,
and the inherent inaccuracies of the model itself.
Eventually,
these errors could introduce inaccuracies into the optimization process.
Despite this potential for discrepancy between the surrogate model and the actual scenario,
the comparisons we make between different optimization algorithms remain valid, provided that the surrogate model used is consistently applied across all optimization procedures.

\section{Societal Impacts}
\label{sec:societal_impacts}

This research contributes to the field of materials science, specifically focusing on nanophotonic structures and their design problems. As such, it is important to consider the potential societal impacts, both negative and positive, that may arise from the practical application of our findings.
On one hand, the manufacturing, fabrication, and disposal of materials and nanophotonic structures could lead to environmental pollution, affecting air, water, and land quality. It is crucial that these pollution issues are addressed proactively and managed effectively to mitigate any adverse effects on the environment and society.
On the other hand, our work has the potential to significantly benefit the renewable energy sector, particularly in the advancement of solar cell technology. By contributing to the development of more efficient solar cells, we are aiding the transition towards a more sustainable energy landscape, which could result in a reduction of greenhouse gas emissions and a move towards a carbon-neutral society.




\end{document}